%% file: main.tex
\documentclass[a4paper,11pt,notitlepage]{report}
\usepackage[left=1in,right=1in,top=1in,bottom=1in,footskip=.25in]{geometry}
\usepackage[utf8]{inputenc}
\usepackage[dvipsnames]{xcolor}
\usepackage{isomath,amsmath,amsfonts,graphicx}
\usepackage{setspace}
\usepackage{mleftright}
\usepackage{algorithm}
\usepackage{algpseudocode}
\usepackage{cite}
\usepackage{hyperref}
\usepackage{tocbibind}
\usepackage{url}
\usepackage{booktabs}
\usepackage{subcaption}
\usepackage{footnote}
\usepackage{tablefootnote}
\usepackage{float}
\usepackage{bbm}
\usepackage{bm}
\usepackage{tikz}
\usepackage{pgfplots}
\pgfplotsset{compat=1.18}
\usetikzlibrary{pgfplots.groupplots}
\usepackage{acro}

\newcommand{\vecs}[1]{\vectorsym{#1}}
\newcommand{\mats}[1]{\matrixsym{#1}}
\newcommand{\tens}[1]{\tensorsym{#1}}

\newcommand{\subsubsubsection}[1]{\vspace{0.1in}\noindent{\bf #1:}}

\DeclareMathAlphabet\mathbfcal{OMS}{cmsy}{b}{n}
\DeclareMathOperator*{\argmax}{arg\,max}
\DeclareMathOperator*{\argmin}{arg\,min}

\begin{document}

\pagenumbering{roman}

\input{text/0a_title}
\clearpage

\input{text/0b_ack}
\clearpage

\tableofcontents
\clearpage
\listoffigures
\clearpage
\listoftables
\clearpage

\input{text/0c_abs}
\clearpage

\pagenumbering{arabic}

\input{text/1_intro}
\clearpage
\input{text/2_back}
\clearpage
\input{text/3_rnn_awe}
\clearpage
\input{text/4_rnn_qbe}
\clearpage
\input{text/5_ctc_a2w}
\clearpage
\input{text/6_multi_awe}
\clearpage
\input{text/7_multi_qbe}
\clearpage
\input{text/8_seg_a2w}
\clearpage
\input{text/9_joint_a2w}
\clearpage
\input{text/10_ssl_awe}
\clearpage
\input{text/11_conc}
\clearpage

\bibliographystyle{IEEEtran}
\bibliography{refs}

\end{document}

%% file: text/0a_title.tex
\definecolor{ttic_blue}{RGB}{55,93,137}
\begin{tikzpicture}[overlay,remember picture]
\draw[line width=1mm][ttic_blue]($(current page.north west)+(0.9in,-0.9in)$)rectangle($(current page.south east)+(-0.9in,0.9in)$);
\end{tikzpicture}
\begin{spacing}{1.25}
\begin{center}
\Large
\MakeUppercase{{\bf Neural approaches to\\spoken content embedding}}\\
\normalsize
\vspace{0.25in}
\MakeUppercase{by\\Shane Settle}\\
\vspace{1in}
A thesis submitted\\
in partial fulfillment of the requirements for\\
the degree of\\
\vspace{0.25in}
Doctor of Philosophy in Computer Science\\
\vspace{0.25in}
at the\\
\vspace{0.25in}
\MakeUppercase{Toyota Technological Institute at Chicago}\\
Chicago, Illinois\\
\vspace{0.25in}
September, 2023\\
\vspace{1in}
Thesis Committee:\\
Karen Livescu (Thesis Advisor)\\
Kevin Gimpel\\
Kartik Audhkhasi\\
Herman Kamper\\
\end{center}
\end{spacing}
\thispagestyle{empty}

%% file: text/0b_ack.tex
\chapter*{Acknowledgements}
\addcontentsline{toc}{chapter}{Acknowledgements}

Thank you to the teachers, mentors, and friends at UCLA and prior, for helping to build the foundation I needed to succeed on this path.\\

\noindent Thank you to the community of professors, colleagues, and friends I have made both at TTIC and on internships, for the insights, guidance, and laughs along the way.\\

\noindent Thank you to my advisor Karen Livescu, for the wisdom you shared, the patience you showed, and the time you spent to make me the researcher that I am today.\\

\noindent And thank you to my parents and family, for the unyielding support you continue to offer along this greater journey of life.\\

\noindent I am grateful for the experiences I have had, and I am excited for what is to come next.

%% file: text/0c_abs.tex
\begin{abstract}
\addcontentsline{toc}{chapter}{\abstractname}
Comparing spoken segments is a central operation to speech processing. Traditional approaches in this area have favored frame-level dynamic programming algorithms, such as dynamic time warping, because they require no supervision, but they are limited in performance and efficiency. As an alternative, acoustic word embeddings---fixed-dimensional vector representations of variable-length spoken word segments---have begun to be considered for such tasks as well. These embeddings can be learned discriminatively such that they are similar for speech segments corresponding to the same word, while being dissimilar for segments corresponding to different words. Acoustic word embedding models also speed up segment comparison, which reduces to a dot product between segment embedding vectors. However, the current space of such discriminative embedding models, training approaches, and their application to real-world downstream tasks is limited.

We start by considering ``single-view" training losses where the goal is to learn an acoustic word embedding model that separates same-word and different-word spoken segment pairs. Then, we consider ``multi-view" contrastive losses. In this setting, acoustic word embeddings are learned jointly with embeddings of character sequences to generate {\it acoustically grounded} embeddings of written words, or acoustically grounded word embeddings; such embeddings have been used to improve speech retrieval, recognition, and spoken term discovery.

In this thesis, we contribute new discriminative acoustic word embedding (AWE) and acoustically grounded word embedding (AGWE) approaches based on recurrent neural networks (RNNs). We improve model training in terms of both efficiency and performance. We take these developments beyond English to several low-resource languages and show that multilingual training improves performance when labeled data is limited. We apply our embedding models, both monolingual and multilingual, to the downstream tasks of query-by-example speech search and automatic speech recognition. Finally, we show how our embedding approaches compare with and complement more recent self-supervised speech models.
\end{abstract}

%% file: text/1_intro.tex
\chapter{Introduction}

Speech technology research produces many useful and important tools. Among them, speech-to-text systems are one prominent success story. However, performance improvements to automatic speech recognition, the task of transcribing speech into text, has largely been enabled by data intensive techniques that require significant labeled resources and linguistic expertise. When such resources are available, success in recognition has meant success in other speech tasks as well, such as query-by-example or keyword search, for which conventional recognition training recipes can be re-purposed. However, for low- and zero-resource data regimes where these search tasks can be most crucial, success in speech recognition often remains elusive, necessitating the use of different techniques developed with these resource restrictions in mind. This thesis investigates one such technique in the discriminative embedding of spoken segments into fixed-dimensional vectors~\cite{maas2012word,levin2013fixed,bengio2014word,kamper2016cnn_awe}. 

\begin{figure}
\footnotesize
\begin{minipage}{0.45\textwidth}
  \centering
  \includegraphics[width=0.925\linewidth]{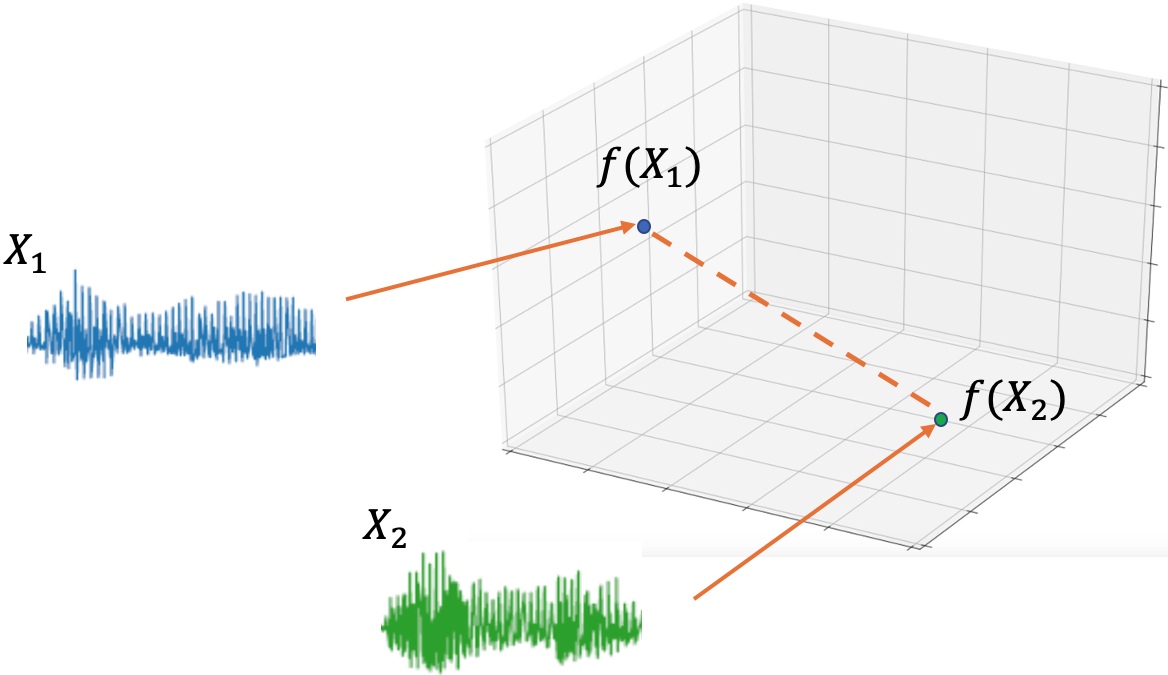}
  \caption*{(a)}
\end{minipage}%
\begin{minipage}{0.55\textwidth}
  \scriptsize
  \centering
  \includegraphics[width=0.825\linewidth]{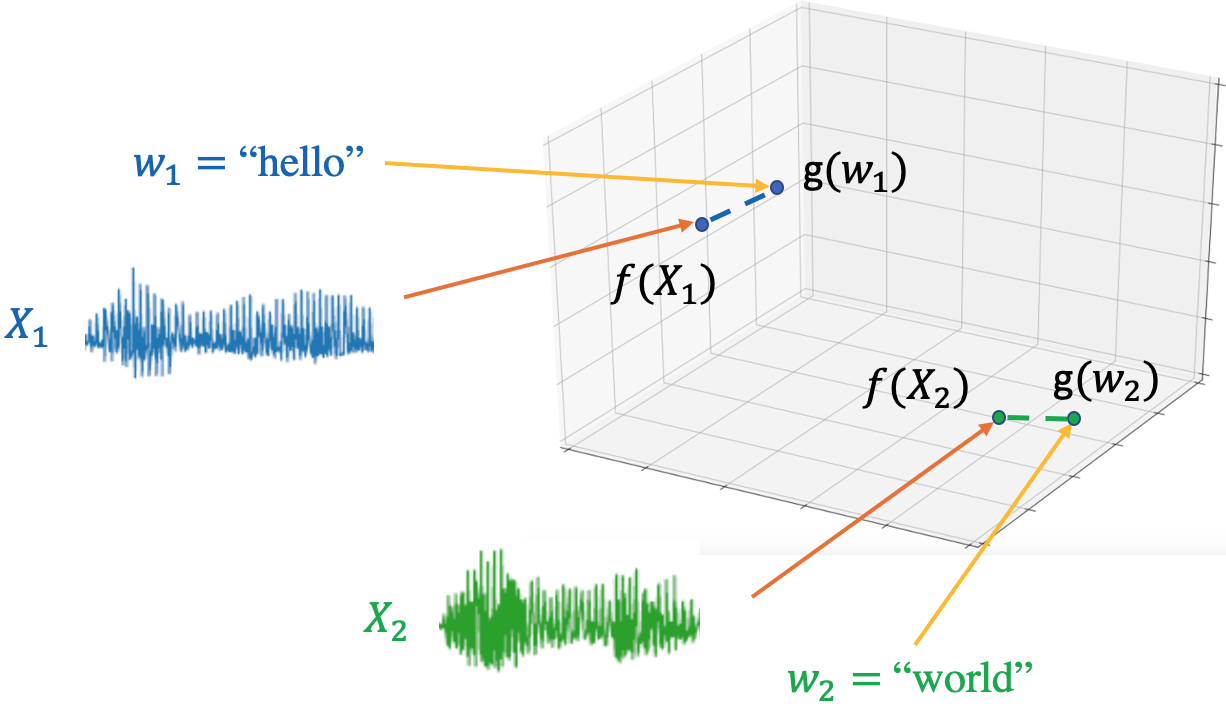}
  \caption*{(b)}
\end{minipage}
\caption{Illustrations of embedding spaces learned to model acoustic-phonetic similarity. In (a), our embedding space is mapped to by a single function that takes in speech segments and produces vector embeddings, while (b) is an embedding space mapped to by two functions accepting different inputs, one from the acoustic domain as in (a) and another from the textual domain.}
\label{fig:embs}
\end{figure}

The goal is to learn an embedding space for application to spoken language tasks such as speech recognition and search that respects acoustic-phonetic content but disregards nuisance variations (e.g., tempo, dialect, gender, noise). Our methods are effective even when training on limited data, in challenging acoustic conditions, and across several diverse languages. In addition to our contributions to acoustic embedding techniques, we also showcase their utility with applications to two downstream tasks: query-by-example speech search and automatic speech recognition.

We propose approaches that predominately operate on the level of spoken word segments, termed {\it acoustic word embeddings} (Figure~\ref{fig:embs}a). Acoustic word embeddings are vector representations of variable-length spoken word segments. The goal is to embed spoken segments corresponding to the same word close together, while spoken segments of different words are mapped farther apart. We consider this the ``single-view" approach as there is only one embedding model, and it operates on speech segments. For settings requiring comparison between spoken segments and written words, we also investigate ``multi-view" learning of acoustic word embeddings alongside embeddings of their written word labels (Figure~\ref{fig:embs}b). The multi-view approach uses multiple embedding models that each support a different ``view" of the data. For example, in Figure~\ref{fig:embs}b, the function $f$ embeds spoken word segments from their acoustic sequences, while the function $g$ embeds written words from their character sequences. By embedding spoken word segments close to their written word labels, spoken segments with the same word label will also be close. While embeddings of written words are also a common tool in natural language processing (NLP), there they often capture word meaning; our objective is instead to learn to represent acoustic-phonetic similarity, or how a word sounds, so we will refer to our written word embeddings as {\it acoustically grounded} word embeddings.

The first downstream application we consider is {\it query-by-example} speech search. Query-by-example is the task of searching for a spoken query in a collection of speech recordings. This requires direct comparison between audio segments. Traditional low-resource query-by-example search systems rely on a frame-level dynamic programming algorithm, called dynamic time warping, to measure similarity between spoken queries and spoken segments of the search collection. However, dynamic time warping has known shortcomings~\cite{rabiner1978dtw} both in discriminative performance and in efficiency. Representing variable-length spoken segments as fixed-dimensional embeddings~\cite{levin2015srails} simplifies segment comparison to a dot product, which is more efficient to compute and more easily optimized end-to-end.

Our second application area is automatic speech recognition. A subset of work in automatic speech recognition aims to implicitly learn a joint acoustic, pronunciation, and language model by word-level training under a unified, end-to-end framework, called {\it acoustic-to-word} speech recognition. This approach avoids explicit modeling and prediction of subword structure, which still poses challenges~\cite{livescu2012subword} despite decades of research. However, a significant obstacle to acoustic-to-word recognition models is learning to represent rare and out-of-vocabulary words. After successful application to query-by-example search in the low- and zero-resource domain, we use multi-view acoustic and acoustically grounded word embedding methods to supplement standard recognition training approaches by targeting the representation challenges of acoustic-to-word recognition.

In summary, our contributions span improvements to acoustic word embedding and acoustically grounded word embedding methods, their extension to multilingual modeling, as well as their successful application to the downstream tasks of query-by-example speech search and acoustic-to-word speech recognition where they outperform prior work.

\section{Contributions}
\label{ch:intro:contrib}

\subsection*{Improvements to acoustic embedding modeling~{\scriptsize \cite{settle2016rnn_awe,hu2020multilingual}}}
First, we propose new acoustic word embedding models based on recurrent neural networks (RNNs), compare several types of RNN-based embeddings trained using different objectives, and find our best models outperform prior approaches on acoustic word discrimination~\cite{settle2016rnn_awe}. Second, we make performance and efficiency improvements to multi-view joint training of acoustic word embedding (AWE) and acoustically grounded word embedding (AGWE) models. We improve performance by using phones, rather than characters, as input to the AGWE model, and by exploring negative sampling techniques to train better models more efficiently than in prior work~\cite{hu2020multilingual}.

\subsection*{Multilingual acoustic and written word embedding~{\scriptsize \cite{hu2020multilingual}}}
For target languages with limited labeled data, we find that pretraining AWE and AGWE models on several non-target languages and then fine-tuning on a small amount of data from the target language produces higher-quality embeddings than training on target language data alone~\cite{hu2020multilingual} (as measured by word discrimination performance). Also, if no data is available for fine-tuning, our zero-shot acoustic word discrimination performance far exceeds dynamic time warping. Multilingual embedding models are enabled by our use of phones in place of characters as input to the AGWE model. We then obtain further improvements when using distinctive features to represent phones, rather than simple phone labels, in cases where many phones in the target language are unseen during training~\cite{hu2020multilingual}.

\subsection*{Embedding-based query-by-example search~{\scriptsize \cite{settle2017query,hu2021ase}}}
We apply our single-view neural acoustic word embedding model to query-by-example search~\cite{settle2017query}, achieving large improvements over prior work~\cite{levin2015srails}. Second, we extend embedding training to multi-word spans, and apply our span embeddings to query-by-example search on a number of unseen languages~\cite{hu2021ase}. Not only does this approach~\cite{hu2021ase} achieve performance improvements over prior work~\cite{hou2015nni_qbe,proencca2016segmented,leung2016toward}, but it also represents the first application of both multilingual embedding models {\it and} multi-word span embedding models to query-by-example speech search. 

\subsection*{Acoustic-to-word speech recognition~{\scriptsize \cite{settle2019_a2w,shi2021seg}}}
We improve acoustic-to-word speech recognition by directly addressing the challenges of infrequent and out-of-vocabulary words. First, we use our multi-view acoustic and acoustically grounded word embedding models to initialize acoustic-to-word models prior to recognition training~\cite{settle2019_a2w}, resulting in performance improvements over prior comparable approaches. Next, we develop the first end-to-end training of whole-word segmental models, which also outperform prior acoustic-to-word systems~\cite{shi2021seg}. Then, we extend both of these approaches with joint training of embedding and recognition models, significantly improving performance and outperforming prior work~\cite{audhkhasi2017a2w,audhkhasi2018a2w,yu2018multistage} on acoustic-to-word recognition. Our best jointly trained word-level model is competitive with similar capacity subword models on conversational speech data from English (Switchboard-$300$h~\cite{audhkhasi2019forget}) as well as several low-resource languages (Babel~\cite{inaguma2019transfer,conneau2020unsupervised}).

\subsection*{Extracting acoustic word embeddings with self-supervised models}
We investigate whether pretrained self-supervised models such as Wav2Vec2~\cite{baevski2020wav2vec2}, HuBERT~\cite{hsu2021hubert}, and WavLM~\cite{chen2022wavlm} can be used out-of-the-box to embed spoken word segments discriminatively. We compare the AWEs derived from self-supervised pretrained models to AWEs trained using our RNN-based multi-view method. We find that WavLM can offer competitive results with our approach, but performance varies depending on the similarity between the data domain used for model pretraining and that used in our acoustic word discrimination evaluations. Furthermore, we find that using our supervised multi-view training approach on top of the pretrained frame-level representations offers significant improvement over either approach on its own.\footnote{Since our work predates self-supervised transformer models, this chapter serves to contextualize our prior work within the more recent literature. Our comparisons target the acoustic word discrimination task, but further investigation of downstream tasks is left to future work.}

\section{Thesis outline}

In Chapter~\ref{ch:back}, we cover preliminaries to introduce concepts essential for understanding the methods, tasks, and language of the thesis as well as the related work necessary to contextualize our contributions described in Section~\ref{ch:intro:contrib}. These contributions are then discussed at length throughout Chapters~\ref{ch:rnn_awe}-~\ref{ch:ssl_awe}. Chapters~\ref{ch:rnn_awe} and~\ref{ch:rnn_qbe} focus on the recurrent neural network (RNN)-based acoustic word embedding methods, first applied to the proxy task of acoustic word discrimination (Chapter~\ref{ch:rnn_awe}), and then to the downstream task of query-by-example search (Chapter~\ref{ch:rnn_qbe}). Next, Chapter~\ref{ch:ctc_a2w} explores the first application of our work to improve acoustic-to-word speech recognition performance through updates to multi-view pretraining of acoustic and acoustically grounded word embedding (AWE+AGWE) models. Chapters~\ref{ch:multi_awe} and~\ref{ch:multi_qbe} explore the extension of multi-view acoustic and acoustically grounded word embedding (AWE+AGWE) modeling to multilingual training and evaluation on a variety of low- and zero-resource languages. Chapter~\ref{ch:multi_awe} focuses on multilingual extension and proxy evaluation using the acoustic and cross-view word discrimination tasks, while Chapter~\ref{ch:multi_qbe} investigates application to downstream multilingual query-by-example search. Chapters~\ref{ch:seg_a2w} and~\ref{ch:joint_a2w} build upon our prior work in Chapter~\ref{ch:ctc_a2w} incorporating our embedding models to acoustic-to-word recognition training. Chapter~\ref{ch:seg_a2w} discusses our introduction of segmental modeling to acoustic-to-word recognition. Chapter~\ref{ch:joint_a2w} extends the acoustic-to-word work from Chapters~\ref{ch:ctc_a2w} and~\ref{ch:seg_a2w} by introducing the multi-view embedding loss both in pretraining but also during recognition training as a joint objective. Additionally, Chapter~\ref{ch:joint_a2w} compares acoustic-to-word modeling with {\it static} versus {\it dynamic} word-level lexicons. The final chapter covering our contributions is Chapter~\ref{ch:ssl_awe}, which compares our learned multi-view embedding models with representations obtained from recent popular self-supervised learning approaches.

%% file: text/2_back.tex
\chapter{Background}
\label{ch:back}

Many speech processing tasks---such as automatic speech recognition or spoken term detection---hinge on associating segments of speech signals with word labels. In systems developed for such tasks, words are often broken into subword units such as phones, and models are built for the individual units. An alternative is to consider each word segment as a single unit. These whole-word approaches enable us to consider more flexible features derived over longer time spans, and avoid some of the challenges of subword modeling~\cite{ostendorf1999beads,livescu2012subword}.

In this chapter, we start by defining several preliminaries (Section~\ref{ch:back:prelims}) that cover some of the key terms and methods we will reference throughout the thesis. Then, we describe the tasks we will be evaluating our methods on, including both downstream tasks (Section~\ref{ch:back:prelims:goals}) and proxy tasks (Section~\ref{ch:back:prelims:proxies}). Finally, we discuss related work (Section~\ref{ch:back:related}) to contextualize our contributions (Section~\ref{ch:intro:contrib}) within the literature.

\section{Preliminaries}
\label{ch:back:prelims}

This section begins with a description of frame-level acoustic features since they will serve as the input to many of the systems we describe later. Then, we detail the dynamic time warping (DTW) algorithm since it is an essential baseline which we reference and compare with throughout the thesis. Next, we define the primary concepts of this thesis: acoustic word embeddings (AWEs) and acoustically grounded word embeddings (AGWEs). Last, we introduce the tasks that will be used to measure the performance of our methods and benchmark them against prior work. These include both the goal downstream tasks (Section~\ref{ch:back:prelims:goals}) as well as common task-agnostic proxy tasks (Section~\ref{ch:back:prelims:proxies}).

\subsection{Frame-level acoustic features}
\label{ch:back:prelims:frames}

Speech data is stored in the form of raw waveforms. These are one dimensional vectors tracking changes in air pressure collected by a microphone. By adjusting the rate at which we sample these points, we can capture how sound intensity changes over time with a higher or lower resolution. Common sampling rates for the datasets we will be considering include $8$ and $16$ kilohertz (kHz) where a $16$ kHz sampling rate implies that $16,000$ samples are recorded by the microphone per second. Rather than using this waveform as input to our models, we preprocess them into frame-level acoustic features. First, we perform a framing operation where an analysis window slides across the waveform, typically with a width of $25$ ms and a slide of $10$ ms. In the case of a $5$ second audio snippet collected at a $16$ kHz sampling rate, this windowing process takes a 1-D waveform (a vector) with $80,000$ samples, converts it into a 2-D matrix that is $400$ samples by $500$ frames (i.e. samples from a $25$ ms window shifted every $10$ ms), and applies a windowing operation (such as a Hamming window) to each frame. Next, the fast fourier tranform (FFT) is applied framewise to decompose the windowed signal into frequency coefficients. Depending on the task, a variety of features can now be produced from these coefficients. Our work will primarily use log-Mel spectra or mel-frequency cepstral coefficients, which are computed from the power spectrum derived from the frequency coefficients output by the FFT. 

\begin{algorithm}
\caption{Dynamic time warping (DTW) algorithm returning the cost of the optimal alignment between two variable-length sequences. Inputs $\mats{X}$ and $\mats{Y}$ are length $N$ and $M$ speech segments, respectively, with $D$-dimensional acoustic feature frames.}
\label{algo:dtw}
\begin{algorithmic}[1]
\Require{$\mats{X} := \{\mats{X}_t\}_{t=1}^{N}, \mats{Y} := \{\mats{Y}_t\}_{t=1}^{M} \text{ where } \mats{X}_t \in \mathbb{R}^D, \mats{Y}_t \in \mathbb{R}^D$}
\Ensure{$\mats{C}_{N, M} :=$ frame-level DTW sequence alignment cost}\\

\Function{Dist}{$\vecs{a}, \vecs{b}$} \Comment{Defines frame-level distance (e.g. cosine).}
    \State \Return {$1 - \frac{\vecs{a}^T \vecs{b}}{\Vert \vecs{a} \Vert \Vert \vecs{b} \Vert}$}
\EndFunction\\

\Function{MoveSet}{$i, j$} \Comment{Defines permissible moves to index $(i,j)$.}
    \State \Return {\{$(i-1,j),(i,j-1),(i-1,j-1)$\}}
\EndFunction\\

\Function{Init}{$N,M$} \Comment{Initialize cost matrix $\mats{C}$.}
\For{$i \gets 0$ to $N$}
    \For{$j \gets 0$ to $M$}  
        \State {$\mats{C}_{i, j} \gets \infty$}
    \EndFor
\EndFor
\State {$\mats{C}_{0, 0} \gets 0$}
\State \Return {$\mats{C}$}
\EndFunction\\

\Function{DTW}{$\mats{X}, \mats{Y}$}
\State {$N \gets length(\mats{X}), M \gets length(\mats{Y})$}
\State {$\mats{C} \gets$ \Call{Init}{$N, M$}}
\For{$i \gets 1$ to $N$}
    \For{$j \gets 1$ to $M$}
        \State {$d \gets$ \Call{Dist}{$\mats{X}_i, \mats{Y}_j$}}
        \State {$\mathcal{M} \gets$ \Call{MoveSet}{$i,j$}} 
        \State {$\mats{C}_{i, j} \gets d + \displaystyle \min_{i', j' \in \mathcal{M}} \mats{C}_{i',j'}$}
    \EndFor
\EndFor
\State \Return {$\mats{C}_{N,M}$}
\EndFunction
\end{algorithmic}
\label{ch:back:algo:dtw}
\end{algorithm}

\subsection{Dynamic time warping}
\label{ch:back:prelims:dtw}

Dynamic time warping (DTW) is a dynamic programming algorithm (Algorithm~\ref{algo:dtw}) designed to measure the similarity between two sequences of differing lengths, and it is a powerful baseline when comparing speech segments~\cite{vintsyuk1968speech,sakoe1978dynamic}. In Algorithm~\ref{ch:back:algo:dtw}, $\mats{X}$ (similarly $\mats{Y}$) is a speech segment represented as a sequence of frame-level acoustic features $\{\vecs{x}_t\}_{t=1}^T$ where $\vecs{x}_t \in \mathbb{R}^D$ such that $\mats{X}$ is effectively a matrix in $\mathbb{R}^{T \times D}$. However, any two speech segments $(\mats{X}, \mats{Y})$ are very unlikely to be of the same length, even if they contain the exact same word or phrase and are spoken by the same speaker. The flexibility of DTW enables measuring the similarity between any two speech segments by finding the best alignment cost between their frame-level acoustic features; however, this comes at the cost of a runtime complexity that is quadratic in sequence length (i.e. $\mathcal{O}(NM)$ when looking at Algorithm~\ref{algo:dtw}). The DTW algorithm has been used within a number of speech systems ranging from early work on automatic speech recognition~\cite{dewachter2007template,heigold2012investigations} to query-by-example speech search~\cite{hazen2009query_posteriorgram_templates,zhang2009unsupervised_spoken_keyword_spotting,zhang2011piecewise_posteriorgram_dtw,zhang2012fast,jansen2012rails,mantena2013speedup_dtw_hierarchical_kmeans,leung2016toward} and word discrimination~\cite{carlin2011rapid_eval_spoken_term_detect,levin2013fixed,jansen2013weak_topdown,kamper2015unsupervised_weak_topdown}. Several methods of prior work explore improvements to DTW either from incorporating higher quality frame-level features~\cite{jansen2013weak_topdown,kamper2015unsupervised_weak_topdown,xu2016approximate,leung2016toward} or efficiency from approximation techniques to speed up segment comparisons~\cite{jansen2012rails,xu2016approximate,leung2016toward}. Methods that use DTW for segment comparison rely on learning high quality acoustic features, but any further adjustments come from tuning the choice of frame-level distance metric as well as the permitted move set.

\begin{figure}
\footnotesize
\centering
\includegraphics[width=0.6\linewidth]{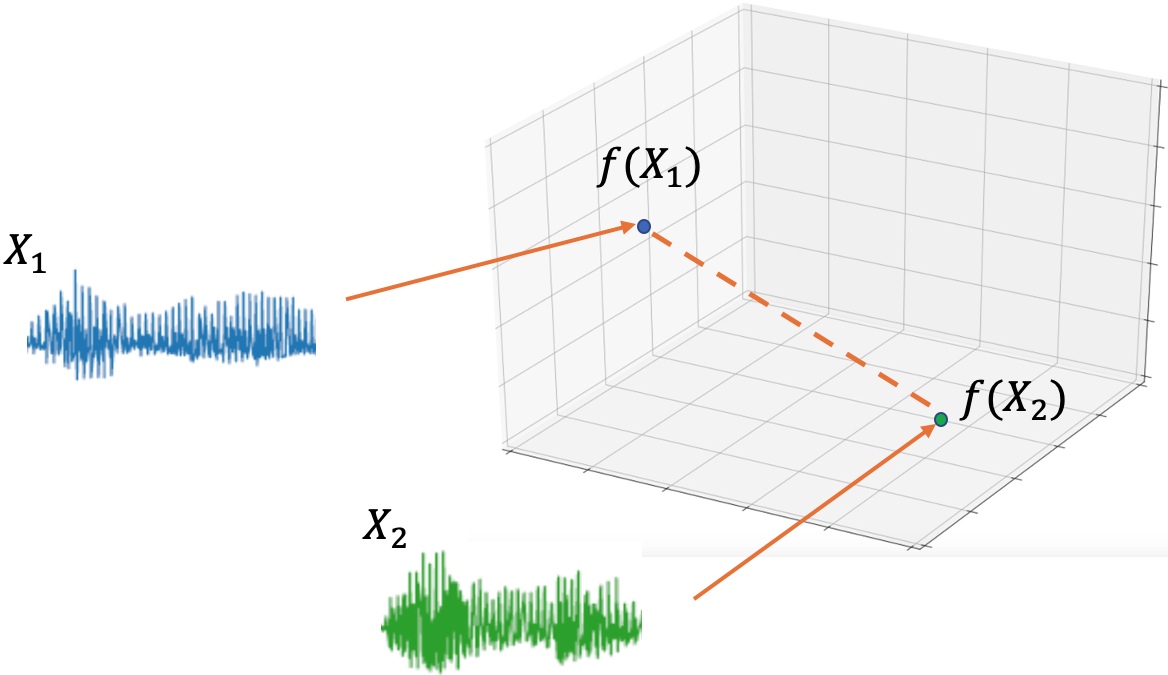}
\caption{Illustration of acoustic word embeddings (AWEs). Spoken word segments $\mats{X}_1$ and $\mats{X}_2$ are mapped into the acoustic word embedding space defined by the function $f$.}
\label{ch:back:fig:awe_grid}
\end{figure}

\subsection{Acoustic word embeddings}
\label{ch:back:prelims:awes}

An AWE model (Figure~\ref{ch:back:fig:awe_grid}) is a function $f$ where the input is a variable-length speech segment $\mats{X} := \{\vecs{x}_t\}_{t=1}^T$ with frame-level acoustic features $\vecs{x}_t \in \mathbb{R}^D$, and the output is a fixed-dimensional vector $f(\mats{X}) \in \mathbb{R}^d$. AWE models will ideally embed two acoustic segments corresponding to the same word close together, while embedding segments corresponding to different words farther apart. Once word segments are represented via fixed-dimensional embeddings, computing pairwise segment distances is as simple as measuring a cosine or Euclidean distance between their embeddings. In practice, the input $\mats{X}$ can be any acoustic segment, whether or not it corresponds to a single spoken word. However, since word-level segmentations~\footnote{These segmentations can be obtained with varying levels of supervision, so they may not always correspond precisely to the ground-truth word boundaries.} are used for training, evaluation, or both, we will refer to the learned function $f$ as an AWE model, and its outputs as AWEs.

\begin{figure}
\centering
\includegraphics[width=0.65\linewidth]{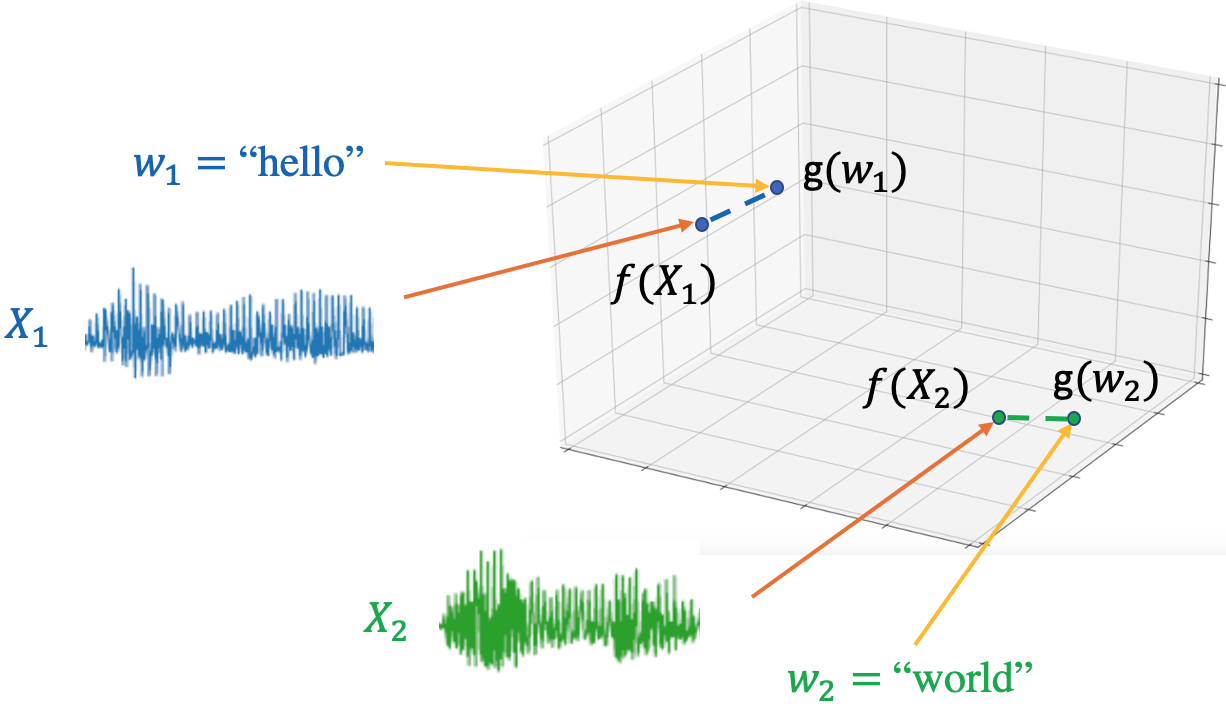}
\caption{Illustration of acoustic word embeddings (AWEs) and acoustically grounded word embeddings (AGWEs). Spoken word segments $\mats{X}_1$ and $\mats{X}_2$ are mapped into AWEs by the function $f$, and the corresponding words $w_1$ and $w_2$ are mapped into AGWEs using their character sequences.}
\label{ch:back:fig:agwe_grid}
\end{figure}

\subsection{Acoustically grounded word embeddings}
\label{ch:back:prelims:agwes}

While AWE models produce a unique embedding for each instance of a spoken word, some tasks necessitate comparison between spoken word instances and written word labels (Figure~\ref{ch:back:fig:agwe_grid}). For such purposes, we want to learn not only embeddings of spoken words (i.e. AWEs), but also embeddings of written words that capture how they sound when spoken. We refer to these word embeddings as {\it acoustically grounded word embeddings} (AGWEs).

Our definition of an AGWE model has two parts. First, the model is a function $g$ where the input is a sequence of characters $\vecs{c}$, and the output is a fixed-dimensional vector $g(\vecs{c}) \in \mathbb{R}^d$. Second, the parameters of this function are learned jointly with speech data (such as AWEs) for acoustic grounding.

\begin{figure}[H]
\centering
\includegraphics[width=0.795\linewidth]{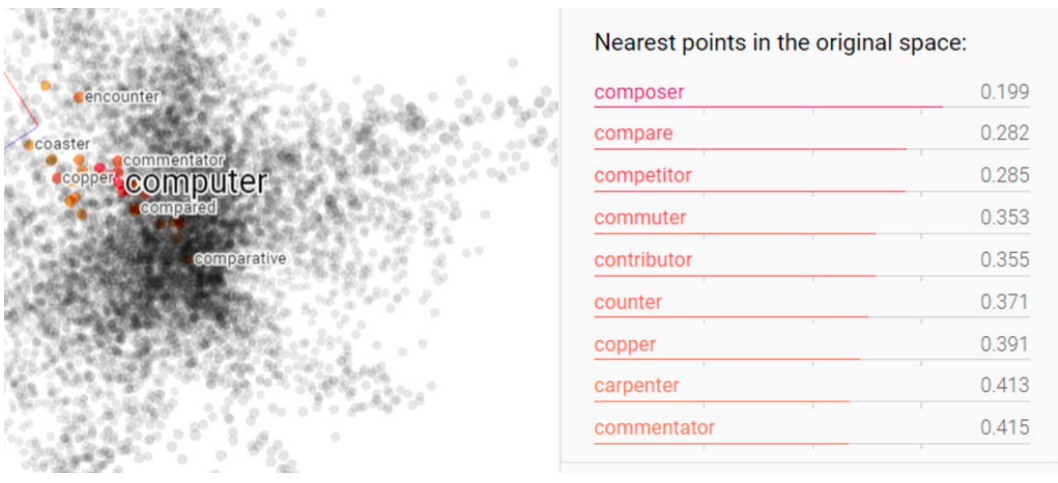}
\caption{Nearest neighbors of the word ``computer" in a learned acoustically grounded word embedding space; figure credit to Karen Livescu (from \href{https://projector.tensorflow.org}{TensorBoard Embedding Projector}).}
\label{ch:back:fig:acoustic_embs}
\end{figure}

\begin{figure}[H]
\centering
\includegraphics[width=0.795\linewidth]{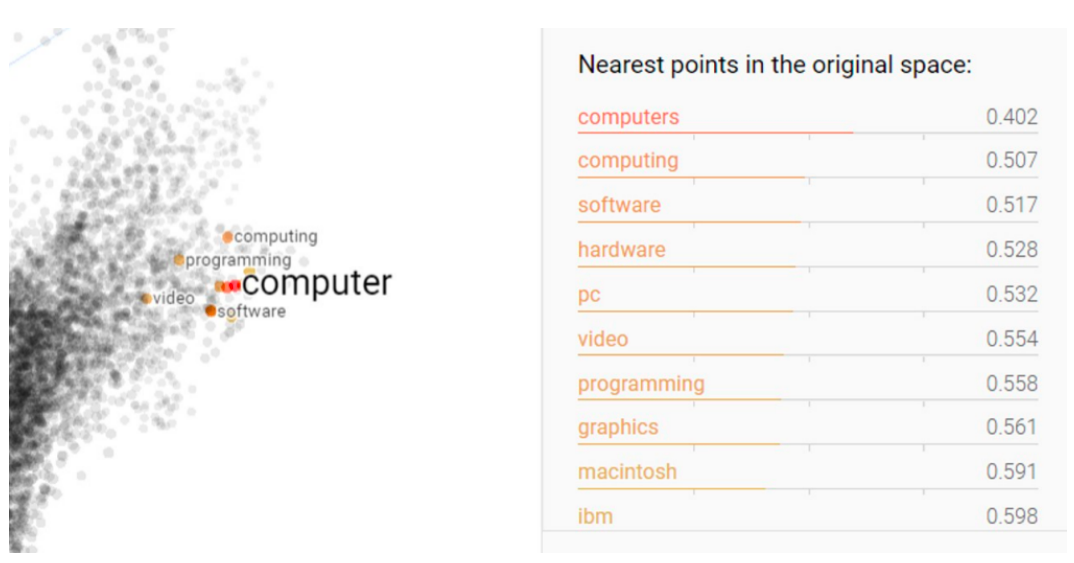}
\caption{Nearest neighbors of the word ``computer" in a learned semantic word embedding space; figure credit to Karen Livescu (from \href{https://projector.tensorflow.org}{TensorBoard Embedding Projector}).}
\label{ch:back:fig:semantic_embs}
\end{figure}

We choose this definition carefully. The first part gives our AGWE model $g$ flexibility. Not only can the AGWE model accept any character sequence $\vecs{c}$ supported by the model's character vocabulary, but it also enables information sharing between words with character overlap. The second part ensures a distinction between AGWEs and popular written word embeddings from natural language processing (NLP) that are learned from text alone such as word2vec~\cite{mikolov2013efficient}, GloVe~\cite{pennington2014glove}, and FastText~\cite{bojanowski2017enriching}. 
These alternative semantic word embedding spaces used in NLP are structured with the goal that distances between vector representations reflect differences in word meaning (Figure~\ref{ch:back:fig:semantic_embs}). However, the downstream goal tasks we consider in this thesis instead require reasoning about the acoustic-phonetic content of written words, so we structure our learned embedding space accordingly (Figure~\ref{ch:back:fig:acoustic_embs}).

\section{Downstream tasks}
\label{ch:back:prelims:goals}

We now describe our downstream applications of interest. These will be the two end goal tasks that we use to test our representations. In addition, we will also benchmark our work in the context of the literature using proxy tasks detailed later in Section~\ref{ch:back:prelims:proxies}.

\subsection{Query-by-example speech search}
\label{ch:back:prelims:qbe}

\begin{figure}[H]
\centering
\includegraphics[width=0.675\linewidth]{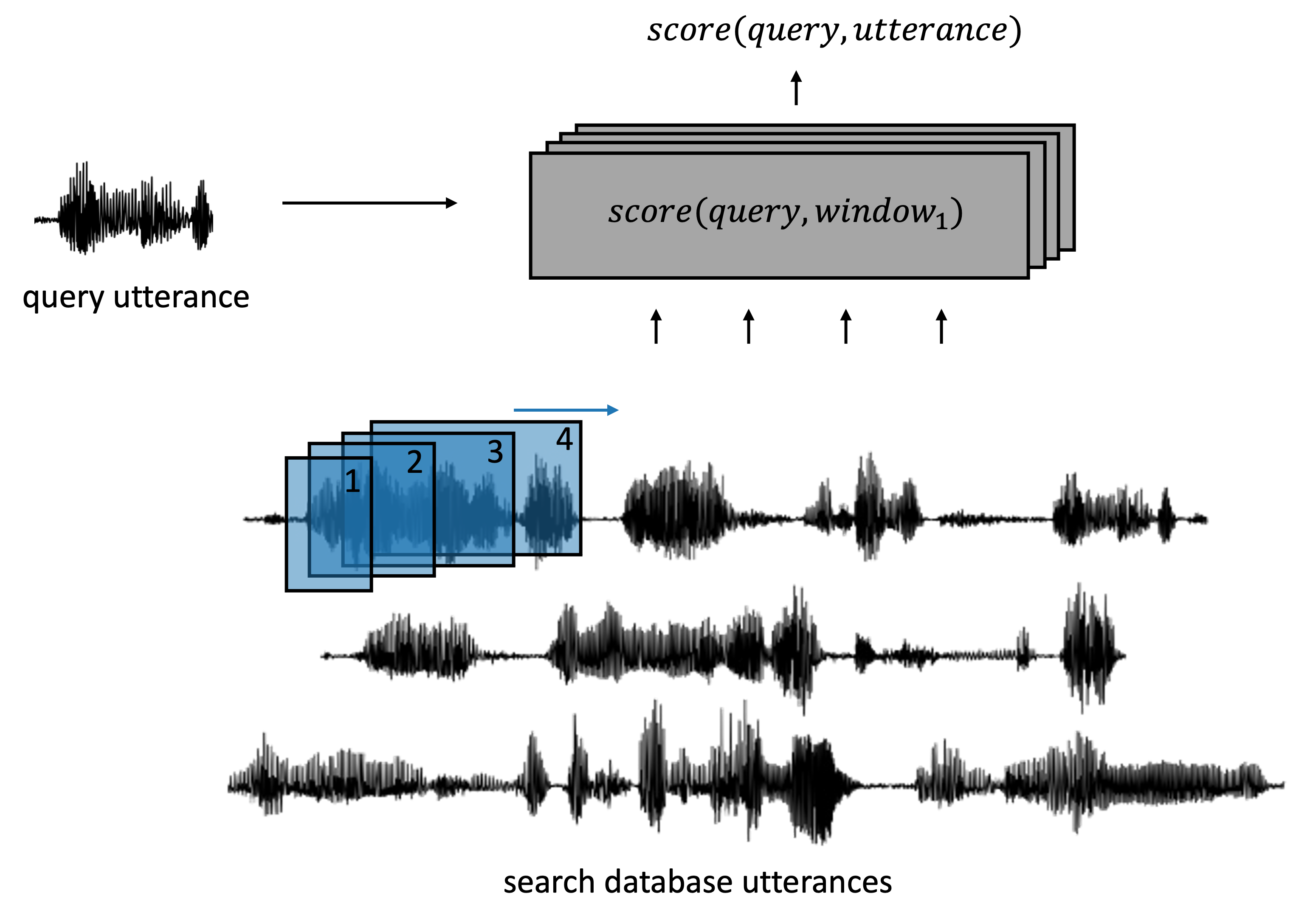}
\caption{The query utterance is scored against a number of windowed segments from each search database utterance to determine matches.}
\label{ch:back:fig:qbe}
\end{figure}

Query-by-example speech search (QbE) (Figure~\ref{ch:back:fig:qbe}) is the task of searching for a spoken query term (a word or phrase) in a collection of speech recordings~\cite{shen2009comparison,parada2009query,hazen2009query_posteriorgram_templates,zhang2011piecewise_posteriorgram_dtw,jansen2012rails,mantena2013speedup_dtw_hierarchical_kmeans,szoke2015query}, which involves directly matching audio segments. 
This is distinct from keyword search (KWS) or spoken term detection (STD) (Section~\ref{ch:back:addl:kws}) where the search terms are given as text~\cite{fiscus1970spoken_term_detection}. The text-free QbE speech search task arises naturally when search terms may be out-of-vocabulary~\cite{shen2009comparison,parada2009query}, in hands-free conditions, or in low- or zero-resource settings~\cite{szoke2015query}.

Specifics of the QbE task can vary depending on the desired use case, but the most basic example is to return a set of utterance IDs from the search collection that are most likely to contain matches to the spoken query. System performance is then evaluated on whether or not those returned utterances truly contain query matches. Some variants of the QbE search task involve localizing the query within the flagged spoken utterances as well, but that is not part of the task we consider here. Also, the specifics of what constitutes a match can vary. Some settings want only ``exact" matches, which is where the text corresponding to the spoken query word or phrase must occur identically within the utterance transcript. Other settings relax this requirement in favor of ``approximate" matches such that the utterance transcript can contain different word forms or a reordering of the text corresponding to the query and still qualify as a match. Since benchmark rules, conditions, and metrics vary, we leave further details to the relevant experiment sections.

\subsection{Automatic speech recognition}
\label{ch:back:prelims:asr}

Automatic speech recognition (ASR) is the task of mapping continuous-valued speech signals to discrete linguistic units. While these units can be smaller, such as phones or characters, or larger such as whole words, the evaluation metric most commonly used for ASR is word error rate (WER). Traditional systems~\cite{waibel1989asr,gauvain2003conversational,chen2006advances,matsoukas2006advances,jurafsky2009speech,saon2015ibm_asr,xiong2016achieving} for automatic speech recognition (ASR) are modular pipelines composed of separately trained and optimized components including acoustic, pronunciation, and language models. In the last several years, an alternative system for ASR has been gaining popularity that focuses on replacing the modular training approaches of traditional systems with conceptually simpler end-to-end methods, which allow for joint optimization of a single objective.

\begin{figure}[b]
    \centering
    \includegraphics[width=0.875\linewidth]{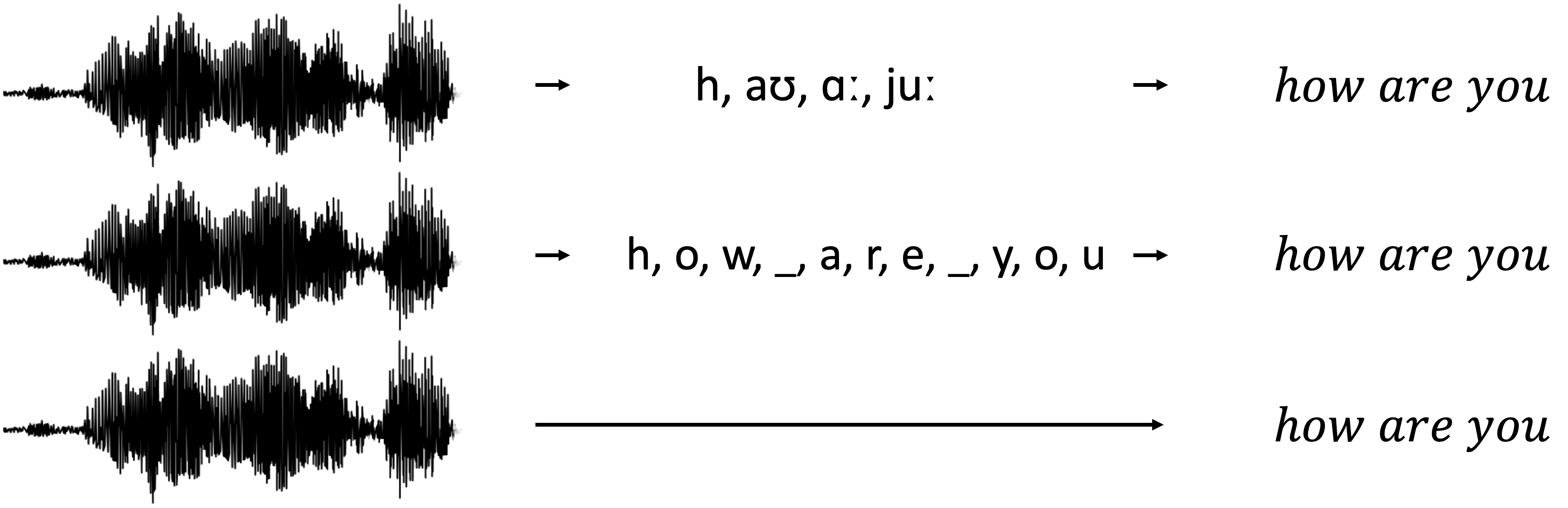}
    \caption{While recognition models are predominantly evaluated with word error rate (WER), training approaches often learn to map from input acoustic features to subwords, such as phones (top) or characters (middle), with word-level mapping done afterwards. We consider instead a training approach that learns to map directly from input acoustics to words (bottom).}
    \label{ch:back:fig:asr}
\end{figure}

In principle, these neural models could map acoustics directly to words. However, to achieve performance comparable with traditional approaches, these newer systems are still typically trained to predict subword units such as phones and characters (Figure~\ref{ch:back:fig:asr}a-b) or intermediate ``wordpiece"~\cite{rao2017exploring_rnn_transducer,chiu2018sota_asr_with_seq2seq,sanabria2018hierarchical,krishna2018hierarchical} units. Since performance is measured predominantly at the word-level, additional modules composing subword predictions with beam search and language modeling frameworks are required at inference time to convert subwords to words. One end-to-end training approach which has gained popularity in the last several years, and which we pay particular focus to in this thesis, is {\it connectionist temporal classification}. 

\subsubsection{Connectionist temporal classification (CTC)}

In this setting, we have a dataset of transcribed utterances $\mathcal{D} = \{(\mats{X}^{(n)}, \vecs{L}^{(n)})\}_{n=1}^{N}$ where example $n$ is a spoken utterance consisting of a continuous-valued acoustic feature vector sequence $\mats{X}^{(n)} = \{\vecs{x}_1, \vecs{x}_2, \dots, \vecs{x}_T\}$ and its transcript is the discrete label sequence $\vecs{L}^{(n)} = \{l_1, l_2, \dots, l_K\}$ whose elements are supported by the vocabulary $\mathcal{V}$. Since the utterance length $T$ is generally not equal to the transcript length $K$, we cannot na\"ively apply a frame-level loss such as cross entropy. Instead, we introduce a new token $\epsilon$ to our output vocabulary $\mathcal{V}$, referred to as the ``blank" symbol, which gives us a new prediction vocabulary $\mathcal{V} \cup \{\epsilon\}$. This allows for the definition of a {\it set} of length-$T$ sequences $\Omega(\vecs{L})$ supported by the augmented vocabulary $\mathcal{V} \cup \{\epsilon\}$ that can be deterministically reduced to $\vecs{L}$ by (1) first removing consecutive repeated output predictions and then (2) deleting the $\epsilon$ tokens. 

\begin{figure}
\centering
\includegraphics[width=0.5\linewidth]{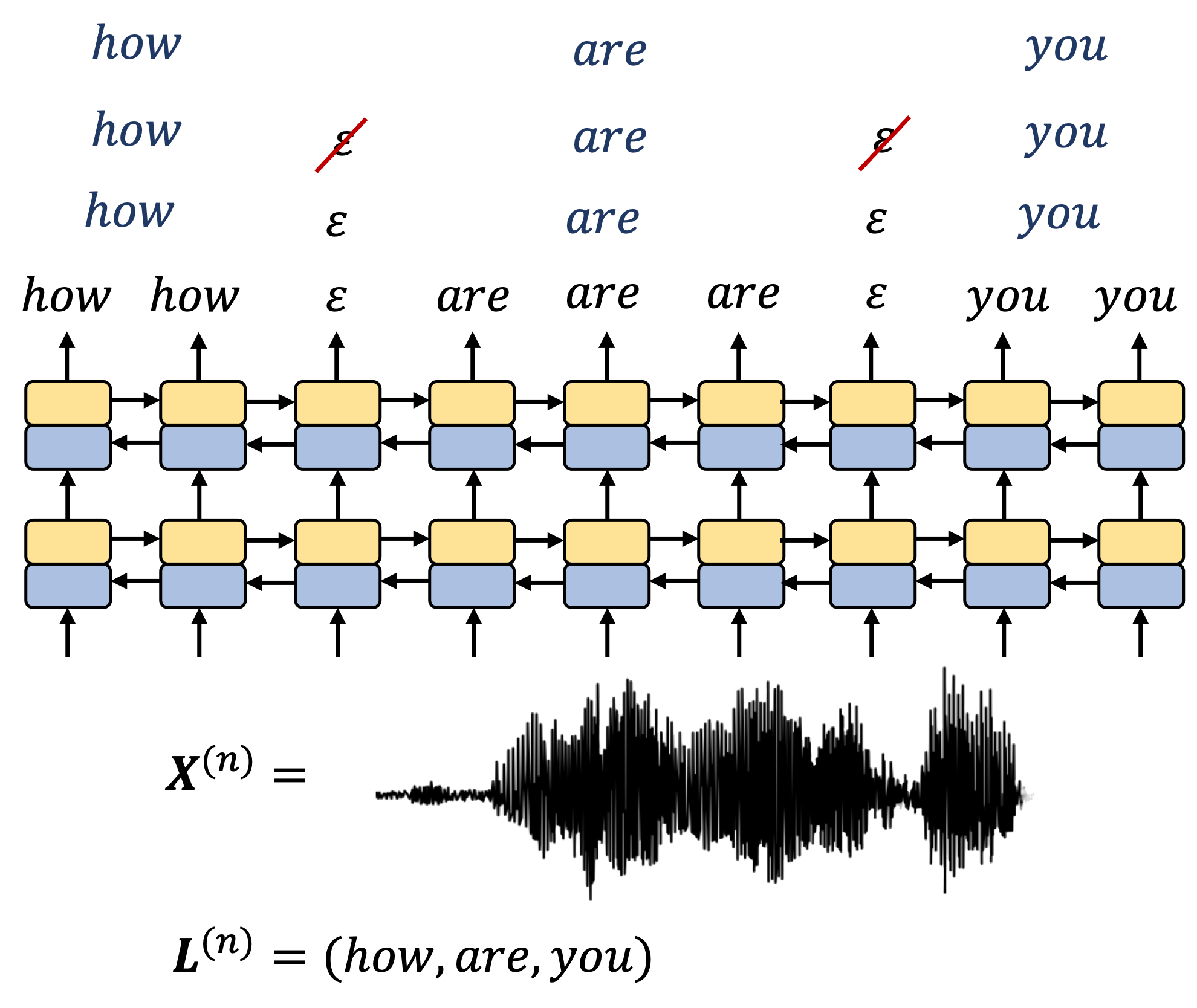}
\caption{Illustration of the word-level CTC-based speech recognizer, including the deterministic decoding process. $(\mats{X}^{(n)}, \mats{L}^{(n)})$ is an example utterance-transcript pair.}
\label{ch:back:fig:ctc}
\end{figure}

With these ideas in mind, we can now use the forward-backward algorithm to efficiently compute the conditional probability of the label sequence $\vecs{L}$ given the input sequence $\mats{X}$:
\begin{flalign*}
P(\vecs{L} \vert \mats{X}) = \displaystyle \sum_{\vecs{L}' \in \Omega(\vecs{L})} P(\vecs{L}' \vert \mats{X}) = \displaystyle \sum_{\vecs{L}' \in \Omega(\vecs{L})} \prod_{t=1}^T P(l'_t \vert \mats{X})
\end{flalign*}
\noindent During training, we can similarly calculate the log loss
\begin{flalign}
\mathcal{L}_{ctc}(\mats{X}, \vecs{L}) = - \log \displaystyle \sum_{\mats{L}' \in \Omega(\mats{L})} e^{s(\mats{L}')}
\label{eq:ctc}
\end{flalign}
\noindent where $s(\mats{L}')$ is the score
\begin{flalign*}
s(\mats{L}') &= \sum_{t=1}^T \mleft( f(\mats{X})_{t,l'_t} - \log \displaystyle \sum_{v \in \mathcal{V} \cup \{\epsilon\}} e^{f(\mats{X})_{t,v}} \mright)
\end{flalign*}
\noindent  for each sequence $\mats{L}' \in \Omega(\mats{L})$ calculated using the frame-level outputs from our acoustic model $f$. At inference time, the model predictions are post-processed following the deterministic reduction described above and depicted in Figure~\ref{ch:back:fig:ctc}.

\section{Proxy tasks}
\label{ch:back:prelims:proxies}

While the primary goal of acoustic model training and research is to improve performance on downstream speech tasks valuable to users, performing and evaluating on these end tasks can often be cumbersome, resource intensive, and challenging for quick iterative development. To streamline experimentation, we use proxy tasks that can be evaluated quickly, while being similar to the downstream applications we care about. We describe two of these proxy tasks here: acoustic word discrimination~\cite{carlin2011rapid_eval_spoken_term_detect} and cross-view word discrimination~\cite{he2017multiview}.

\begin{figure}
\centering
\begin{minipage}{.5\textwidth}
  \centering
  \includegraphics[width=0.5\linewidth]{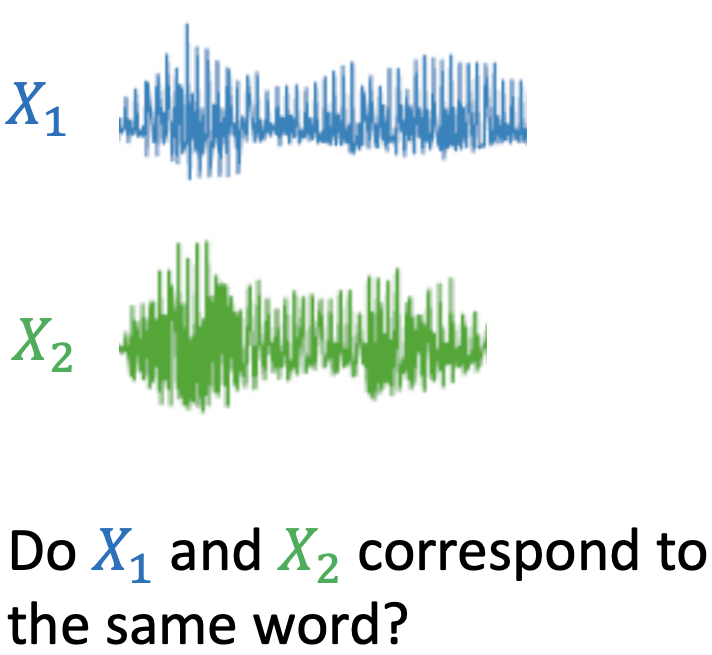}
  \caption*{(a) Acoustic word discrimination}
\end{minipage}%
\begin{minipage}{.5\textwidth}
  \centering
  \includegraphics[width=0.5\linewidth]{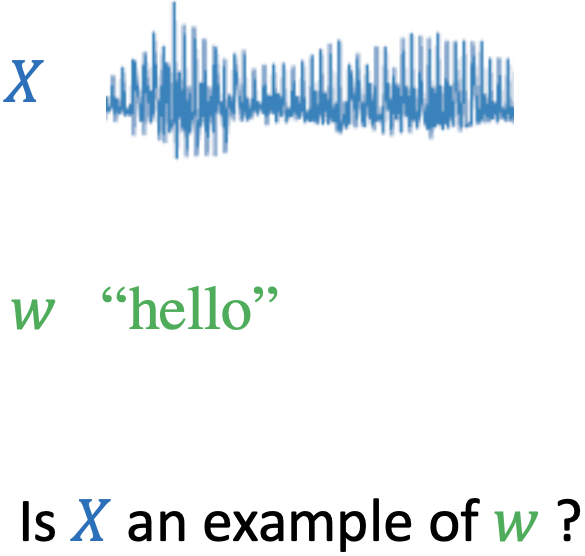}
  \caption*{(b) Cross-view word discrimination}
\end{minipage}
\caption{Illustrations of the (a) acoustic word discrimination and (b) cross-view word discrimination proxy tasks introduced by Carlin {\it et al.}~\cite{carlin2011rapid_eval_spoken_term_detect} and He {\it et al.}~\cite{he2017multiview}, respectively.}
\label{ch:back:fig:proxies}
\end{figure}

\subsection{Acoustic word discrimination}
\label{ch:back:prelims:acoustic_ap}

Acoustic word discrimination is the task of determining whether a pair of acoustic segments $(\mats{X}_1, \mats{X}_2)$ correspond to the same word (Figure~\ref{ch:back:fig:proxies}a). Carlin {\it et al.}~\cite{carlin2011rapid_eval_spoken_term_detect} propose the acoustic word discrimination proxy task to allow for rapid prototyping of DTW-based spoken term discovery techniques.

Prior work uses this proxy task to evaluate both frame-level acoustic features~\cite{carlin2011rapid_eval_spoken_term_detect,jansen2013weak_topdown,kamper2015unsupervised_weak_topdown} and AWEs~\cite{levin2013fixed,kamper2016cnn_awe,yuan2016learning_from_bnf,he2017multiview}. When learning frame-level acoustic features, the distance between a pair of segments is computed using DTW, but to evaluate the quality of segment-level representations (i.e., AWEs), a vector distance, such as Euclidean or cosine distance, between embeddings is used. To perform the task evaluation, the distance between each pair of acoustic segments $(\mats{X}_i, \mats{X}_j)$ in a test set of segment pairs is computed. A pair of segments is predicted ``same" (i.e. a match) if their distance falls below a threshold value and ``different" otherwise.

\begin{flalign}
&& precision &= \frac{\textrm{TP}}{\textrm{TP} + \textrm{FP}} & recall &= \frac{\textrm{TP}}{\textrm{TP} + \textrm{FN}}&
\label{eq:precision-recall}
\end{flalign}
\noindent Equations~\ref{eq:precision-recall}a and~\ref{eq:precision-recall}b show formulas for computing precision and recall, respectively, from sets of true positives (TP), false positives (FP), and false negatives (FN) output by the model. A true positive (TP) is a positive example correctly predicted as positive, a false positive (FP) is a negative example incorrectly predicted as positive, and a false negative (FN) is a positive example incorrectly predicted as negative. As the distance threshold value is increased, more segment pairs are classified as ``same" pairs such that there will be fewer and fewer false negatives until recall is 1. The precision-recall curve presents model precision at different distance threshold values as recall ranges from 0 to 1. Results are then reported as average precision (AP), which is the area under the precision-recall curve.

\subsection{Cross-view word discrimination}
\label{ch:back:prelims:crossview_ap}

Cross-view word discrimination is the task of determining whether an acoustic segment $\mats{X}_i$ is a spoken instance of word label $w_j$ (Figure~\ref{ch:back:fig:proxies}b). He {\it et al.}~\cite{he2017multiview} introduce this task, inspired by acoustic word discrimination, to measure the quality of not only acoustic word embeddings (AWEs) but also acoustically grounded word embeddings (AGWEs). He {\it et al.} propose the use of ``cross-view word discrimination" as a proxy task for spoken term detection (STD)~\cite{fiscus1970spoken_term_detection} where the query is text-based rather than spoken. This proxy task is used to evaluate the quality of our joint embedding space, and serves to indicate the potential of our AWE and/or AGWE models to be used within a system that compares acoustic segments with text (i.e. spoken term detection or whole-word automatic speech recognition).

To perform the evaluation, the distance between each AGWE vector and each AWE vector in a test set of pairs is computed using simple vector distance. As with the acoustic word discrimination task, a pair is considered a match if the distance between the pair of embeddings falls below a threshold value, and the results are reported as average precision (AP) when varying this threshold (Section~\ref{ch:back:prelims:acoustic_ap}).

While acoustic word discrimination is used in this thesis as a tuning criterion for query-by-example speech search (QbE)~\cite{settle2017query,yuan2018learning_awe_for_qbe}, cross-view word discrimination is used when pretraining AWE and AGWE models for application to automatic speech recognition~\cite{settle2019_a2w,shi2021seg}.

\section{Related work}
\label{ch:back:related}

We now describe work related to our own. We discuss developments modeling acoustic word embeddings (AWEs) and acoustically grounded word embeddings (AGWEs) as well as progress in their application to the downstream tasks. 

\subsection{Acoustic word embeddings}
\label{ch:back:related:awes}

Maas {\it et al.}~\cite{maas2012word} propose the earliest method we are aware of to explicitly learn {\it acoustic word embeddings} (AWEs). Maas {\it et al.} train a convolutional neural network (CNN) that takes a speech segment as input and outputs a continuous-valued embedding vector. Then, the embedding model defines a feature function within a segmental conditional random field~\cite{zweig2009segmental} rescoring system for whole-word ASR.

Levin {\it et al.}~\cite{levin2013fixed} develop unsupervised approaches for AWEs using a template-based approach by computing DTW distances between the input segment and a set of reference acoustic segments to construct embeddings. These representations have been subsequently used in several applications including query-by-example speech search~\cite{levin2015srails}, lexical clustering~\cite{kamper2014unsupervised_lexical}, and unsupervised speech recognition~\cite{kamper2015unsupervised_small_vocab_asr}. Levin {\it et al.} also improve performance further by incorporating supervised labeling of the reference segments.

While effective in zero-resource settings, template-based AWE performance is limited when some labeled resources are available. Kamper {\it et al.}~\cite{kamper2016cnn_awe} trains AWE models with several feed-forward neural network architectures and different objectives, comparing them with prior work~\cite{levin2013fixed,kamper2015unsupervised_weak_topdown,jansen2013weak_topdown} on the acoustic word discrimination task~\cite{carlin2011rapid_eval_spoken_term_detect}, highlighting the potential for discriminatively trained AWE models to replace DTW. This work demonstrates the importance of choosing the appropriate network architecture, and the performance benefits of using contrastive triplet hinge loss over word classification. While Kamper {\it et al.}~\cite{kamper2016cnn_awe} considers convolutional neural network (CNN) and fully-connected (DNN) architectures, another option for AWE modeling is to use recurrent neural networks (RNNs). Successfully used for speech recognition~\cite{lu2015study_rnn_lvsr}, RNNs are structured to deal with variable-length sequences, while capturing longer-term temporal information. Following Kamper {\it et al.}~\cite{kamper2016cnn_awe}, our work in Settle and Livescu~\cite{settle2016rnn_awe} focuses on ablating several RNN-based model structures and objectives, while comparing to prior work on the same acoustic word discrimination task~\cite{carlin2011rapid_eval_spoken_term_detect}.

The only work applying RNNs to AWE modeling prior to our own is that of Chen {\it et al.}~\cite{chen2015qbe_kws_lstm} and Chung {\it et al.}~\cite{chung2016audio}. Chen {\it et al.} trains a long short-term memory (LSTM) network for word classification, then uses the hidden state vectors as embeddings in a query-by-example (QbE) task. However, the setting is very specific, with few queries and speaker-dependent training. Chung {\it et al.} train unsupervised single-layer RNN auto-encoders (AEs) to produce embeddings for their own word discrimination task. As in Chung {\it et al.}~\cite{chung2016audio}, later work from Holzenberger {\it et al.}~\cite{holzenberger2018unsupervised_awe} and Kamper {\it et al.}~\cite{kamper2019truly_unsupervised_awe} explore using RNN-based auto-encoders to learn fully unsupervised AWEs where AE techniques outperform a variety of clever downsampling methods, while being nearly competitive with DTW. Kamper {\it et al.}~\cite{kamper2019truly_unsupervised_awe} found RNN-based correspondence auto-encoders could learn better embeddings than vanilla auto-encoders, provided segment pair annotations are available (either through unsupervised term discovery or from oracle word-level segmentations). While AEs are powerful in the unsupervised and weakly-supervised settings, discriminative training when some form of labels are available has been found to consistently, and significantly, outperform AE-based approaches (see Table~\ref{ch:multi_awe:tab:baselines} in Chapter~\ref{ch:multi_awe}). 

While we are most interested in learning representations that are acoustically discriminative, there exist several approaches that aim to learn {\it semantic} representations directly from speech. These include Speech2Vec~\cite{chung2018speech2vec} and contextual AWEs~\cite{palaskar2019learned}. These studies found that one can learn reasonable semantic embeddings from acoustic data that is on par with or better than the embeddings learned from the associated transcripts using typical text techniques~\cite{mikolov2013efficient}.

While much of early work on AWEs focuses on English~\cite{carlin2011rapid_eval_spoken_term_detect,levin2013fixed,kamper2016cnn_awe}, several studies have also branched out beyond English and even to multilingual modeling. Yuan {\it et al.}~\cite{yuan2016learning_from_bnf} use pretrained multilingual bottleneck features as input to improve low-resource AWE models, though they are trained and evaluated only on English. Holzenberger {\it et al.}~\cite{holzenberger2018unsupervised_awe} and Kamper {\it et al.}~\cite{kamper2019truly_unsupervised_awe} present results on both Xitsonga and English, though training and evaluation is still monolingual for each. Yang and Hirschberg~\cite{yang2019linguistically_informed_awe} explore using additional linguistic knowledge in AWE training to capture approximate matches and improve both in- and out-of-domain performance, and they consider training on English and evaluating on Sinhala.

Concurrent with our own multilingual work in Hu {\it et al.}~\cite{hu2020multilingual}, Kamper {\it et al.}~\cite{kamper2020multilingual} is the first work to explicitly explore multilingual AWE training and evaluation. Kamper {\it et al.} explore multilingual AWE training specifically for zero-resource languages, including unsupervised approaches trained on a zero-resource language of interest and supervised models trained on multiple additional languages and applied to a zero-resource language. Follow up work to Kamper {\it et al.}~\cite{kamper2020multilingual} continues along this direction~\cite{van2020improving,matusevych2020analyzing_AE_AWEs,kamper2021improved_multilingual,jacobs2021acoustic}, including Matusevych {\it et al.}~\cite{matusevych2020analyzing_AE_AWEs}, which analyzes auto-encoder AWEs by looking at properties beyond word identity that may be learned implicitly during training such as speaker identity, word segment duration, and number of phones per word.

\subsection{Jointly trained acoustic and acoustically grounded word embeddings}
\label{ch:back:related:agwes}

Bengio and Heigold~\cite{bengio2014word} propose the earliest example of joint training acoustic word embedding (AWE) and acoustically grounded word embedding (AGWE) models. In this work, Bengio and Heigold jointly train an AWE model and an AGWE model that represents words as bags of letter n-grams. They train the models with a contrastive triplet hinge loss to encourage AWEs to be close to the AGWE for their corresponding written word label in the learned embedding space. The embeddings are then combined with a traditional speech recognition system for word-level rescoring. Since the AGWE model derives word representations from letter n-grams, it allows for recognition of words outside the training vocabulary. However, the model structure is constrained, with their CNN-based AWE model requiring truncation or padding to a fixed-size input, and the AGWE model using letter n-grams but no other sequence information. 

Following our work on RNN-based single-view AWE approaches~\cite{settle2016rnn_awe} (Chapter~\ref{ch:rnn_awe}), He {\it et al.}~\cite{he2017multiview} introduce RNN-based multi-view joint training of AWE and AGWE models. In this work, our RNN-based approach is extended to both AWEs and AGWEs, but rather than contrasting embeddings of acoustic sequences directly, the ``same" embedding pair in the triplet loss now consists of both an AWE and an AGWE. This contrastive training procedure is similar to that of Bengio and Heigold~\cite{bengio2014word}, but He {\it et al.} further explore the impact of additional loss terms. This multi-view method outperforms all prior work on the acoustic word discrimination task of Carlin {\it et al.}~\cite{carlin2011rapid_eval_spoken_term_detect}, but it relies on full supervision to get word labels rather than segment pair information alone~\cite{jansen2013weak_topdown,kamper2015unsupervised_weak_topdown,kamper2016cnn_awe}.

As mentioned above in Section~\ref{ch:back:related:awes}, our work in Hu {\it et al.}~\cite{hu2020multilingual} (Chapter~\ref{ch:multi_awe}), concurrent with the multilingual AWE approach of Kamper {\it et al.}~\cite{kamper2020multilingual}, complements Kamper {\it et al.} by jointly learning both AWEs and AGWEs across a number of languages simultaneously, thus widening the range of tasks and languages to which our models apply. The only prior work to similarly consider training AGWE models on non-English languages is from Audhkhasi {\it et al.}~\cite{audhkhasi2017asr_free_kws} who jointly train AWE and AGWE models using multilingual bottleneck input features~\cite{cui2015multilingual_representations} for keyword search, but model training and evaluation is monolingual (Georgian).

\subsection{Applications of acoustic and acoustically grounded word embeddings}
\label{ch:back:related:downstream}

We investigate the downstream application of learned AWEs and AGWEs to query-by-example speech search (QbE) and automatic speech recognition (ASR).

\subsubsection{Query-by-example speech search}
\label{ch:back:related:qbe}

In settings where significant transcribed speech resources are available, QbE methods involve much the same process as training a full speech recognition system~\cite{allauzen2004general,parada2009query}, but in low- and zero-resource settings different methods are required, commonly relying on dynamic time warping (DTW) to compute audio segment similarity~\cite{hazen2009query_posteriorgram_templates,zhang2009unsupervised_spoken_keyword_spotting}. However, DTW-based QbE search can be slow and performance dependent on the frame-level representation quality~\cite{rabiner1978dtw}. Some researchers have focused on speeding up computation time through approximation~\cite{zhang2011piecewise_posteriorgram_dtw,jansen2012rails,mantena2013speedup_dtw_hierarchical_kmeans}, while others have worked to improve frame-level representations~\cite{jansen2013weak_topdown,kamper2015unsupervised_weak_topdown}, including learned phonetic posteriorgrams~\cite{hazen2009query_posteriorgram_templates,zhang2009unsupervised_spoken_keyword_spotting,lafarga2015elirf,lopez-otero2015gtm-uvigo_qbe} as well as supervised or unsupervised bottleneck features (BNFs)~\cite{cui2015multilingual_representations,proencca2016segmented,hou2015nni_qbe,leung2016toward,chen2016unsupervised_bnf_qbe}. Further difficulties are also introduced as use cases get more complicated since it can be difficult to modify DTW-based search to address a variety of query settings well, such as covering both exact and approximate match scenarios. As a result, the best systems fuse many~\cite{leung2016toward} (sometimes dozens~\cite{hou2015nni_qbe,proencca2016segmented}) of systems together.

An alternative to DTW-based search, which we explore in this thesis, is to use acoustic word embeddings (AWEs) to represent variable-duration speech segments as fixed-dimensional vectors and directly measure similarity between them via a simple vector distance. The first work to use AWEs for QbE is that of Levin {\it et al.}~\cite{levin2015srails}, which uses a template-based AWE function~\cite{levin2013fixed} to embed queries and segments from the search collection as fixed-dimensional vectors. Putative hits (matches) correspond to those segments in the search collection that are closest to the query in the embedding space. This work demonstrates the speed benefits of embedding-based search over a DTW-based system~\cite{jansen2012rails} while also matching or exceeding its performance. However, the growing body of work on AWEs~\cite{kamper2016cnn_awe,settle2016rnn_awe,he2017multiview} shows that neural network-based methods can outperform template-based embeddings on the acoustic word discrimination task. In Chapter~\ref{ch:rnn_qbe}, we present our work in Settle {\it et al.}~\cite{settle2017query}, which applies our own RNN-based AWEs from Chapter~\ref{ch:rnn_awe} to QbE. We demonstrate that neural AWEs~\cite{settle2016rnn_awe} can achieve large performance improvements over the templated-based approach of Levin {\it et al.}~\cite{levin2013fixed} not only on the word discrimination task~\cite{levin2013fixed} but also within a QbE system~\cite{levin2015srails}.

As detailed in Sections~\ref{ch:back:prelims:awes} and~\ref{ch:back:prelims:agwes}, early work on AWEs and AGWEs focuses on monolingual training and evaluation, especially artificially low-resource English data~\cite{carlin2011rapid_eval_spoken_term_detect,levin2013fixed,kamper2016cnn_awe,settle2016rnn_awe,he2017multiview}, though some work explores application to Georgian~\cite{audhkhasi2017asr_free_kws} and Xitsonga~\cite{holzenberger2018unsupervised_awe,kamper2019truly_unsupervised_awe}. Despite the supporting evidence for AWEs as an effective tool in the low- and zero-resource regime when used for acoustic word discrimination~\cite{kamper2016cnn_awe,settle2016rnn_awe,he2017multiview} and monolingual QbE~\cite{levin2015srails,settle2017query,yuan2018learning_awe_for_qbe}, our work in Hu {\it et al.}~\cite{hu2021ase} offers the first application of multilingual AWE training to multilingual QbE search. Several benchmark datasets and tasks developed for comparing multilingual QbE systems have been dominated by DTW methods~\cite{leung2016toward,proencca2016segmented,hou2015nni_qbe}, but the challenge of multilingual QbE is evident in the engineering complexity of these highest performing systems. We give the first comparison of multilingual embedding-based techniques~\cite{hu2021ase} with DTW-based approaches on the QUESST 2015~\cite{szoke2015query} multilingual query-by-example speech search benchmark. 

\subsubsection{Whole-word speech recognition}
\label{ch:back:related:asr}

Within the area of speech recognition, we will focus in particular on recent approaches to whole-word speech recognition that train word-level prediction models end-to-end~\cite{soltau2016a2w,audhkhasi2017a2w,audhkhasi2018a2w,yu2018multistage}. These systems are commonly referred to as being {\it acoustic-to-word} (A2W) systems as they are trained to map input acoustic representations directly to word-level units (Figure~\ref{ch:back:fig:asr}c). Modeling whole-word units is not entirely new to speech recognition since such methods were applied early on to isolated digit recognition~\cite{levinson1983isolated_digit}, were later developed as rescoring tools first with template-based approaches~\cite{heigold2012investigations}, and then more recently using acoustic word embedding (AWE) techniques~\cite{maas2012word,bengio2014word}. These newer end-to-end training approaches, however, allow for joint learning of the acoustic, pronunciation, and language models under a trainable word-level objective.

By performing prediction at the level of whole words, the quality of the learned representations for these different words can be impacted significantly by the frequency with which each word is seen in the training set. In the extreme case, words that do not appear in the training set at all cannot even be predicted as their representations do not get learned. Our models and training objectives will ideally be devised in such a way that words are learned efficiently, their representations are shared, and those words unseen during training can still be predicted through vocabulary extension.

The first work to produce results using an A2W model that matches the performance of a subword level model trained on the same data is Soltau {\it et al.}~\cite{soltau2016a2w}. However, they train on over $125$k hours of speech data, nearly two orders of magnitude more data than is often used for subword-based ASR systems, and they use a $100$k word vocabulary to minimize the issue of recognizing words beyond the training vocabulary. Following Soltau {\it et al.}, Audhkhasi {\it et al.}~\cite{audhkhasi2017a2w} aim to approach parity with subword models, but while operating in a more modest resource regime. Audhkhasi {\it et al.} make progress in this direction by introducing first phone-level pretraining of the acoustic model encoder, and then using pretrained semantic word embeddings (GloVe~\cite{pennington2014glove}) to initialize the word-level prediction layer. Later work by Audhkhasi {\it et al.}~\cite{audhkhasi2018a2w} reduces the performance gap with subword systems further with subtle updates to the model architecture and training details as well as a ``spell-and-recognize" component to allow for recovery from out-of-vocabulary (OOV) predictions. Yu {\it et al.}~\cite{yu2018multistage} also explore the impact of model structure, layer initialization, and regularization on A2W performance. They introduce hierarchical character and phone pre-training, curriculum learning to grow the vocabulary over time, joint training with multiple ASR objectives, and data augmentation to great effect and improving over Audhkhasi {\it et al.}~\cite{audhkhasi2018a2w}. Li {\it et al.}~\cite{li2017acoustic} propose integration of character prediction, and then a mixed-unit approach in follow-up work~\cite{li2018advancing} that incorporates only a certain number of high-frequency words into the output vocabulary, and decomposes all other words into subword sequences to avoid OOVs. Considering even larger units, Gaur {\it et al.}~\cite{gaur2019acoustic_phrase} explore the use of both word- and phrase-level units by growing the output vocabulary to include common multi-word phrases.

Concurrent with some of our work, Collobert {\it et al.}~\cite{collobert2019wordlevel} introduces A2W model training with a ``dynamic" lexicon approach. Instead of learning a separate vector representation in a lookup table for each word in the vocabulary, they construct each word vector on the fly from a letter-to-word encoder (i.e. an AGWE model). Similar to prior multi-view AWE and AGWE training~\cite{he2017multiview}, word representations in Collobert {\it et al.} are learned from subword sequences, so the representations can be more effectively shared and learned from limited data. Collobert {\it et al.}~\cite{collobert2020word}, in an update to the original paper~\cite{collobert2019wordlevel}, incorporate a transformer architecture to build a strong A2W system without complicated mixed-unit outputs that remains memory efficient during training by sub-sampling the word-level lexicon.

While early work in this area focuses on A2W models using connectionist temporal classification (CTC)~\cite{soltau2016a2w,audhkhasi2017a2w,audhkhasi2018a2w,yu2018multistage,li2017acoustic}, sequence-to-sequence style models have also been considered~\cite{collobert2019wordlevel,collobert2020word,palaskar2018_s2s}. Chapter~\ref{ch:seg_a2w} details our work in Shi {\it et al.}~\cite{shi2021seg} that proposes another A2W training approach using whole-word segmental models. Segmental models have a long history in speech recognition research as phone-level acoustic models~\cite{ostendorf1996hmm,glass2003probabilistic,zweig2009segmental,zweig2012_seg,he2012_sc,hamid2013_dsnn,tang2014_losses,lu2016_srnn} and for second-pass whole-word rescoring~\cite{zweig2009segmental,maas2012word,bengio2014word}, but ours is the first end-to-end whole-word segmental model.

In Chapter~\ref{ch:joint_a2w}, we compare baseline ``static" and ``dynamic"~\cite{collobert2020word} lexicon approaches, while using multi-view AWE and AGWE objectives to pretrain or joint train A2W recognition models. Chapter~\ref{ch:joint_a2w} focuses on conversational speech in the low to modest to data regime. We find that joint training enables our A2W models to perform comparably with many subword recognition systems on a variety of low-resource languages, and even some of the more recent self-supervised transformer-based models.

\subsection{Additional tasks}
\label{ch:back:addl}

Below we acknowledge additional relevant tasks for our methods, though we do not consider these tasks in this thesis. 

\subsubsection{Minimal-pair ABX}
\label{ch:back:addl:abx}

Minimal-pair ABX is described by Schatz {\it et al.}~\cite{schatz2013evaluating_abx,schatz2014evaluating_abx_2} as another framework for evaluation of acoustic features, which aims to improve upon the interpretability of the word discrimination task introduced by Carlin {\it et al.}~\cite{carlin2011rapid_eval_spoken_term_detect}. The minimal-pair ABX task involves three example speech segments A, B, and X. By comparing speech segments A and B against another segment X, the goal is to isolate subtle differences and correctly identify the minimally different pair, either A-X or B-X. There are three such differences tested: phoneme across talker (PaT), phoneme across context (PaC), and talker across phoneme (TaP). Phoneme across talker (PaT) is a setting where the speaker is the same for A and B but not for X, while the phone sequence of X matches either A or B. This aims to match phonetic information across speakers. Phoneme across context (PaC) is a setting where the speaker is the same for all segments (A, B, and X), but X shares part of its phone sequence with either A or B. This aims to match specific phonetic information across different surrounding contexts.
Talker across phoneme (TaP) is a setting where the phone sequence is the same for A and B but not for X, while the speaker of X matches either A or B. This aims to match segments based on speaker identity rather than phonetic information.

We focus instead on the acoustic and cross-view word discrimination tasks as they are more in line with our tasks of interest, namely operating on the word-level or above for use in query-by-example speech search and whole-word prediction for speech recognition.

\subsubsection{Keyword search}
\label{ch:back:addl:kws}

Keyword search (or spoken term detection) involves text-based queries rather than spoken queries as in query-by-example. This can restrict its application areas and use cases since the queries must be a sequence of characters~\cite{audhkhasi2017asr_free_kws} or, in some cases, may even be restricted further to a fixed vocabulary of keywords~\cite{palaz2016keyword}.

We focus on QbE rather than keyword search since we are predominantly interested in application to the low- and zero-resource data. Also, we choose to use ASR to showcase the power of our multi-view approach.

\subsubsection{Spoken term discovery}

This is the process of searching for repeated words or phrases within a speech corpus. This task is even more challenging than query-by-example search, but has been approached in similar ways where you could imagine considering segments within the speech corpus to be queries and search for these segments within the speech corpus itself to look for repeats~\cite{zhang2009unsupervised_spoken_keyword_spotting,zhang2011piecewise_posteriorgram_dtw,jansen2012rails}.

The challenging nature of this task lead to the need for the development of proxy tasks such as the acoustic word discrimination task of Carlin {\it et al.}~\cite{carlin2011rapid_eval_spoken_term_detect} and the minimal-pair ABX task of Schatz {\it et al.}~\cite{schatz2013evaluating_abx,schatz2014evaluating_abx_2}. Rather than perform this task itself, we use these proxy tasks for the development and iteration of our algorithm and choose to showcase them instead with the (related) task of query-by-example search.

%% file: text/3_rnn_awe.tex
\chapter{Discriminative acoustic word embeddings: recurrent neural network-based approaches}
\label{ch:rnn_awe}

\begin{figure}[H]
\centering
\includegraphics[width=0.55\linewidth]{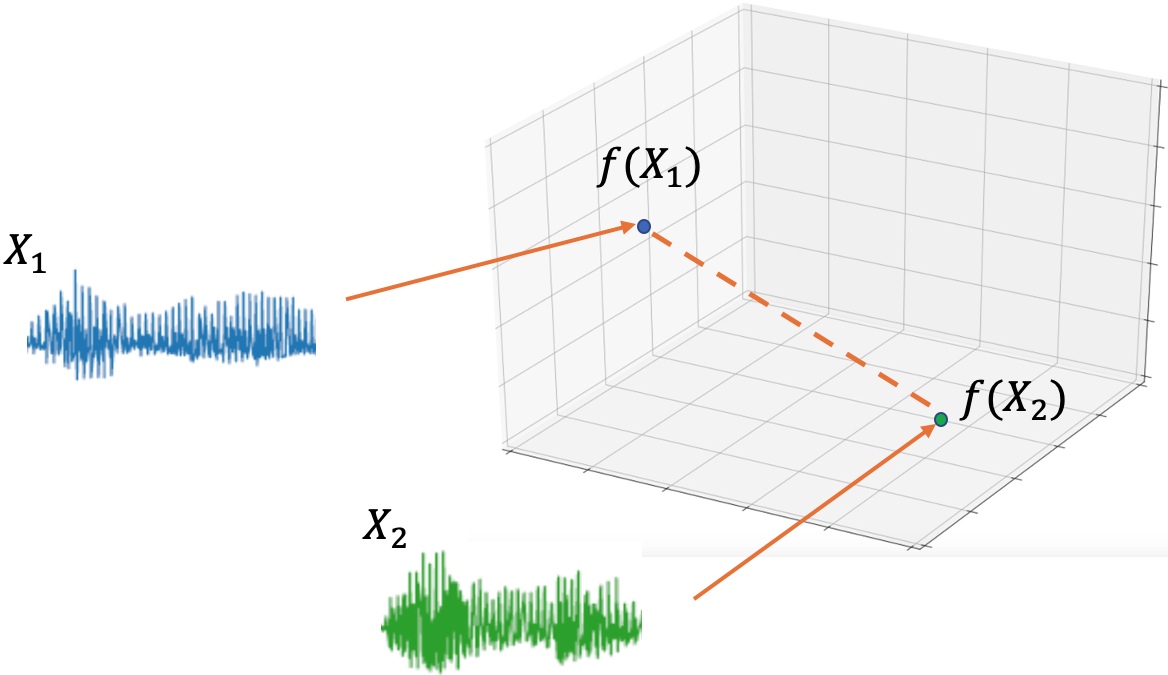}
\caption{Speech segments $\mats{X}_1$ and $\mats{X}_2$ are mapped into fixed-dimensional embeddings by the acoustic word embedding (AWE) model $f$.}
\label{ch:rnn_awe:fig:awe_grid}
\end{figure}

Acoustic word embeddings (AWEs)---fixed-dimensional vector representations of variable-length spoken word segments---have started to be considered for downstream speech tasks such as query-by-example (QbE) search. An AWE model (Figure~\ref{ch:rnn_awe:fig:awe_grid}) is a function $f$ where the input is a variable-length speech segment $\mats{X}$, and the output is a fixed-dimensional vector $f(\mats{X}) \in \mathbb{R}^d$. Such AWE models can be trained discriminatively such that they map acoustic segments into a learned vector space with the goal that segments corresponding to the same word are close together, while segments corresponding to different words are farther apart. Once acoustic segments are represented via these fixed-dimensional embeddings, pairwise segment distances can be measured efficiently by a vector distance between their embeddings.

In this chapter, we describe improvements to training and modeling of AWEs using recurrent neural networks (RNNs) with evaluation on the acoustic word discrimination proxy task introduced by Carlin {\it et al.}~\cite{carlin2011rapid_eval_spoken_term_detect} and described earlier in Section~\ref{ch:back:prelims:acoustic_ap}. These contributions are published in Settle and Livescu~\cite{settle2016rnn_awe}.

\section{Introduction}

Many speech processing tasks, such as automatic speech recognition or spoken term detection, hinge on associating segments of speech signals with word labels. In most systems developed for such tasks, words are broken down into subword units such as phones, and models are built for the individual units. An alternative, which has been considered by some researchers, is to consider each entire word segment as a single unit, without assigning parts of it to subword units. One motivation for the use of whole-word approaches is that they avoid the need for subword models. This is helpful since, despite decades of work on subword modeling~\cite{ostendorf1999beads}, it still poses challenges~\cite{livescu2012subword}. A second motivation is that considering whole words at once allows us to reason about more flexible sets of features over longer time spans.

Whole-word approaches typically involve, at some level, template matching, and often boil down to making decisions about whether two segments are examples of the same word or not. For example, in template-based speech recognition~\cite{dewachter2007template,heigold2012investigations}, word scores are computed from dynamic time warping (DTW) distances between an observed segment and training segments of the hypothesized word. In query-by-example (QbE) search, putative matches are typically found by measuring the DTW distance between the audio query and speech segments of the search database~\cite{metze2013spoken,anguera2012speaker,zhang2012fast,szoke2015coping}. DTW-based methods are flexible as they do not require any available labeled data, but segment comparison with these systems is quite expensive (Algorithm~\ref{algo:dtw}). Also, in instances where some labeled data may be available, most DTW-based approaches incorporate labeled data through indirect training of frame-level acoustic features on other tasks, which limits how effectively these resources can be leveraged.

An alternative to DTW that has started to be explored is the use of acoustic word embeddings (AWEs), or vector representations of spoken word segments. AWEs are representations that can be learned from data, ideally such that the embeddings of two segments corresponding to the same word are close, while embeddings of segments corresponding to different words are far apart. Once word segments are represented via fixed-dimensional embeddings, computing pairwise segment distances is as simple as measuring a cosine or Euclidean distance between their two vectors.

The earliest work on learning AWE models is split between neural network-based~\cite{maas2012word,bengio2014word} and template-based~\cite{levin2013fixed,voinea2014word} approaches. Maas {\it et al.}~\cite{maas2012word} and Bengio and Heigold~\cite{bengio2014word} use convolutional neural network (CNN)-based AWEs to generate scores for word segments within larger automatic speech recognition (ASR) systems. Maas {\it et al.} train CNNs to predict continuous-valued embeddings of the word labels, and uses the resulting embeddings to define feature functions in a segmental conditional random field~\cite{zweig2009segmental} ASR rescoring system. Bengio and Heigold also develop CNN-based embeddings for lattice rescoring, but with a contrastive loss to separate embeddings of a given word from embeddings of other words. Levin {\it et al.}~\cite{levin2013fixed} develop unsupervised embeddings by representing each word as a vector of DTW distances to a collection of reference word segments. This representation is subsequently seen in several applications targeting low-resource domains such as query-by-example search~\cite{levin2015srails}, lexical clustering~\cite{kamper2014unsupervised_lexical}, and unsupervised speech recognition~\cite{kamper2015unsupervised_small_vocab_asr}. Voinea {\it et al.}~\cite{voinea2014word} also use template-based representations shown to be robust on digit classification that are invariant to specific phone transformations.

Most similar to our own work, Kamper {\it et al.}~\cite{kamper2016cnn_awe} compare several types of AWEs on the Carlin {\it et al.}~\cite{carlin2011rapid_eval_spoken_term_detect} acoustic word discrimination task related to query-by-example search. Kamper {\it et al.} find that CNN-based embeddings trained with a contrastive loss outperform other DNN and CNN embeddings~\cite{kamper2016cnn_awe}, the reference vector approach of Levin {\it et al.}~\cite{levin2013fixed}, and prior methods that perform DTW on top of pretrained frame-level acoustic features~\cite{carlin2011rapid_eval_spoken_term_detect,jansen2013weak_topdown,kamper2015unsupervised_weak_topdown}. 

In this chapter, we use the same word discrimination task~\cite{carlin2011rapid_eval_spoken_term_detect} and two of the same training losses as Kamper {\it et al.}, but develop new embedding models based on recurrent neural networks (RNNs).~\footnote{It is worth noting that this work was completed in the pre-transformer era.} RNNs are a natural model class for acoustic word embeddings since they can handle arbitrary-length sequences. There have now been a number of approaches compared on this same task and conversational data~\cite{levin2013fixed,carlin2011rapid_eval_spoken_term_detect,kamper2015unsupervised_weak_topdown,jansen2013weak_topdown}, but none use RNN-based approaches. The only prior work which does consider RNN-based AWEs is that of Chen {\it et al.}~\cite{chen2015qbe_kws_lstm} and Chung {\it et al.}~\cite{chung2016audio}. Chen {\it et al.} learns a long short-term memory (LSTM) RNN for word classification and uses the resulting hidden state vectors as word embeddings in a query-by-example task. This setting is quite specific, however, with a small number of queries and speaker-dependent training. Chung {\it et al.}~\cite{chung2016audio} train single-layer RNN autoencoders in an unsupervised way to produce embeddings for their own word discrimination and query-by-example tasks applied to read speech, but do not compare with prior work. Our work instead focuses on the supervised setting, and directly compares several types of RNN-based embeddings with prior work~\cite{levin2013fixed,carlin2011rapid_eval_spoken_term_detect,kamper2015unsupervised_weak_topdown,jansen2013weak_topdown} on the same word discrimination task and challenging spontaneous speech data from Carlin {\it et al.}~\cite{carlin2011rapid_eval_spoken_term_detect} where we find our best models achieve sizable improvements over prior work. 

\section{Approach}

\begin{figure}
\centering
\includegraphics[width=0.7\linewidth]{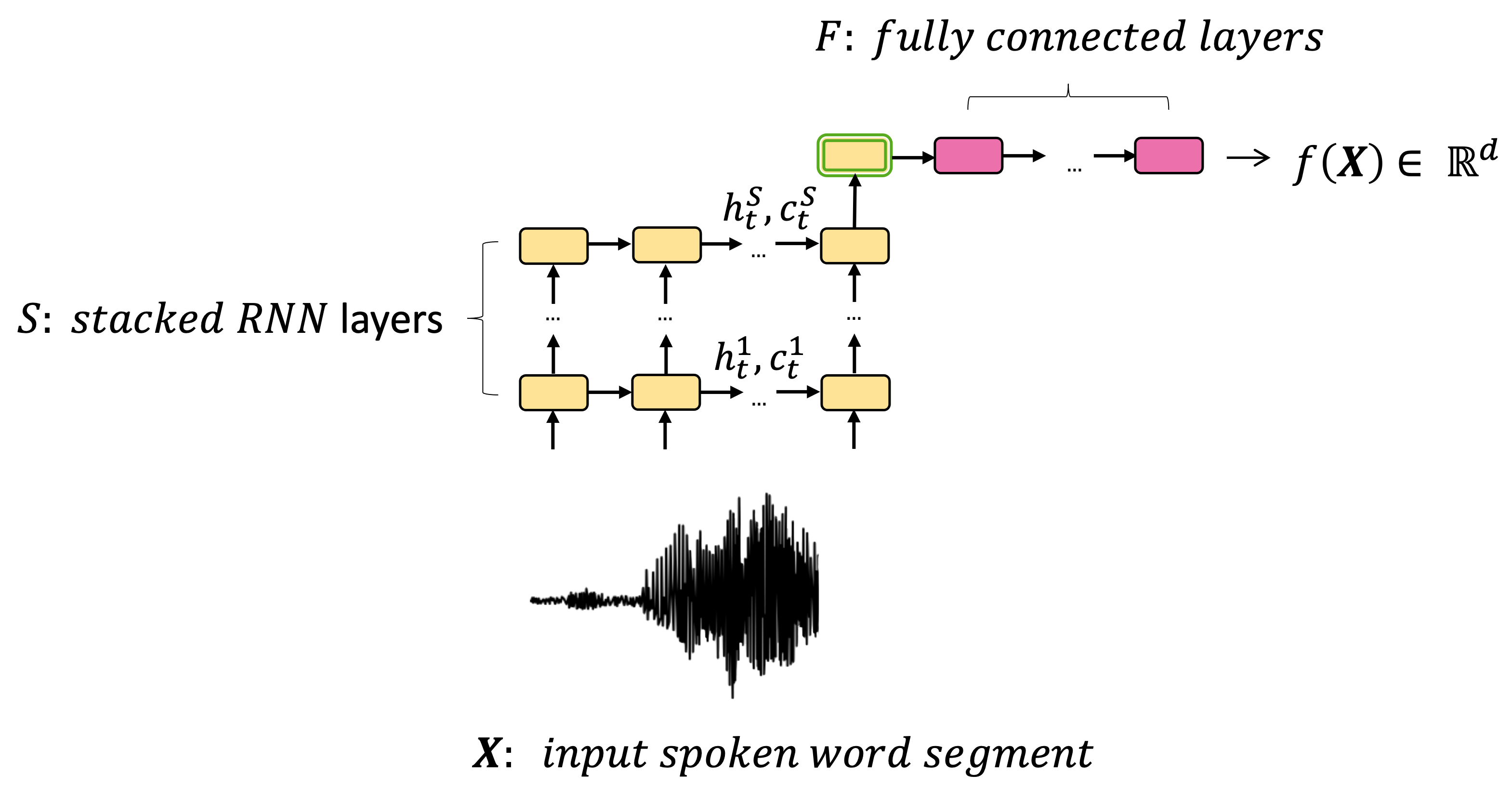}
\caption{LSTM-based acoustic word embedding model. For GRU-based models, the structure is the same, but the LSTM cells are replaced with GRU cells, and there is no cell activation vector; the recurrent connections only carry the hidden state vector $\mathbf{h}_t^l$.}
\label{ch:rnn_awe:fig:arch}
\end{figure}

An acoustic word embedding (AWE) model takes as input a speech segment (typically corresponding to an isolated spoken word), $\mats{X} = \{\vecs{x}_t\}_{t=1}^T$, where each $\vecs{x}_t$ is a vector of frame-level acoustic features, and outputs a fixed-dimensional vector representing the segment, $f(\mats{X})$. The basic embedding model structure we use is shown in Figure~\ref{ch:rnn_awe:fig:arch}. The model consists of a deep RNN with $S$ stacked layers, whose final hidden state vector is passed as input to a set of $F$ fully-connected layers. The output of the final fully-connected layer is the embedding $f(\mats{X}) \in \mathbb{R}^d$. The RNN hidden state at each time frame can be viewed as a representation of the input seen thus far, and its value in the last time frame $T$ could itself serve as the final word embedding. The fully-connected layers are added to account for any additional transformation that may improve the discriminative quality of the representation.

Within this class of embedding models, we focus on Long Short-Term Memory (LSTM)~\cite{hochreiter1997lstm} and Gated Recurrent Unit (GRU)~\cite{chung2014gru} networks. These are both types of RNNs that include a mechanism for selectively retaining or discarding information at each time frame when updating the hidden state, in order to better utilize long-term context. Both of these RNN variants have been used successfully in speech recognition~\cite{graves2013speech,sak2014lstm_lvsr,chorowski2015attention,lu2015study_rnn_lvsr}.

In an LSTM RNN, at each time frame both the hidden state $\mathbf{h}_t$ and an associated ``cell memory" vector $\mathbf{c}_t$, are updated and passed on to the next time frame. In other words, each forward edge in Figure~\ref{ch:rnn_awe:fig:arch} can be viewed as carrying both the cell memory and hidden state vectors. The updates are modulated by the values of several gating vectors, which control the degree to which the cell memory and hidden state are updated in light of new information in the current frame.  For a single-layer LSTM network, the updates are as follows:
\begin{align*}
    \mathbf{i}_t &=&& \sigma( \mathbf{W_i} [\mathbf{x}_t, \mathbf{h}_{t-1}] + \mathbf{b_i}) & \;\;\; & \textrm{input gate}\\
    \mathbf{f}_t &=&& \sigma( \mathbf{W_f} [\mathbf{x}_t, \mathbf{h}_{t-1}] + \mathbf{b_f}) & \;\;\; & \textrm{forget gate}\\
    \mathbf{\widetilde{c}}_t &=&& \tanh( \mathbf{W_c} [\mathbf{x}_t, \mathbf{h}_{t-1}] + \mathbf{b_c}) & \;\;\; & \textrm{candidate cell memory}\\
    \mathbf{c}_t &=&& \mathbf{i}_t \odot \mathbf{\widetilde{c}}_t + \mathbf{f}_t \odot \mathbf{c}_{t-1} & \;\;\; & \textrm{cell memory} \\
    \mathbf{o}_t &=&& \sigma( \mathbf{W_o} [\mathbf{x}_t, \mathbf{h}_{t-1}] + \mathbf{b_o}) & \;\;\; & \textrm{output gate} \\
    \mathbf{h}_t &=&& \mathbf{o}_t \odot \tanh(\mathbf{c}_t) & \;\;\; & \textrm{hidden state}
\end{align*}
\noindent where $\mathbf{h}_t, \mathbf{c}_t, \mathbf{\widetilde{c}}_t, \mathbf{i}_t, \mathbf{f}_t$, and $\mathbf{o}_t$ are all vectors of the same dimensionality, $\mathbf{W_i}, \mathbf{W_o}, \mathbf{W_f}$, and $\mathbf{W_c}$ are learned weight matrices of the appropriate sizes, $\mathbf{b_i}, \mathbf{b_o}, \mathbf{b_f}$ and $\mathbf{b_c}$ are learned bias vectors, $\sigma(\cdot)$ is a componentwise logistic activation, and $\odot$ is a Hadamard (componentwise) product.

Similarly, in a GRU network, at each time step a GRU cell determines what components of old information are retained, overwritten, or modified in light of the next step in the input sequence.  The output from a GRU cell is only the hidden state vector.  A GRU cell uses a reset gate $\mathbf{r}_t$ and an update gate $\mathbf{u}_t$ as described below for a single-layer network:
\begin{align*}
    \mathbf{r}_t &=&& \sigma( \mathbf{W_r} [\mathbf{x}_t, \mathbf{h}_{t-1}] + \mathbf{b_r}) & \;\;\; & \textrm{reset gate}\\
    \mathbf{u}_t &=&& \sigma( \mathbf{W_u} [\mathbf{x}_t, \mathbf{h}_{t-1}] + \mathbf{b_u}) & \;\;\; & \textrm{update gate}\\
    \mathbf{\widetilde{h}}_t &=&& \tanh( \mathbf{W_h} [\mathbf{x}_t, \mathbf{r}_t \odot \mathbf{h}_{t-1}] + \mathbf{b_h}) & \;\;\; & \textrm{candidate hidden}\\
    \mathbf{h}_t &=&& \mathbf{u}_t \odot \mathbf{h}_{t-1} + (1 - \mathbf{u}_t) \odot \mathbf{\widetilde{h}}_t & \;\;\; & \textrm{hidden state}
\end{align*}
\noindent where $\mathbf{r}_t, \mathbf{u}_t, \mathbf{\widetilde{h}}_t$, and $\mathbf{h}_t$ are all the same dimensionality, $\mathbf{W_r}, \mathbf{W_u}$, and $\mathbf{W_h}$ are learned weight matrices of the appropriate size, and $\mathbf{b_r}$, $\mathbf{b_u}$ and $\mathbf{b_h}$ are learned bias vectors.

All of the above equations refer to single-layer networks. In a deep network, with multiple stacked layers, the same update equations are used in each layer, with the state, cell, and gate vectors replaced by layer-specific vectors $\mathbf{h}_t^l, \mathbf{c}_t^l,$ and so on for layer $l$. For all but the first layer, the input $\mathbf{x}_t$ is replaced by the hidden state vector from the previous layer $\mathbf{h}_t^{l-1}$. For the fully-connected layers, we use rectified linear unit (ReLU)~\cite{nair2010rectified} activation, except for the final layer which depends on the form of supervision and loss used in training.

\subsection{Training objectives}

\begin{figure}[H]
\centering
\includegraphics[width=0.65\linewidth]{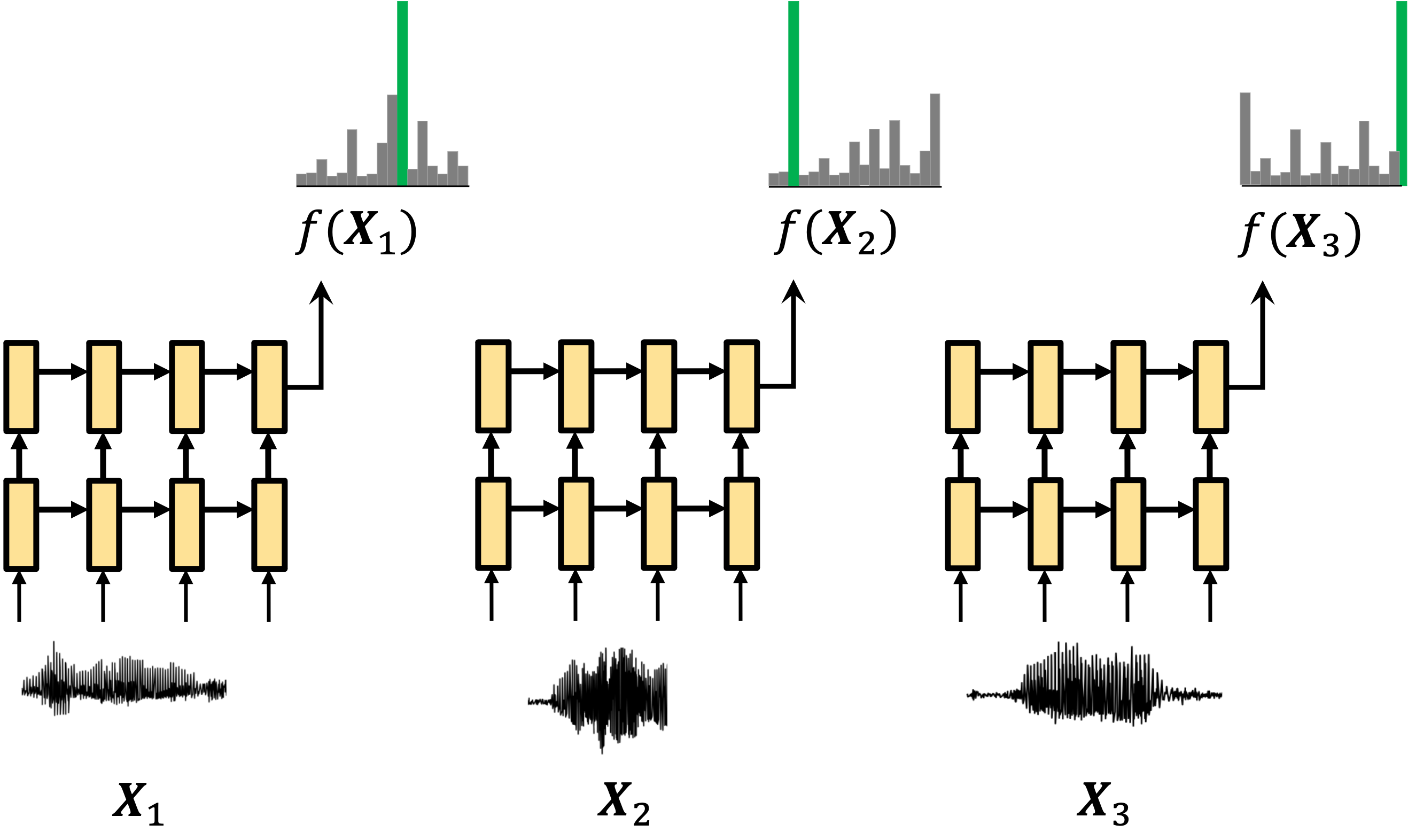}
\caption{Acoustic word embedding model training with a classification objective.}
\label{ch:rnn_awe:fig:word_classifier_objective}
\end{figure}

\begin{figure}[H]
\centering
\includegraphics[width=0.675\linewidth]{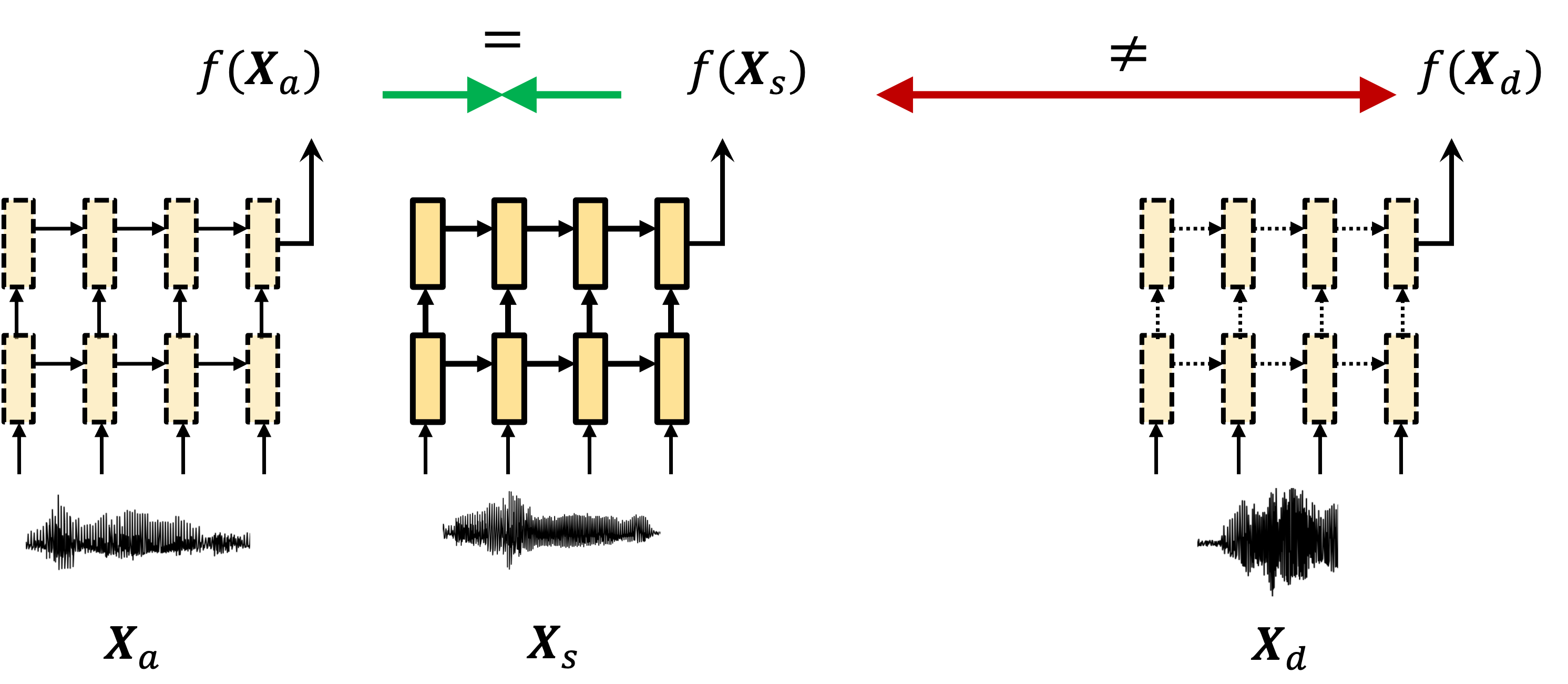}
\caption{Acoustic word embedding model training with a contrastive triplet objective using ``Siamese" networks with shared weights.}
\label{ch:rnn_awe:fig:contrastive_triplet_objective}
\end{figure}

We train the RNN-based embedding models using a set of pre-segmented spoken words using the two training approaches illustrated in Figures~\ref{ch:rnn_awe:fig:word_classifier_objective} and~\ref{ch:rnn_awe:fig:contrastive_triplet_objective}. These are inspired by prior work~\cite{kamper2016cnn_awe,bengio2014word} but with some differences in the details.

\subsubsection{Word classifier}

Our first approach is to use the word labels of the training segments and train the networks for word classification. The embedding model $f$ includes a Softmax non-linearity following the final linear layer with output dimensionality equal to the training vocabulary size $\mathcal{V}$.

Here we are limited to the subset of the training set vocabulary that has a sufficient number of segments per word to train a good classifier. In this setting, the embedding dimensionality $d$ of $f(\mats{X}) \in \mathbb{R}^d$ is equal to this training vocabulary size. Kamper {\it et al.}~\cite{kamper2016cnn_awe} studies decoupling the AWE dimensionality from the vocabulary size by introducing a bottleneck layer, but this flexibility comes at the cost of performance so we do not incorporate that here. The word classifier model is trained end-to-end with a word-level cross entropy loss
\begin{equation*}
    \mathcal{L}_{ce}(\mats{X}, v) =  \displaystyle - \sum_{i=1}^{\vert \mathcal{V} \vert} \mathbbm{1}_{\{y_i = v\}}(y) \log p(y_i| \mats{X}) = - \log f(\mats{X})_i
\end{equation*}
\noindent where $f(\mats{X})$ is a vector of posterior probabilities such that $f(\mats{X}) \in \mathbb{R}^{\vert \mathcal{V}\vert}$, $\sum_{i=1}^{\vert \mathcal{V} \vert} f(\mats{X})_i = 1$, and $f(\mats{X})_i \in [0, 1] \text{  } \forall i$. During evaluation, an embedding $f(\mats{X})$ corresponds to the posterior probabilities over words in $\mathcal{V}$ given input spoken word segment $\mats{X}$. Although our labeled data is limited, the goal is that the trained models will be useful even when applied to spoken examples of words not seen in the training data. For words not seen in training, their embeddings should correspond to some measure of similarity of the new word to the training words, measured via the posterior probabilities of the previously seen words. In our experiments, we examine this assumption by analyzing performance on words that appear in the training data compared to those that do not.

\subsubsection{Contrastive learning with triplet hinge loss and ``Siamese" networks}

Related work~\cite{jansen2013weak_topdown,synnaeve2014phonetic_embedding,kamper2015unsupervised_weak_topdown} considers a lower resource alternative using acoustic segment pair information as opposed to full supervision based on text transcripts. In this lower resource setting, training data is presented as pairs of audio segments labeled either ``same" or ``different". This allows for training with a contrastive ranking loss to make ``same" pair segments closer together, while making ``different" pair segments farther apart. Unlike training a word classifier, learning with the contrastive triplet hinge loss directly optimizes relative distance between ``same" and ``different" spoken word pairs.

There are multiple types of contrastive losses, but here we focus on the triplet hinge loss in Kamper {\it et al.}~\cite{kamper2016cnn_awe} and Settle and Livescu~\cite{settle2016rnn_awe}, which considers only pairs of acoustic segments. During training, we input three spoken word segments $(\mats{X}_a, \mats{X}_s, \mats{X}_d)$ through a trio of ``Siamese" networks~\cite{bromley1993signature_verification_siamese} (i.e.~networks with shared weights). Siamese networks have the benefit that, for a given training dataset, they measure a loss and gradient for every pair of data points. The first two input segments $\mats{X}_a$ and $\mats{X}_s$ are referred to as the ``anchor" and ``same" segments, respectively, as they share a word label $v_a$. The third segment $\mats{X}_d$, referred to as the ``different" segment, is a segment with a different word label from $\mats{X}_a$. The network is trained to optimize the cos-hinge triplet loss
\begin{equation}
\mathcal{L}_\textrm{cos-hinge}(\mats{X}_a, \mats{X}_s, \mats{X}_d) = \left[m + d(f(\mats{X}_a), f(\mats{X}_s)) - d(f(\mats{X}_a), f(\mats{X}_d))\right]_+
\label{ch:rnn_awe:eq:cos-hinge}
\end{equation}
\noindent where $m$ is a margin hyperparameter and $d({\bf a}, {\bf b})$ is the cosine distance between ${\bf a}$ and ${\bf b}$ defined $d({\bf a}, {\bf b}) := 1-\frac{{\bf a}\cdot {\bf b}}{\Vert {\bf a} \Vert\Vert {\bf b} \Vert}$. The speech segment $\mats{X}_d$ is randomly sampled from the set of negative examples $\mathcal{X}'(\mats{X}_a, v_a)$ defined as 
\begin{flalign*}
    \mathcal{X}'(\mats{X}, v) &:= \{\mats{X}' \text{   } \vert \text{   } v' \in \mathcal{V} / v, \text{  } (\mats{X}', v') \in \mathcal{D}\}
\end{flalign*}
\noindent where $\mathcal{V}$ is the training vocabulary and $\mathcal{D}$ is the training set of spoken segments and their word labels. In this setting, we use cosine distance (rather than the euclidean distance) following prior work~\cite{kamper2016cnn_awe}. Unlike cross entropy training, $\mathcal{L}_\textrm{cos-hinge}$ gives us the flexibility to define our output embedding dimension irrespective of the vocabulary size. Also, for tasks such as word discrimination and query-by-example search, this ranking-based training loss directly aligns with our end objective.

\section{Experimental setup}

Our end goal is to improve performance on downstream tasks requiring accurate word discrimination. In this work, we use an intermediate task that directly tests our ability to discriminate ``same" and ``different" word pairs, and allows us to compare to a variety of prior work. Specifically, we use the acoustic word discrimination task of Carlin {\it et al.}~\cite{carlin2011rapid_eval_spoken_term_detect}. 

This task is similar to the query-by-example task where the word segmentations are known. The evaluation consists of determining, for each pair of evaluation segments, whether they are examples of the same or different words, and measuring performance via the average precision (AP). We do this by measuring the cosine similarity between their acoustic word embeddings and declaring them to be the same if the distance is below a threshold. By sweeping the threshold, we obtain a precision-recall curve from which we compute the AP. We will often refer to this performance measure as ``acoustic AP". For further details, see the earlier discussion in Section~\ref{ch:back:prelims:acoustic_ap}.  

\subsection{Data}

The data is drawn from the Switchboard conversational English corpus~\cite{godfrey1992switchboard}. The word segments range from $50$ to $200$ frames in length, which is an approximate duration of $0.5$ to $2$ seconds. The frame-level acoustic input features are $39$-dimensional MFCCs+$\Delta$+$\Delta \Delta$ generated using the Kaldi toolkit~\cite{povey2011kaldi}. We use the same train, development, and test partitions as in prior work~\cite{carlin2011rapid_eval_spoken_term_detect,levin2013fixed,jansen2013weak_topdown,kamper2015unsupervised_weak_topdown,kamper2016cnn_awe}, and the same acoustic features as in~\cite{kamper2016cnn_awe}, for as direct a comparison as possible. The train set contains approximately $10$k example segments, while the development and test sets each contain approximately $11$k segments (corresponding to $60$ million pairs when computing acoustic AP). As in~\cite{kamper2016cnn_awe}, when training the classifier-based embeddings, we use a subset of the training set containing all word types with a minimum of $3$ occurrences, which reduces the training set size to approximately $9$k segments. When using contrastive training with Siamese networks, the training set consists of all (approximately $100$k) ``same" word pairs. For each such training pair, we randomly sample a third example belonging to a different word type, as required for the $\mathcal{L}_{\textrm{cos-hinge}}$ loss. 

All models were implemented in Torch~\cite{torch7}.

\subsection{Classification network details}

Our classifier-based embeddings use LSTM or GRU networks with $2$--$4$ stacked layers and $1$--$3$ fully-connected layers, and the final embedding dimensionality is $1061$, equal to the number of unique word labels in the training set. The recurrent hidden state dimensionality is fixed at $512$ and dropout~\cite{srivastava2014dropout} between stacked recurrent layers is used with probability $p=0.3$. The fully-connected hidden layer dimensionality is fixed at $1024$.  Rectified linear unit (ReLU) non-linearities and dropout with $p=0.5$ are applied between the fully-connected layers, but between the final recurrent hidden state output and the first fully-connected layer no non-linearity or dropout is applied. These settings were determined through experiments on the development set.

The classifier network is trained with a cross entropy loss and optimized using stochastic gradient descent (SGD) with Nesterov momentum \cite{nesterov1983method}. The learning rate is initialized at $0.1$ and is reduced by a factor of $10$ according to a heuristic.~\footnote{If 99\% of the current epoch's average batch loss is greater than the running average of batch losses over the last $3$ epochs, this is considered a plateau; if there are $3$ consecutive plateau epochs, then the learning rate is reduced.} When reducing the learning rate no longer improves development set acoustic AP, training stops and the model checkpoint from the epoch corresponding to the the best development set acoustic AP is chosen. Results for these experiments are labeled ``Classifier".

\subsection{Siamese network details}

For experiments with contrastive training using Siamese networks, we initialize (warm-start) the networks with the tuned classification network, remove the final softmax activation layer, and replace it with a linear layer of size equal to the desired embedding dimensionality. We use a margin of $0.4$ in the cos-hinge loss, and we explore a range of embedding dimensionalities from $8$ to $2048$.

In training the Siamese networks, each training mini-batch consists of $2B$ triplets. $B$ triplets are of the form $(\mats{X}_a,\mats{X}_s,\mats{X}_d)$ where $\mats{X}_a$ and $\mats{X}_s$ are examples of the same class (i.e. a pair from the $100$k ``same" word pair set) and $\mats{X}_d$ is a randomly sampled example from a different class. Then, for each of these $B$ triplets $(\mats{X}_a,\mats{X}_s,\mats{X}_d)$, an additional triplet $(\mats{X}_s,\mats{X}_a,\mats{X}_d)$ is added to the mini-batch to allow all segments to serve as anchors. This is a slight departure from earlier work~\cite{kamper2016cnn_awe}, which we found to improve stability in training and performance on the development set. We optimize the contrastive model using SGD with Nesterov momentum for $15$ epochs. The learning rate is initialized to $0.001$ and dropped every $3$ epochs until no improvement is seen on the development set. The final model is taken from the epoch with the highest development set acoustic AP. Results for these experiments are labeled ``Siamese".

\subsubsection{Negative sampling}

In preliminary experiments, we compare two methods for choosing the negative examples $\mats{X}_d$ during training, a uniform sampling approach and a non-uniform one. In the case of uniform sampling, we sample $\mats{X}_d$ uniformly at random from the full set of training examples with labels different from $\mats{X}_a$. This sampling method requires only word-pair supervision. In the case of non-uniform sampling, $\mats{X}_d$ is sampled in two steps. First, we construct a distribution $P_{v|label({\bm X}_a)}$ over word labels $v$ and sample a different label from $label({\bm X}_a)$. Second, we sample an example uniformly from within the subset of segments with the chosen label $v$. The goal of this method is to speed up training by targeting word label pairs with spoken segments that violate the margin constraint. 

To perform non-uniform sampling, we construct the multinomial PMF $P_{v|label({\bm X}_a)}$ by maintaining an $\vert \mathcal{V} \vert \times \vert \mathcal{V} \vert$ matrix $\mathbf{S}$ where $\vert \mathcal{V} \vert$ is the number of unique word labels in training. Each word label corresponds to an integer $i$ $\in$ [1, $\vert \mathcal{V} \vert$] representing a row in $\mathbf{S}$, and the row values of $\mathbf{S}$ are similarity scores. We retrieve the desired row PMF by normalizing by its sum. At the start of each epoch, we initialize $\mathbf{S}$ with $0$'s along the diagonal and $1$'s elsewhere (which reduces to uniform sampling). For each training triplet  $(\mats{X}_a,\mats{X}_s,\mats{X}_d)$, we update $\mathbf{S}$ for both $(i,j) = (label(\mats{X}_a),label(\mats{X}_d))$ and $(i,j) = (label(\mats{X}_d),label(\mats{X}_a))$:
\[	s_{i,j} \mathrel{+}=  \begin{cases}
				cos(f(\mats{X}_a),f(\mats{X}_d)) & d(f(\mats{X}_a),f(\mats{X}_d)) \leq d(f(\mats{X}_a),f(\mats{X}_s)) + m^* \\
                0 & \text{otherwise}
   			\end{cases}
\]
\noindent The PMFs $P_{v|label({\bm X}_a)}$ are updated after the forward pass of an entire mini-batch. The constant $m^*$ enforces a potentially stronger constraint than is used in the $\mathcal{L}_{\textrm{cos hinge}}$ loss, in order to promote diverse sampling. In all experiments, we set $m^*=0.6$. This is a heuristic approach, and it would be interesting to consider various alternatives. Preliminary experiments showed that the non-uniform sampling method outperformed uniform sampling, and in the following we report results with non-uniform sampling.

\section{Results}

\begin{table}[H]
\centering
\caption{Development set acoustic AP with classifier-based embeddings.}
\begin{tabular}{cccc}
\toprule
RNN depth $S$ & DNN depth $F$ & GRU dev AP & LSTM dev AP\\
\midrule
$2$ & $1$ & $0.21$ & $0.24$ \\
$3$ & $1$ & $0.25$ & $0.24$ \\
$4$ & $1$ & $0.30$ & $0.27$ \\
$3$ & $2$ & $0.41$ & $0.42$ \\
$3$ & $3$ & $0.45$ & $0.52$ \\
\bottomrule
\end{tabular}
\label{tab:classifier_arch}
\end{table}

\subsection{Effect of model structure}

Table~\ref{tab:classifier_arch} shows classifier-based embedding performance as we vary the number of stacked layers $S$, the number of fully-connected layers $F$, and the type of RNN cell used (LSTM vs.~GRU). The best performance in this experiment is achieved by the LSTM network with $S=F=3$. However, performance still seems to be improving with additional layers, suggesting that we may be able to further improve performance by adding even more layers of either type. However, we fix the model to $S=F=3$ to allow for more experimentation and analysis.

Table~\ref{tab:classifier_arch} reveals an interesting trend. When only one fully-connected layer is used, GRU networks outperform LSTMs given a sufficient number of stacked layers. Once we add more fully-connected layers, the LSTMs outperform the GRUs. In the first few lines of Table~\ref{tab:classifier_arch}, we use $2$, $3$, and $4$ layer stacks of LSTMs and GRUs while holding fixed the number of fully-connected layers at $F=1$. There is clear utility in stacking additional layers; however, even with 4 stacked layers the RNNs still underperform the CNN-based embeddings of~\cite{kamper2016cnn_awe} until we begin adding fully-connected layers.

After exploring a variety of stacked RNNs, we fix the stack to $3$ layers and vary the number of fully-connected layers. The value of each additional fully-connected layer is clearly greater than that of adding stacked recurrent layers when considering the classifier approach. All networks trained with $2$ or $3$ fully-connected layers obtain more than $0.4$ AP on the development set, while stacked RNNs with $1$ fully-connected layer are $0.3$ AP or less. This may raise the question of whether some simple fully-connected model may be all that is needed; however, previous work~\cite{kamper2016cnn_awe} has shown that this approach is not competitive, and convolutional or recurrent layers are needed to summarize arbitrary-length segments into a fixed-dimensional representation.

\subsection{Effect of embedding dimensionality}

For the ``Siamese" approach, we varied the output embedding dimensionality, as shown in Figure~\ref{ch:rnn_awe:fig:varydim_varymin}. This analysis shows that the embeddings learned by the Siamese RNN network are quite robust to reduced dimensionality, outperforming the classifier model for all dimensionalities $32$ or higher and outperforming previously reported development set performance with CNN-based embeddings~\cite{kamper2016cnn_awe} for all dimensionalities $\ge 16$.

\subsection{Effect of training vocabulary}

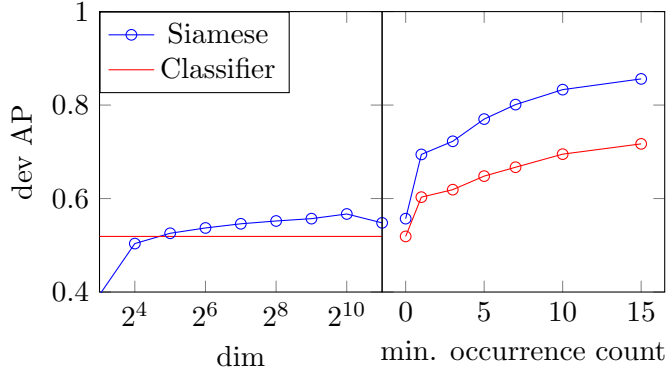
\begin{figure}
\centering
\begin{tikzpicture}
\begin{groupplot}[
    group style={
        group name= myplots,
        group size=2 by 1,
        horizontal sep=0pt,
        ylabels at= edge left,
        yticklabels at= edge left
    },
    width=5.3cm,
    height=5.3cm,
    ylabel=dev AP,
    ymin=0.4,
    ymax=1.0
]
\nextgroupplot[
    xmode= log,
    log basis x=2,
    xlabel= dim,
    legend style={at={(0,1)},anchor=north west},
    xmin=8,
    xmax=2048,
]
\addplot[mark=o, blue] coordinates {(8,0.3940) (16,0.5037) (32,0.5257) (64,0.5371)(128,0.546) (256,0.552) (512,0.557) (1024, 0.567) (2048,0.548)};
\addlegendentry{Siamese}
\addplot[mark=none, red] coordinates {(8,0.519) (16,0.519) (32,0.519) (64,0.519)(128,0.519) (256,0.519) (512,0.519) (1024, 0.519) (2048,0.519)};
\addlegendentry{Classifier}
\nextgroupplot[
xlabel= min. occurrence count,
legend style = {at={(0,1)},anchor=north west}
]
\addplot [blue,mark=o] coordinates {(0,0.557) (1,0.6944) (3,0.7225) (5,0.7700) (7,0.801) (10,0.833) (15,0.856)};
\addplot [red,mark=o] coordinates {(0,0.519) (1,0.603) (3,0.619) (5,0.648) (7,0.667) (10, 0.695) (15,0.717)};
\end{groupplot}
\end{tikzpicture}
\caption{Effect of embedding dimensionality (left) and occurrences in training set (right).}
\label{ch:rnn_awe:fig:varydim_varymin}
\end{figure}

We might expect the learned embeddings to be more accurate for words that are seen in training than for ones that are not. Figure~\ref{ch:rnn_awe:fig:varydim_varymin} measures this effect by showing performance as a function of the number of occurrences of development set words in the training set. Indeed, both model types are much more successful for in-vocabulary words, and their performance improves the higher the training frequency of the words. Performance increases more quickly for the Siamese network than for the classifier as training frequency increases. This may be due to the fact that, if a word type occurs at least $k$ times in the classifier training set, then it occurs at least $2 \times \binom{k}{2}$ times in the Siamese paired training data.

\begin{table}[H]
\centering
\caption{Test set acoustic AP compared with prior work; * indicates frame dimension for DTW approaches.}
\begin{tabular}{lcc}
\toprule
Model & $d$ & test AP\\
\midrule
MFCCs+DTW \cite{kamper2016cnn_awe} & $39^*$ & $0.21$\\
Corr. autoencoder+DTW~\cite{kamper2015unsupervised_weak_topdown} & $100^*$ & $0.47$\\
\midrule
Classifier DNN~\cite{kamper2016cnn_awe} & $1061$ & $0.30$\\
Classifier CNN~\cite{kamper2016cnn_awe} & $1061$ & $0.53$\\
Siamese CNN~\cite{kamper2016cnn_awe} & $1024$ & $0.55$\\
Classifier LSTM (ours) & $1061$ & $0.62$\\
Siamese LSTM (ours) & $1024$ & $0.67$\\
\bottomrule
\end{tabular}
\label{ch:rnn_awe:tab:test_acoustic_ap}
\end{table}

\subsection{Final test-set evaluations}

Based on development set results, our final embedding models are LSTM networks with $3$ stacked layers and $3$ fully-connected layers, and with output dimensionality of $1024$ in the case of Siamese networks. Final test set results are given in Table~\ref{ch:rnn_awe:tab:test_acoustic_ap}. We include a comparison with the best prior results on this task from~\cite{kamper2016cnn_awe}, including results using standard DTW on the input MFCCs (copied from~\cite{kamper2016cnn_awe}) as well as the best prior results using DTW with frame-level features learned with correlated autoencoders~\cite{kamper2015unsupervised_weak_topdown}. Both ``Classifier" and ``Siamese" LSTM embedding models outperform all prior results on this task of which we are aware.~\footnote{Yuan {\it et al.}~\cite{yuan2016learning_from_bnf} have recently been able to improve AP on this test set even further with CNN embeddings when using a large set of additional (cross-lingual) training data. We omit these above as we do not consider these results to be comparable because of their reliance on additional data.} Next, we analyze the effects of different model design choices, and visualize the learned AWEs learned with the Classifier and Siamese training methods.

\subsection{Visualization of embeddings}

\begin{figure}
\centering
\includegraphics[height=8cm,width=15cm]{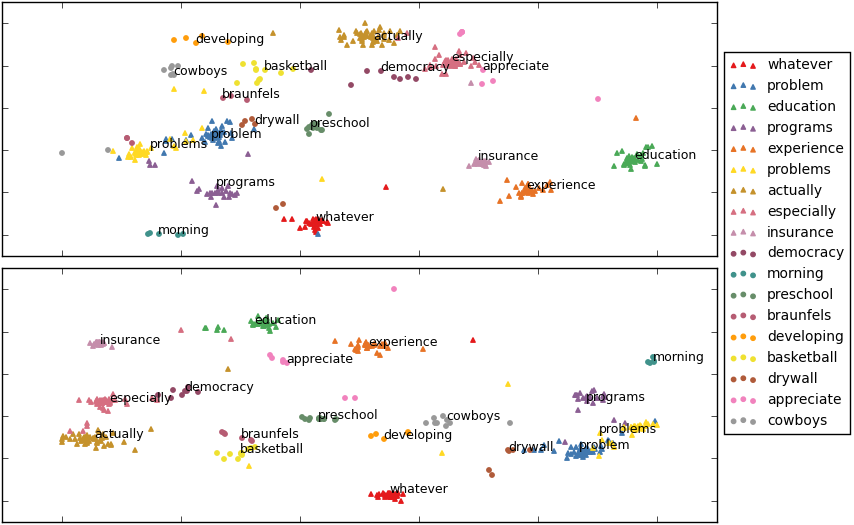}
\caption{t-SNE~\cite{maaten2008visualizing} visualization of word embeddings from the development set produced by the classifier (top) vs.~Siamese (bottom) models.  Word labels seen at training time are denoted by triangles and word labels unseen at training time are denoted by circles.}
\label{ch:rnn_awe:fig:tsne}
\end{figure}

In order to gain a better qualitative understanding of the differences between Classifier and Siamese embeddings as well as the learned embedding space more generally, we plot a two-dimensional visualization of some of our learned embeddings via t-SNE~\cite{maaten2008visualizing} in Figure~\ref{ch:rnn_awe:fig:tsne}. For both Classifier and Siamese embeddings, there is a marked difference in the quality of clusters formed by embeddings of words that were previously seen vs.~previously unseen in training.  However, the Siamese network embeddings appear to have better relative distances between word clusters with similar and dissimilar pronunciations. For example, the word \texttt{programs} appears equidistant from \texttt{problems} and \texttt{problem} in the classifier-based embedding space, but in the Siamese embedding space \texttt{problems} falls between \texttt{problem} and \texttt{programs}. Similarly, the cluster for \texttt{democracy} shifts with respect to \texttt{actually} and \texttt{especially} to better respect differences in pronunciation. More study of learned embeddings, using more data and word types, is needed to confirm such patterns in general. Improvements in unseen word embeddings from the Classifier embedding space to the Siamese embedding space (such as for \texttt{democracy}, \texttt{morning}, and \texttt{basketball}) are a likely result of optimizing the model for relative distances between words.

\section{Conclusion}

Our main finding is that RNN-based AWEs outperform prior approaches, as measured via the acoustic word discrimination task from Carlin {\it et al.}~\cite{carlin2011rapid_eval_spoken_term_detect}. Our best results are obtained when using deep LSTM RNNs with both several stacked recurrent layers and several fully-connected layers, as well as optimization done with a contrastive Siamese loss. Our experiments suggest that the models could still be improved with additional layers. In addition, we have found that, for the purposes of classifier-based AWE training, fully-connected layers are very important and have a more significant effect per layer than stacked layers.

These experiments represent an initial exploration of sequential neural models for AWEs. There are a number of directions for further work. For example, while our analyses suggest that Siamese networks are better than classifier-based models at embedding previously unseen words, our best embeddings are still much poorer for unseen words. Improvements in this direction may come from larger training sets, or may require new models that better model the shared structure between words.  Other directions for future work include additional forms of supervision and training, as well as application to downstream tasks. Many of these will be addressed in the thesis. For example, improved shared structure can be seen with our follow-up to multi-view training~\cite{he2017multiview} in Chapters~\ref{ch:ctc_a2w} and~\ref{ch:multi_awe}, and downstream application of our embedding models to query-by-example search can be seen in Chapters~\ref{ch:rnn_qbe} and~\ref{ch:multi_qbe} and to speech recognition in Chapters~\ref{ch:ctc_a2w},~\ref{ch:seg_a2w}, and~\ref{ch:joint_a2w}.

%% file: text/4_rnn_qbe.tex
\chapter{Query-by-example search with discriminative neural acoustic word embeddings}
\label{ch:rnn_qbe}

\begin{figure}[H]
\centering
\includegraphics[width=0.8\linewidth]{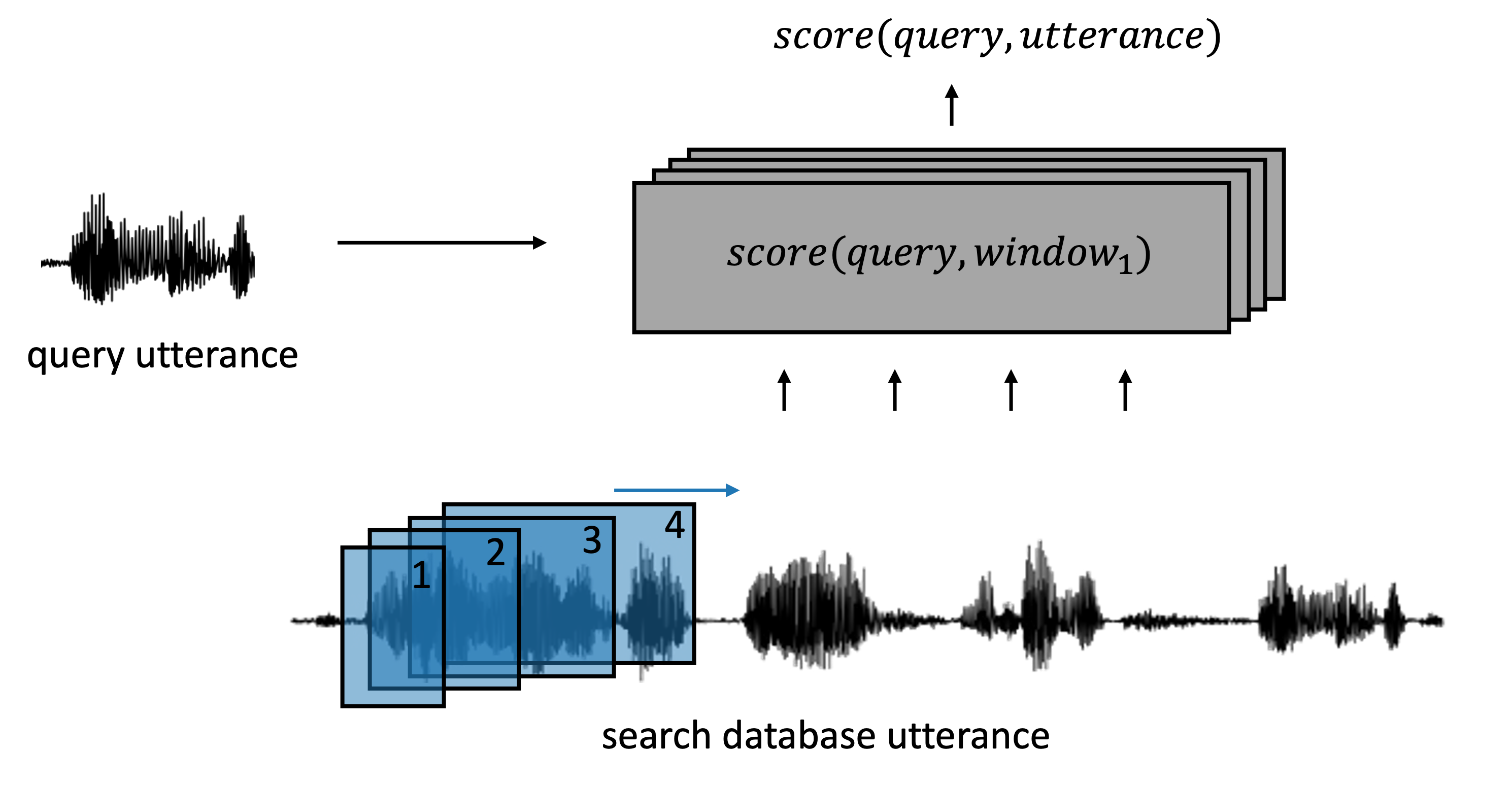}
\caption{Illustration of query-by-example speech search where a spoken query is compared to windows of an utterance within the search audio database.}
\label{ch:rnn_qbe:qbe_basic}
\end{figure}

Query-by-example (QbE) speech search often uses dynamic time warping (DTW) for comparing queries and segments within a search utterance. Recent work has shown that comparing speech segments by representing them as fixed-dimensional vectors---acoustic word embeddings (AWEs)---and measuring their vector distance (e.g., cosine distance) can discriminate between words more accurately than DTW-based approaches. We consider an approach to QbE search that embeds both the query and database segments according to a neural model, followed by nearest-neighbor search to find the matching segments. Earlier work on embedding-based QbE~\cite{levin2015srails} with template-based AWEs~\cite{levin2013fixed} achieves competitive performance with DTW while being much faster. Our recurrent neural network (RNN)-based embeddings trained to optimize word discrimination, achieve substantial improvements in performance and run-time efficiency over previous approaches. In this chapter, we detail our next contribution using RNN-based AWEs in downstream application to query-by-example (QbE) speech search. This work is published in Settle {\it et al.}~\cite{settle2017query}, and showcases that performance improvements seen in acoustic word discrimination translate well to QbE search.

\subsubsubsection{Collaboration} This work relies on collaboration with Keith Levin. Keith is the first author of Levin {\it et al.}~\cite{levin2015srails}, which introduces the Segmental Randomized Acoustic Indexing and Logarithmic-Time Search (S-RAILS) system. This is the first embedding-based QbE search system, and it uses templated-based AWEs introduced by Keith's prior work~\cite{levin2013fixed}. S-RAILS~\cite{levin2015srails} is our primary baseline, and Keith provides the same approximate nearest neighbor search system for this work to isolate the comparison to the quality of the learned embeddings.

\section{Introduction}
Query-by-example (QbE) speech search is the task of searching for a spoken query term (a word or phrase) in a collection of speech recordings. Unlike keyword search and spoken term detection, where the search terms are given as text, QbE involves matching audio segments directly. This task arises naturally when the search terms may be out-of-vocabulary~\cite{shen2009comparison,parada2009query}, under hands-free conditions, or in low- or zero-resource settings~\cite{szoke2015query}.

For QbE in high-resource settings, one can train a model to map the audio query to a sequence of subword units, such as phonemes, and search for this sequence in a lattice built from the search collection~\cite{allauzen2004general,parada2009query}. This approach requires very significant resources, however, since it involves much the same process as training a full speech recognition system. In contrast, approaches to this task in low-resource settings typically use dynamic time warping (DTW) to determine the similarity between audio segments. Early approaches to low-resource QbE perform DTW alignment of the query against a search collection either exactly~\cite{hazen2009query_posteriorgram_templates,zhang2009unsupervised_spoken_keyword_spotting} or approximately~\cite{zhang2011piecewise_posteriorgram_dtw,jansen2012rails,mantena2013speedup_dtw_hierarchical_kmeans}.

An alternative to DTW for QbE, which we explore in this chapter, is to represent variable-duration speech segments as fixed-dimensional vectors and directly measure similarity between them via a simple vector distance. In this approach, the query is embedded using an \textit{acoustic word embedding} (AWE) function, which produces a vector representation of the query. All potential segments in the search collection are then represented as vectors as well using the same embedding function. The putative hits (matches) correspond to those segments in the search collection that are closest to the query in the fixed-dimensional embedding space. This type of approach requires preprocessing steps for learning the embedding function and generating the embeddings of the search collection. Then, at test time, efficient approximate nearest-neighbor search can be used to further speed up computation.

In prior work, Levin {\it et al.}~\cite{levin2013fixed} proposes a template-based AWE approach. For each speech segment, a {\it reference vector} is defined as the vector of DTW alignment costs to a set of template segments. Dimensionality reduction (based on Laplacian eigenmaps~\cite{belkin2003laplacian_eigenmaps}) is then applied to the reference vector to obtain an embedding in $\mathbb{R}^d$. Levin {\it et al.} shows that using embedding representations can greatly speed up search compared to a purely DTW-based system~\cite{jansen2012rails}, while matching or improving performance. This template-based embedding approach was subsequently used for unsupervised speech recognition in~\cite{kamper2015unsupervised_small_vocab_asr} and, more importantly, the full QbE system of~\cite{levin2015srails}, which we consider to be a baseline in our experiments. Their template-based embedding approach does not require any labeled supervision, but in many practical settings a limited amount of training data may be available. In this work, we consider this low-resource setting. Following from Chapter~\ref{ch:rnn_qbe}, we train RNN-based neural AWE models to discriminate between words given a limited (roughly $100$ minute) training set.

We build on a growing body of work on neural network-based AWEs~\cite{chung2016audio,kamper2016cnn_awe,settle2016rnn_awe,he2017multiview,audhkhasi2017asr_free_kws}. In several of these studies, neural approaches are shown to far outperform template-based embeddings (such as those used in~\cite{levin2015srails}) on the acoustic word discrimination task~\cite{carlin2011rapid_eval_spoken_term_detect}. Here, we use the neural embedding approach of Settle and Livescu~\cite{settle2016rnn_awe}, based on Siamese training of RNN-based AWEs, and incorporate these into a complete QbE system using the embedding-based approach of Levin {\it et al.}~\cite{levin2015srails}. We show that these neural acoustic word embeddings, trained only on a small amount of labeled data, achieve large improvements in QbE performance.

\section{Approach}

Embedding-based QbE search consists of an embedding method and a nearest neighbor search component. We first describe our neural acoustic word embedding (NAWE)\footnote{To differentiate our approach from the template-based AWE approach of Levin {\it et al.}~\cite{levin2015srails}, we refer to our work as neural AWE or ``NAWE".} approach, and then give details of the embedding-based QbE search system in which the embeddings are used.

\subsection{Neural acoustic word embeddings (NAWEs)}
\label{ch:rnn_qbe:approach:nawes}

An acoustic word embedding (AWE) function $f$ maps a variable-length speech segment $\mats{X}$ to a single embedding vector $f(\mats{X}) \in \mathbb{R}^d$, and once embedded, these speech segments can be compared by computing the vector distance between their embeddings, rather than using DTW.

Recently, neural acoustic word embeddings (NAWEs) have been proposed as an alternative~\cite{chung2016audio,kamper2016cnn_awe,settle2016rnn_awe,he2017multiview,audhkhasi2017asr_free_kws}. In this recent work, NAWEs have achieved much better performance than the template-based approach of~\cite{levin2013fixed}, but only in an isolated-word discrimination task that can be seen as a proxy for QbE~\cite{carlin2011rapid_eval_spoken_term_detect}. Here, we specifically focus on the NAWE approach developed in~\cite{settle2016rnn_awe}, where it was shown that embeddings based on Long Short-Term Memory (LSTM)~\cite{hochreiter1997lstm} networks outperform competing neural methods. However, here we apply these NAWEs in a complete embedding-based QbE system.

We use the concatenation of the hidden representations from a deep bidirectional LSTM network as our embedding function, i.e.\ $f(\mats{X}) = [\overrightarrow{h_T}; \overleftarrow{h_1}]$, where $\overrightarrow{h_T}, \overleftarrow{h_1}$ refer to the final hidden state vector from the forward and backward LSTMs, respectively. Following from Chapter~\ref{ch:rnn_awe}, this LSTM is trained using the Siamese weight-sharing scheme~\cite{bromley1993signature_verification_siamese} depicted previously in Figure~\ref{ch:rnn_awe:fig:contrastive_triplet_objective} with a contrastive triplet loss~\cite{chopra2005learning_similarity_face,socher2014grounded} defined as 
\begin{equation}
\mathcal{L}_\textrm{cos-hinge}(\mats{X}_a, \mats{X}_s) = \left[m + d(f(\mats{X}_a), f(\mats{X}_s)) - \min_{\mats{X}_d \in \mathcal{X}'(\mats{X}_a, v_a)} d(f(\mats{X}_a), f(\mats{X}_d))\right]_+
\label{ch:rnn_qbe:eq:most-cos-hinge}
\end{equation}
\noindent where $m$ is a margin hyperparameter, $d({\bf a}, {\bf b}) = 1-\frac{{\bf a}\cdot {\bf b}}{\Vert {\bf a} \Vert\Vert {\bf b} \Vert}$, and 
\begin{flalign*}
    \mathcal{X}'(\mats{X}, v) &:= \{\mats{X}' \text{  } \vert \text{  } v' \in \mathcal{V} / v, \text{  } (\mats{X}', v') \in \mathcal{D}\}
\end{flalign*}
\noindent where $\mathcal{V}$ is the training vocabulary and $\mathcal{D}$ is the training set of speech segments and their word labels. Equation~\ref{ch:rnn_qbe:eq:most-cos-hinge} is used for contrastive training where segments $\mats{X}_a$ and $\mats{X}_s$ are the ``anchor" and ``same" segments, respectively, as they share a word label $v_a$. Here, there is an important update to Equation~\ref{ch:rnn_awe:eq:cos-hinge} from Chapter~\ref{ch:rnn_awe}, which is that now the ``different" segment $\mats{X}_d$ is defined as the most offending example from a set of negative examples (i.e. examples with word label different from $v_a$). Now, rather than sampling a single negative example uniformly at random as in Kamper {\it et al.}~\cite{kamper2016cnn_awe} or according to confusion statistics as in Chapter~\ref{ch:rnn_awe}, Equation~\ref{ch:rnn_qbe:eq:most-cos-hinge} uses the acoustic example $\mats{X}'$ whose embedding $f(\mats{X}')$ most violates the margin constraint. For efficiency, we will in practice use an approximation by randomly subsampling from the full dataset $\mathcal{D}$ to derive the set $\mathcal{X}'(\mats{X}_a, v_a)$ from which we find the most offensive example $\mats{X}'$. We find this approach to significantly improve both our model's word discrimination performance as well as the rate of model convergence. 

\subsection{Embedding-based QbE}
\label{ch:rnn_qbe:approach:srails}

\begin{figure}
\centering
\includegraphics[width=0.8\linewidth]{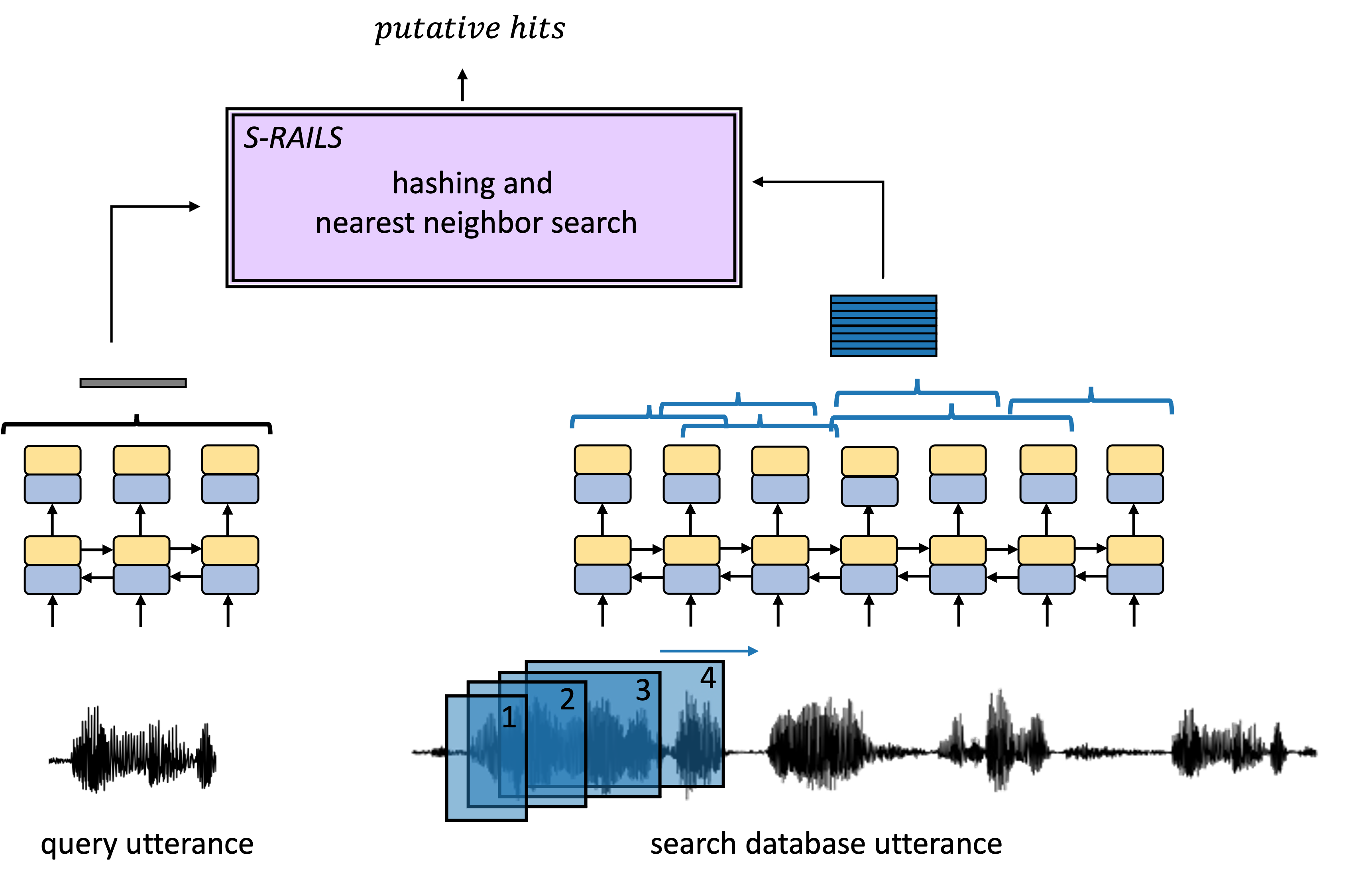}
\caption{Illustration of our embedding-based query-by-example search pipeline where the details of S-RAILS are applied within the ``hashing and nearest neighbor search" module after the query and spoken segments of each utterance have been embedded.}
\label{ch:rnn_qbe:fig:qbe_srails}
\end{figure}

Our system (Figure~\ref{ch:rnn_qbe:fig:qbe_srails}) needs to quickly retrieve from a large collection those segments nearest to a given spoken query. For this, we use the Segmental Randomized Acoustic Indexing and Logarithmic-Time Search (S-RAILS) system~\cite{levin2015srails}, an embedding-based QbE approach. Although S-RAILS was first applied using the template-based AWE method, it is agnostic to the embedding type, and here we apply it to our neural AWEs.

S-RAILS is a simple platform for performing approximate nearest neighbor search over vectors, relying on locality-sensitive hashing (LSH)~\cite{indyk1998approximate_nearest_neighbors,charikar2002similarity_from_rounding_algo}. Let $\mathcal{S} = \{\vecs{s}_1,\vecs{s}_2,\dots,\vecs{s}_N\} \in \mathbb{R}^d$ be embeddings of segments of the search collection. LSH is a method for representing vectors in $\mathbb{R}^d$ as bit vectors, which we refer to as {\it signatures}, such that if two vectors $s_i,s_j \in \mathcal{S}$ are close under the cosine distance, then their signatures $\sigma_i,\sigma_j \in \{0,1\}^b$ will agree in most of their entries (i.e. are close under Hamming distance).

S-RAILS uses LSH to replace the comparatively expensive $d$-dimensional cosine distance between embeddings with a fast approximation. S-RAILS arranges the signatures of segments of the search collection $\mats{\Sigma} = \{ \vecs{\sigma}_i : 1 \le i \le N \}$ into a lexicographically sorted list $\mats{\Sigma}^{\star}$. Then, given a query vector $\vecs{q} \in \mathbb{R}^d$, we map $\vecs{q}$ to its LSH signature $\vecs{\sigma}_q \in \{0,1\}^b$, and find its location in the sorted signature list $\mats{\Sigma}^{\star}$ in $\mathcal{O}(\log b)$ time. A set of (approximate) near neighbors to $\vecs{q}$ can be read off this list by looking at the $B$ entries appearing before $\vecs{\sigma}_q$ and the $B$ entries after $\vecs{\sigma}_q$. Since bits appearing earlier in the signatures $\vecs{\sigma}_i,\vecs{\sigma}_j$ have more influence on whether or not two vectors $s_i,s_j$ from the search database are judged as similar, S-RAILS performs this lexicographic lookup under $P$ different permutations of the bits to ameliorate this effect.

This gives the S-RAILS algorithm three hyperparameters: signature length $b$, beamwidth $B$, and number of permutations $P$. Increasing any of these parameters tends to improve performance either because it increases the fidelity of our approximation to the cosine distance (in the case of $b$ and $P$) or because it improves recall (in the case of $B$). However, such improvements come at the cost of increased memory required to store the index and permuted lists (in the case of $b$ and $P$) and increased runtime (in the case of $B$ and, to a lesser extent, $b$ and $P$). All told, building the index requires $\mathcal{O}( P b N \log N )$ time in the worst case, and querying the index requires $\mathcal{O}( B + P b \log N )$ time.

\section{Experimental setup}
\label{ch:rnn_qbe:setup}

We use data from the Switchboard corpus of conversational English telephone speech~\cite{godfrey1992switchboard}. For training the NAWE model, we use a training set consisting of approximately $10$k word segments covering less than 2 hours of speech taken from conversation sides distinct from those used to extract the query set and the evaluation collection for QbE. The size of this training set is comparable to those used for training in prior work on AWEs~\cite{kamper2015unsupervised_weak_topdown,jansen2013weak_topdown,settle2016rnn_awe}. As acoustic features, we use $39$-dimensional MFCC+$\Delta$+$\Delta\Delta$s. 

For the QbE task setup, we partition Switchboard into a $37$-hour set from which we draw our query terms, a $48$-hour development search collection on which we tune the hyperparameters of S-RAILS, and a $433$-hour evaluation search collection. These partitions are identical to those used in prior work on QbE from Jansen {\it et al.}~\cite{jansen2012rails} and Levin {\it et al.}~\cite{levin2015srails}, which serve as our primary baselines for comparison. We use a set of $43$ query word types previously used in~\cite{jansen2012rails,levin2015srails}, which were chosen subject to the constraints that the median word segment duration of each word type across the entire corpus is at least $0.5$ seconds and the orthographic representation of each word type has at least $6$ characters~\cite{jansen2012rails,levin2013fixed,levin2015srails}. Each word type appears $20$ to $162$ times in the query set, $2$ to $188$ times in the development search collection, and $39$ to $1386$ times in the evaluation search collection.

For our NAWE model (Section~\ref{ch:rnn_qbe:approach:nawes}), we use a stacked $3$-layer bidirectional LSTM with $256$ hidden units in each direction such that the output embeddings of our model are $512$-dimensional. Dropout is applied with probability $0.3$ between LSTM layers. When using the contrastive loss $\mathcal{L}_{\textrm{cos-hinge}}$, we set the margin hyperparameter to $m=0.5$ and we sample $k=10$ negative instances per anchor segment. We use minibatch optimization with a batch size of $32$ and the Adam~\cite{kingma2014adam} optimizer with learning rate $\eta = 0.001$, $\beta_1 = 0.9$, $\beta_2 = 0.999$, and $\epsilon=10^{-8}$. These parameters are tuned based on development set acoustic word discrimination~\cite{carlin2011rapid_eval_spoken_term_detect} performance. For our QbE evaluation experiments, all models train for $100$ epochs.

We evaluate search results using three metrics from prior work, including two recall-focused metrics in \emph{figure-of-merit} (FOM) and \emph{oracular term weighted value} (OTWV) as well as one precision-focused metric in \emph{precision at 10} (P@10). FOM is the recall averaged over the ten operating points at which the false alarm rate per hour of search audio is equal to $1,2,\dots,10$. OTWV is a query-specific weighted difference between the recall and the false alarm rate (further explanation can be found in~\cite{miller2007rapid_spoken_term_detect}). P@10 is the fraction of the ten top-scoring results that are correct matches to the query.

Since each spoken instance of a query word type varies, we compute the median and maximum metric scores over all examples of each query word type, and then report the unweighted arithmetic mean of the median and maximum scores across the $43$ query word types.

\section{Results}

We first present QbE performance on the development search data in order to show how performance differs across parameter settings, and then give evaluation results. In this section, we refer to the QbE system that employs the original template-based embeddings simply as S-RAILS, and to the system with neural embeddings as S-RAILS+NAWE.

\begin{table}[H]
\centering
\caption{Effect of signature length $b$ when permutations $P=16$ and beamwidth $B=2000$.}
\begin{tabular}{ccccccc}
\toprule
    & \multicolumn{3}{c}{{\bf Median}} & \multicolumn{3}{c}{{\bf Maximum}}\\
$b$ & FOM & OTWV & P@10   & FOM & OTWV & P@10\\
\midrule
128  & 62.1 & 37.4 & 42.1 & 81.7 & 60.8 & 83.8\\
256  & 67.2 & 42.6 & 48.6 & 83.0 & 65.4 & 84.9\\
512  & 68.2 & 44.8 & 52.6 & 83.6 & 65.9 & 84.9\\
1024 & 69.1 & 46.5 & 54.5 & 84.1 & 66.7 & 84.8\\
2048 & 70.4 & 48.3 & 54.5 & 85.0 & 66.8 & 86.0\\
\bottomrule
\end{tabular}
\label{ch:rnn_qbe:tab:vary-b}
\end{table}

\begin{table}[H]
\centering
\caption{Effect of permutations $P$ when signature length $b=1024$ and beamwidth $B=2000$.}
\begin{tabular}{ccccccc}
\toprule
    & \multicolumn{3}{c}{{\bf Median}} & \multicolumn{3}{c}{{\bf Maximum}}\\
$P$ & FOM & OTWV & P@10   & FOM & OTWV & P@10\\
\midrule
4  & 48.8 & 33.2 & 45.2 & 75.2 & 59.0 & 83.0\\
8  & 60.9 & 41.0 & 50.3 & 80.3 & 63.8 & 85.0\\
16 & 69.1 & 46.5 & 54.5 & 84.1 & 66.7 & 84.8\\
\bottomrule
\end{tabular}
\label{ch:rnn_qbe:tab:vary-P}
\end{table}

\begin{table}[H]
\centering
\caption{Effect of beamwidth $B$ when signature length $b=1024$ and permutations $P=16$.}
\begin{tabular}{ccccccc}
\toprule
    & \multicolumn{3}{c}{{\bf Median}} & \multicolumn{3}{c}{{\bf Maximum}}\\
$B$ & FOM & OTWV & P@10   & FOM & OTWV & P@10\\
\midrule
	1000  & 65.8 & 44.8 & 53.4 & 83.0 & 65.6 & 85.0\\
	2000  & 69.1 & 46.5 & 54.5 & 84.1 & 66.7 & 84.8\\
	10000 & 74.6 & 49.5 & 54.2 & 86.3 & 67.9 & 84.8\\
\bottomrule
\end{tabular}
\label{ch:rnn_qbe:tab:vary-B}
\end{table}

\subsection{Development set performance}

\begin{figure}[t]
  \centering
  \includegraphics[width=0.8\linewidth]{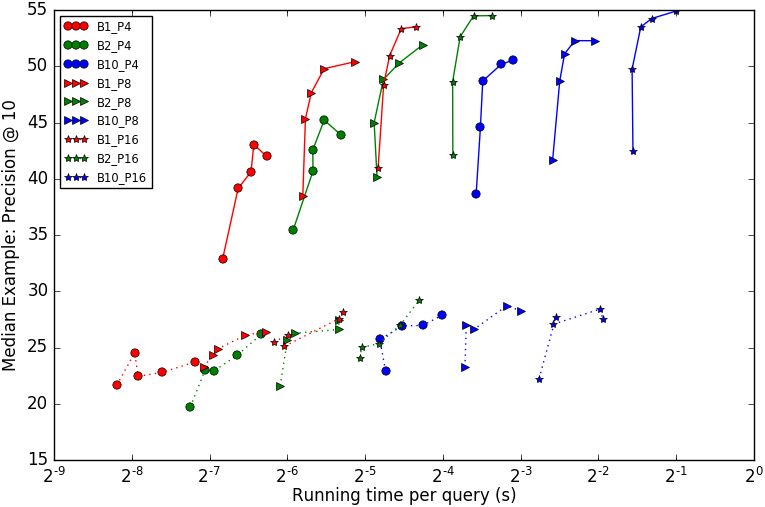}
  \caption{Median P@10 of S-RAILS (dotted) and S-RAILS+NAWE (solid) on the development search collection.  Each sequence of connected points indicates results for a fixed permutation number (P) and beamwidth (B) while signature length (b) varies from 128 to 2048.  In the legend, ``B$x$\_P$y$'' indicates beamwidth of $1000x$ and number of permutations $y$.}  
  \label{ch:rnn_qbe:fig:p_at_10_vs_runtime}
\end{figure}

Figure~\ref{ch:rnn_qbe:fig:p_at_10_vs_runtime} shows development set performance in terms of median P@10 for the baseline S-RAILS system using template-based embeddings and our S-RAILS+NAWE. Tables~\ref{ch:rnn_qbe:tab:vary-b},~\ref{ch:rnn_qbe:tab:vary-P}, and~\ref{ch:rnn_qbe:tab:vary-B} show development set performance for S-RAILS+NAWE as the signature length $b$, permutations $P$, and beamwidth $B$ are varied, respectively.

Figure~\ref{ch:rnn_qbe:fig:p_at_10_vs_runtime} shows that neural embeddings improve the performance of S-RAILS by large margins at all running time operating points. This figure also shows that increased signature length yields much larger improvements in P@10 for S-RAILS+NAWE than it does for the baseline S-RAILS system. Significant improvements in P@10 can be seen when holding fixed any combination of settings for $P$ and $B$. Our performance on P@10 saturates with signatures around $1024$ bits, while S-RAILS' saturates, for the most part, at $256$ bits.

Again in contrast to the S-RAILS system, our method responds strongly to increases in the number of permutations used. In both Figure~\ref{ch:rnn_qbe:fig:p_at_10_vs_runtime} and Table~\ref{ch:rnn_qbe:tab:vary-P}, adjustment to this parameter improves performance consistently across signature lengths. This is to be expected as the neural embeddings provide a better measure of speech segment distances, since the increased number of permutations helps provide a more exact estimate of the cosine distance. We note that performance as measured in Table~\ref{ch:rnn_qbe:tab:vary-P} has not plateaued in any of the Median Example metrics. Further increasing $P$ may further improve these metrics, but this incurs a large cost in memory.

Figure~\ref{ch:rnn_qbe:fig:p_at_10_vs_runtime} and Table~\ref{ch:rnn_qbe:tab:vary-B} show that, except for cases with short signatures and few permutations, increasing beamwidth does not improve P@10 performance, while incurring significant cost. To obtain higher precision systems, it is more important to use computational resources for increasing the number of permutations ($P$) or using longer signatures ($b$). However, the higher beamwidths do help to significantly improve recall as seen in the FOM score.

\subsection{Evaluation set performance}

\begin{table}[H]
\small
\centering
\caption{Comparison with prior work on the evaluation set; time is per query.}
\begin{tabular}{cccccccc}
\toprule
System  & Time (s)  & \multicolumn{3}{c}{{\bf Median}}  & \multicolumn{3}{c}{{\bf Maximum}}\\
        &           & FOM & OTWV & P@10                 & FOM & OTWV & P@10\\
\midrule
RAILS~\cite{jansen2012rails}    & 24.7 &  6.7 &  2.7 & 44.0 & 20.7 & 10.4 & 84.4\\
S-RAILS~\cite{levin2015srails}  & 0.08 & 24.5 & 14.4 & 34.5 & 46.2 & 26.6 & 87.4\\
S-RAILS+NAWE (ours)             & 0.38 & 43.3 & 22.4  & 60.2 & 65.4 & 43.3 & 95.1\\ 
\bottomrule
\end{tabular}
\label{ch:rnn_qbe:tab:eval}
\end{table}

Based on development set results, we find that an operating point of $16$ permutations, beamwidth of $2000$, and signature length of $1024$ is close to optimal, in terms of both performance and query speed, for both the baseline S-RAILS and S-RAILS+NAWE. We use these settings for the final evaluation shown in Table~\ref{ch:rnn_qbe:tab:eval}. Besides the S-RAILS baseline, we also compare to RAILS~\cite{jansen2012rails}, a DTW-based system that is optimized for speed using LSH to get approximate frame-level near neighbor matches. RAILS evaluation scores are reproduced from Jansen {\it et al.}~\cite{jansen2012rails}. We find that our approach improves significantly over both RAILS and S-RAILS in terms of all performance metrics at this operating point. Note that, based on Figure~\ref{ch:rnn_qbe:fig:p_at_10_vs_runtime}, the improvements should hold at most operating points, including ones with much higher query speeds. The biggest gains from S-RAILS+NAWE are seen in the Median Example results, where there is a relative improvement over S-RAILS of more than $55\%$ across all measures. In terms of FOM and OTWV, we see relative improvements of over $40\%$ in the Maximum Example case. Although the baselines obtain good P@10, we still improve on them, going from $87.1\%$ to $95.1\%$.

\begin{figure}[H]
  \centering
  \includegraphics[width=\linewidth]{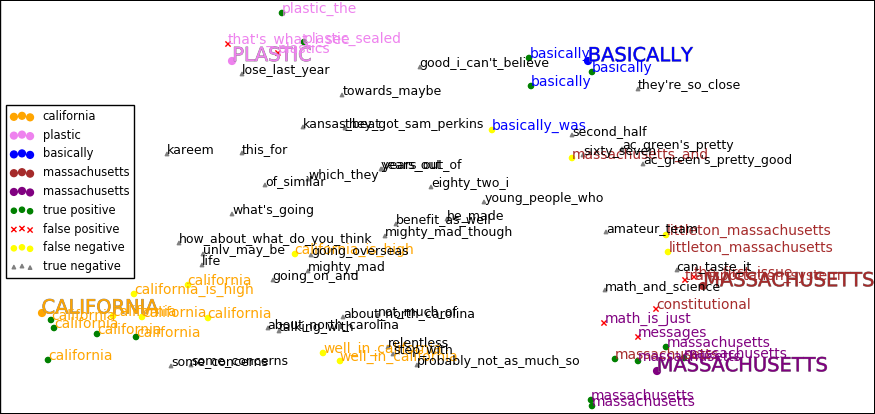}
  \caption{Embeddings of queries and their top hits, visualized in two dimensions using t-SNE~\cite{maaten2008visualizing}.  Queries are shown in capital letters.  The top several hits for each query are shown in the same color as the query. Random segments from the search collection and their associated transcriptions are shown in gray to provide additional context.}
  \label{ch:rnn_qbe:fig:queries_and_top_hits}
\end{figure}

For a qualitative view, Figure~\ref{ch:rnn_qbe:fig:queries_and_top_hits} visualizes several queries and their top hits in the evaluation collection. This visualization shows some expected properties. For example, the two ``Massachusetts" queries and their top hits are embedded close together. Two of the false alarms for ``Massachusetts" are the similar-sounding ``messages" and ``math is just", while the somewhat more distant ``math and science" is (correctly) not retrieved.

\section{Conclusion}

We have presented an approach to query-by-exmaple (QbE) speech search using neural acoustic word embeddings (NAWEs), demonstrating the ability of these embedding models to improve over previous methods on a realistic downstream task. The neural embeddings are learned from a very limited set of data, and one interesting future direction is to study the limits of the approach as the amount of training data is varied. Another interesting aspect of the approach is that the neural embeddings are learned from speech segments that have been pre-segmented at word boundaries, but they are then applied for embedding arbitrary segments that may or may not (and usually do not) correspond to words.  It is encouraging that this approach works despite the lack of non-word examples in the training data, and an interesting avenue for future work is to attempt to further improve performance by explicitly training on both word, non-word, and multi-word segments. In Chapter~\ref{ch:multi_qbe}, we explicitly explore this idea by training on multi-word segments.

%% file: text/5_ctc_a2w.tex
\chapter{Acoustically grounded word embeddings for acoustic-to-word speech recognition}
\label{ch:ctc_a2w}

\begin{figure}[H]
\centering
\includegraphics[width=0.65\linewidth]{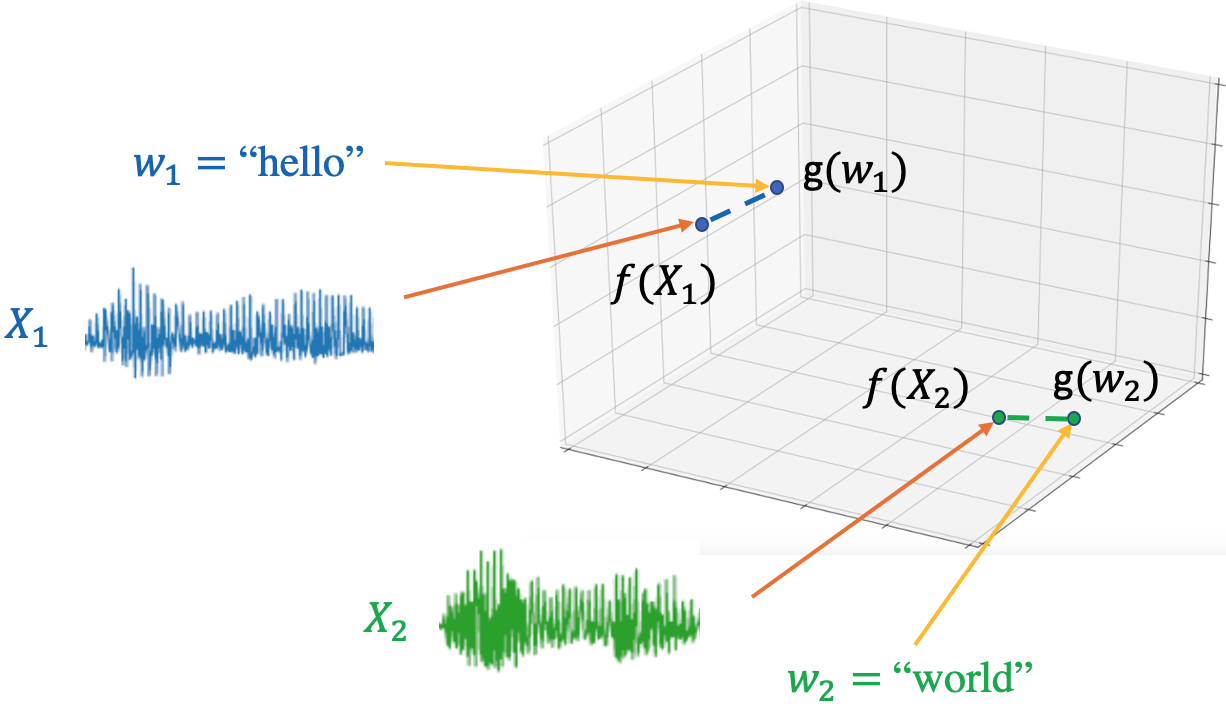}
\caption{Speech segments and written words mapped into a joint embedding space by AWE model $f$ and AGWE model $g$.}
\label{ch:ctc_a2w:fig:agwe_grid}
\end{figure}

While modeling whole-word units is not new to automatic speech recognition (ASR), early methods consider only simple small vocabulary tasks like isolated digit recognition~\cite{levinson1983isolated_digit}. Later work on large vocabulary ASR only use word-level modeling for rescoring purposes, applying either template-based approaches~\cite{heigold2012investigations} or acoustic word embedding (AWE) techniques~\cite{maas2012word,bengio2014word}. More recently, acoustic-to-word (A2W) speech recognition systems~\cite{soltau2016a2w,audhkhasi2017a2w,li2017acoustic} have been introduced, which refer specifically to whole-word models trained {\it end-to-end} for word-level prediction. The goal of A2W models is to learn a mapping from input acoustic representations directly to word-level units, which makes them potentially simpler to train and decode than subword systems. However, this comes at a performance cost relative to subword approaches when training data is limited~\cite{soltau2016a2w,audhkhasi2017a2w,li2017acoustic} or when needing to recognize many words outside the training vocabulary. In this chapter, we describe our contributions published in Settle {\it et al.}~\cite{settle2019_a2w} to improve A2W recognition performance by addressing these limitations.

\section{Introduction}
\label{ch:ctc_a2w:sec:intro}

End-to-end automatic speech recognition (ASR) replaces the modular training approaches of traditional ASR systems with conceptually simpler methods. Instead of requiring separately trained acoustic, pronunciation, and language models, neural network-based models allow for joint optimization under a single objective such as connectionist temporal classification (Section~\ref{ch:back:prelims:asr}). In principle, such neural models could map acoustics directly to words, but to achieve performance comparable with traditional methods, these systems are still typically trained to predict subword units~\cite{rao2017exploring_rnn_transducer,chiu2018sota_asr_with_seq2seq,sanabria2018hierarchical,krishna2018hierarchical}, relying on additional decoders and externally trained language models to produce word-level transcriptions. Acoustic-to-word (A2W) systems~\cite{soltau2016a2w,audhkhasi2017asr_free_kws,audhkhasi2018a2w,li2018advancing,yu2018multistage} avoid this need to decode from subwords to words but introduce new challenges.

\begin{figure}
\centering
\includegraphics[width=0.675\linewidth]{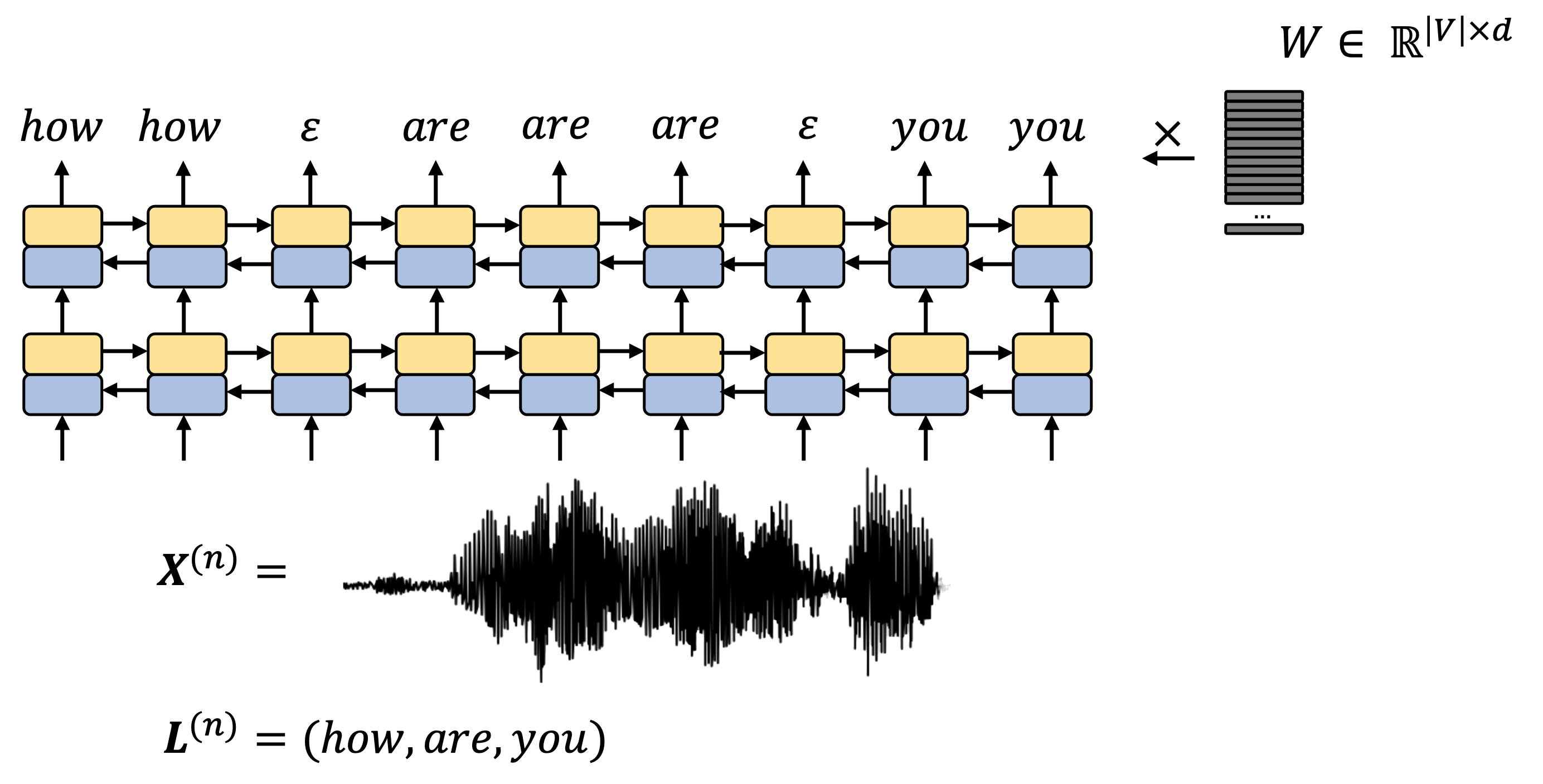}
\caption{Illustration of CTC-based acoustic-to-word (A2W) speech recognition.}
\label{ch:ctc_a2w:fig:recognizer}
\end{figure}

In an A2W model, the prediction layer weight matrix consists of one row vector per word in the vocabulary, i.e. a word embedding matrix. Most rows in this large matrix are associated with very few training examples since the majority of words are rare, and some words cannot even be predicted (unlike in subword models) as they do not appear at all in the training vocabulary. Prior work finds that A2W models either require very large amounts of training data~\cite{soltau2016a2w} or careful training recipes~\cite{audhkhasi2017a2w,audhkhasi2018a2w,yu2018multistage} when using limited amounts of training data. Methods have also been developed to address the out-of-vocabulary (OOV) prediction problem using mixed-unit outputs, including the spell-and-recognize model~\cite{audhkhasi2018a2w} trained to predict a word's character sequence followed by the word itself as well as the multi-task approach of Li {\it et al.}~\cite{li2017acoustic} that uses both character- and word-based CTC training.

In our work, we are most interested in A2W models that produce word-level output predictions only. 
Our method starts by jointly learning an acoustic word embedding (AWE) function---mapping an acoustic signal to a fixed-dimensional vector---and an {\it acoustically grounded} word embedding (AGWE) function---mapping a character sequence to a vector---with the multi-view word embedding approach of He {\it et al.}~\cite{he2017multiview} updated for A2W recognition. After this multi-view pretraining, the learned AWE and AGWE models are used to initialize components of the A2W recognizer. The AWE model serves as the initialization of the backbone acoustic encoder, while the row vectors of the word-level prediction layer matrix are initialized with embeddings produced by forwarding each word's character sequence through the AGWE model. The AGWE model enables better generalization to rare (and even unseen) words by learning to construct word representations from their character sequences, even when learning from very little data (e.g., approximately 100 minutes)~\cite{kamper2016cnn_awe,settle2016rnn_awe,he2017multiview,settle2017query}.

We find A2W recognition performance is consistently improved by initializing the acoustic encoder with the AWE model and the prediction layer with AGWEs. Additional benefits are also seen when regularizing toward AGWEs during recognizer training. Finally, we offer a simple method to extend the vocabulary at test-time by adding rows using the pretrained AGWE model, which benefits performance when the training vocabulary is particularly limited. To do this most effectively, we freeze the rows of the prediction layer after initialization such that the AGWE space is consistent before and after ASR training.

\begin{figure}[b]
\centering
\includegraphics[width=0.8\linewidth]{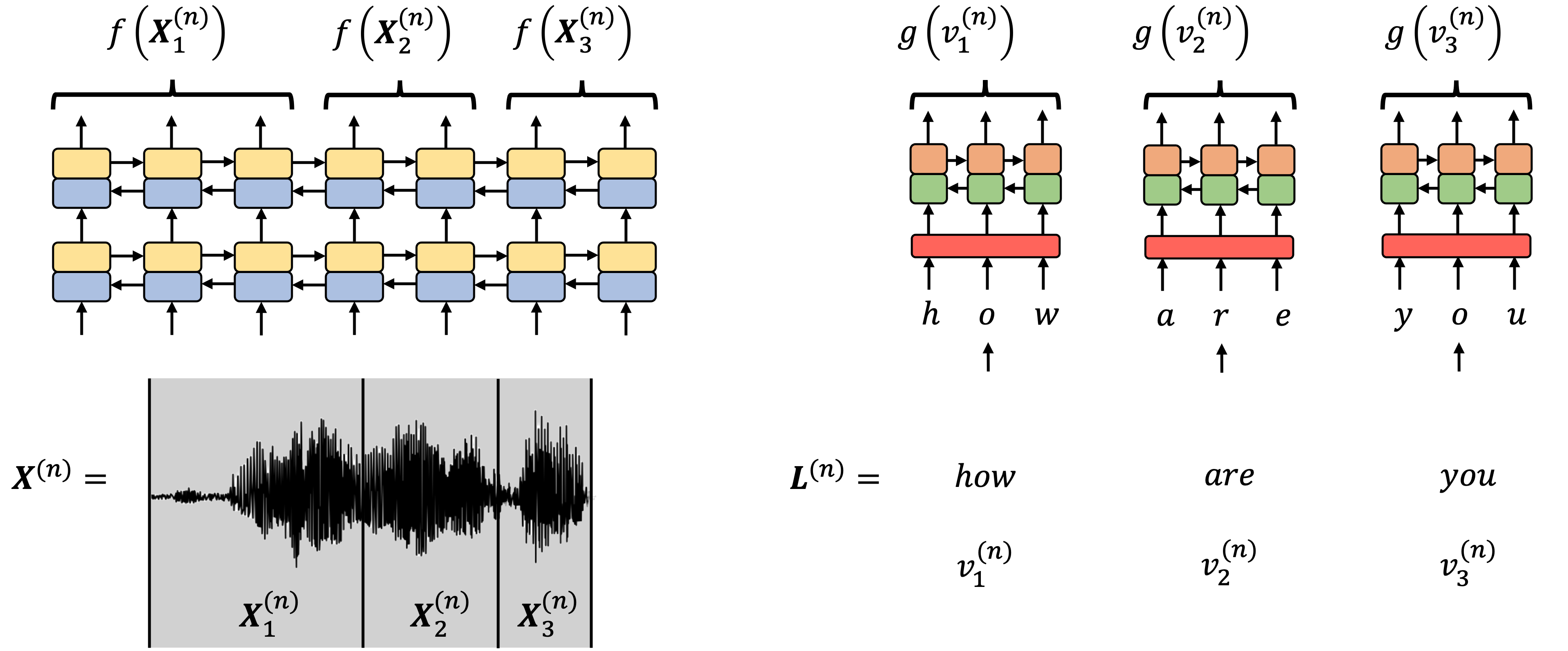}
\caption{Illustration of AWEs (left) and AGWEs (right) produced from spoken segments and written word labels $\mats{X}_i^{(n)}$  and $\vecs{v}_i^{(n)}$, respectively, during multi-view training where utterance $\mats{X}^{(n)}$ has corresponding word label sequence $\mats{L}^{(n)}$.}
\label{ch:ctc_a2w:fig:agwe}
\end{figure}

\section{Approach}
\label{ch:ctc_a2w:sec:approach}

Our A2W recognizer is composed of a single stacked recurrent neural network (RNN), typically a bidirectional long short-term memory network (BiLSTM), trained with the connectionist temporal classification (CTC) loss (Section~\ref{ch:back:prelims:asr}) to map frame-level input acoustic sequences to words. The final (pre-softmax) weight layer consists of one vector per word in the vocabulary, and can therefore also be viewed as an embedding matrix of those words. 

Word embeddings, or continuous-valued vector representations of words, are a common tool in natural language processing (NLP) where they are often trained to represent the semantic meaning of words~\cite{deerwester1990indexing,bengio2003neural,mnih2007three,mikolov2013efficient,pennington2014glove}. In fact, the earlier work on A2W-based speech recognition from Audhkhasi {\it et al.}~\cite{audhkhasi2017a2w} initializes the final layer weights of their A2W model with GloVe word embeddings~\cite{pennington2014glove}. In our work, we investigate the effect of initializing the final weight layer instead with externally trained word embeddings as well as encouraging these weight vectors to remain close to their initialization. However, rather than using semantic word embeddings as in~\cite{audhkhasi2017a2w}, we consider whether {\it acoustically grounded} word embeddings (AGWEs)---that is, embeddings that are trained jointly with AWEs to encode acoustic-phonetic similarity rather than semantic similarity---may be more helpful (see Section~\ref{ch:back:related:agwes}).

Several AWE approaches have been developed for learning to encode either acoustic~\cite{maas2012word,levin2013fixed,chen2015qbe_kws_lstm,kamper2016cnn_awe,chung2016audio,settle2016rnn_awe,audhkhasi2017asr_free_kws} or semantic~\cite{chung2018speech2vec} information, or both~\cite{chen2018phonetic}. Other work considers training both AWE and AGWE models to embed written words into a vector space encoding their acoustic-phonetic content~\cite{bengio2014word,ghannay2016evaluation_awe,audhkhasi2017asr_free_kws,he2017multiview}. Our approach, depicted in Figure~\ref{ch:ctc_a2w:fig:agwe}, follows the latter and is based on He {\it et al.}~\cite{he2017multiview}, where two embedding functions are learned jointly, one for acoustic signals (spoken words) and one for character sequences (written words).

These two embedding models, $f$ and $g$, are trained jointly to map acoustic sequences $\mats{X}$ and character sequences $\vecs{c}$, respectively, to fixed-dimensional vectors in $\mathbb{R}^d$. The AWE model $f$ consists of a stacked BiLSTM, pooling over the output hidden states, and a projection layer mapping to $f(\mats{X}) \in \mathbb{R}^d$. The AGWE model consists of a character embedding layer, a single-layer BiLSTM, concatenation of the final hidden states from each direction, and a projection layer, giving the (acoustically grounded) textual embedding vector $g(\vecs{c}) \in \mathbb{R}^d$. The fully-connected projection layer used for both models is shared. In Equation~\ref{ch:ctc_a2w:eq:multiview} below we use the word $v$ as input to $g$ for notational convenience, but in practice $v$ is first converted to its character sequence $\vecs{c} = \textsc{char}(v)$ before being fed to $g$, i.e. $g(v) = g(\textsc{char}(v)) = g(\vecs{c})$.

During multi-view AWE+AGWE pretraining, our objective $\mathcal{L}_{emb}$ consists of the best loss term combination from He {\it et al.}~\cite{he2017multiview}\footnote{Note that the loss terms are numbered $0$ and $2$, which follows the notation used within He {\it et al.}~\cite{he2017multiview} and will remain consistently written this way throughout the thesis.}:
\begin{flalign}
\mathcal{L}_{emb}(\tens{X}, \tens{L}) = \sum_{n=1}^{N}\sum_{i=1}^{\vert \mats{L}^{(n)} \vert} \mathcal{L}_0(\mats{X}_i^{(n)}, v_i^{(n)}) + \mathcal{L}_2(\mats{X}_i^{(n)}, v_i^{(n)}) 
\label{ch:ctc_a2w:eq:multiview}
\end{flalign}
\noindent where $\mats{X}^{(n)}_i$ is the $i^{th}$ spoken word segment within $\mats{X}^{(n)}$ and $v^{(n)}_i$ is its word label, $\mats{L}^{(n)}$ is the utterance transcript, and the individual loss terms $\mathcal{L}_0$ and $\mathcal{L}_2$ are defined
\begin{flalign*}
\mathcal{L}_0(\mats{X}, v) &= \frac{1}{\vert \mathcal{N}_0(\mats{X}, v) \vert} \sum_{v' \in \mathcal{N}_0(\mats{X}, v)} \left[m + d(f(\mats{X}), g(v)) - d(f(\mats{X}), g(v'))\right]_+ \\
\mathcal{L}_2(\mats{X}, v) &= \frac{1}{\vert \mathcal{N}_2(\mats{X}, v) \vert} \sum_{\mats{X}' \in \mathcal{N}_2(\mats{X}, v)} \left[m + d(f(\mats{X}), g(v)) - d(g(v), f(\mats{X}'))\right]_+
\end{flalign*}
\noindent where $d(\vecs{a}, \vecs{b}) = 1-\frac{\vecs{a}\cdot \vecs{b}}{\Vert \vecs{a} \Vert\Vert \vecs{b} \Vert}$, $m$ is a margin hyperparameter, and the hard negative sampling sets $\mathcal{N}_0(\mats{X}, v)$ and $\mathcal{N}_2(\mats{X}, v)$ are defined
\begin{flalign*}
\mathcal{N}_0(\mats{X}, v) &:= \left\{
    v'\ \middle\vert
    \begin{array}{c}
    v' \in \mathcal{V} / v
    \end{array}
\right\}\\
\mathcal{N}_2(\mats{X}, v) &:= \left\{
    \mats{X}'\ \middle\vert
    \begin{array}{c}
    \textsc{label}(\mats{X}') \in \mathcal{V} / v
    \end{array}
\right\}
\end{flalign*}
\noindent with $\mathcal{V}$ the training vocabulary. For efficiency, hard negative sampling is performed only over the mini-batch. This means that in Equation~\ref{ch:ctc_a2w:eq:multiview} $N$ is the number of utterances in a batch and the vocabulary $\mathcal{V}$ that defines the negative sampling sets consists of the unique words in that batch. Additionally, we limit the size of these negative sample sets to $k$ elements. 
\begin{flalign*}
\mathcal{N}_0^k(\mats{X}, v) &:= \left\{
    v_1', \dots, v_k'\ \middle\vert
    \begin{array}{c}
        v_i' = \displaystyle \argmin_{v' \in \mathcal{N}_0(\mats{X}, v)} d(f(\mats{X}), g(v'))
    \end{array}
\right\}\\
\mathcal{N}_2^k(\mats{X}, v) &:= \left\{
    \mats{X}_1', \dots, \mats{X}_k'\ \middle\vert
    \begin{array}{c}
        \mats{X}_i' = \displaystyle \argmin_{\mats{X}' \in \mathcal{N}_2(\mats{X}, v)} d(g(v), f(\mats{X}'))
    \end{array}
\right\}
\end{flalign*}
We find this form of negative sampling improves both cross-view word discrimination performance as well as the rate of model convergence during training.

\begin{figure}[H]
\centering
\includegraphics[width=0.85\linewidth]{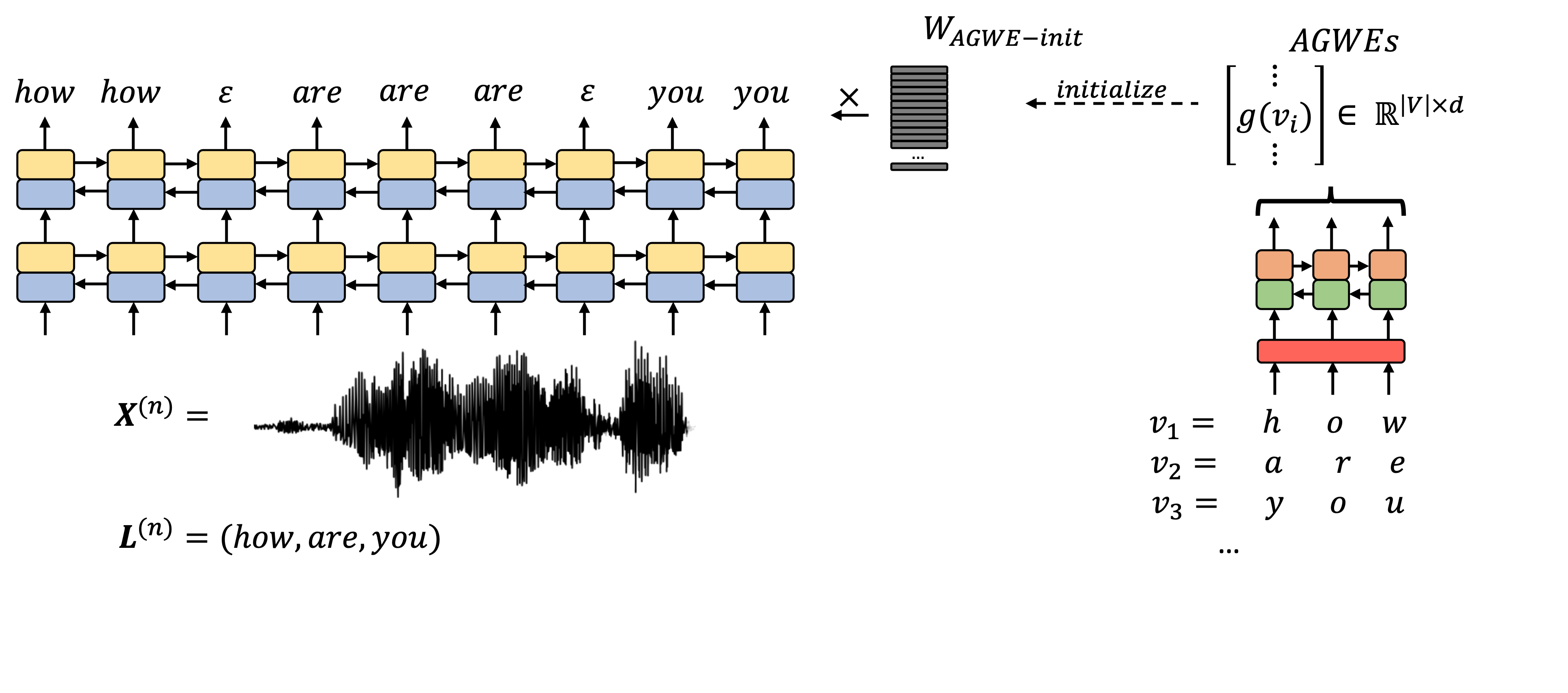}
\caption{CTC-based A2W recognizer with AGWE model $g$ used for (1) initialization of the prediction layer $\mats{W}$.}
\label{ch:ctc_a2w:initialize}
\end{figure}

\begin{figure}
\centering
\includegraphics[width=0.85\linewidth]{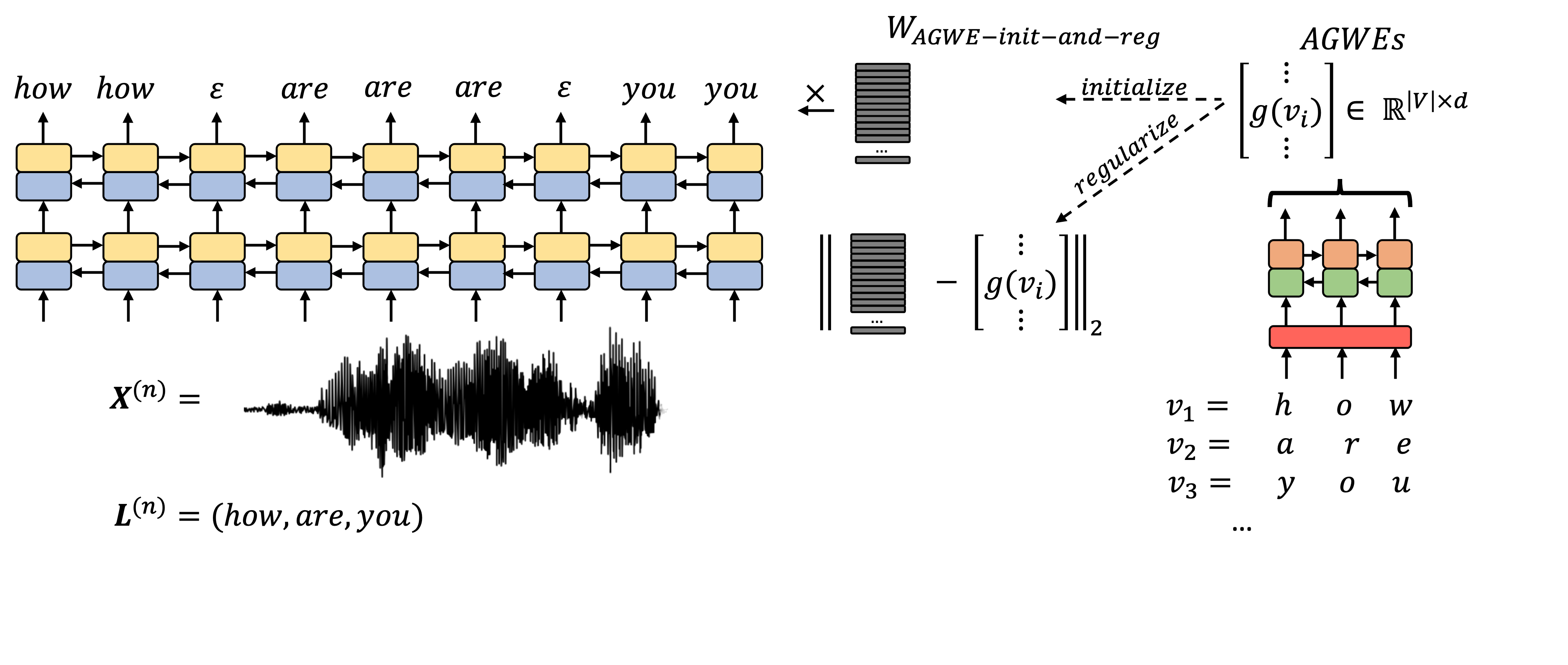}
\caption{CTC-based A2W recognizer with AGWE model $g$ used for (2) regularization of $\mats{W}$ after (1) initialization.}
\label{ch:ctc_a2w:regularize}
\end{figure}

\begin{figure}
\centering
\includegraphics[width=0.85\linewidth]{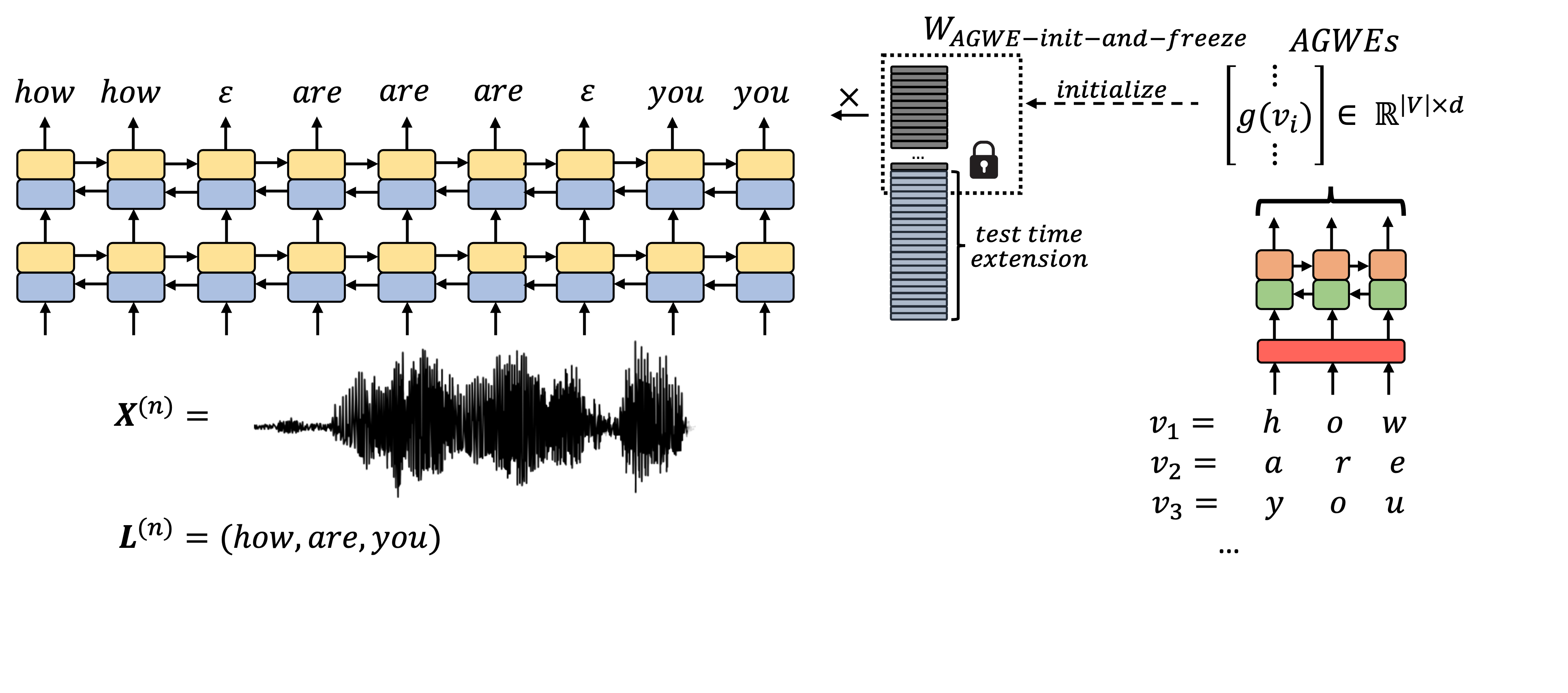}
\caption{CTC-based A2W recognizer with prediction layer $\mats{W}$ frozen after initialization by AGWE model $g$. Then, at test-time $\mats{W}$ can be extended using $g$ to grow the vocabulary $\mathcal{V}$.}
\label{ch:ctc_a2w:freeze}
\end{figure}

After training the AWE and AGWE models following Equation~\ref{ch:ctc_a2w:eq:multiview}, we use the AWE model $f$ to initialize the recurrent acoustic encoder, and we use the AGWE model $g$ to initialize and constrain the prediction layer. We explore using the AGWEs in a few ways: (1) to {\bf initialize} the rows of the prediction layer and then train as usual (Figure~\ref{ch:ctc_a2w:initialize}); (2) to {\bf regularize} the row weights after (1) initialization (Figure~\ref{ch:ctc_a2w:regularize}); and (3) to {\bf freeze} the row weights after (1) initialization (Figure~\ref{ch:ctc_a2w:freeze}).

For the regularization approach, we train the recognizer with a training objective that is a weighted average of the baseline recognizer loss and an $L_2$ regularization loss:
\begin{flalign}
\mathcal{L}_{asr}(\tens{X}, \tens{L}) 
&=  (1-\lambda) \mathcal{L}_{ctc} (\tens{X}, \tens{L}) + \lambda \sum_{v \in \cup_{n=1}^N \mats{L}^{(n)}} \Vert g(v) - \mats{W}_v \Vert_2
\label{ch:ctc_a2w:eq:multitask}
\end{flalign}
\noindent where $(\tens{X}, \tens{L})$ is a batch of $N$ utterance-transcript pairs, $g$ is the character sequence embedding function (i.e. AGWE model), $\mats{W}_v$ is the row of the prediction layer weight matrix $\mats{W}$ corresponding to word $v$, and $\mathcal{L}_{ctc}$ is the CTC loss used for the baseline recognizer (Section~\ref{ch:back:related:asr}).

Freezing the prediction layer after initialization can be viewed as an extreme case of optimizing for CTC while strictly adhering to the embedding initialization. Since $g$ can be applied to arbitrary character sequences, this approach allows us to easily extend the vocabulary at decoding time using a simple adaptation (Figure~\ref{ch:ctc_a2w:freeze}). When decoding, we first predict from the training vocabulary. Then, if we predict an $<$UNK$>$ token, we ``rescore" it by replacing $<$UNK$>$ with a prediction from our extended vocabulary. We show that  our learned embedding function $g$ can accurately extend to arbitrary unseen words without additional training.

\section{Experimental setup}

We use the standard $300$h Switchboard corpus of conversational English speech. Speaker-independent $40$-dimensional log-Mel spectra features are computed with the addition of $\Delta$s+$\Delta\Delta$s and frame stacking with rate of $2$. This results in $240$-dimensional input acoustic features. We explore three different vocabulary sizes during training of $4$k, $10$k, and $20$k that correspond to minimum word occurrence counts in the training set of $25$, $6$, and $2$, respectively. Out-of-vocabulary (OOV) extension experiments rescore with a $34$k word vocabulary.

For training the AWE and AGWE models, character sequences are composed of symbols from a $35$-character vocabulary, including $26$ English letters, $6$ punctuation symbols ({[]}$<>$-'), and $3$ sounds ({[NOISE]}, {[VOCALIZED-NOISE]},{[LAUGHTER]}). Words containing digits are spelled out using this vocabulary (e.g. ``7-11" is spelled ``SEVEN-ELEVEN"). Unspoken parts of partial words are in {[]} with partial starts and ends denoted by $<$ and $>$ (e.g. ``$<$[YO]UR" and ``YO[UR]$>$"), respectively. Acoustic word segment boundaries are obtained from word-level forced alignments produced by a competitive ASR system for Switchboard. When conducting word embedding training, word segments shorter than $6$ frames and words outside the training vocabulary are omitted from the embedding loss computation. During pre-processing, if these restrictions omit all words from an utterance, then the utterance itself is removed. The same set of utterances is then used for training both the word embeddings and the recognizer.

\subsection{Acoustically grounded word embeddings}

The acoustic view, or AWE, model is composed of a $6$-layer BiLSTM with $512$ hidden units (per direction per layer), which is initialized with a BiLSTM trained as a phone CTC recognizer. The BiLSTM takes full utterances as input, but to produce a word embedding for a given word segment within an utterance, the hidden state outputs corresponding to frames in that segment are averaged to produce a single $1024$-dimensional vector. The character view, or AGWE, model includes an initial character embedding layer that maps each of the $35$ characters to a $64$-dimensional vector, followed by a $1$-layer BiLSTM with $512$ hidden units (per direction). A projection layer is used to transform the $1024$-dimensional vectors output by the a character view models to $256$-dimensional embeddings. Unlike CTC, the loss is not calculated for each utterance in isolation. Since the multi-view objective (Equation~\ref{ch:ctc_a2w:eq:multiview}) involves sampling the $k$ most offending examples with respect to each acoustic segment and each character sequence, distributing across 4 GPUs limits samples to those present in only $16$ utterances. However, we find that distributing speeds up training without degradation in discriminative performance. We start with $k=15$ negatives and reduce to $k=5$ over the first $300$ mini-batches. 
As in He {\it et al.}~\cite{he2017multiview}, we evaluate the quality of our embeddings using a cross-view word discrimination task (see Section~\ref{ch:back:prelims:crossview_ap}) applied to the development set.

\subsection{Acoustic-to-word recognition model}

The A2W model consists of a $6$-layer BiLSTM ($512$ hidden units per direction per layer), a $256$-dimensional linear projection layer, and a prediction layer with dimension given by the vocabulary size. Dropout ($p=0.25$) is used between BiLSTM layers. The baseline A2W system uses a phone CTC initialization for the BiLSTM, while feed-forward layers are randomly initialized as in~\cite{audhkhasi2018a2w}. During A2W training, development set WER is used to measure performance.

\begin{table}[!htbp]
\centering
\begin{tabular}{@{}lc@{}}
    \toprule
    \multicolumn{1}{c}{Initialization} &
    \multicolumn{1}{c}{SWB development WER}\\
    \midrule
	Phone CTC init & 17.5\\
	Phone CTC init + AGWE init & 17.6\\
	AWE init + AGWE init & 16.7\\
    \bottomrule
\end{tabular}
\caption{SWB development (held-out) word error rate (WER) for different methods of initializing the A2W model. All use a $10$k word vocabulary.}
\label{ch:ctc_a2w:tab:init}
\end{table}

For word embedding integration experiments, the A2W backbone encoder is initialized with the AWE model's BiLSTM and the $256$-dimensional projection layer from multi-view training. The prediction layer is then initialized with unit-normalized AGWEs for each word in the vocabulary (with the exception of $<$BLANK$>$ and $<$UNK$>$, which are randomly initialized). We find that using the AWE model to initialize the backbone acoustic encoder is essential to see improvement over the baseline system when incorporating AGWE initialization of the prediction layer (Table~\ref{ch:ctc_a2w:tab:init}).

Regularization experiments penalize deviation ($L_2$ distance) of the prediction layer weights from the AGWE initialization. This penalty is only applied with respect to those words present in a given mini-batch. The hyperparameter $\lambda \in \{0.1, 0.25, 0.5, 0.75, 0.9, 0.99\}$ is tuned to manage the trade-off between the CTC objective and this penalty (Equation~\ref{ch:ctc_a2w:eq:multitask}). At the extreme end of regularization, we experiment with freezing the prediction layer after initialization, including the randomly initialized $<$BLANK$>$ and $<$UNK$>$ tokens. By freezing the prediction layer throughout training, we can acquire new embeddings for any OOV words at test time by running their character sequences through the AGWE model. We conduct experiments with OOV prediction by concatenating these new word embeddings to the prediction layer, and rescoring with an extended vocabulary whenever $<$UNK$>$ is predicted.

\subsection{Training}

A batch size of $64$ utterances is used, split across 4 GPUs, for both multi-view AWE+AGWE pretraining as well as A2W recognition training. The same learning rate reduction scheme is used for both training setups, which is as follows: if the held-out performance fails to improve after 4 epochs, the learning rate is decayed by a factor of $10$ and the model is reset to the previous best. The AGWE model is trained using the Adam optimizer~\cite{kingma2014adam} with an initial learning rate of $0.0005$ with additional parameters $\beta_1 = 0.9$, $\beta_2 = 0.999$, and $\epsilon = 10^{-8}$. We train the A2W recognition model using stochastic gradient descent (SGD) with Nesterov momentum~\cite{nesterov1983method}, and we use an initial learning rate of $0.02$ and momentum of $0.9$. Training stops when the held-out performance stops improving and the learning rate is $< 10^{-8}$. All experiments were conducted using the PyTorch toolkit~\cite{pytorch}.

\begin{table}[!htbp]
\centering
\begin{tabular}{@{}ccccc@{}}
\toprule
\multicolumn{1}{c}{Vocab} &
\multicolumn{1}{c}{AGWE} &
\multicolumn{3}{c}{CTC}
\\
\cmidrule{3-5}	& 		  & Baseline	& Initialized	& Regularized\\
\midrule
4K  			      & 0.89	& 0.49		  & 0.72 			  & 0.76 		\\
10K 			      & 0.88	& 0.28		  & 0.64			  & 0.73		\\
20K 			      & 0.86	& 0.16 		  & 0.63			  & 0.60		\\
\bottomrule
\end{tabular}
\caption{Cross-view word discrimination performance measured via average precision (AP). We compare the discriminative quality of AGWEs learned by multi-view pretraining versus the word embeddings (i.e. rows of the prediction layer weight matrix) implicitly learned through CTC training alone.}
\label{ch:ctc_a2w:tab:crossview_ap}
\end{table}

\section{Results}
\label{ssec:results}
In Table~\ref{ch:ctc_a2w:tab:crossview_ap}, we compare the quality of acoustically grounded word embeddings (AGWE) trained explicitly using the multi-view objective against CTC-based embeddings given by the prediction-layer weights after CTC training. The significantly better AP of AGWE shows that they capture discriminative information that is not discovered implicitly by CTC training. By using these pretrained embeddings to initialize our model and regularize CTC training, the prediction layer is better able to retain this word discrimination ability.

\begin{table}[!htbp]
\centering
\begin{tabular}{@{}lccc@{}}
\toprule
\multicolumn{1}{c}{System} & \multicolumn{3}{c}{Vocab}\\
\cmidrule{2-4}	& $4$K 				& $10$K 				& $20$K\\
\midrule
Baseline		& 16.4/25.7 			& 14.8/24.9 			& 14.7/24.3\\
Initialized 	& 15.6/{\bf 25.3} 	& 14.2/{\bf 24.2} 	& 13.8/24.0\\
Regularized 	& 15.5/25.4			& {\bf 14.0}/24.5	& {\bf 13.7/23.8}\\
Frozen			& 15.6/25.6			& 14.6/24.7			& 14.2/24.7\\
\indent+OOV rescoring	& {\bf 15.0/25.3} 	& 14.4/24.5			& 14.2/24.7\\
\midrule
Curriculum~\cite{yu2018multistage} & - & - & 13.4/24.2\\
Curriculum+Joint CTC/CE~\cite{yu2018multistage} & - & - & 13.0/23.4\\
\bottomrule
\end{tabular}
\caption{Results (\% WER) on the SWB/CH evaluation sets.  The best result for each data set and each vocabulary size is boldfaced.}
\label{ch:ctc_a2w:tab:eval}
\end{table}

Table~\ref{ch:ctc_a2w:tab:eval} shows evaluation results on the Switchboard (SWB) and CallHome (CH) test sets. We find that both embedding initialization and regularization improve over the baseline WERs.  For the regularized model, the values of $\lambda$ (tuned on held-out data) are $0.25$, $0.25$, and $0.5$ for the $4$K, $10$K, and $20$K vocabularies, respectively.  However, all values of $\lambda$ yield improvements on the held-out set over the baseline.

Although outperformed by the embedding-initialized and regularized models, the model trained with a frozen prediction layer (``Frozen'') also consistently outperforms the baseline (with the exception of the $20$K CH evaluation). As discussed in Section~\ref{ch:ctc_a2w:sec:approach}, the Frozen model also allows for straightforward vocabulary extension at test-time. Whenever the $<$UNK$>$ token is predicted, we replace $<$UNK$>$ with a prediction made from the added word-level vocabulary. Table~\ref{ch:ctc_a2w:tab:eval} shows that rescoring using the Frozen model to extend the vocabulary improves performance considerably when using the smallest vocabulary. For the $4$K vocabulary recognizer, this approach results in an absolute WER reduction of $1.4\%$ over the baseline model. We also note that the relative improvement seen by adding OOV rescoring to the $10$K vocabulary Frozen model is similar to that offered by the spell-and-recognize system in~\cite{audhkhasi2018a2w} without the need for additional training.

Recent work~\cite{yu2018multistage} shows strong results using a multi-stage A2W approach including curriculum learning from the $10$K to the $20$K vocabulary, joint CTC/cross entropy (CE) training, and data augmentation through speed perturbation. In Table~\ref{ch:ctc_a2w:tab:eval}, we report results from their most comparable setups, i.e. {\it curriculum} and {\it curriculum+joint CTC/CE}~\cite{yu2018multistage}. Future work may improve further upon these results by combining embedding regularization with the curriculum learning and augmentation techniques from~\cite{yu2018multistage}.

\begin{table}[H]
\footnotesize\centering
    \begin{tabular}{@{}l@{}}
        \toprule
        \textbf{REF:} some REMINDERS for me as we are talking\\
        \textbf{HYP (1st pass):} some $<$UNK$>$ for me as we are talking\\
        \textbf{HYP (rescoring):} some \textbf{REMINDERS} for me as we are talking\\
        \midrule
        \textbf{REF:}   fair and speedy TRIAL\\
        \textbf{HYP (1st pass):}   fair and speedy $<$UNK$>$\\
        \textbf{HYP (rescoring):}   fair and speedy \textbf{TRIAL}\\
        \midrule
		\textbf{REF:} but those LOANS ARE so much cheaper\\
		\textbf{HYP (1st pass):} 	but those $<$UNK$>$ so much cheaper\\
        		\textbf{HYP (rescoring):} but those \textbf{LOANER} so much cheaper\\
		\midrule
		\textbf{REF:} one particular CAREGIVER AND then that one\\
        \textbf{HYP (1st pass):} one particular CARE $<$UNK$>$ then that one\\
		\textbf{HYP (rescoring):} one particular CARE \textbf{GIVER} then that one\\
		\midrule
		\textbf{REF:} bring two CANTEENS  just to make sure\\
        \textbf{HYP (1st pass):} bring two $<$UNK$>$ just to make sure\\
		\textbf{HYP (rescoring):} bring two \textbf{CAMPING'S} just to make sure\\
        \bottomrule
    \end{tabular}
	\caption{Successes and failures in OOV prediction with the Frozen+OOV rescoring model.}
\label{ch:ctc_a2w:tab:analysis}
\end{table}

Inspecting outputs from the Frozen+OOV rescoring model, we find that when the A2W system produces an $<$UNK$>$ prediction in place of a single word, we often accurately recover the correct word within the top hypotheses, as seen in the first two rows of Table~\ref{ch:ctc_a2w:tab:analysis}. The majority of remaining mistakes correspond to the first-pass model predicting $<$UNK$>$ in place of multiple words or part of a word. In such cases, we cannot recover the correct word, but we find that many predictions are reasonable phonetic matches for the ground truth. For example, in the third example in Table~\ref{ch:ctc_a2w:tab:analysis}, the first-pass model combines two words ``LOANS ARE'' into a single $<$UNK$>$, and the rescoring model produces a close phonetic match, ``LOANER''. Other typical examples include splitting up compound words such as ``CAREGIVER" and words that are outside the extended $34$k vocabulary such as ``CANTEENS".

\section{Conclusion}

We have introduced techniques for using pretrained AWE and AGWE models for improving A2W CTC-based speech recognition models. We find that consistent performance improvements are obtained when incorporating embeddings through initialization, regularization, and out-of-vocabulary test-time prediction. By regularizing the recognizer prediction layer toward the embeddings, we obtain $0.8$-$1\%$ and $0.4$-$0.5\%$ absolute WER improvements for the Switchboard and CallHome data sets, respectively, and if we also rescore with an expanded vocabulary to resolve OOVs, then in the small-vocabulary ($4$k-word) case we can improve the WER by a total of $1.4\%$ absolute on Switchboard.

There are a number of interesting directions for future work, many of which we describe later in this thesis. These include tighter integration of the embedding and recognizer training through the development of a whole-word segmental modeling approach to ASR (Chapter~\ref{ch:seg_a2w}) as well as investigation of joint word embedding and recognition training (Chapter~\ref{ch:joint_a2w}). Another promising direction, considering our results with small vocabulary sizes, is to apply these ideas to the recognition of low-resource languages, which we include at the end of Chapter~\ref{ch:joint_a2w}.

%% file: text/6_multi_awe.tex
\chapter{Multilingual joint training of acoustic and acoustically grounded word embeddings}
\label{ch:multi_awe}

\begin{figure}[H]
  \centering
  \includegraphics[width=0.65\linewidth]{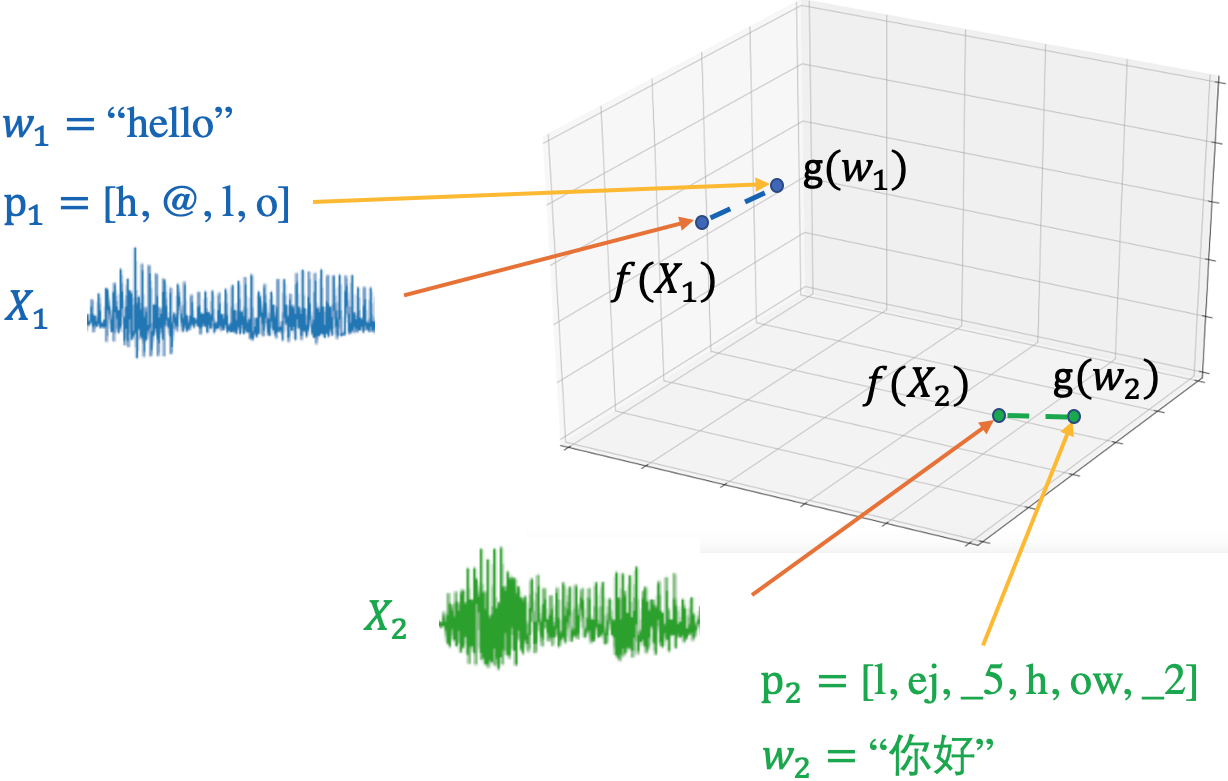}
  \caption{Speech segments and written words are mapped into a multilingual embedding space by AWE model $f$ and AGWE model $g$, respectively.}
\label{ch:multi_awe:fig:multilingual_agwe_grid}
\end{figure}

Previously in Chapter~\ref{ch:ctc_a2w}, acoustic word embeddings (AWEs)---vector representations of spoken word segments---are learned jointly with embeddings of character sequences, to generate acoustically grounded representations of written words, i.e. acoustically grounded word embeddings (AGWEs). Now, we extend these ideas to multilingual joint training of an AWE model and an AGWE model using phonetically transcribed data from a variety of low- and zero-resource languages. These pretrained models can be applied to unseen zero-resource target languages, or fine-tuned on data from low-resource target languages. In this chapter, we describe two contributions. The first (1) is improved joint training and modeling of AWEs and AGWEs by exploring different discrete linguistic units as input to the AGWE model, i.e. characters vs. phones vs. distinctive features. These benefits are seen both in acoustic and cross-view word discrimination results. The second (2) is an extension of AWE+AGWE methods to multilingual training and evaluation where our experiments cover a variety of languages, supervision settings, and resource levels. These contributions are published in Hu {\it et al.}~\cite{hu2020multilingual}.

\subsubsubsection{Collaboration} This work relies on collaboration with Yushi Hu, who performs the majority of the experiments with the processed data, code, and guidance provided by myself.

\section{Introduction}

\begin{figure}[H]
  \centering
  \includegraphics[width=0.7\linewidth]{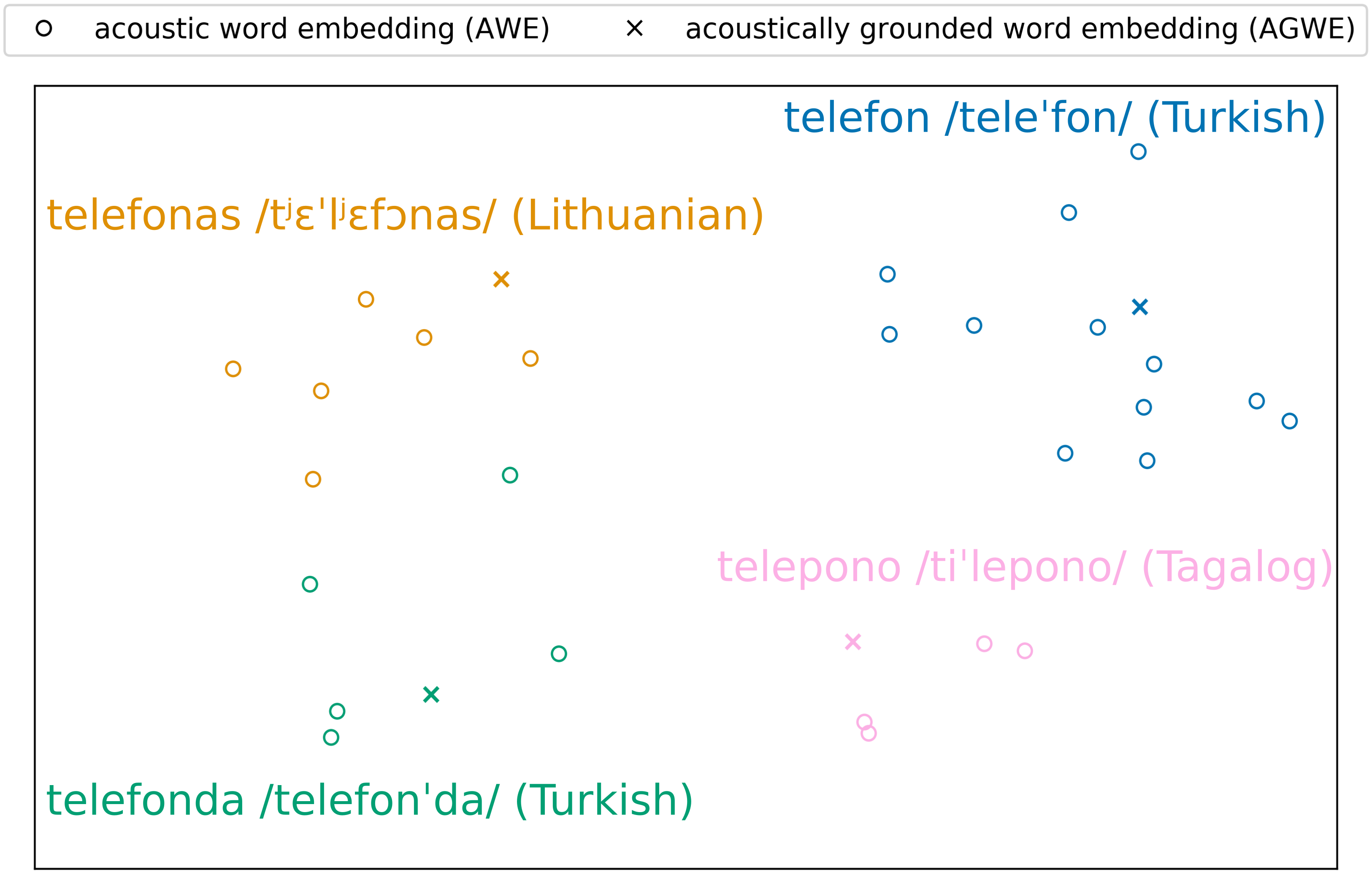}
  \caption{Examples of jointly trained acoustic and written word embeddings for the Turkish word ``telefon" and its nearest neighbors, visualized with t-SNE~\cite{maaten2008visualizing}.}
\label{ch:multi_awe:fig:telephone_emb_cb_cropped}
\end{figure}

Acoustic word embeddings (AWEs) are an attractive tool when reasoning about whole word segments, as they provide a compact representation of spoken words and can be used to efficiently measure similarity between segments (Section~\ref{ch:back:prelims:awes}). Some tasks involve measuring similarity between spoken and written words. For such purposes it can be useful to jointly learn both embeddings of spoken words and embeddings of written words that represent the word's phonetic content, which we refer to as {\it acoustically grounded word embeddings} (AGWEs) (Section~\ref{ch:back:related:agwes}). For example, AWEs and AGWEs have been used together for spoken term detection~\cite{audhkhasi2017asr_free_kws}, where a written query word can be compared to speech segments directly via vector similarity between their embeddings; and to whole-word speech recognition, where such embeddings have been used either for rescoring~\cite{bengio2014word} or to initialize an end-to-end neural model (Chapter~\ref{ch:ctc_a2w}). While AWEs and AGWEs can be learned jointly~\cite{bengio2014word,he2017multiview} to embed spoken and written words in the same vector space (Figure~\ref{ch:multi_awe:fig:telephone_emb_cb_cropped}), prior work largely focuses on English.

In this work, we study joint learning of AWEs and AGWEs for multiple languages simultaneously, in particular low-resource languages. Recent related work~\cite{kamper2020multilingual} has begun to explore multilingual AWEs, specifically for zero-resource languages, including unsupervised approaches trained on a zero-resource language of interest and supervised models trained on multiple additional languages and applied to a zero-resource language. Our work complements this prior work by exploring, in addition to the zero-resource regime, a number of low-resource settings, and the trade-off between performance and data availability. In addition, we learn not only AWEs but also AGWEs, thus widening the range of tasks to which our models apply.

Our approach is based on joint embedding training with a multi-view contrastive loss~\cite{he2017multiview,settle2019_a2w}; however, we use phonetic pronunciations as supervision, rather than characters, to avoid the difficulties introduced by multiple written alphabets. Then, to better model rare or unseen phones, we further explore using distinctive features as an alternative to phone labels.

We find that in low-resource settings where target language data is limited, pretraining embedding models first on multiple non-target languages and then fine-tuning on a small amount of available target language data produces much higher-quality embeddings than training on target language data alone. In addition, word discrimination performance is improved when using the phone sequences of written words, rather than character sequences as in prior work. Further benefits can be seen when using distinctive features in place of phones, but only when there are many phones in the target language that are rare or unseen in the 
languages used for pretraining.

\section{Approach}

\begin{figure}
    \centering
    \includegraphics[width=0.85\linewidth]{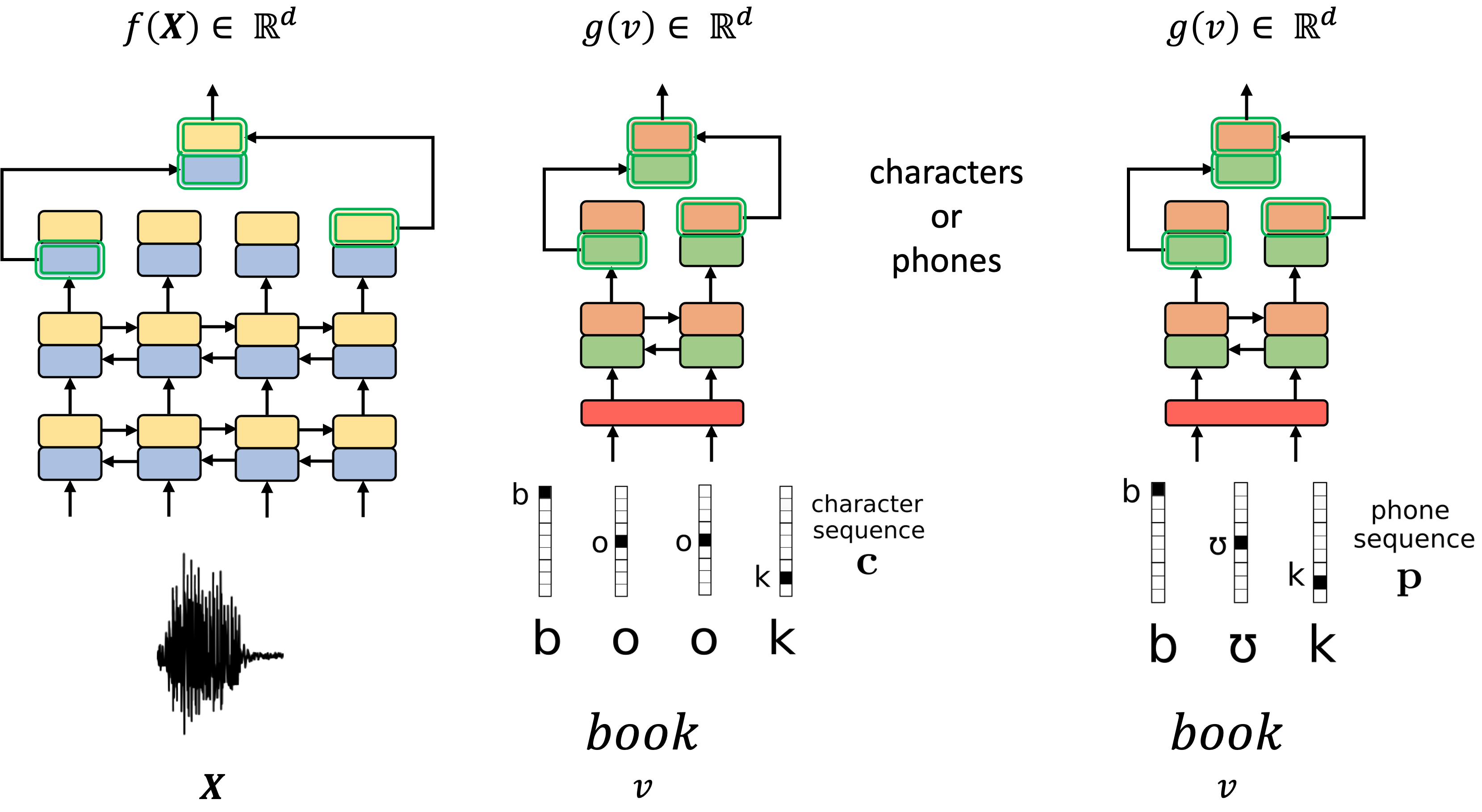}
    \caption{Acoustic word embedding (AWE) model $f$ and example acoustically grounded word embedding (AGWE) models $g$ with either character- or phone-based inputs.}
    \label{ch:multi_awe:fig:multiview_chars_phones_arch}
\end{figure}

Following from Chapter~\ref{ch:ctc_a2w} and prior work~\cite{he2017multiview,settle2019_a2w}, our approach jointly learns an AWE model $f$ and an AGWE model $g$ (Figure~\ref{ch:multi_awe:fig:multiview_chars_phones_arch}). The input to $f$ is again a variable-length spoken segment corresponding to a word; however, the input to $g$ is now a sequence of phones (rather than characters) to allow multilingual training using languages with widely differing writing systems. Specifically, we use the Extended Speech Assessment Methods Phonetic Alphabet (X-SAMPA)~\cite{wells1995xsampa}. We obtain the phonetic transcription from the written one using a pronunciation dictionary. Therefore, we do not require manual phonetic transcriptions, but do require a pronunciation dictionary for each language. Once a multilingual model is trained, however, the {\it acoustic} word embedding model $f$ can embed input spoken word segments coming from any language, seen or unseen during training, regardless of whether or not we have a pronunciation dictionary for it.

\begin{figure}
\centering
\includegraphics[width=0.85\linewidth]{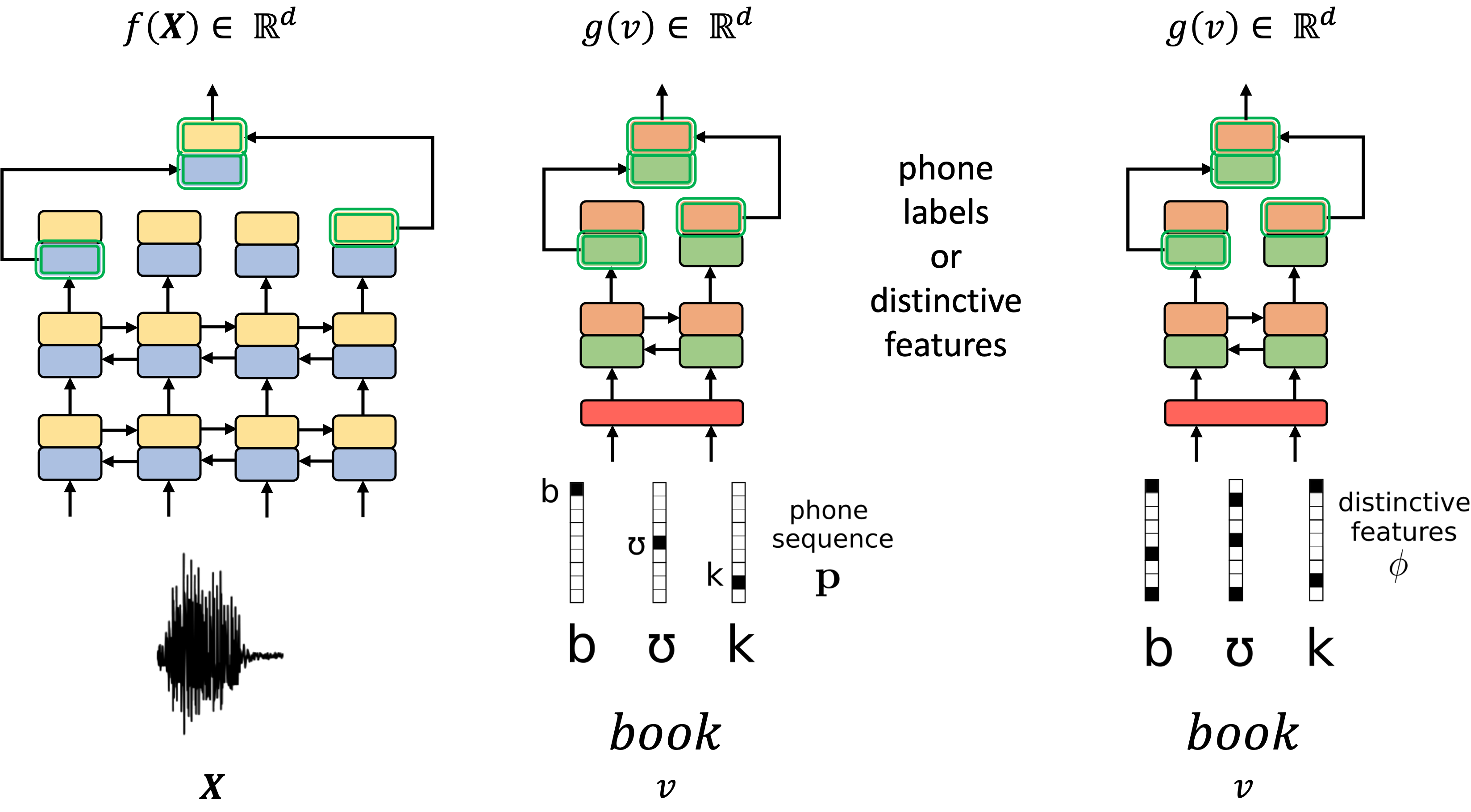}
\caption{Acoustic word embedding (AWE) model $f$ and example acoustically grounded word embedding (AGWE) models $g$ with phone-based inputs using either phone labels directly or their distinctive features.}
\label{ch:multi_awe:fig:multiview_phones_features_arch}
\end{figure}

To further improve cross-lingual transfer and make the best use of multilingual data, we also explore using sequences of distinctive features in place of phone labels. While the X-SAMPA phone set improves over characters, it is far from perfect. Approximately $60\%$ of phones in our $255$-phone set appear in only one of the twelve languages used in our experiments, which means that input embeddings of unseen phones are not learned beyond random initialization. To address this issue, we investigate using distinctive features, such as manner and place features (see PHOIBLE~\cite{phoible}). In this setting, instead of learning input embeddings of our phone labels directly, we map each phone $p_j$ to a vector of its distinctive features ${\bm \phi}_j$ containing $1$ in dimensions corresponding to features that are ``on" and $0$ for features that are ``off" (Figure~\ref{ch:multi_awe:fig:multiview_phones_features_arch}). We then pass this sequence of binary vectors through a linear layer which outputs vectors with the same dimensionality as the phone embeddings in our phone-based model. This step can be viewed as computing phone embeddings as a sum of distinctive feature embeddings. While many phones are language-specific, distinctive features allow us to share even more cross-lingual information since most feature values are used across many languages with the exceptions being click features in Zulu and tone features in Cantonese and Lithuanian.

\subsection{Training details}

The AWE model $f$ is a stacked bidirectional gated recurrent unit (BiGRU) network~\cite{chung2014gru}.~\footnote{While previous related work has used LSTMs~\cite{he2017multiview,settle2019_a2w}, in initial experiments on English we found better performance with GRUs on acoustic word discrimination.} The AGWE model $g$ of the written view consists of a learned phonetic embedding layer followed by a single-layer BiGRU network. For each view, we concatenate the last time step outputs of the top layer in both directions of the BiGRU as the fixed-dimensional embedding. Similar to Equation~\ref{ch:ctc_a2w:eq:multiview} from Chapter~\ref{ch:ctc_a2w}, we train $f$ and $g$ jointly to minimize the following objective $\mathcal{L}_{emb}$. One important distinction, however, is that training is done using {\it isolated} spoken word segments, since our goal is word discrimination, rather than the {\it contextualized} spoken word segments within longer utterances as in Equation~\ref{ch:ctc_a2w:eq:multiview} where our end goal is pretraining for ASR evaluated using utterance-level word error rate (WER).
\begin{flalign}
\mathcal{L}_{emb}(\tens{X}, \tens{V}) = \sum_{n=1}^{N} \mathcal{L}_0(\mats{X}^{(n)}, v^{(n)}) + \mathcal{L}_2(\mats{X}^{(n)}, v^{(n)})
\label{ch:multi_awe:eq:multiview}
\end{flalign}
\noindent where $(\tens{X}, \tens{V})$ is a set of isolated spoken word segment and word label pairs and each individual loss term in $\mathcal{L}_{emb}$ is defined
\begin{flalign*}
\mathcal{L}_0(\mats{X}, v) &= \left(\frac{1}{\vert \mathcal{N}_0(\mats{X}, v) \vert} \sum_{v' \in \mathcal{N}_0(\mats{X}, v)} \left[m + d(f(\mats{X}), g(v)) - d(f(\mats{X}), g(v'))\right]_+ \right)^{0.5}\\
\mathcal{L}_2(\mats{X}, v) &= \left(\frac{1}{\vert \mathcal{N}_2(\mats{X}, v) \vert} \sum_{\mats{X}' \in \mathcal{N}_2(\mats{X}, v)} \left[m + d(f(\mats{X}), g(v)) - d(g(v), f(\mats{X}'))\right]_+ \right)^{0.5}
\end{flalign*}
\noindent where $d(\vecs{a}, \vecs{b}) = 1-\frac{\vecs{a}\cdot \vecs{b}}{\Vert \vecs{a} \Vert\Vert \vecs{b} \Vert}$, $m$ is a margin hyperparameter, and the sets $\mathcal{N}_0(\mats{X}, v)$ and  $\mathcal{N}_2(\mats{X}, v)$ used for negative sampling within objectives $\mathcal{L}_0$ and $\mathcal{L}_2$, respectively, are defined
\begin{flalign*}
\mathcal{N}_0(\mats{X}, v) &:= \left\{
    v'\ \middle\vert
    \begin{array}{c}
    v' \in \mathcal{V} / v
    \end{array}
\right\}\\
\mathcal{N}_2(\mats{X}, v) &:= \left\{
    \mats{X}'\ \middle\vert
    \begin{array}{c}
    \textsc{label}(\mats{X}') \in \mathcal{V} / v
    \end{array}
\right\}
\end{flalign*}
\noindent where $\mathcal{V}$ the training vocabulary. For efficiency, each mini-batch only corresponds to examples of a single language, and negative sampling is performed only over the mini-batch. This means that in Equation~\ref{ch:multi_awe:eq:multiview} $N$ is the number of spoken word segments in a batch and the vocabulary $\mathcal{V}$ used to create the negative sample sets consists only of the (monolingual) unique word labels in the batch. Additionally, we limit the size of the negative sampling sets to just the $k$ within $\mathcal{N}_0(\mats{X}, v)$ or $\mathcal{N}_2(\mats{X}, v)$ that are closest to the acoustic word embedding $f(\mats{X})$ or acoustically grounded word embedding $g(v)$, respectively. 
\begin{flalign*}
\mathcal{N}_0^k(\mats{X}, v) &:= \left\{
    v_1', \dots, v_k'\ \middle\vert
    \begin{array}{c}
        v_i' = \displaystyle \argmin_{v' \in \mathcal{N}_0(\mats{X}, v)} d(f(\mats{X}), g(v'))
    \end{array}
\right\}\\
\mathcal{N}_2^k(\mats{X}, v) &:= \left\{
    \mats{X}_1', \dots, \mats{X}_k'\ \middle\vert
    \begin{array}{c}
        \mats{X}_i' = \displaystyle \argmin_{\mats{X}' \in \mathcal{N}_2(\mats{X}, v)} d(g(v), f(\mats{X}'))
    \end{array}
\right\}
\end{flalign*}

\section{Experimental setup}

We use conversational data from 12 languages, including English data (10 hrs) from a subset of Switchboard~\cite{godfrey1992switchboard} and 11 languages from the IARPA Babel project~\cite{babel_data}: Cantonese (31 hrs), Assamese (19 hrs), Bengali (21 hrs), Pashto (28 hrs), Turkish (32 hrs), Tagalog (30 hrs), Tamil (23 hrs), Zulu (23 hrs), Lithuanian (15 hrs), Guarani (15 hrs), and Igbo (12 hrs). The development and test sets are about $1$-$3$ hours per language. We use Kaldi~\cite{povey2011kaldi} (Babel recipe \texttt{s5d}) to compute input acoustic features, train HMM/GMM triphone models, and extract word alignments for the Babel languages. The acoustic features are $117$-dimensional, consisting of $36$-dimensional log-Mel spectra + $3$-dimensional (Kaldi default) pitch features. Training, development, and test sets are constructed from non-overlapping conversation sides. Babel training sets include segments with duration $25$--$500$ frames corresponding to words occurring $\geq3$ times in training; development and test sets include segments of $50$--$500$ frames with no word frequency restrictions. English development and test sets are the same as in prior related work~\cite{carlin2011rapid_eval_spoken_term_detect,levin2013fixed,kamper2015unsupervised_weak_topdown,kamper2016cnn_awe,settle2016rnn_awe,he2017multiview}. The English training set contains the same conversation sides as in prior work; however, for consistency within multilingual experiments we use word segments satisfying the same duration and frequency restrictions as for the Babel language training sets, which does not exactly match the prior AWE work. To compare with prior work on English (Table~\ref{ch:multi_awe:tab:baselines}), we train a separate set of models on the same training set as in prior work. For distinctive feature-based experiments, features for a given phone are retrieved from the PHOIBLE database~\cite{phoible}.

We consider the following experimental settings: {\bf single} (train and test on the target language), {\bf unseen} (train on the 11 non-target languages, and then test on the unseen target language), and {\bf fine-tune} (train on the 11 non-target languages, and then fine-tune and test on the target language). We vary the amount of training data for {\bf single} and {\bf fine-tune} experiments among $10$min, $60$min, and ``all" (the entire training set for that target language). The {\bf unseen} setting is a zero-resource setting. To tune hyperparameters in \textbf{unseen} experiments, an average evaluation score from the development sets of the 11 training languages is used such that the target language is not seen until test evaluation.

\subsection{Evaluation}

AWEs and AGWEs can be used for a variety of downstream tasks. Here we use a task-agnostic evaluation approach similar to prior work~\cite{jansen2013jhu,carlin2011rapid_eval_spoken_term_detect,levin2013fixed,kamper2016cnn_awe,settle2016rnn_awe,he2017multiview}, including two ``proxy" tasks: acoustic word discrimination and cross-view word discrimination. Acoustic word discrimination is the task of determining whether a pair of acoustic segments $(\mats{X}_i, \mats{X}_j)$ correspond to the same word, while cross-view word discrimination is the task of determining whether an acoustic segment and word label $(\mats{X}_i, w_j)$ correspond to the same word. Further details of these tasks are given in Section~\ref{ch:back:prelims:proxies}. We refer to the performance measures as ``acoustic AP" for acoustic word discrimination, and ``cross-view AP" for the cross-view task. In addition, we measure how well our embeddings capture phonetic similarity between words. Given a pair of speech segments, we calculate the distance between their acoustic embeddings and a phonetic distance~\cite{jyothi-livescu-2014-revisiting} between their phone sequences. We report the Spearman's rank correlation between the embedding distances and phonetic distances for all speech segment pairs in a given test set.

\subsection{Hyperparameters}

Hyperparameters are tuned on the small Switchboard subset from prior work~\cite{levin2013fixed,kamper2015unsupervised_weak_topdown,settle2016rnn_awe,he2017multiview,kamper2016cnn_awe}, then the same model architecture is used for all languages. The AWE model is a $4$-layer BiGRU (with $0.4$ dropout rate between layers), while the AGWE model consists of an input embedding layer and a 1-layer BiGRU. Both recurrent models use 512 hidden units per direction per layer and output 1024-dimensional embeddings. When encoding sequences of either phones or phonetic features in the AGWE model, the embedding layer maps input representations to 64-dimensional vectors as depicted in Figure~\ref{ch:multi_awe:fig:multiview_phones_features_arch}. There are $255$ X-SAMPA phones and $38$ distinctive features. Some distinctive features can take on more than 2 values, so we represent each value of each distinctive feature separately, giving $101$ learned feature embeddings.

During training, the margin $m$ in Equation~\ref{ch:multi_awe:eq:multiview} is set to $0.4$, and negative sampling sets are of size $k=20$. We perform mini-batch optimization with Adam~\cite{kingma2014adam}, with batch size $256$ and initial learning rate $0.0005$. The learning rate is decayed by a factor of $10$ if the cross-view AP on the development set(s) fails to improve over 5 epochs. Training stops when the learning rate drops below $10^{-8}$. All experiments use the PyTorch toolkit~\cite{pytorch}.

\section{Results}

\subsection{Comparison with prior work on English}

\begin{table}[H]
  \centering
  \small
  \caption{Test set performance of several embedding approaches on the English acoustic and cross-view word discrimination tasks. The numbers reported are average precision (AP).}
  \begin{tabular}{lcc}
    \toprule
    \textbf{Method}      & \textbf{Acoustic}  &\textbf{Cross-view}\\
    \midrule
    \textbf{100-minute training set}\\
    $\,\,$ MFCCs + DTW \cite{kamper2016cnn_awe}                  & 0.21\\
    $\,\,$ CAE + DTW \cite{kamper2015unsupervised_weak_topdown}    &  0.47\\
    $\,\,$ Phone posteriors + DTW \cite{carlin2011rapid_eval_spoken_term_detect}      &  0.50\\
    $\,\,$ Siamese CNN \cite{kamper2016cnn_awe}&   0.55\\
    $\,\,$ Supervised CAE-RNN~\cite{kamper2020multilingual}& 0.58\\
    $\,\,$ Siamese LSTM  \cite{settle2016rnn_awe}  &   0.67\\
    $\,\,$ Multi-view LSTM~\cite{he2017multiview}~\tablefootnote{\cite{he2017multiview} calculates cross-view AP differently and is not comparable.}  &   0.81\\
    $\,\,$ Our multi-view GRU (chars)        &    0.81 & 0.71\\
    $\,\,$ Our multi-view GRU (phones)        & \textbf{0.84}   & \textbf{0.77}\\
    $\,\,$ Our multi-view GRU (features)  &  \textbf{0.83} &  \textbf{0.76}\\
    \midrule
    \textbf{10-hour training set} \\
    $\,\,$ Our multi-view GRU (phones)   &  \textbf{0.88}  & \textbf{0.81}\\
   $\,\,$  Our multi-view GRU (features)  & \textbf{0.87} & \textbf{0.81}\\
    \midrule
    \textbf{135-hour training set} \\
    $\,\,$ Our multi-view GRU (phones)  & \textbf{0.89} & \textbf{0.86}\\
    $\,\,$ Our multi-view GRU (features) & \textbf{0.89} & \textbf{0.86}\\
    \bottomrule
  \end{tabular}
\label{ch:multi_awe:tab:baselines}
\end{table}

A number of previously reported results on AWEs have used a particular small $100$ minute subset of Switchboard for training~\cite{levin2013fixed,kamper2015unsupervised_weak_topdown,settle2016rnn_awe,he2017multiview,kamper2016cnn_awe}. To compare with this work, we also train using this same subset. From Table~\ref{ch:multi_awe:tab:baselines}, we find our models outperform all previous methods, including (our implementation of) CAE-RNN from recent work on multilingual AWEs~\cite{kamper2020multilingual}.

We also find that representing the written word as a phone sequence improves over the character-based input representation used in prior work. For the remaining experiments, we use our multi-view GRU-based models with phonetic representations of written words. Between the two phonetic representations (phone-based and feature-based), results are almost identical, with a slight edge for the phone-based representation on acoustic word discrimination. However, for multilingual experiments, we will largely use the feature-based representation as it allows us to embed previously unseen phones, and improves multilingual cross-view performance (see Section~\ref{ch:multi_awe:results:phone_vs_df}).

Finally, we compare results of our models trained on different training set sizes. We find the acoustic AP plateaus at the $10$-hour training set, with the results of the model trained on only $100$ minutes being not far behind.

\subsection{Evaluation of multilingual acoustic word embeddings}

\begin{figure}[H]
\centering
\includegraphics[width=0.75\linewidth]{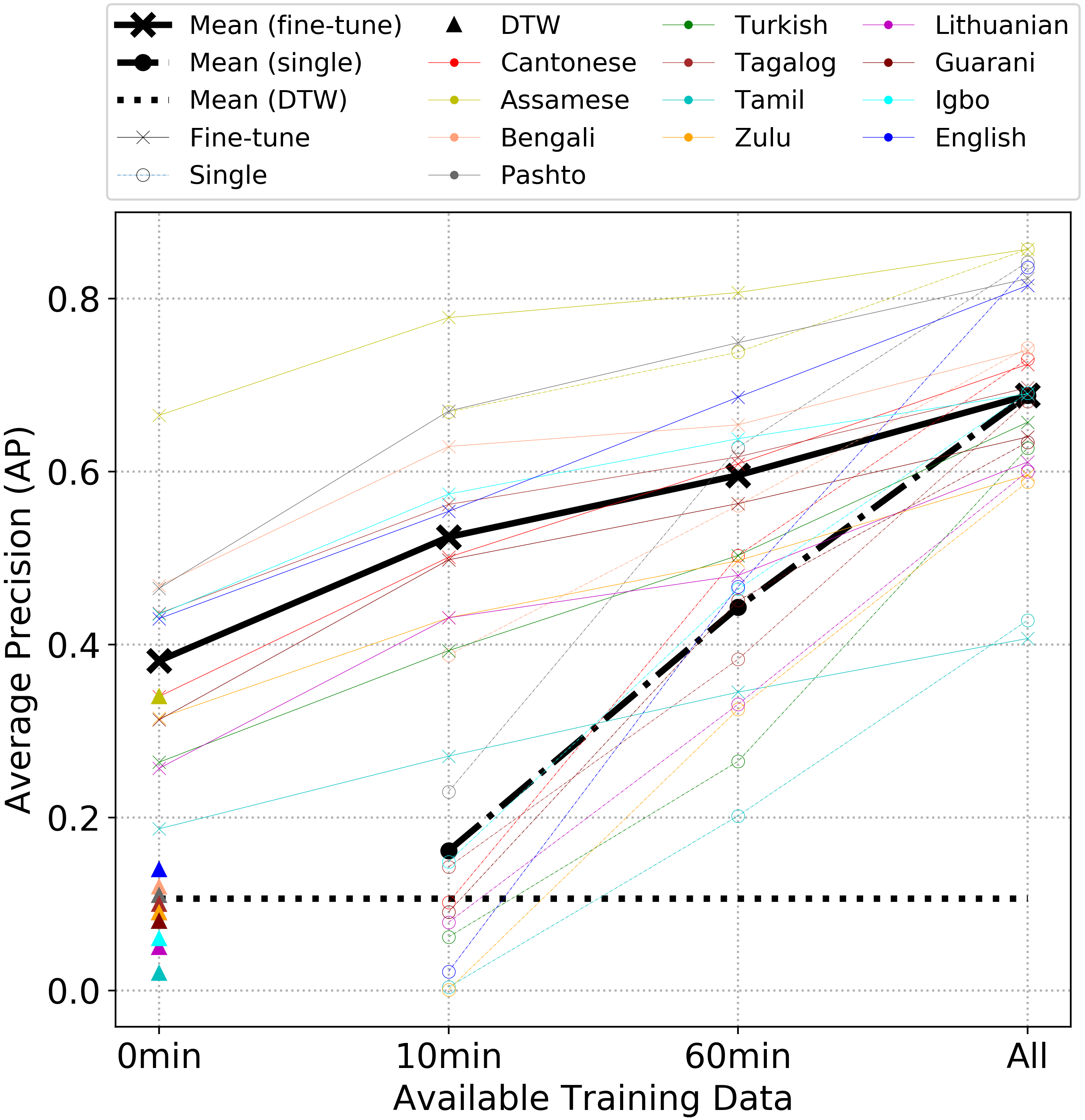}
\caption{Test set acoustic AP for models trained with distinctive feature inputs on varying amounts of target language data. Note that the x-axis is not linear. ``0 min" refers to the {\bf unseen} training setting, i.e.~the {\bf fine-tune} setting with no target language training data. ``All" refers to all training data available for the target language. Mean acoustic APs across languages are given by thick black lines; language-specific results are given by thin colored lines.}
\label{ch:multi_awe:fig:acoustic_ap_h_cropped}
\end{figure}

Figure~\ref{ch:multi_awe:fig:acoustic_ap_h_cropped} gives our main acoustic AP results for distinctive feature-based models across the 12 languages in the three training settings.~\footnote{In terms of acoustic AP, there is little difference between the phone-based and distinctive feature-based models.} These results indicate that, when resources are limited in the target language, multilingual pretraining offers clear benefits. Fine-tuning a multilingual model on 10 minutes of target language data can outperform training on 60 minutes from the target language alone. Furthermore, if 60 minutes of data is available in the target language, multilingual pretraining cuts the performance gap between training on just that 60 minutes alone and the full $10$-$30$ hour training set by more than half. The {\bf unseen} model (equivalent to {\bf fine-tune} with 0 minutes of target language data) is a zero-resource model with respect to the target language since it is trained on the other 11 languages and never sees examples from the target language until test evaluation. Our {\bf unseen} models significantly outperform the unsupervised DTW baselines---confirming results of other recent work in the zero-resource setting~\cite{kamper2020multilingual}---as well as the {\bf single}-10min models, while offering performance closer to the {\bf single}-60min models.

\subsection{Phonetic vs.~distinctive feature supervision}
\label{ch:multi_awe:results:phone_vs_df}

\begin{figure}[H]
\centering
\includegraphics[width=0.75\linewidth]{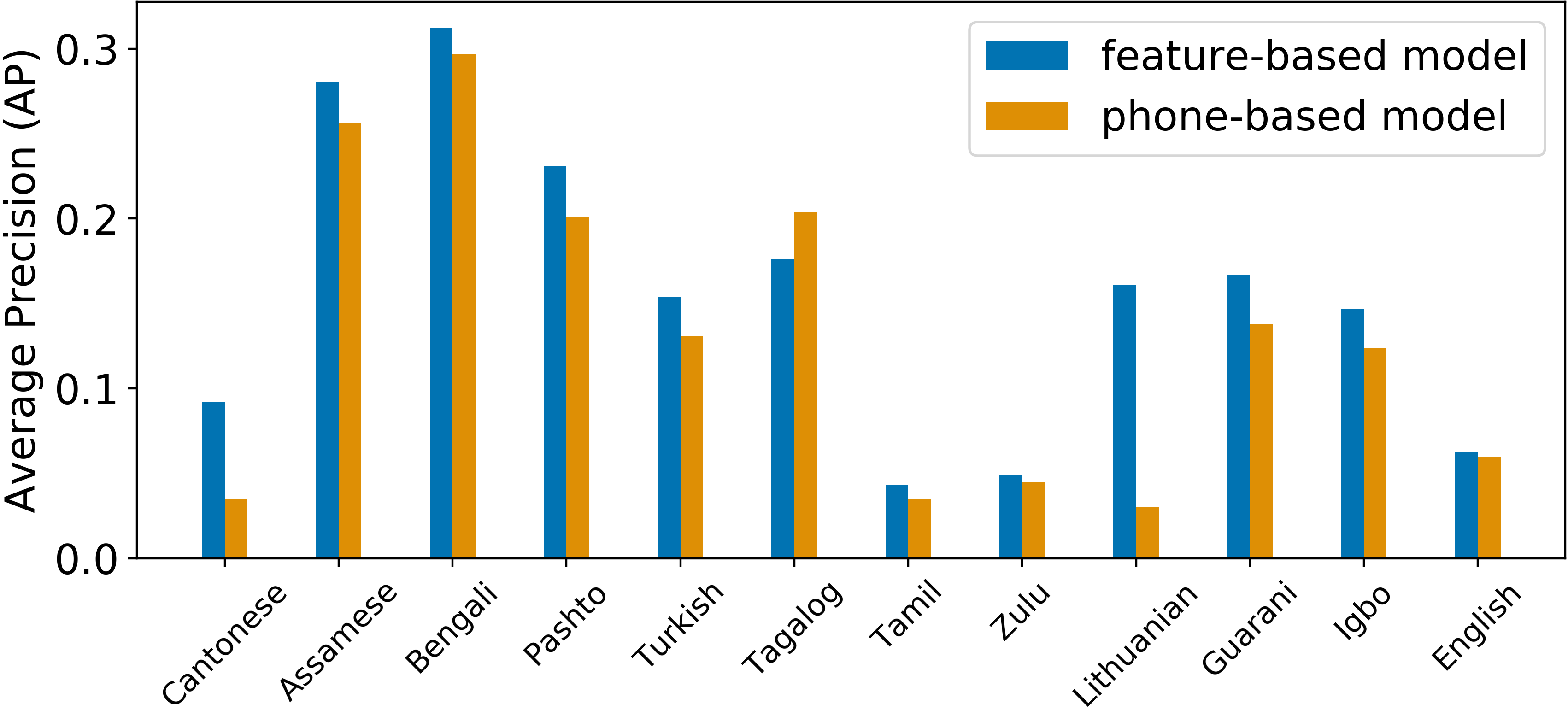}
\caption{Test set cross-view AP in the {\bf unseen} setting.}
\label{ch:multi_awe:fig:cross_ap_unseen_cb_cropped}
\end{figure}

While acoustic AP is largely unaffected by the choice of phone vs.~distinctive feature supervision, Figure~\ref{ch:multi_awe:fig:cross_ap_unseen_cb_cropped} shows that {\bf unseen} models typically benefit in terms of cross-view AP from using distinctive features over phones.

The two languages with the largest improvement from use of distinctive features are Cantonese and Lithuanian. The Cantonese data includes a large number of diphthongs that are unseen in other languages, so their embeddings cannot be learned in the phone-based model, but the features of those diphthongs are shared with phones in other languages. In the Lithuanian data, vowels are paired with their tones, making these phones unique to Lithuanian and again making it impossible to learn the vowel embeddings from other languages using phone-based supervision. All distinctive features, except for the Lithuanian-specific tone features themselves, are shared with other languages, making it easier for the feature-based model to learn good embeddings.

\begin{figure}[H]
\centering
\includegraphics[width=0.65\linewidth]{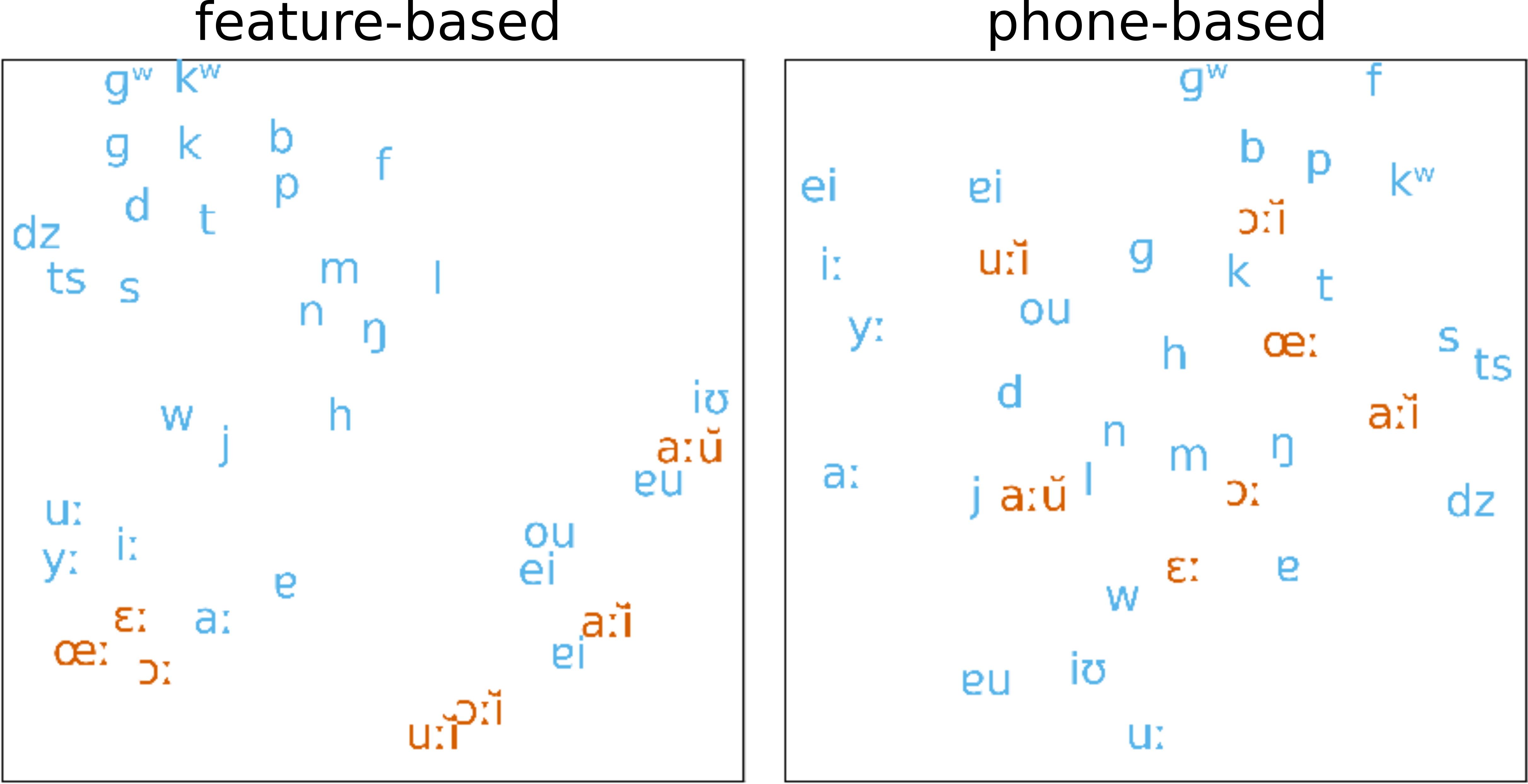}
\caption{t-SNE~\cite{maaten2008visualizing} visualizations of Cantonese phone embeddings from \textbf{unseen} models supervised with distinctive features (left) and phones (right).
\textcolor[rgb]{0,0.5,0.7}{Blue} phones appear in other languages; \textcolor[rgb]{0.87,0.56,0.02}{orange} phones are unique to Cantonese.}
\label{ch:multi_awe:fig:phone_comparison_cropped}
\end{figure}

In addition to generating embeddings of spoken and written words, our written embedding models also include a learned embedding for each phone. Figure~\ref{ch:multi_awe:fig:phone_comparison_cropped} visualizes Cantonese phone embeddings taken from a model trained on the other 11 languages. The model trained using distinctive features is able to infer reasonable embeddings for the phones that are unique to Cantonese and unseen in other languages, placing them near similar phones in the embedding space, while the phone-based model is forced to use (random) initial embeddings.

\subsection{Word Similarity}

\begin{table}[H]
  \centering
  \caption{Word similarity results on English using character, phone, and distinctive feature sequence supervision, given as Spearman's rank correlation $\rho$ between embedding cosine distances and feature-based edit distances computed using the method of~\cite{jyothi-livescu-2014-revisiting}. Specifically, we compute the word edit distances using code available at \texttt{http://www.ifp.illinois.edu/~pjyothi/nbhd/}.}
  \begin{tabular}{lcc}
    \toprule
    Supervision & $\rho$ (AWE) & $\rho$ (AGWE)\\
    \midrule
    Characters & 0.20 & 0.25\\
    Phones & 0.20 & 0.26\\
    Distinctive features & \textbf{0.21} & \textbf{0.28}\\
    \bottomrule
  \end{tabular}
\label{ch:multi_awe:tab:correlation}
\end{table}

As another measure of the quality of the learned embeddings, Table~\ref{ch:multi_awe:tab:correlation} shows the Spearman's rank correlation between embedding distances and a measure of phonetic distance for all word pairs in the English development set. The model trained with distinctive features consistently has slightly higher correlation for both the AWEs and AGWEs compared with models trained using character or phone sequence supervision.

\section{Conclusion}

We have presented a multilingual approach for jointly learning AWE and AGWEs for low- and zero-resource languages. Multilingual pretraining offers significant benefits when we have only a small amount of (or no) labeled training data for the target language. By using distinctive features to encode the pronunciations of written words, we improve cross-lingual transfer by allowing phones unseen during training to share information with similar phones seen in the training set. Chapter~\ref{ch:multi_qbe} will apply our learned embeddings to the downstream task of query-by-example speech search, while expanding evaluation to a larger unseen language set.

%% file: text/7_multi_qbe.tex
\chapter{Acoustic span embeddings for multilingual query-by-example search}
\label{ch:multi_qbe}

Query-by-example (QbE) speech search is the task of matching spoken queries to utterances within a search collection. In low- or zero-resource settings, QbE search is often addressed with approaches based on dynamic time warping (DTW). Recent work has found that methods based on acoustic word embeddings (AWEs) can improve both search performance and speed. However, prior work on AWE-based QbE primarily focuses on English data and single-word queries. In this chapter, we generalize AWE training to spans of words, producing acoustic span embeddings (ASE), and we explore the application of ASE modeling to QbE with arbitrary-length queries in multiple unseen languages. We consider the commonly used setting where we have access to labeled data in other languages (in our case, several low-resource languages) distinct from the unseen test languages. We evaluate our approach on the benchmarks from the query-by-example search on speech task (QUESST) 2015, and find that multilingual ASE-based search is both much faster than DTW-based search and outperforms the best previously published results. These contributions are published in Hu {\it et al.}~\cite{hu2021ase}.

\subsubsubsection{Collaboration} This work relies on collaboration with Yushi Hu, who performs the experiments with the preprocessed data, code framework, and guidance provided by myself.

\section{Introduction}
\label{sec:intro}

The goal of query-by-example (QbE) speech search is to match spoken queries to utterances in a speech database~\cite{parada2009query,szoke2015coping,hazen2009query_posteriorgram_templates}. In low-resource settings, QbE methods often use DTW for audio segment comparison~\cite{hazen2009query_posteriorgram_templates,zhang2009unsupervised_spoken_keyword_spotting,zhang2011piecewise_posteriorgram_dtw,mantena2013speedup_dtw_hierarchical_kmeans}. DTW is a dynamic programming approach to determine the similarity between two audio segments by finding their best frame-level alignment cost~\cite{vintsyuk1968speech,sakoe1978dynamic} (Section~\ref{ch:back:prelims:dtw}), but DTW-based search can be very slow. Many such systems rely on speech activity detection (either energy-based or using a trained phoneme recognizer) as a preprocessing step~\cite{leung2016toward,xu2016approximate,proencca2016segmented} and approximate nearest neighbor indexing methods~\cite{jansen2012rails} to improve search efficiency. Frame-level representations are also important when using DTW~\cite{chen2016unsupervised_bnf_qbe,leung2016toward}, and a great deal of DTW-based QbE work focuses on learning better frame-level acoustic features, including the use of phonetic posteriorgrams~\cite{hazen2009query_posteriorgram_templates,zhang2009unsupervised_spoken_keyword_spotting,lafarga2015elirf,lopez-otero2015gtm-uvigo_qbe} and supervised or unsupervised bottleneck features (BNFs)~\cite{proencca2016segmented,hou2015nni_qbe,leung2016toward,chen2016unsupervised_bnf_qbe}. More challenging query definitions (e.g., approximate matches) and adverse speech conditions (e.g., noisy or reverberant audio), further complicate DTW-based QbE search, motivating data augmentation techniques as well as system ensembling~\cite{leung2016toward,proencca2016segmented,hou2015nni_qbe}. Modifying DTW-based search to cover both exact and approximate query matches well can be challenging, so the best setups on benchmark QbE tasks often fuse many such systems together~\cite{leung2016toward,proencca2016segmented}. Additionally, symbolic systems based on finite-state transducers built from predicted phone sequences have been explored to address approximate matches and improve speed, and these improve performance only when combined with more standard DTW systems~\cite{leung2016toward,xu2016approximate}. 

An alternative to DTW-based search systems is to use acoustic word embeddings (AWE)~\cite{levin2013fixed,kamper2016cnn_awe,chung2016audio,he2017multiview,kamper2019truly_unsupervised_awe,yuan2018learning_awe_for_qbe,holzenberger2018unsupervised_awe}---vector representations of spoken word segments---and compare segments using vector distances between their embeddings. 
A number of AWE approaches have been developed, primarily for low- and zero-resource settings, including both supervised~\cite{kamper2016cnn_awe,settle2016rnn_awe,he2017multiview} and unsupervised~\cite{levin2013fixed,audhkhasi2017asr_free_kws,kamper2019truly_unsupervised_awe,holzenberger2018unsupervised_awe} approaches. AWE-based approaches to QbE search, first demonstrated using template-based AWEs~\cite{levin2015srails} and later using discriminative neural network-based AWEs~\cite{settle2017query} (Chapter~\ref{ch:rnn_qbe}), can greatly speed up search, even over speed-optimized DTW~\cite{jansen2012rails}, while also improving performance. 

Recent work on AWEs has begun to consider the use of multilingual training. For example, multilingual BNFs have been used to improve low-resource AWE models on English~\cite{yuan2016learning_from_bnf,yuan2018learning_awe_for_qbe}. Other work has found that AWEs trained on multilingual data can transfer well to additional zero- or low-resource languages when evaluated on word discrimination~\cite{kamper2020multilingual,hu2020multilingual}. However, to our knowledge no prior work explores embedding-based QbE search in a multilingual setting. 

In this chapter, we extend the multilingual AWEs of~\cite{hu2020multilingual} (Chapter~\ref{ch:multi_awe}) for application to multilingual QbE search. Several benchmark datasets and tasks have been developed for comparing QbE systems, notably the MediaEval series of challenges~\cite{metze2013spoken,anguera2014query,szoke2015query}. We focus on the query-by-example search on speech task (QUESST) from MediaEval 2015~\cite{szoke2015query}, which is the most recent QbE task from the MediaEval series and includes challenging acoustic conditions, several low-resource target languages, and a variety of query settings. Our approach outperforms all prior work on this benchmark, while also being much faster than DTW-based search. In particular, we make two primary contributions. First, we demonstrate how embedding-based QbE can be successfully applied to multiple unseen languages after training on languages with available labeled data. Second, we extend the idea of acoustic word embedding to multi-word spans to better handle queries containing an arbitrary numbers of words.

\section{Approach}

\begin{figure}
  \centering
  \includegraphics[width=0.75\linewidth]{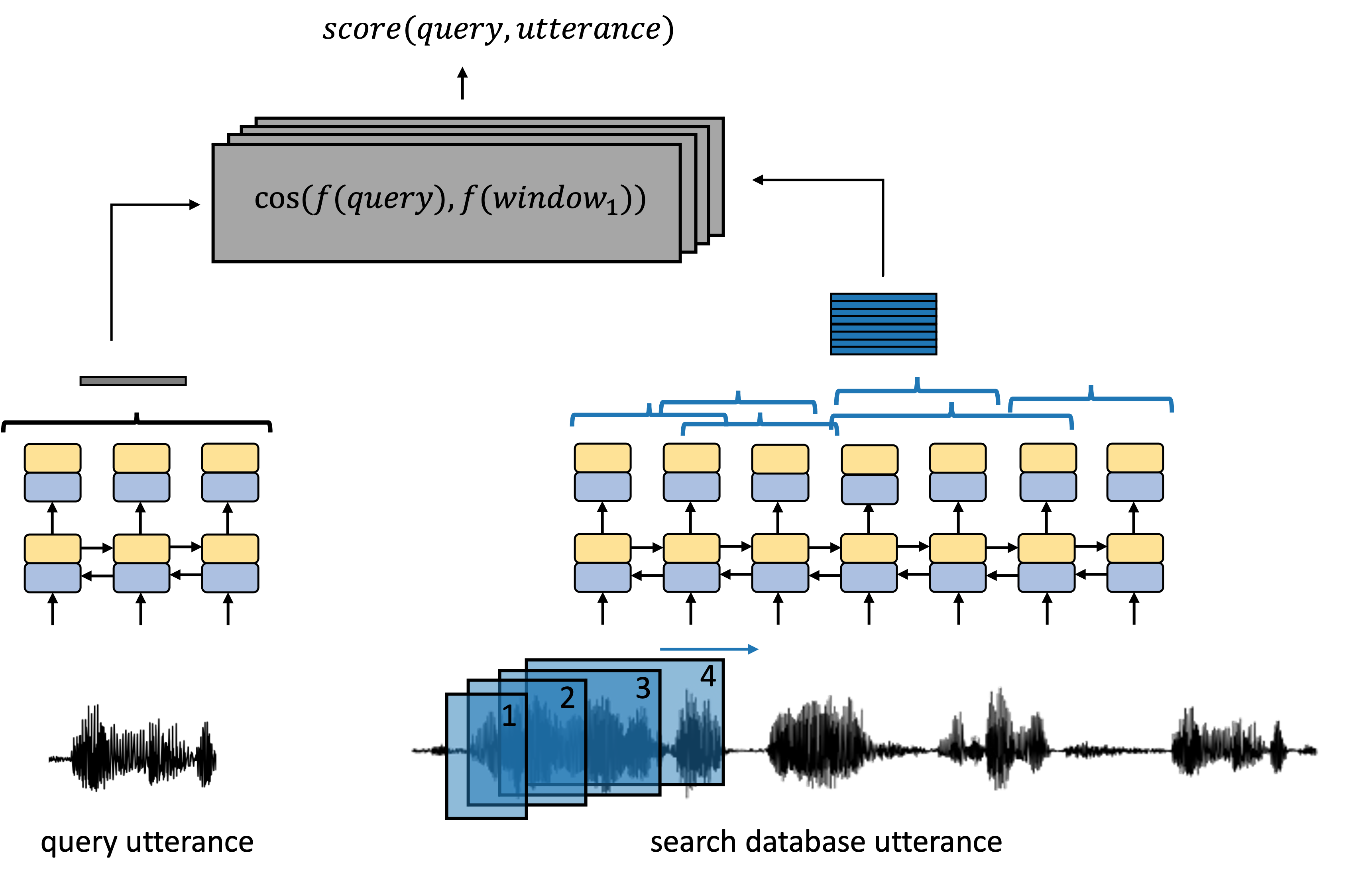}
  \caption{Illustration of the embedding-based QbE search pipeline, which compares query embeddings to embeddings of windowed segments of each utterance in the search collection using cosine distance.}
\label{ch:multi_qbe:fig:qbe_windows}
\end{figure}

Our embedding-based QbE approach, shown in Figure~\ref{ch:multi_qbe:fig:qbe_windows}, consists of a multilingual embedding model and a search component. The embedding model maps segments of speech---both the query and windowed segments in the search collection---to vectors, and the search component finds matches between embedding vectors. The embedding model is trained on data from a set of languages, and can then be used for QbE in any language; here we use low-resource languages to train the embedding model, and perform QbE on a disjoint set of unseen languages. The overall approach is very simple, using no speech activity detection, fine-tuning for the QbE task, or special-purpose components to accommodate approximate matches; we simply use the trained embedding models out of the box, plugging them into the parameter-free search component. 

The following subsections describe each component. For the embedding model, we begin with acoustic word embeddings (AWEs) as has been done in prior work on embedding-based QbE~\cite{levin2015srails,settle2017query,yuan2018learning_awe_for_qbe}. We then extend the approach to accommodate spans of multiple words, resulting in {\it acoustic span embeddings} (ASEs). Finally, we describe the search component.

\subsection{Contextual acoustic word embedding model}
\label{ch:multi_qbe:approach:awe}

We take as our starting point an approach that jointly learns acoustic word embeddings (AWEs) and acoustically grounded word embeddings (AGWEs) using a multi-view contrastive loss~\cite{he2017multiview,settle2019_a2w,hu2020multilingual}. The AWE model sees the acoustic ``view" of the training data (spoken segments) and the AGWE model sees the written ``view" of the data (character or phone sequences). While we do not use the AGWE model for QbE search, we jointly train both models since this approach has produced the highest quality AWE models in prior work as measured by the acoustic word discrimination task~\cite{carlin2011rapid_eval_spoken_term_detect}. We slightly alter the approach from Chapter~\ref{ch:multi_awe} (similarly to Chapter~\ref{ch:ctc_a2w}) to embed spoken word segments within their utterance context. Using contextual AWEs rather than isolated AWEs improves QbE performance~\cite{yuan2018learning_awe_for_qbe}, simplifies our extension to multi-word spans (Section~\ref{ch:multi_qbe:approach:ase}), and allows us to more efficiently embed the search collection and incoming queries at inference time.

When training the AWE model, we are given a training set of $N$ spoken utterances $\tens{X}$ where each utterance $\mats{X} \in \mathbb{R}^{T \times D}$ has a corresponding word-level alignment $\mathcal{A}$ used to extract spoken word segments. A word alignment $\mathcal{A} = \{(s_1, e_1, v_1), ..., (s_L, e_L, v_L)\}$ consists of tuples $(s_i, e_i, v_i)$ indicating start frame, end frame, and word label, respectively, for each spoken word in the utterance.

\begin{figure}
    \centering
    \includegraphics[width=0.85\linewidth]{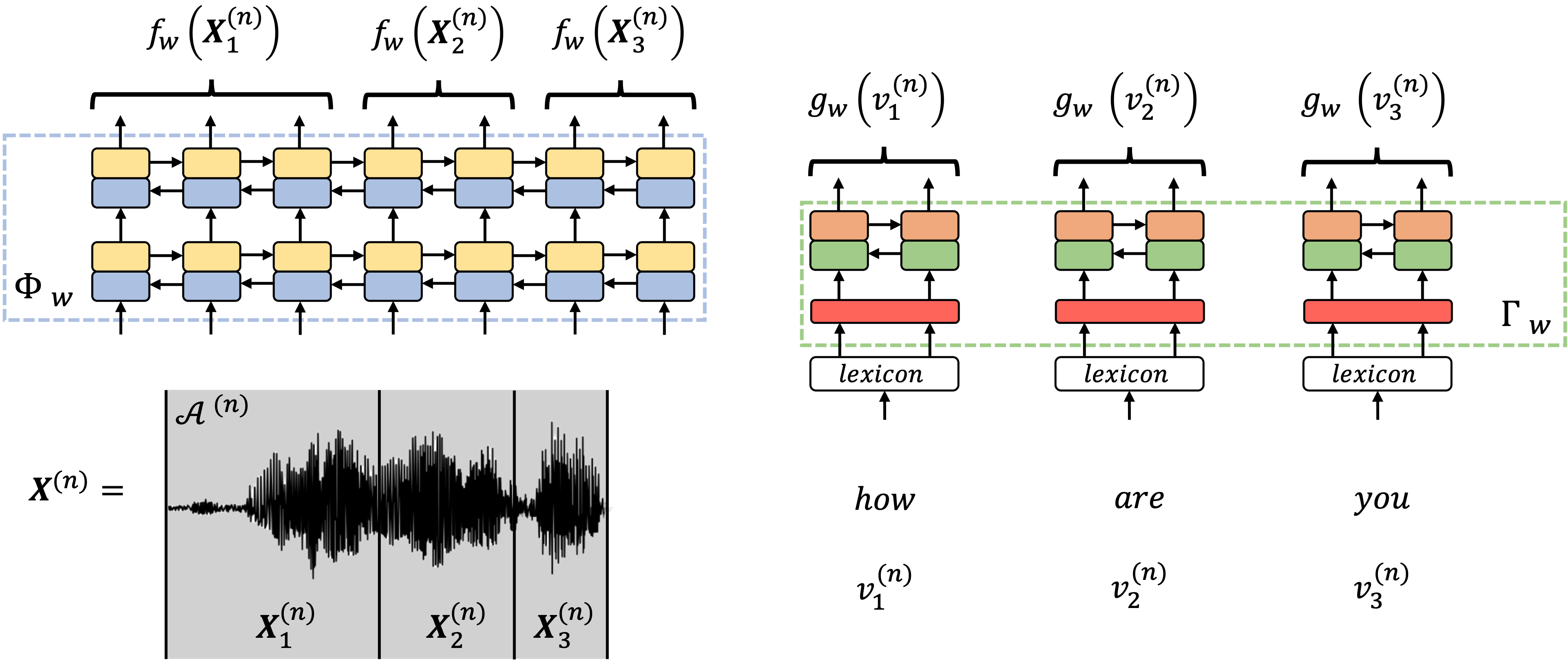}
    \caption{Contextualized acoustic word embedding (AWE) model $f_w$ and acoustically grounded word embedding (AGWE) model $g_w$. $\mats{X}^{(n)}$ is the $n^{th}$ utterance of the batch with word label sequence $(v_1^{(n)}, v_2^{(n)}, v_3^{(n)})$ derived from the corresponding alignment $\mathcal{A}^{(n)}$.}
    \label{ch:multi_qbe:fig:awe}
\end{figure}

Figure~\ref{ch:multi_qbe:fig:awe} outlines the structure of the embedding models. The acoustic-view model $f_w$ consists of an utterance encoder $\Phi_w$ and a pooling function $G$. The embedding of the $i^\textrm{th}$ segment in $\mats{X}$ is given by the encoder outputs pooled over the corresponding segment frames:
\begin{flalign*}
    f(\mats{X}_i) &= f(\mats{X}, \mathcal{A}_i) = G(\Phi_w(\mats{X}), s_i, e_i)
\end{flalign*}
\noindent where $\Phi_w(\mats{X}) \in \mathbb{R}^{T \times d}$, $\mathcal{A}_i = (s_i, e_i, v_i)$, and $G$ is applied over frames $s_i$ to $e_i$. For the encoder $\Phi_w$, we use a bidirectional recurrent network, so output features within $[s_i, e_i]$ also depend on the full utterance context. We consider two pooling functions for $G$, either {\it concatenation} of the starting and ending encoder states or a {\it mean} over all encoder states in the segment:
\begin{flalign*}
    G(\Phi_w(\mats{X}), s_i, e_i) &= \left[ \Phi_w(\mats{X})_{e_i}^{\rightarrow}; \Phi_w(\mats{X})_{s_i}^{\leftarrow} \right]\tag{concat}\\
    G(\Phi_w(\mats{X}), s_i, e_i) &= \frac{1}{e_i - s_i + 1} \sum_{j={s_i}}^{e_i} \Phi_w(\mats{X})_j\tag{mean}
\end{flalign*}
where $[{\bf u}; {\bf v}]$ indicates concatenation of vectors ${\bf u}$ and ${\bf v}$ and, for bidirectional network outputs, $\rightarrow$ and $\leftarrow$ index the final forward and backward hidden states, respectively.

The AGWE model $g_w$ consists of a pronunciation lexicon $\text{Lex}(\cdot)$ mapping words to phones, an encoder $\Gamma_w$, and a pooling function (concatenation)\footnote{Note that we define the AGWE model here to contain the pronunciation lexicon and accept words as input, which is different from its definition in Chapter~\ref{ch:multi_awe}. This distinction is only for notational convenience, and the underlying model structure is unchanged.}. The approach can be relaxed to not require a pronunciation lexicon, using instead the written character sequence, but prior work finds an advantage to phones~\cite{hu2020multilingual}, particularly in the context of multilingual training. As in prior work~\cite{hu2020multilingual,he2017multiview}, the embedding of the word $v_i$ is defined as
\begin{flalign*}
    g(v_i) &= \left[ \Gamma_w(\vecs{p}_i)_l^{\rightarrow}; \Gamma_w(\vecs{p}_i)_1^{\leftarrow} \right]
\end{flalign*}
\noindent where $\vecs{p}_i = \text{Lex}(v_i)$, $l = \text{len}(\vecs{p}_i)$, and $\Gamma_w(\vecs{p}_i) \in \mathbb{R}^{l \times d}$.

During training, we minimize the following objective consisting of three contrastive loss terms (updated slightly from Chapter~\ref{ch:multi_awe} to include $\mathcal{L}_1$ from He {\it et al}~\cite{he2017multiview}):
\begin{flalign}
    \mathcal{L}_{emb}(\tens{X}, \mathbfcal{A}) &= \sum_{n=1}^{N}\sum_{i=1}^{\vert \mathcal{A}^{(n)} \vert} \sum_{obj=0}^{2} \mathcal{L}_{obj}(\mats{X}_i^{(n)}, v_i^{(n)})
    \label{ch:multi_qbe:eq:multiview}
\end{flalign}
\noindent where $\mathcal{A}^{(n)}$ is the word-level alignment for utterance $\mats{X}^{(n)}$ and $\mats{X}^{(n)}_i$ is a spoken word segment with word label $v^{(n)}_i$ extracted using this alignment. Semi-hard negative sampling~\cite{schroff2015facenet} is used for all three terms of the loss:
\begin{flalign*}
\mathcal{L}_0(\mats{X}, v) &= \frac{1}{\vert \mathcal{N}_0(\mats{X}, v) \vert} \sum_{v' \in \mathcal{N}_0(\mats{X}, v)} \left[m + d(f(\mats{X}), g(v)) - d(f(\mats{X}), g(v'))\right]_+ \\
\mathcal{L}_1(\mats{X}, v) &= \frac{1}{\vert \mathcal{N}_1(\mats{X}, v) \vert} \sum_{v' \in \mathcal{N}_1(\mats{X}, v)} \left[m + d(f(\mats{X}), g(v)) - d(g(v), g(v'))\right]_+\\
\mathcal{L}_2(\mats{X}, v) &= \frac{1}{\vert \mathcal{N}_2(\mats{X}, v) \vert} \sum_{\mats{X}' \in \mathcal{N}_2(\mats{X}, v)} \left[m + d(f(\mats{X}), g(v)) - d(g(v), f(\mats{X}'))\right]_+
\end{flalign*}
\noindent where $d(\vecs{a}, \vecs{b}) = 1-\frac{\vecs{a}\cdot \vecs{b}}{\Vert \vecs{a} \Vert\Vert \vecs{b} \Vert}$, $m$ is a margin hyperparameter, and the sets $\mathcal{N}_{obj}(\mats{X}, v)$ used for negative sampling within each corresponding objective $\mathcal{L}_{obj}$ are defined
\begin{flalign*}
\mathcal{N}_0(\mats{X}, v) &:= \left\{
    v'\ \middle\vert
    \begin{array}{c}
    v' \in \mathcal{V} / v, \hspace{0.5cm} d(f(\mats{X}), g(v')) > d(f(\mats{X}), g(v))
    \end{array}
\right\}\\
\mathcal{N}_1(\mats{X}, v) &:= \left\{
    v'\ \middle\vert
    \begin{array}{c}
    v' \in \mathcal{V} / v, \hspace{0.5cm} d(g(v), g(v')) > d(f(\mats{X}), g(v))
    \end{array}
\right\}\\
\mathcal{N}_2(\mats{X}, v) &:= \left\{
    \mats{X}'\ \middle\vert
    \begin{array}{c}
    \textsc{label}(\mats{X}') \in \mathcal{V} / v, \hspace{0.5cm} d(g(v), f(\mats{X}')) > d(f(\mats{X}), g(v))
    \end{array}
\right\}
\end{flalign*}
\noindent where $N$ in Equation~\ref{ch:multi_qbe:eq:multiview} is the number of utterances and $\mathcal{V}$ is the whole training vocabulary. In practice, this objective is applied only over a mini-batch such that $N$ represents the number of utterances in the batch and $\mathcal{V}$ represents the unique words in the batch. Also, we limit the size of the negative sampling sets to just $k$ such embeddings within each $\mathcal{N}_{obj}(\mats{X}, v)$ as follows.
\begin{flalign*}
\mathcal{N}_0^k(\mats{X}, v) &:= \left\{
    v_1', \dots, v_k'\ \middle\vert
    \begin{array}{c}
        v_i' = \displaystyle \argmin_{v' \in \mathcal{N}_0(\mats{X}, v)} d(f(\mats{X}), g(v'))
    \end{array}
\right\}\\
\mathcal{N}_1^k(\mats{X}, v) &:= \left\{
    v_1', \dots, v_k'\ \middle\vert
    \begin{array}{c}
        v_i' = \displaystyle \argmin_{v' \in \mathcal{N}_1(\mats{X}, v)} d(g(v), g(v'))
    \end{array}
\right\}\\
\mathcal{N}_2^k(\mats{X}, v) &:= \left\{
    \mats{X}_1', \dots, \mats{X}_k'\ \middle\vert
    \begin{array}{c}
        \mats{X}_i' = \displaystyle \argmin_{\mats{X}' \in \mathcal{N}_2(\mats{X}, v)} d(g(v), f(\mats{X}'))
    \end{array}
\right\}
\end{flalign*}

\section{Contextual acoustic span embedding (ASE) model}
\label{ch:multi_qbe:approach:ase}

Next, we extend the idea of jointly trained AWE+AGWE to jointly trained acoustic span embeddings (ASE) and acoustically grounded span embeddings (AGSE) to better model spans of multiple words that may appear in queries and search utterances. As before, we train both embedding models jointly, but use only the ASE model for QbE.

The training data for ASE+AGSE learning consists of spans of consecutive word segments, constructed from the word-aligned AWE+AGWE training data by (randomly) merging neighboring word segments. Specifically, given the training utterance $\mats{X}$ with word-level alignment $\mathcal{A}$, we merge any given pair of consecutive word segments $i$ and $j$ with probability $p$, to generate a span-level alignment $\mathcal{A}'$
\begin{flalign*}
    \mathcal{A}' &= \{\ldots, (s_i, e_j, {\bf v}_{i:j}), \ldots\} = \{\ldots, (s_{i'}, e_{i'}, {\bf v}_{i'}), \ldots\}
\end{flalign*}
\noindent where $\vert \mathcal{A}' \vert \leq \vert \mathcal{A} \vert$ and $\mathcal{A}'$ consists of tuples $(s_{i'}, e_{i'}, \vecs{v}_{i'})$ indicating start frame, end frame, and multi-word label sequence, respectively, for each span in the utterance after merging. 

\begin{figure}
    \centering
    \includegraphics[width=0.85\linewidth]{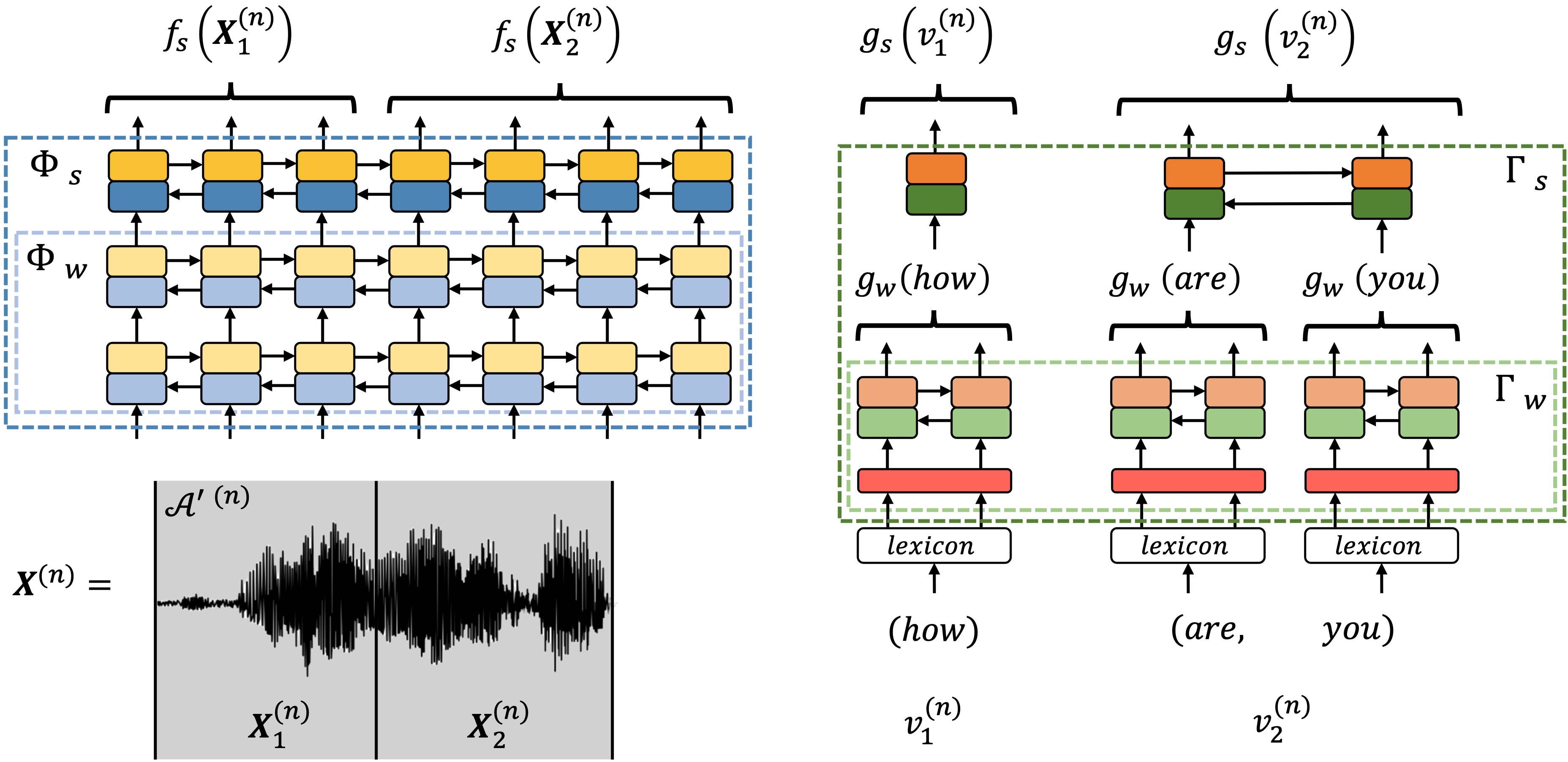}
    \caption{Model architecture illustration of both the acoustic span embedding (ASE) model $f_s$ and the acoustically grounded span embedding (AGSE) model $g_s$. $\mats{X}^{(n)}$ is the $n^{th}$ utterance of the batch and span label sequence $(v_1^{(n)}, v_2^{(n)})$ is derived from the corresponding span-level alignment $\mathcal{A}'^{(n)}$.}
    \label{ch:multi_qbe:fig:ase}
\end{figure}

As shown in Figure~\ref{ch:multi_qbe:fig:ase}, the ASE function $f_s$ has the same form as the AWE function $f_w$, consisting of an encoder $\Phi_s$ and a pooling function $G$:
\begin{flalign*}
    f(\mats{X}_{i'}) &= f(\mats{X}, \mathcal{A}_{i'}) = G(\Phi_s(\mats{X}), s_{i'}, e_{i'})
\end{flalign*}
\noindent where $\Phi_s(\mats{X}) \in \mathbb{R}^{T \times d}$, $\mathcal{A}'_{i'} = (s_{i'}, e_{i'}, \vecs{v}_{i'})$, and $G$ is applied over the window $[s_{i'}, e_{i'}]$.

The written-view model (the acoustically grounded span embedding model) $g_s$ consists of an encoder $\Gamma_s$ and another pooling function (concatenation):
\begin{flalign*}
    g_s(\vecs{v}_{i'}) &= \left[ \Gamma_s(\vecs{v}_{i'})_{l'}^{\rightarrow}; \Gamma_s(\vecs{v}_{i'})_1^{\leftarrow} \right]
\end{flalign*}
\noindent where $\vecs{v}_{i'} = (v_i, \ldots, v_j) ,l' = j - i + 1$, and $\Gamma(\vecs{v}_{i'}) \in \mathbb{R}^{l' \times d}$.

We jointly train $f_s$ and $g_s$ by minimizing the same contrastive loss as in Equation~\ref{ch:multi_qbe:eq:multiview} with modifications to the input alignments:
\begin{flalign}
   \mathcal{L}_{emb}(\tens{X}, \mathbfcal{A}') &= \sum_{n=1}^{N}\sum_{i'=1}^{\vert {\mathcal{A}'}^{(n)} \vert} \sum_{obj=0}^{2} \mathcal{L}_{obj}(\mats{X}_{i'}^{(n)}, {\bf v}_{i'}^{(n)})
    \label{ch:multi_qbe:eq:multiview_span}
\end{flalign}
\noindent where the loss terms are defined as in Section~\ref{ch:multi_qbe:approach:awe} but where $g$ is $g_s$, $f$ is $f_s$, and $\mathcal{V}$ is the set of multi-word spans $\vecs{v}$. ASE+AGSEs can be trained from scratch using this objective, but in this work we first pretrain word-level models $f_w, g_w$ and use them to initialize the lower layers of $\Phi_s, \Gamma_s$.

One potential pitfall of a span-level contrastive loss would seem to be that negative examples may be ``too easy". In preliminary experiments, we explored alternative objective functions taking into account not only same and different span pairs but also the degree of difference between them, but these did not outperform the simple contrastive loss described here.

\subsection{Embedding-based QbE}
\label{ssec:embqbe}

\begin{figure}
\centering
\includegraphics[width=0.85\linewidth]{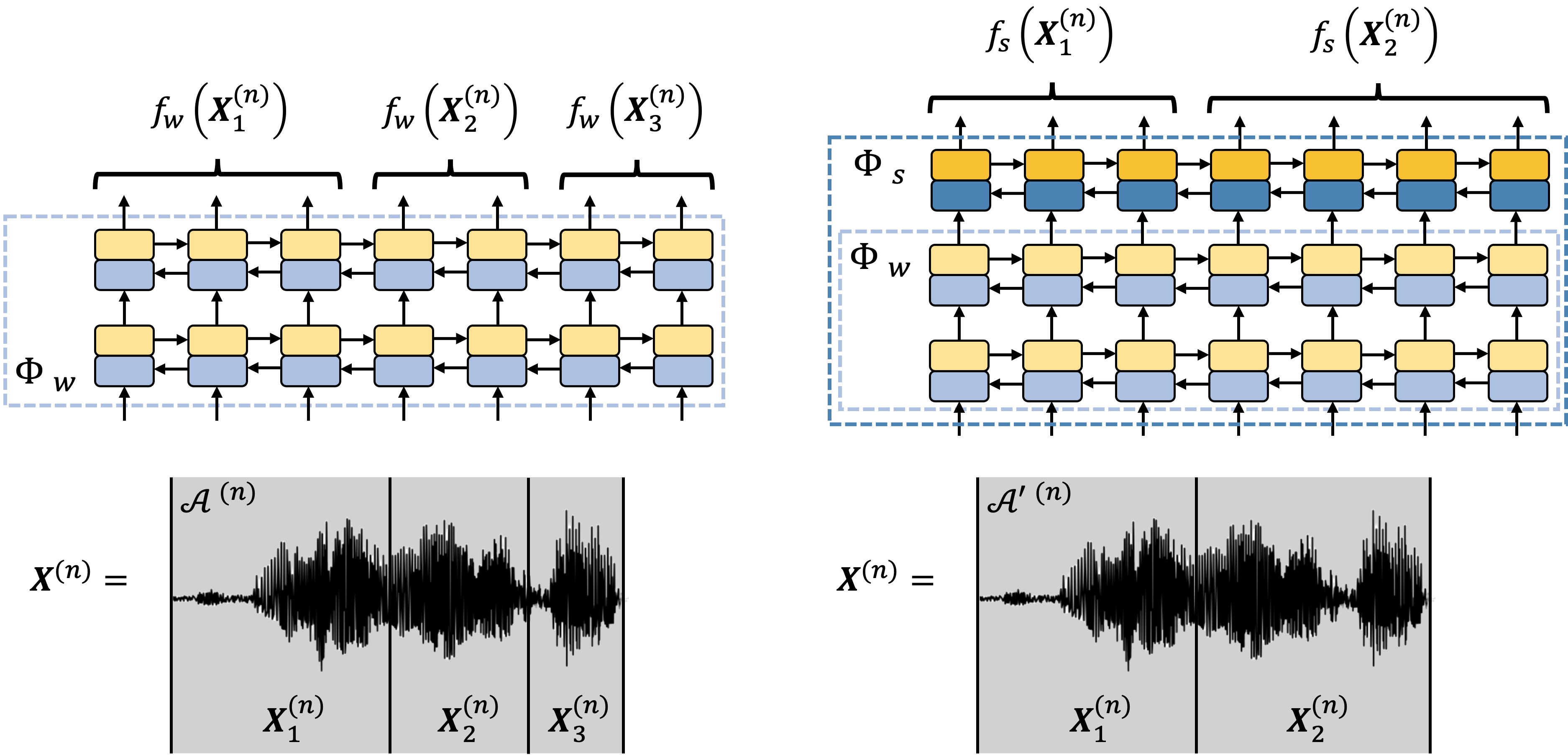}
\caption{Embedding-based QbE search only uses the ``acoustic-view" embedding model, i.e. the AWE model $f_w$ (left) or the ASE model $f_s$ (right).}
\label{ch:multi_qbe:fig:awe_ase_only}
\end{figure}

Given a pretrained acoustic embedding model (either $f_w$ for AWE or $f_s$ for ASE, see Figure~\ref{ch:multi_qbe:fig:awe_ase_only}), we first build an index of utterances in the search collection by embedding all possible segments that meet certain minimal constraints. Similarly to~\cite{yuan2018learning_awe_for_qbe}, we use a simple sliding window algorithm with several window sizes to get possible matching segments. Each windowed segment is then mapped to an embedding vector using one of our AWE or ASE embedding models. Contextual AWE and ASE models simplify this process, since each utterance can be forwarded just once through the encoder network ($\Phi_w$ or $\Phi_s$) and the embedding for each windowed segment in the utterance is then computed by applying the pooling function $G$.

At query time, given an audio query, we embed the query with the same pretrained embedding model, and then compute a detection score for each utterance in the search collection via cosine similarity between the corresponding embeddings:
\begin{equation}
    \textrm{score}(\vecs{q},\mats{S}) = \underset{\vecs{s} \in \Sigma(\vecs{q}, \mats{S})}{\max}{\frac{f(\vecs{s}) \cdot f(\vecs{q})}{\Vert f(\vecs{s}) \Vert_2 \Vert f(\vecs{q}) \Vert_2}}
\label{ch:multi_qbe:eq:score}
\end{equation}
\noindent where $f$ is the acoustic-view embedding model (either AWE or ASE), $\vecs{q}$ is the spoken query segment, $\mats{S}$ is the utterance from the search collection, and $\Sigma(\vecs{q}, \mats{S})$ is the set of all windowed segments $\vecs{s} \in \mats{S}$ of similar length to $\vecs{q}$. We compute this score for all possible segments in the search collection, but it is possible to speed up the search via approximate nearest neighbor search (as in previous work Chapter~\ref{ch:rnn_qbe}). In this work, we are not concerned with obtaining the fastest QbE system possible, so we simply use exhaustive search; but as we will see this exhaustive embedding-based search is still much faster than DTW-based alternatives.

\section{Experimental setup}

\subsection{Embedding models}

We train our embedding models (Figures~\ref{ch:multi_qbe:fig:awe} and~\ref{ch:multi_qbe:fig:ase}) on conversational speech from $12$ languages including $22$ hours of English from Switchboard~\cite{godfrey1992switchboard} and $20$-$120$ hours from each of $11$ languages in the IARPA Babel project~\cite{babel_data}: Cantonese (122 hrs), Assamese (50 hrs), Bengali (51 hrs), Pashto (68 hrs), Turkish (68 hrs), Tagalog (74 hrs), Tamil (56 hrs), Zulu (52 hrs), Lithuanian (33 hrs), Guarani (34 hrs), and Igbo (33 hrs). We use the Kaldi Babel recipe \texttt{s5d}~\cite{povey2011kaldi} to compute acoustic features and extract word alignments for all Babel languages. We use $39$-dimensional acoustic features consisting of $36$-dimensional log-Mel spectra and $3$-dimensional (Kaldi default) pitch features. All other details are the same as in Hu {\it et al.}~\cite{hu2020multilingual} (see Chapter~\ref{ch:multi_awe}). In some experiments, we augment the data with SpecAugment~\cite{park2019specaugment}, using one frequency mask ($m_F=1$) with width $F \sim \mathcal{U}\{0, 9\}$ and one time mask ($m_T=1$) with width $T \sim \mathcal{U}\{0, \frac{t_{min}}{2}\}$, where $t_{min}$ is the length of the shortest word segment in the utterance such that no word is completely masked.

For AWE+AGWE training, the acoustic-view model ($f_w$) encoder $\Phi_w$ is a $4$-layer bidirectional gated recurrent unit (BiGRU~\cite{chung2014gru}) network with dropout rate $0.4$, and the pooling function $G$ is either mean or concatenation. The written-view model ($g_w$) encoder $\Gamma_w$ is a phone embedding layer followed by a $1$-layer BiGRU. For ASE+AGSE training, the acoustic-view model ($f_s$) encoder $\Phi_s$ is a $6$-layer BiGRU network with dropout rate $0.4$. The written-view model ($g_s$) encoder $\Gamma_s$ is a $1$-layer BiGRU on top of an AGWE submodule that has the same structure as $g_w$. Before span-level training, we first train AWE+AGWE models $f_w$ and $g_w$ to be used as initialization. The bottom $4$ layers of $\Phi_s$ are initialized with $\Phi_w$ (from $f_w$) and are kept fixed during training. The bottom word-level submodule in $\Gamma_s$ is also initialized with $g_w$ and is not updated during training either. All recurrent models use $256$ hidden dimensions per direction per layer and generate $512$-dimensional embeddings.

During ASE+AGSE training, we randomly convert each word-level alignment $\mathcal{A}$ (length $L$) into a span-level alignment $\mathcal{A}'$ on-the-fly by modifying the original list of word boundaries $\mathcal{B}_{\mathcal{A}} = \{(e_1, s_2), \ldots, (e_{L-1}, s_{L})\}$. First, we sample how many word boundaries from $\mathcal{B}_{\mathcal{A}}$ to remove ($r \sim \mathcal{U}(\frac{L-1}{2}, L - 1)$), and then we select $r$ boundary tuples from $\mathcal{B}_{\mathcal{A}}$ at random. We remove each tuple $(e_i, s_{i+1})$ by merging segments $i$ and $i+1$ such that $\{\ldots, (s_i, e_i, v_i), (s_{i+1}, e_{i+1}, v_{i+1}), \ldots\}$ becomes $\{\ldots, (s_i, e_{i+1}, {\bf v}_{i:i+1}), \ldots\}$, giving us $\mathcal{A}'$. The margin $m$ in Equation~\ref{ch:multi_qbe:eq:multiview} is $0.4$ and the mini-batch size is approximately $30000$ acoustic frames. We start with sampling $k=64$ negatives and gradually reduce to $k=20$. We use Adam~\cite{kingma2014adam} to perform mini-batch optimization, with an initial learning rate of $0.0005$ and an $L_2$ weight decay parameter of $0.0001$. All other settings are the same as in Chapter~\ref{ch:multi_awe}. Hyperparameters are tuned using cross-view word discrimination performance (Section~\ref{ch:back:prelims:crossview_ap}) on the development sets as in~\cite{hu2020multilingual}.

\subsection{Embedding-based QbE system}

We preprocess each utterance in the search collection with a sliding window approach to generate a set of possible segments, and then we embed the segments using either our AWE or ASE model. Specifically, we consider the set of possible windowed segments to be windows of size $\{12, 15, 18, \ldots, 30, 36, 42, 48, \ldots, 120\}$ frames with shifts of $5$ frames between windows. We embed the query (length $l_q$) using the same acoustic embedding model and compute the cosine similarity between the query embedding and the embeddings of all windowed segments with length between $\frac{2}{3}l_q$ and $\frac{4}{3}l_q$.

\subsection{Evaluation}

We evaluate our models on the query-by-example search on speech task (QUESST) at MediaEval 2015~\cite{szoke2015query}. The task uses a dataset containing speech in $6$ languages (Albanian, Czech, Mandarin, Portuguese, Romanian, and Slovak). There are $18$ hours of test utterances in the search collection, $445$ development queries, and $447$ evaluation queries. The data includes artificially added noise and reverberation with equal amounts of clean, noisy, reverberated, and noisy+reverberated speech. The task includes three sub-task types: ${\bf T1}$ (exact match only), ${\bf T2}$ (allow for word reordering and lexical variations), and ${\bf T3}$ (same as ${\bf T2}$, but for conversational queries in context).

\subsubsection{Evaluation metrics}
\label{ssec:metrics}

We use the official QUESST 2015 grading scripts~\cite{szoke2015query}. The QbE system outputs a score (in our case, a cosine similarity) for each query ($\vecs{q}$) and utterance ($\mats{S}$) pair. A threshold $\theta$ can be set such that if $score(\vecs{q}, \mats{S}) > \theta$, then the utterance $\mats{S}$ is considered a hit for query $\vecs{q}$. The evaluation script varies $\theta$ to compute normalized cross entropy $C_{nxe}$ and term-weighted value $TWV$ as defined in~\cite{metze2013spoken,szoke2015query}. In QUESST 2015, $C_{nxe}$ is considered the primary metric. The final results are reported as $min C_{nxe} = \displaystyle \min_\theta C_{nxe}$ and $max TWV = \displaystyle \max_\theta TWV$, that is the best values achieved for the metrics by varying the threshold $\theta$.

Specifically, the normalized cross entropy $C_{nxe}$ is the ratio between the cross entropy of the QbE system output scores (measured against the binary ground truth) and the cross entropy of random scoring. It ranges between $[0,1]$, and lower $C_{nxe}$ values are better. The term weighted value $TWV$ is based on hard decisions and is defined as $1- (P_{miss}(\theta) + \beta P_{fa}(\theta))$ where $\theta$ is a threshold, $P_{miss}$ is the miss rate, $P_{fa}$ is the false alarm rate, and $\beta$ is a term weighing the cost of misses vs.~false alarms. $TMV$ ranges from $-\beta$ to $1$, and higher values are better. The official script uses $\beta=12.49$.

\section{Results}
\label{sec:results}

\begin{table}
\small
\centering
\caption{QUESST 2015 performance on development set measured by $min C_{nxe}$ and $max TWV$. Training languages are separated into in- and out-of-domain.}
\begin{tabular}{lcrrrccc}
\toprule
{\bf Method} & {\it models} & \multicolumn{2}{c}{{\it langs}} & {\it labeled}               & {\it Augment} & \multicolumn{2}{c}{$min C_{nxe} \downarrow$ / $max TWV \uparrow$} \\
             & \#           & in & out                        & \multicolumn{1}{c}{hours}   &               & \multicolumn{1}{c}{dev} & \multicolumn{1}{c}{eval} \\
\midrule
\underline{Top prior results~\tablefootnote{Prior work uses phone recognizer-based SAD systems.}}&&&&&&&\\
BNF+DTW~\cite{proencca2016segmented}                & 36 & 2 & 4  & $>$384  & noise         & 0.78 / 0.23   & 0.79 / 0.21 \\
BNF+DTW~\cite{hou2015nni_qbe}                       & 66 & 2 & 15 & $>$643  & noise+reverb  & 0.76 / 0.29   & 0.75 / 0.27 \\
Exact match fusion~\cite{leung2016toward}       & 2 & 0 & 2 & 423 & noise+reverb &  0.80 / 0.26 &\\
Partial match+symbolic~\cite{leung2016toward}   & 2 & 0 & 1 & 260 & noise+reverb &  0.78 / 0.23 &\\
Fusion of above two~\cite{leung2016toward}      & 4 & 0 & 2 & 423 & noise+reverb & 0.72 / 0.32 &\\
\midrule
\underline{Our systems}&&&&&&&\\
AWE (concat)                                    & 1 & 0 & 12 & 664 & &  0.85 / 0.08 &\\
AWE (mean)                                      & 1 & 0 & 12 & 664 & &  0.80 / 0.10 &\\
AWE (mean)                                      & 1 & 0 & 12 & 664 & SpecAug &  0.78 / 0.14 &\\
ASE (concat)                                    & 1 & 0 & 12 & 664 & &  0.75 / 0.19 &\\
ASE (mean)                                      & 1 & 0 & 12 & 664 & &  0.73 / 0.24 &\\
ASE (mean)                                      & 1 & 0 & 12 & 664 & SpecAug &  \textbf{0.71} / \textbf{0.26} & \textbf{0.69} / \textbf{0.25}\\
ASE (mean+concat)                               & 2 & 0 & 12 & 664 & SpecAug & \textbf{0.67} / \textbf{0.32} & \textbf{0.66} / \textbf{0.30}\\
\bottomrule
\end{tabular}
\label{ch:multi_qbe:tab:baselines}
\end{table}

\subsection{Comparison with prior work}
\label{ssec:comparison_prior}

Table~\ref{ch:multi_qbe:tab:baselines} compares our work with the top QUESST 2015 submissions as well as the best follow-up work~\cite{leung2016toward}. Both the AWE and ASE models we use here are $6$ layers for a fair comparison.

Many QUESST systems are trained on labelled data (both in- and out-of-domain with respect to the target languages) and use augmentation techniques to limit mismatch with the test data. In addition, speech activity detection (SAD) and system fusion significantly improve performance, and all of the top-performing QUESST systems use both.

Our AWE (mean) model with SpecAugment is competitive with some of the top prior $min C_{nxe}$ results, but there is a clear benefit to span-level embeddings. Although our approach is much simpler, our best single system, ASE (mean) with SpecAugment, outperforms all prior work on the primary metric $min C_{nxe}$. On the $max TWV$ metric, this system matches all but two~\cite{hou2015nni_qbe,leung2016toward} of the fusion models in prior work. To compete with these fusion systems, we create our own simple fusion model by summing the scores output by our two ASE (mean) and ASE (concat) systems (both trained with SpecAugment), which gives us an additional performance boost, matching $max TWV$ of the best prior fusion models and outperforming them in $min C_{nxe}$ by a wide margin.

\subsection{Dependence on query sub-task}

\begin{figure}
 \centering
 \includegraphics[width=0.85\linewidth]{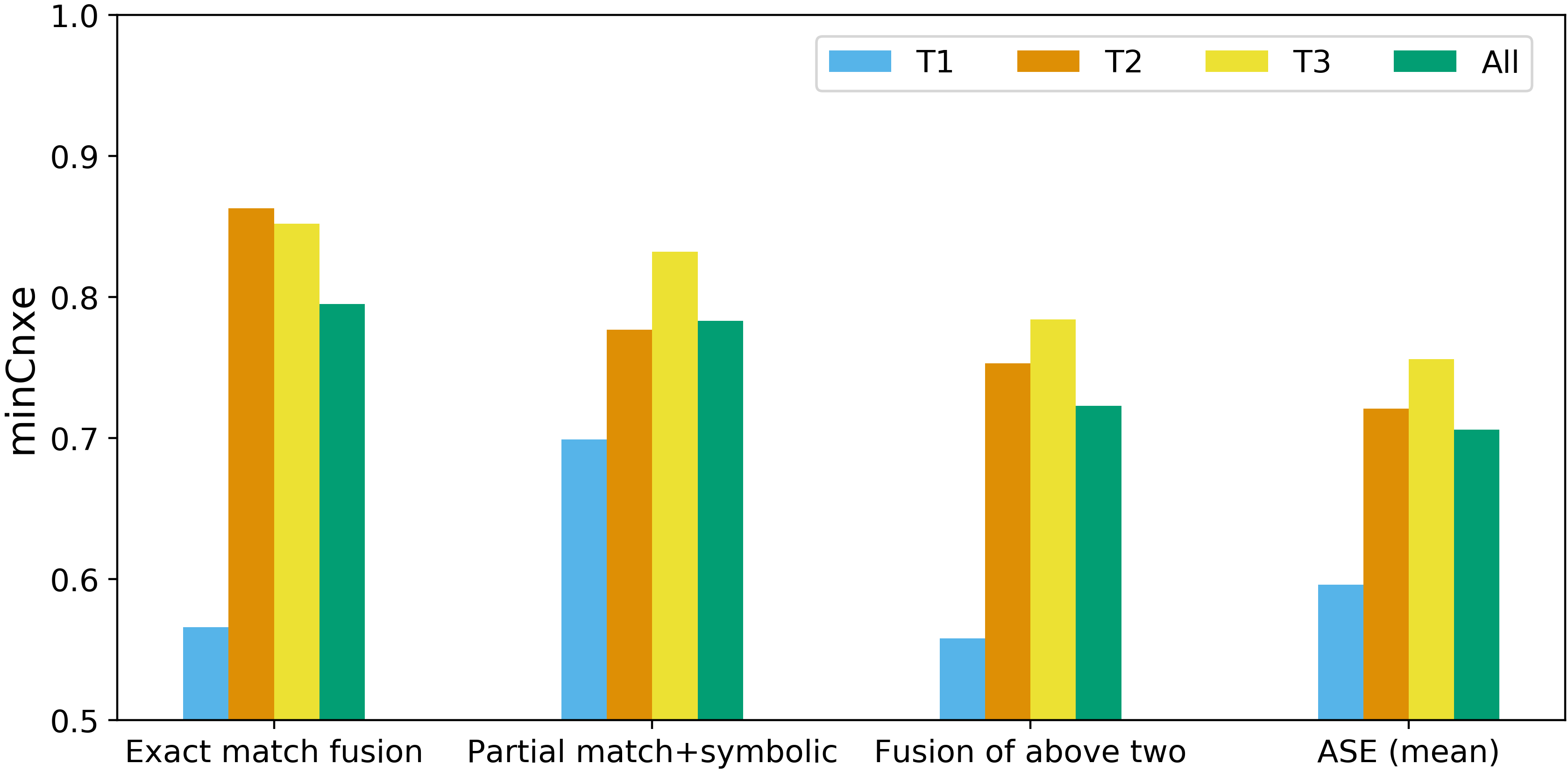}
 \caption{QUESST 2015 sub-task performance measured by development set $min C_{nxe}$ ($\downarrow$).}
\label{ch:multi_qbe:fig:type_of_queries}
\end{figure}

Figure~\ref{ch:multi_qbe:fig:type_of_queries} compares $min C_{nxe}$ performance across the QUESST sub-tasks, for both our best single model, ASE (mean) with SpecAugment, and the best fusion models from prior work~\cite{leung2016toward} seen in Table~\ref{ch:multi_qbe:tab:baselines}. The exact DTW system does best in the exact match sub-task (${\bf T1}$), but cannot generalize well to approximate matches (${\bf T2}$ and ${\bf T3}$). The partial-match system has the opposite behavior. This illustrates the importance of model fusion in DTW-based systems as each individual system can specialize to a particular sub-task. Meanwhile, our best single system is competitive with the others on exact match (${\bf T1}$) and outperforms all of them on approximate match tasks. This suggests that ASE models are better at accommodating lexical variations and word re-ordering than DTW-based systems without any special-purpose partial-match handling, and without sacrificing too much performance on exact matches.

\subsection{Contribution of segment embeddings}

Our approach differs from prior work in a number of respects, including choice of training data and newer neural network techniques. One may ask how much of the performance improvements are due to these differences, rather than due to our segment embedding-based approach.  To address this question, we perform a simple comparison on the proxy task of acoustic word discrimination, which is often used to compare speech representations~\cite{carlin2011rapid_eval_spoken_term_detect,kamper2016cnn_awe} (see Section~\ref{ch:back:prelims:acoustic_ap}).  Specifically, we use the same task and multilingual development set of Hu {\it et al.}~\cite{hu2020multilingual} described in Chapter~\ref{ch:multi_awe}. We compare performance on this task using when using AWEs versus when using DTW applied to the frame-level hidden state outputs from the final layer of our AWE models. Both approaches use the same network trained on the same data, so the only difference is whether we measure segment distances using the pooled embeddings output from our model or DTW on the frame-level outputs. On this task, the AWE approach has an acoustic average precision (AP) of $0.57$, whereas the DTW approach has an AP of $0.49$.  This comparison demonstrates the improvement due to embedding segments as vectors rather than aligning them via DTW.

\subsection{Run time}
\label{ch:multi_qbe:results:run_time}

Table~\ref{ch:multi_qbe:tab:run_time} compares the average per-query run times of our implementations of ASE-based and DTW-based QbE search. In the DTW-based system, we use a sliding window approach with a window size of $90$ frames and a shift of $10$ frames. This results in each query being compared with $600$k windowed segments, while our ASE-based QbE system on average compares each query with $4$M windowed segments. %as described in Section~\ref{ssec:qbe_setup}. 
We use $39$-dimensional filterbank+pitch features and $512$-dimensional BiGRU hidden states from the ASE model as inputs to the DTW-based QbE system. All systems are tested on a single thread of an Intel Core i7-9700K CPU. As expected, the ASE-based approach is more than two orders of magnitude faster than the DTW-based approaches. Although the DTW system here is not identical to any of the %actual 
QUESST 2015 systems, this rough comparison gives an idea of the speed difference. In addition, for embedding-based systems used in a real search setting, approximate nearest neighbor can be used to further speed up the search~\cite{levin2015srails,settle2017query}.

\begin{table}[H]
\centering
\small
\caption{Run times on the QUESST 2015 development set.}
\begin{tabular}{lcr}
\toprule
{\bf Method}    & \# of comparisons & Run-time \\ 
                & (per query)       & (s / query) \\
\midrule
DTW on Filterbank features & $600$k & 486\\
DTW on ASE hidden states  & $600$k & 847\\
\textbf{ASE-based QbE} & \hspace{.1in} $4$M & \textbf{5}\\
\bottomrule
\end{tabular}
\label{ch:multi_qbe:tab:run_time}
\end{table}

\section{Conclusion}

We have presented a simple embedding-based approach for multilingual query-by-example search that outperforms prior work on the QUESST 2015 challenge, while also being much more efficient. We verify that multilingual acoustic word embedding (AWE) models can be effective for query-by-example search on unseen target languages, and we further improve performance by extending the approach to multi-word spans using acoustic span embeddings (ASE). Our embeddings are jointly trained with embeddings of the corresponding written words, but for QbE we ignore the written embeddings.  In future work, it would be interesting to explore the use of both the learned models to allow for search by either spoken or written query.

%% file: text/8_seg_a2w.tex
\chapter{Whole-word segmental speech recognition with acoustic word embeddings}
\label{ch:seg_a2w}

Segmental models for speech recognition are sequence prediction models in which hypothesis scores are based on partitioning an utterance into variable-length segments. Segmental models are computationally challenging as the number of segmentation paths is proportional to the vocabulary size, which is orders of magnitude larger when using whole words than when using subword units like phones. We consider segmental models for end-to-end whole-word (i.e. ``acoustic-to-word") speech recognition where the feature vectors used to score segments are derived from acoustic embedding models. 

In this chapter, we describe contributions published in Shi {\it et al.}~\cite{shi2021seg}. The first (1) is Bowen Shi's extension of forward-backward and Viterbi decoding to the GPU for efficient training and inference of end-to-end whole-word segmental models. The second (2) is a simple scoring function for variable-length segments (similar to acoustic word embedding functions in prior work~\cite{settle2016rnn_awe,he2017multiview,hu2020multilingual}) that reduces space complexity and allows for seemless multi-view pretraining similar to Chapter~\ref{ch:ctc_a2w}. We include details of both contributions to properly contextualize and describe our work, but contribution (2) is the primary focus of this thesis. We pretrain acoustic word embedding (AWE) and acoustically grounded word embedding (AGWE) models to initialize the whole-word segmental model parameters before speech recognition training. We find that the most significant reduction in word error rate (WER) comes from pretraining the acoustic segment function first as an AWE model, and additional smaller gains can be seen when initializing the word-level prediction layer with AGWEs. 

\subsubsubsection{Collaboration} This work relies on collaboration with Bowen Shi, who performs all speech recognition experiments with my contributions focused on acoustic word embedding training and integration with the segmental model-based acoustic-to-word speech recognizer.

\section{Introduction}

End-to-end whole-word speech recognition models are often referred to as ``acoustic-to-word" (A2W) models since they learn to map input acoustic frame features directly to words. Much of the work on these A2W models~\cite{soltau2016a2w,audhkhasi2017a2w,audhkhasi2018a2w,li2018advancing,yu2018multistage,settle2019_a2w,gaur2019acoustic_phrase} is based on connectionist temporal classification (CTC)~\cite{graves2006connectionist} as seen in Chapter~\ref{ch:ctc_a2w}, where the word sequence probability is defined as the product of frame-level probabilities. There has also been recent work on encoder-decoder A2W models~\cite{collobert2019wordlevel,palaskar2018_s2s} that implicitly focus on ``soft segments" via an attention mechanism, but neither approach explicitly models multi-frame segments corresponding to words. We propose such a method using whole-word segmental models where the sequence probability is computed by aggregating over {\it segment} scores instead of {\it frame} probabilities. 

Segmental models have a long history in speech recognition research, but they have been used primarily for phonetic recognition or as phone-level acoustic models~\cite{ostendorf1996hmm,glass2003probabilistic,zweig2009segmental,zweig2012_seg,he2012_sc,hamid2013_dsnn,tang2014_losses,lu2016_srnn}. There has also been work on whole-word segmental models for second-pass rescoring~\cite{zweig2009segmental,maas2012word,bengio2014word}, but to our knowledge our approach is the first to address A2W segmental models.

The key ingredient in our approach is to define segment scoring in terms of dot products between vector embeddings of acoustic segments and a weight layer of word label embeddings. This model form allows for (1) the efficient re-use of feature functions for a reduced memory cost and (2) the initialization of the recognizer components using pretrained AWE and AGWE models, following their successful application to both speech recognition~\cite{settle2019_a2w} and speech search~\cite{settle2017query,audhkhasi2017asr_free_kws,hu2021ase}. This pretraining provides significant benefits, and results in segmental models that outperform the best prior A2W models on conversational speech recognition~\cite{audhkhasi2017a2w,audhkhasi2018a2w,yu2018multistage,li2017acoustic,li2018advancing,settle2019_a2w}.

\section{Approach}

\begin{figure}
\center
\includegraphics[width=0.675\linewidth]{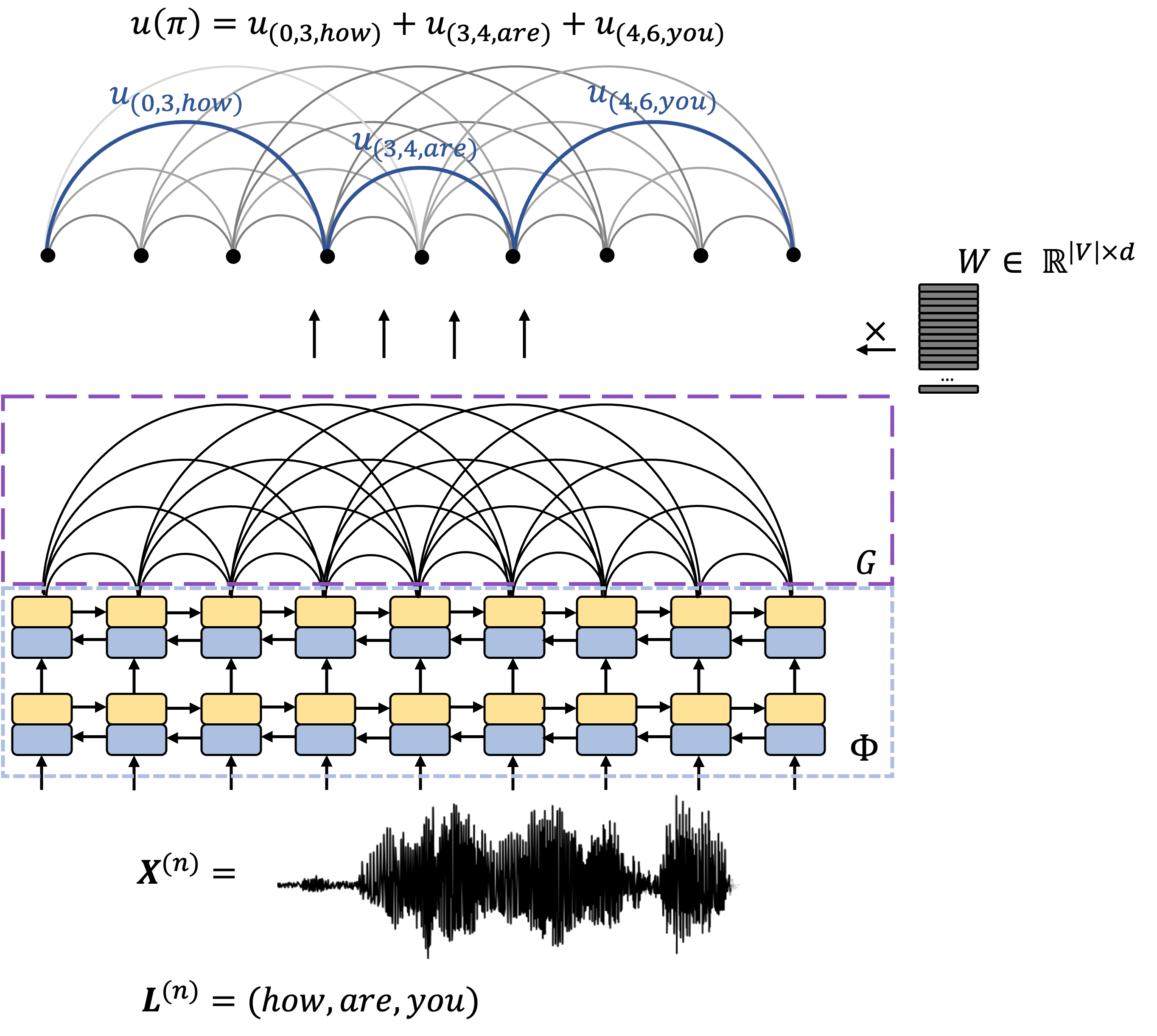}
\caption{Whole-word segmental model for speech recognition with acoustic encoder network $\Phi$, pooling module $G$, and word-level prediction layer $W$. $\mats{X}^{(n)}$ and $\mats{L}^{(n)}$ are an example utterance-transcript pair.}
\label{ch:seg_a2w:fig:model}
\end{figure}

Segmental models compute the score of a hypothesized label sequence by combining scores of multi-frame speech segments rather than frame-level scores (Figure~\ref{ch:seg_a2w:fig:model}). This presents a challenge since we must consider the set of all possible frame-level segmentations (up to a maximum segment size). Whole-word segmental modeling is then additionally complex since we must efficiently score these segments against a large word-level vocabulary.

We start with the segmental model formulation, and follow with our whole-word segment scoring function as well as common segment scoring functions used by prior work for subword-based modeling. Then, we describe the details of model training and decoding, and conclude with our multi-view pretraining approach.

\subsection{Segmental Model Formulation}

Let an utterance $\mats{X} = \{\vecs{x}_1, \vecs{x}_2, ..., \vecs{x}_T\}$ be a sequence of input acoustic frames and its transcript $\mats{L} = \{l_1, l_2, ..., l_K\}$ be the output label sequence. A segmentation $\pi$ with respect to an utterance-transcript pair $(\mats{X}, \mats{L})$ is given by a sequence of tuples 
\begin{flalign*}
\pi :=  \{(t_1, s_1, l_1), (t_2, s_2, l_2), ..., (t_K, s_K, l_K)\} 
\end{flalign*}
\noindent indicating start timestep\footnote{Frame $x_t$ is the acoustic signal between timesteps $t-1$ and $t$.} $t_k$, end timestep $t_k+s_k$, and label $l_k$ for each segment, and follows the set of constraints:
\begin{flalign*}
t_1 &= 0\\
t_K + s_K &= T\\
t_k + s_k &= t_{k+1} \text{ where } s_k > 0 \text{ for all } k \in \left[1, K\right]
\end{flalign*}
A segment score function assigns a score $u_{t,s,v}$ to each valid segment $(t, s, v)$ to form a score tensor
\begin{flalign*}
\tens{U} \in \mathbb{R}^{T \times S \times V}    
\end{flalign*}
\noindent where $S$ and $V$ denote the maximum segment size and vocabulary size, respectively. The score of a particular segmentation $\pi$ is given by summing up its component segment scores:
\begin{flalign*}
u(\pi) = \sum_{(t,s,v)\in\pi}u_{t,s,v}
\end{flalign*}

\subsubsection{Whole-word segment scoring}

Our whole-word approach derives the segment score tensor $\tens{U}$ from dot products between embeddings of variable-length spoken segments and embeddings of the word labels. The score of spoken segment $\mats{X}_{t:t+s}$ with respect to a particular word $v$ is
\begin{flalign}
u_{t,s,v} &= {\mats{W}_v}^T f(\mats{X}_{t:t+s}) + b_v
\label{ch:seg_a2w:score_func}
\end{flalign}
\noindent where $f$ is the acoustic segment embedding function mapping a segment $\mats{X}_{t:t+s}$ to a fixed-dimensional embedding $f(\mats{X}_{t:t+s}) \in \mathbb{R}^d$, ${\mats{W}_v}^T$ is a row from the matrix $\mats{W} \in \mathbb{R}^{V \times d}$ composed of embeddings for all words $v$ in the vocabulary, and $b_v$ is the bias on word $v$, which can be interpreted as a log-unigram probability.

As in Chapter~\ref{ch:multi_qbe}, the acoustic segment embedding function 
\begin{flalign}
    f(\mats{X}_{t:t+s}) &:= G(\Phi(\mats{X}), t, t+s) \in \mathbb{R}^d
\label{ch:seg_a2w:segment_func}
\end{flalign}
\noindent is composed of an acoustic encoder network $\Phi$ applied over the whole utterance $\mats{X}$ followed by a pooling module $G$. The encoder $\Phi$ produces frame-level output features $\Phi(\mats{X})$, and the pooling module $G$ maps each variable-length segment with boundaries $t$ and $t+s$ to fixed dimensional embeddings. This allows for substantial feature sharing with $\Phi$, and helps to limit the memory needed to compute segment features to $O(TSd)$. Equation~\ref{ch:seg_a2w:score_func} also simplifies segment scoring to matrix multiplication 
\begin{equation}
\tens{U}_{t,s}=\mats{W} f(\mats{X}_{t:t+s})+ \vecs{b}
\label{ch:seg_a2w:score_func_full}
\end{equation}
\noindent where $\tens{U}_{t,s} \in \mathbb{R}^V$. 
Additionally, since our segment scoring function can be described in terms of dot products between acoustic segment embeddings produced by $f$ and word label embeddings in $W$, we can pretrain these components very effectively using multi-view joint training of AWE and AGWE models as in Chapter~\ref{ch:ctc_a2w}.

\subsubsection{Prior work using subword-based segment scoring}

Prior work on segmental models focuses on subword prediction and largely uses two types of segment score functions: (1) frame classifiers~\cite{hamid2013_dsnn,tang2015_cascade,he2012_sc,tang2017_e2e} and (2) segmental recurrent neural networks (SRNNs)~\cite{lu2016_srnn,kong2016_srnn,tang2017_e2e}. 

Frame classifier-based segment score functions use a mapping from input acoustic frames $\mats{X}$ to frame log-probability vectors $\mats{P}$, which are then pooled (via averaging, sampling, etc.) to get segment scores $u_{t,s,v}$. This method has a multiplicative memory dependence on the vocabulary size of $O(TSV)$, which is a factor $\frac{V}{d}\times$ increase in memory overhead over our approach at $O(TSd)$. In our case, $V$ is the number of words in the vocabulary, which can be orders of magnitude larger than $d$ such that this approach is often prohibitively expensive. 

SRNN-based segment score functions compute the score $u_{t,s,v} = \phi^T f_{\theta}([\mats{H}_t; \mats{H}_{t+s}; \mats{W}_v])$, where $\phi \in \mathbb{R}^d$ is a parameter vector, $f_{\theta}$ is a learned feature function, $\mats{H} = \text{RNN}(\mats{X})$ are frame-level acoustic encoder features, $\mats{W}_v$ is an embedding of vocabulary element $v$, and $[\vecs{a}; \vecs{b}]$ denotes concatenation of $\vecs{a},\vecs{b}$. This method has dependencies on both $d$ and $V$ for an $O(TSdV)$ memory overhead, which again makes it infeasible for word-level vocabulary recognition.

\subsubsection{Training}

Segmental models can be trained in a variety of ways~\cite{tang2017_e2e}. One way, which we adopt here, is to interpret them as probabilistic models and optimize the marginal log loss under that model, which is equivalent to viewing our models as segmental conditional random fields~\cite{zweig2011speech}. Under this view, the model assigns probabilities to paths, conditioned on the input acoustic sequence, by normalizing the path score. 

For convenience, we use the exponentiated form of the score tensor ${\bf U}$ output by our segment score function 
\begin{equation}
\tilde{\bf U} := \exp({\bf U}) \in \mathbb{R}_{>0}^{T \times S \times V}
\end{equation}
\noindent such that the score of segmentation $\pi$ is now computed by the product
\begin{equation}
\tilde{u}(\pi) = \prod_{(t,s,v)\in\pi} \tilde{u}_{t,s,v}
\end{equation}
which we use to define the probability of a segmentation $\pi$ as 
\begin{equation}
p(\pi) := \frac{\tilde{u}(\pi)}{\displaystyle \sum_{\pi'\in\mathcal{P}_{0:T}}\tilde{u}(\pi')}
\end{equation}
where  $\mathcal{P}_{0:T}$ denotes all segmentations of $\mats{X}_{1:T}$. The loss for a given length-$T$ input acoustic sequence $\mats{X}$ and length-$K$ word sequence $\mats{L}$ is the marginal log loss, computed by marginalizing over all possible segmentations
\begin{equation}
\begin{split}
\label{eq:segmental_log_loss}
\mathcal{L}_{seg}(\mats{X}, \mats{L}) &= -\log\displaystyle\sum_{\substack{\pi\in\mathcal{P}_{0:T} \text{,}\\ \Omega(\pi)=\mats{L}}}\tilde{u}(\pi) +\log{\displaystyle\sum_{\pi\in\mathcal{P}_{0:T}}\tilde{u}(\pi)} \\
\end{split}
\end{equation}
\noindent where $\Omega(\pi)$ maps $\pi=\{(t_k,s_k,l_k)\}_{1\leq k\leq K}$ to its label sequence $\mats{L} = \{l_1, l_2, \dots, l_K\}$. %_{1\leq k\leq K}$.
The summations in Equation~\ref{eq:segmental_log_loss} can be efficiently computed with dynamic programming 
\begin{equation}
\begin{split}
\alpha^{(d)}_t &:= \displaystyle\sum_{\pi\in\mathcal{P}_{0:t}} \tilde{u}(\pi) = \displaystyle\sum_{s=1}^S\displaystyle\sum_{v=1}^{V}\tilde{u}_{t-s,s,v}\alpha^{(d)}_{t-s} \\ %[-5pt]
\alpha^{(n)}_{t,y} &:= \hspace{-10pt}\displaystyle\sum_{\substack{\pi\in\mathcal{P}_{0:t} \\ \Omega(\pi_{1:y})=\mats{L}_{1:y}}}\hspace{-10pt} \tilde{u}(\pi) = \displaystyle\sum_{s=1}^S \tilde{u}_{t-s,s, l_y}\alpha^{(n)}_{t-s, y-1}\\
\end{split}
\label{eq:alpha_nom_denom}
\end{equation}
\noindent where $S$ is the maximum segment length and $V$ is the vocabulary size. The loss value follows directly once $\alpha^{(d)}$ and $\alpha^{(n)}$ are computed. 
\begin{equation}
\mathcal{L}_{seg}(\mats{X}, \mats{L})=-\log\alpha^{(n)}_{T,K}+\log{\alpha^{(d)}_T}
\end{equation}
The last summations in Equations~\ref{eq:alpha_nom_denom} can be efficiently implemented on a GPU. In addition, $\alpha_{1:T,y}^{(n)}$ can be computed in parallel given $\alpha_{1:T,y-1}^{(n)}$ such that the overall time complexity\footnote{Number of times $a+b$ is called} of computing the loss is $O(T\log(SV)+K\log(S))$. 

To train with gradient descent, we need to differentiate $\mathcal{L}_{seg}(\mats{X}, \mats{L})$ with respect to $\mats{X}$, which can in principle be done with auto-differentiation toolkits (e.g.~PyTorch~\cite{pytorch}). However, in practice using auto-differentiation to compute the gradient is many times slower than the loss computation. Instead we explicitly implement the gradient computation (Equation~\ref{eq:grad}) using the backward algorithm. We define two backward variables $\beta^{(d)}_t$ and $\beta^{(n)}_{t,y}$ for the denominator and numerator, respectively,
\begin{equation}
\label{eq:beta_nom_denom}
\begin{split}
  \beta^{(d)}_t &:=\displaystyle\sum_{\pi\in\mathcal{P}_{t:T}} \tilde{u}(\pi)=\displaystyle\sum_{s=1}^{S}\displaystyle\sum_{v=1}^{V} \tilde{u}_{t,s,v}\beta^{(d)}_{t+s}\\
  \beta^{(n)}_{t,y} &:=\hspace{-15pt}\displaystyle\sum_{\substack{\pi\in\mathcal{P}_{t:T} \\ \Omega(\pi_{y:K})=\mats{L}_{y:K}}}\hspace{-15pt} \tilde{u}(\pi)=\displaystyle\sum_{s=1}^{S}\displaystyle \tilde{u}_{t,s,l_y}\beta^{(n)}_{t+s,y+1}\\
\end{split}
\end{equation}
and the gradient $\frac{\partial \mathcal{L}_{seg}(\mats{X}, \mats{L})}{\partial \tilde{u}_{t,s, v}}$ is given by 
\begin{equation}
\label{eq:grad}
\begin{split}
&\frac{\partial \mathcal{L}_{seg}(\mats{X}, \mats{L})}{\partial \tilde{u}_{t,s,v}} = -\displaystyle\sum_{k\in \{k|l_{k}=v\}}\frac{\alpha^{(n)}_{t,k}\beta^{(n)}_{t+s,k}}{\alpha^{(n)}_{T,K}} + \frac{\alpha^{(d)}_{t} \beta^{(d)}_{t+s}}{\alpha^{(d)}_T}
\end{split}
\end{equation}
\noindent where $\{k|l_{k}=v\}$ are the indices in $\mats{L}$ where label $v$ occurs.

\subsubsection{Decoding}
Decoding consists of solving $\pi^\star=\argmax_{\pi\in\mathcal{P}_{0:T}}u(\pi)$. This optimization problem can be solved efficiently via the Viterbi algorithm with the recursive relationship:
\begin{equation}
d(t) := \max_{\pi\in\mathcal{P}_{0:t}} u(\pi) = \max_{\substack{1\leq s\leq S\\ 1\leq v\leq V}}[u_{t-s,s,v}+d(t-s)]
\label{eq:viterbi}
\end{equation}
where the last max operation can be parallelized on a GPU such that the overall runtime\footnote{Number of times $\max(a, b)$ is called} of decoding is only $O(T\log(SV))$.

\subsection{Pretraining via acoustic and acoustically grounded word embeddings}
\label{ch:seg_a2w:pretraining}

Many words in our vocabulary at test-time are infrequent or unseen in the training set, which presents an important issue for A2W models. Chapter~\ref{ch:ctc_a2w}, which focuses on prior work in Settle {\it et al.}~\cite{settle2019_a2w}, finds that joint multi-view pretraining of acoustic word embedding (AWE) and acoustically grounded word embedding (AGWE) models offer better parameter initialization for CTC-based A2W models~\cite{settle2019_a2w}. This initialization is even more natural for whole-word segmental models than for the CTC-based approach of Settle {\it et al.}~\cite{settle2019_a2w} since the segments are explicitly intended to model words, so we apply this same idea here.

Our multi-view pretraining follows a similar approach to Chapters~\ref{ch:ctc_a2w} and~\ref{ch:multi_qbe}, in which we jointly learn parameters of a contextual AWE function $f$ and an AGWE function $g$ using a contrastive loss. The contextual AWE function $f$ is defined 
\begin{flalign}
f(\mats{X}_{t:t+s}) &= G(\Phi(\mats{X}), t, t+s)
\label{eq:awe_func}
\end{flalign}
where $f(\mats{X}_{t:t+s}) \in \mathbb{R}^d$ is a contextual representation of the spoken segment from timestep $t$ to $t+s$ within the utterance $\mats{X}$. This function $f$ consists of an utterance-level acoustic encoder $\Phi$ followed by the module $G$ used to pool over the timesteps $t$ to $t+s$. $G$ consists of a pooling operation followed by a single trainable affine transformation and a ReLU nonlinearity. The pooling operation operates on top of the encoded frame-level output features 
\begin{flalign*}
\mats{H} = \Phi(\mats{X}) \in \mathbb{R}^{T \times F}
\end{flalign*}
\noindent and is chosen among concatenation, mean pooling, and attention pooling (with trainable parameter {\bf r}). 
\begin{flalign}
Pool(\mats{H}, t, t+s) &= [\mats{H}_t; \mats{H}_{t+s}] \tag{concatenation}\\ %\label{eq:concat}\\
Pool(\mats{H}, t, t+s) &= \frac{1}{s}\textstyle\sum_{i=1}^{s}\mats{H}_{t+i} \tag{mean}\\ %\label{eq:avg}\\
Pool(\mats{H}, t, t+s) &= \frac{1}{s}\textstyle\sum_{i=1}^{s}{\text{Softmax}({\bf r}^{T}\mats{H}_{t:t+s})}_i \mats{H}_{t+i} \tag{attention}
\end{flalign}
The AGWE model function $g$ takes in a word label $v$, maps $v$ to a subword (e.g., character/phone) sequence using a lexicon, and uses this sequence to produce an embedding vector as output from another BiLSTM network as in Chapters~\ref{ch:ctc_a2w},~\ref{ch:multi_awe}, and~\ref{ch:multi_qbe}.

We train $f$ and $g$ jointly to minimize the objective
\begin{flalign}
\mathcal{L}_{emb}(\tensorsym{X}, \tensorsym{L}) = \sum_{n=1}^{N}\sum_{i=1}^{\vert \mats{L}^{(n)} \vert} \sum_{obj=0}^2 \mathcal{L}_{obj}(\mats{X}_i^{(n)}, v_i^{(n)})
\label{ch:seg_a2w:eq:multiview}
\end{flalign}
\noindent where $\mats{X}^{(n)}_i$ is the $i^{th}$ spoken word segment within $\mats{X}^{(n)}$ and $v^{(n)}_i$ is its word label, $\mats{L}^{(n)}$ is the word-level transcript for utterance $\mats{X}^{(n)}$, and the individual loss terms $\mathcal{L}_{obj}$ are defined
\begin{flalign*}
\mathcal{L}_0(\mats{X}, v) &= \sum_{v' \in \mathcal{N}_0(\mats{X}, v)} \left[m + d(f(\mats{X}), g(v)) - d(f(\mats{X}), g(v'))\right]_+ \\
\mathcal{L}_1(\mats{X}, v) &= \sum_{v' \in \mathcal{N}_1(\mats{X}, v)} \left[m + d(f(\mats{X}), g(v)) - d(g(v), g(v'))\right]_+\\
\mathcal{L}_2(\mats{X}, v) &= \sum_{\mats{X}' \in \mathcal{N}_2(\mats{X}, v)} \left[m + d(f(\mats{X}), g(v)) - d(g(v), f(\mats{X}'))\right]_+
\end{flalign*}
\noindent where $d(\vecs{a}, \vecs{b}) = 1-\frac{\vecs{a}\cdot \vecs{b}}{\Vert \vecs{a} \Vert\Vert \vecs{b} \Vert}$, $m$ is a margin hyperparameter, and sets $\mathcal{N}_{obj}(\mats{X}, v)$ used for semi-hard negative sampling within each objective $\mathcal{L}_{obj}$ are defined
\begin{flalign*}
\mathcal{N}_0(\mats{X}, v) &:= \left\{
    v'\ \middle\vert
    \begin{array}{c}
    v' \in \mathcal{V} / v, \hspace{0.5cm} d(f(\mats{X}), g(v')) > d(f(\mats{X}), g(v))
    \end{array}
\right\}\\
\mathcal{N}_1(\mats{X}, v) &:= \left\{
    v'\ \middle\vert
    \begin{array}{c}
    v' \in \mathcal{V} / v, \hspace{0.5cm} d(g(v), g(v')) > d(f(\mats{X}), g(v))
    \end{array}
\right\}\\
\mathcal{N}_2(\mats{X}, v) &:= \left\{
    \mats{X}'\ \middle\vert
    \begin{array}{c}
    \textsc{label}(\mats{X}') \in \mathcal{V} / v, \hspace{0.5cm} d(g(v), f(\mats{X}')) > d(f(\mats{X}), g(v))
    \end{array}
\right\}
\end{flalign*}
\noindent with $\mathcal{V}$ the training vocabulary. For efficiency, negative sampling is performed only over the mini-batch. This means that in Equation~\ref{ch:seg_a2w:eq:multiview} $N$ is the number of utterances in a batch and the vocabulary $\mathcal{V}$ used to define the negative sampling sets $\mathcal{N}_{obj}$ consists only of words in the batch. Additionally, we limit the size of the negative sampling sets to just $k$ such embeddings within each $\mathcal{N}_{obj}(\mats{X}, v)$ as in previous chapters:
\begin{flalign*}
\mathcal{N}_0^k(\mats{X}, v) &:= \left\{
    v_1', \dots, v_k'\ \middle\vert
    \begin{array}{c}
        v_i' = \displaystyle \argmin_{v' \in \mathcal{N}_0(\mats{X}, v)} d(f(\mats{X}), g(v'))
    \end{array}
\right\}\\
\mathcal{N}_1^k(\mats{X}, v) &:= \left\{
    v_1', \dots, v_k'\ \middle\vert
    \begin{array}{c}
        v_i' = \displaystyle \argmin_{v' \in \mathcal{N}_1(\mats{X}, v)} d(g(v), g(v'))
    \end{array}
\right\}\\
\mathcal{N}_2^k(\mats{X}, v) &:= \left\{
    \mats{X}_1', \dots, \mats{X}_k'\ \middle\vert
    \begin{array}{c}
        \mats{X}_i' = \displaystyle \argmin_{\mats{X}' \in \mathcal{N}_2(\mats{X}, v)} d(g(v), f(\mats{X}'))
    \end{array}
\right\}
\end{flalign*}

After pretraining, we use the AWE and AGWE models $f$ and $g$, respectively, to initialize the A2W segmental model recognizer. The AWE model $f$ {\it becomes} the acoustic segment embedding model, and the AGWE model $g$ can be used to generate embeddings for the word-level prediction layer $\mats{W}$.\footnote{The bias vector $\mats{b}$ is randomly initialized before recognizer training.} Beyond initialization, the pretrained AGWEs can also be used to regularize $\mats{W}$ during recognizer training. In Equation~\ref{ch:seg_a2w:eq:multitask}, we add an $L_2$ regularizer loss during recognizer training that considers for each word $v$ in the batch the distance between the row $\mats{W}_v$ of the prediction layer weight matrix and the corresponding pretrained AGWE $g(v)$:
\begin{flalign}
\mathcal{L}_{asr}(\tens{X}, \tens{L}) 
&=  (1-\lambda_{reg}) \mathcal{L}_{seg} (\tens{X}, \tens{L}) + \lambda_{reg} \sum_{v \in \cup_{n=1}^N \mats{L}^{(n)}} \Vert g(v) - \mats{W}_v \Vert_2
\label{ch:seg_a2w:eq:multitask}
\end{flalign}
\noindent where $(\tens{X}, \tens{L})$ is again a batch of $N$ utterance-transcript pairs, $g$ is the AGWE model, $\mats{W}_v$ is the row of the prediction layer corresponding to word $v$, $\lambda_{reg} \in \left[0, 1\right]$ is a hyperparameter, and $\mathcal{L}_{seg}$ is the segmental modeling recognition loss (Equation~\ref{eq:segmental_log_loss}).

\begin{figure}
\centering
\includegraphics[width=0.875\linewidth]{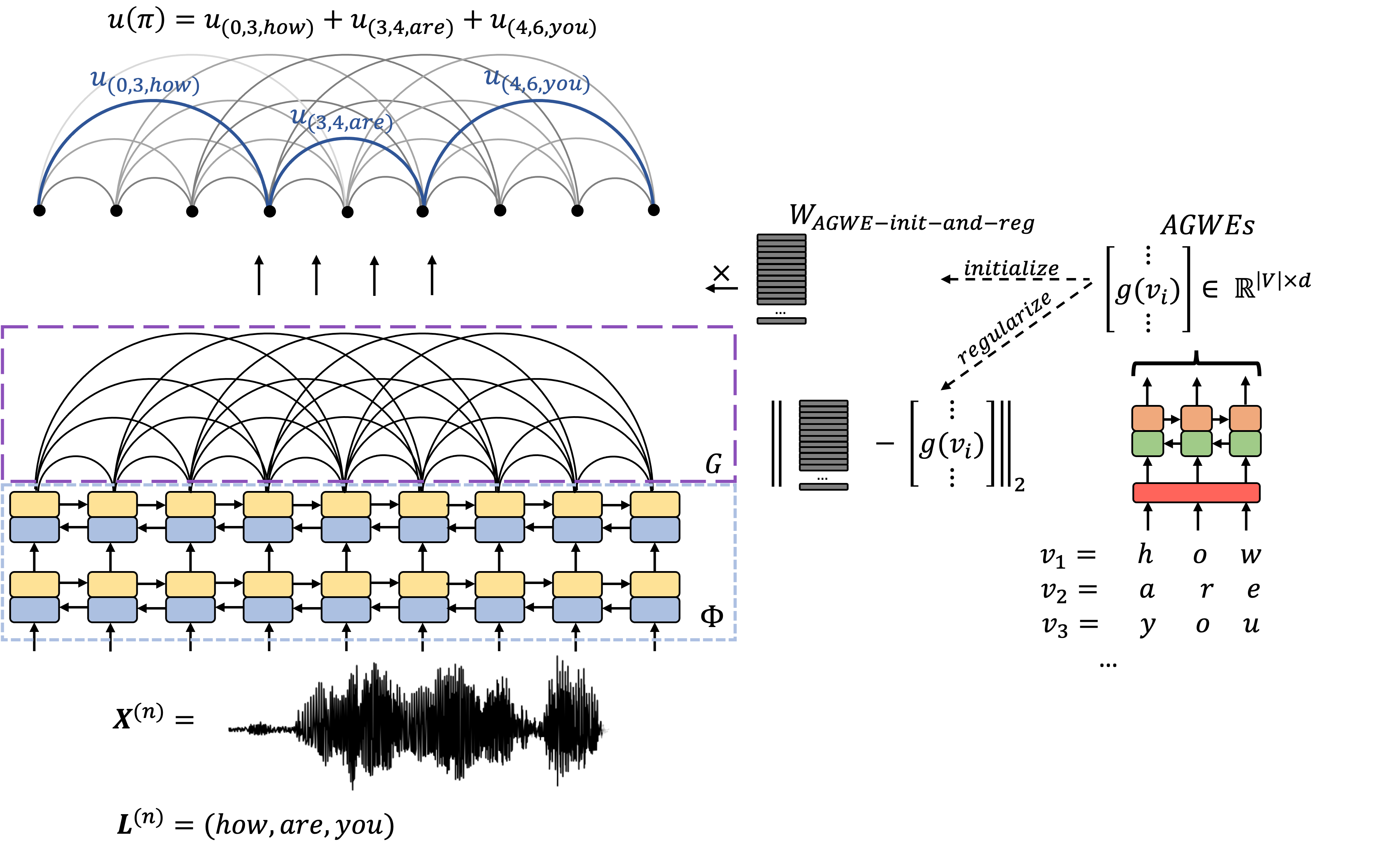}
\caption{Whole-word segmental recognizer with AGWE model $g$ used for regularization after initialization. $\mats{X}^{(n)}$ and $\mats{L}^{(n)}$ are an example utterance-transcript pair.}
\label{ch:agwe_a2w_ctc:pretrain}
\end{figure}

\section{Experimental setup}
\label{ch:seg_a2w:setup}

\subsection{Data}
We conduct experiments on the standard Switchboard-$300$h dataset and data partitioning~\cite{godfrey1992switchboard}. We use $40$-dimensional log-Mel spectra +$\Delta$+$\Delta\Delta$s, extracted with Kaldi~\cite{povey2011kaldi}, as input features. Every two successive frames are stacked and alternate frames dropped, resulting in $240$-dimensional features. We explore $5$k, $10$k, and $20$k vocabularies based on word occurrence thresholds of $18$, $6$, and $2$, respectively.

\subsection{Whole-word segmental model details}
\label{ch:seg_a2w:seg_details}
The acoustic encoder network $\Phi$ consists of a $6$-layer BiLSTM network with $512$ hidden units per direction per layer with dropout added between layers ($0.25$ except when otherwise specified below), a convolutional layer with kernel size $5$, and mean pooling with stride $4$.\footnote{The convolutional layer and mean pooling are used on top of the backbone BiLSTM network to speed up training.} The maximum segment length is $32$ frames, corresponding to a maximum word duration of approximately $2.4s$. We sort training utterances by input length such that similar-length utterances are batched together. For training efficiency, we reduce the maximum segment size per batch to $\min\{2*\max\{\frac{\text{input length}}{\text{\# words}}\}, 32\}$. We use a batch size of $16$ utterances with the Adam optimizer~\cite{kingma2014adam}. The initial learning rate is $0.001$, and it is decreased by a factor of $2$ when development set WER stops decreasing. No language model is used for decoding.

The pooling operation within $G$ is tuned with respect to development set WER among concatenation ($18.2\%$), mean pooling ($18.5\%$), and attention pooling ($19.0\%$). The best performance is obtained with concatenation, which slightly increases the number of parameters but consumes a factor of $\frac{S}{2}$ less memory when computing segment features.

\subsection{AWE+AGWE pretraining details}

We jointly train AWE and AGWE models on the Switchboard-300h training set with the multi-view training approach described in Section~\ref{ch:seg_a2w:pretraining}. In Equation~\ref{ch:seg_a2w:eq:multiview}, we use $m=0.45$, $k=64$ reduced by $1$ per batch until $k=6$, and a variable batch size with utterances totaling up to $20,000$ frames per batch. The AWE model has the same structure as the segmental model such that both are indicated by the same function $f$. The AGWE model $g$ is composed of an input embedding layer mapping $37$ input characters to $32$-dimensional embeddings followed by a 1-layer BiLSTM with 256 hidden units per direction. We use the Adam optimizer~\cite{kingma2014adam} with an initial learning rate of $0.0005$ that is reduced by a factor of $10$ when the development set cross-view average precision (AP) (Section~\ref{ch:back:prelims:crossview_ap}) does not improve for $3000$ steps. Training is stopped when the learning rate drops below $10^{-9}$. Our best embeddings have a cross-view AP of $0.83$ when evaluated on development set word segments and a $20$k vocabulary. After multi-view training, the acoustic view $f$ (our AWE function) and the written view $g$ (our AGWE function) are used to initialize our segmental feature function $f$ and the rows of the predicion layer $\mats{W}$, respectively, in Equation~\ref{ch:seg_a2w:score_func}.

Our pretraining approach is most similar to that used in Settle {\it et al.}~\cite{settle2019_a2w} with three updates: (1) the use of {\it semi}-hard negative sampling~\cite{schroff2015facenet} (replacing hard negative sampling in~\cite{settle2019_a2w}), (2) the inclusion of a third contrastive term $\mathcal{L}_1$ ($obj_1$ in He {\it et al.}~\cite{he2017multiview}), and (3) a convolutional layer followed by average pooling on top of the BiLSTM in the AWE model (Section~\ref{ch:seg_a2w:seg_details}). Changes (1-2) increase word discrimination task performance in prior work on AWEs~\cite{settle2016rnn_awe,kamper2016cnn_awe,he2017multiview}, and (3) improves the efficiency of our segmental model.

\begin{figure}
\centering
\includegraphics[width=0.55\linewidth]{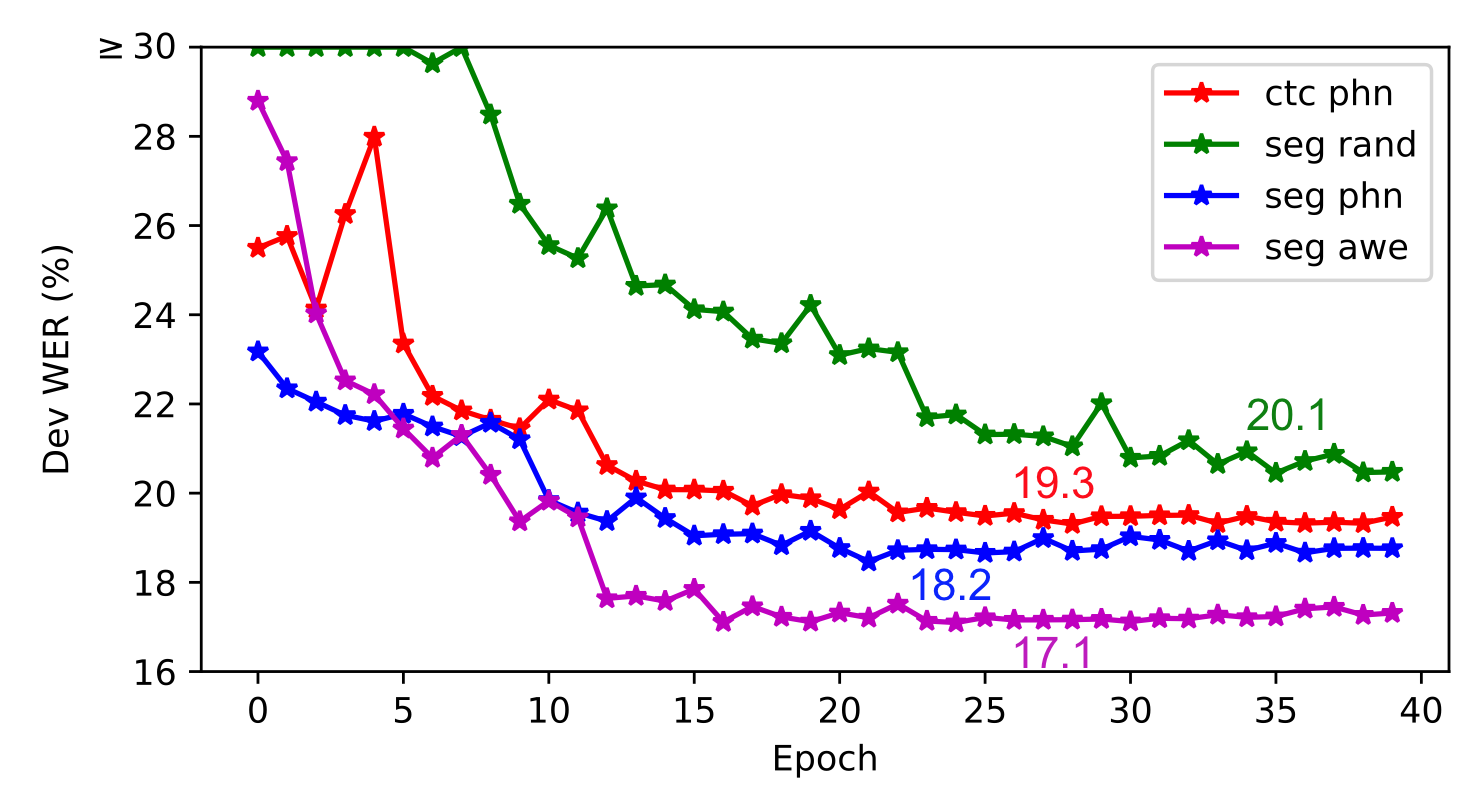}
\caption{Switchboard development set WER vs.~epoch for A2W CTC (ctc) and A2W segmental (seg) models across random (rand), phone CTC (phn) pretraining, and AWE+AGWE (awe) pretraining initializations. A2W models are trained with a $5$k vocabulary; each curve is annotated with its lowest WER.}
\label{fig:ctc_seg_dev}
\end{figure}

\section{Results}

\subsection{Phone CTC pretraining}
\label{ssec:exp_phn}

For our initial baseline experiment, we initialize the segmental model's BiLSTM network by first pretraining with a phone CTC objective\footnote{The phone error rate (PER) of this phone CTC model is $11.0\%$.}, as in prior work on CTC-based A2W models~\cite{audhkhasi2017a2w,audhkhasi2018a2w,yu2018multistage,settle2019_a2w}, but all other trainable parameters are randomly initialized. Then, we perform word-level training with a $5$k vocabulary. Phone CTC pretraining reduces WER by $1.9\%$ (Figure~\ref{fig:ctc_seg_dev}) over a fully randomly initialized model, which is consistent with prior A2W CTC work~\cite{audhkhasi2017a2w}. When both our A2W segmental model and our baseline A2W CTC model are pretrained with phone CTC, the segmental model achieves around $1\%$ lower WER (Figure~\ref{fig:ctc_seg_dev}).

\begin{table}[H]
\centering
\caption{Phone CTC vs.~AWE+AGWE init measured by SWB development WER (\%). ``AWE init" and ``AGWE init" refer to initializing parameters of $f$ and $\mats{W}$, respectively.}
\begin{tabular}{lccc}
\toprule
{\bf System} & \multicolumn{3}{c}{Vocab size}\\
             & $5$k & $10$k & $20$k\\
\midrule
A2W CTC w/ phone CTC init & 19.3 & 18.0 & 17.7 \\
\midrule
A2W Segmental w/ phone CTC init  & 18.2 & 17.9 & 18.0 \\
\hspace{0.25cm} + AGWE init & 18.4 & 18.0 & 18.0 \\
A2W Segmental w/ AWE init & 17.1 & 16.0 & 16.4 \\
\hspace{0.25cm} + AGWE init & 17.1 & 15.8 & 16.5 \\
\hspace{0.25cm} + AGWE $L_2$ reg & \textbf{17.0} & \textbf{15.5} & \textbf{15.6}\\
\bottomrule
\end{tabular}
\label{tab:exp_agwe_init} 
\end{table}

\subsection{Vocabulary size}

Unlike our A2W CTC models and those of prior work~\cite{settle2019_a2w,yu2018multistage}, our segmental models do not necessarily improve with larger vocabulary (Table~\ref{tab:exp_agwe_init}). One possible reason is that word representations in segmental models, especially for rare words, may require additional data as they must be more robust to variations in segment duration and content. As a result, we use larger dropout values as the vocabulary size increases to mitigate overfitting with the best dropout values for the $5$k, $10$k and $20$k vocabularies being $0.25$, $0.35$, and $0.45$, respectively.

\begin{figure}[H]
\centering
\includegraphics[width=0.525\linewidth]{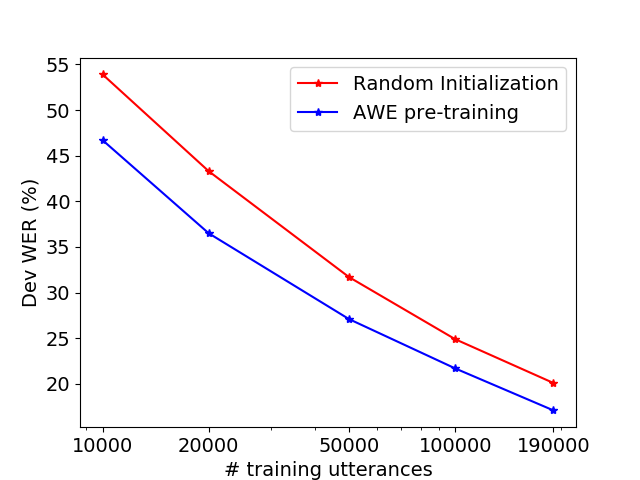}
\caption{Segmental model dev WER, using random vs.~AWE+AGWE pre-training initialization with various ASR training set sizes, using $5$k-word vocabulary.}
\label{fig:wer_utt}
\end{figure}

\subsection{AWE+AGWE pretraining}
\label{ch:seg_a2w:setup:pretrain}

\begin{figure}[H]
\centering
\includegraphics[width=0.675\linewidth]{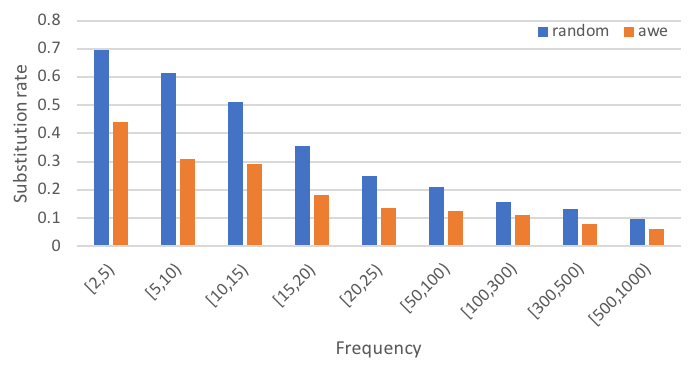}
\caption{Word substitution rates when using random vs.~AWE+AGWE pretraining initialization, for words with different frequencies in the training set. The vocabulary size is $20$k.}
\label{fig:error_frequency}
\end{figure}

Table~\ref{tab:exp_agwe_init} compares development set performance between A2W segmental models using different initializations. Initialization of $f$ with the pretrained AWE model reduces WER by $1$--$2\%$ over phone CTC initialization, but additional initialization of $\mats{W}$ with the pretrained AGWEs alone does not help. However, initializing {\it and} regularizing $\mats{W}$ with respect to the AGWEs is helpful, especially for larger vocabularies. This is consistent with our expectation that since AGWEs are composed from character sequences, they are impacted less by vocabulary size and should improve  recognition of rare words. This intuition is further demonstrated in Figures~\ref{fig:wer_utt} and~\ref{fig:error_frequency}. We note that the optimal $\lambda_{reg}$ in Equation~\ref{ch:seg_a2w:eq:multitask} tends to be larger as the vocabulary size increases, reinforcing the need for more regularization when there are many rare words.

\subsection{Final test-set evaluations}
\label{ssec:final-results}

Table~\ref{tab:exp_eval2000} compares our final results on the Switchboard (SWB) and CallHome (CH) test sets with prior work\footnote{We do not compare with prior segmental models~\cite{tang2014_losses,lu2016_srnn} since, due to their larger memory footprint, we are unable to train them for A2W recognition with a similar network architecture using a typical GPU (e.g., 12GB memory).} on A2W models, including our CTC results described in Chapter~\ref{ch:ctc_a2w}. For the smaller $5$k- and $10$k-word vocabularies, the segmental model improves WER over CTC by around $1\%$ (absolute). Training with SpecAugment~\cite{park2019specaugment} produces an additional gain of approximately $1\%$ across all vocabulary sizes. As noted before, the $10$k-word model outperforms the $20$k-word one. In addition to the issue of rare words, the longer training time with a $20$k-word vocabulary prevents us from tuning hyperparameters as well as for the $5$k and $10$k models. Overall our best model improves over all previous A2W models of which we are aware, while having fewer parameters than the previous best-performing model~\cite{yu2018multistage}.

Despite the improved performance of our A2W segmental models, there is still a gap between our A2W model and systems based on subword units, where the best performance on this task of which we are aware is $6.3\%$ on SWB and $13.3\%$ on CH using a subword-based sequence-to-sequence transformer model~\cite{wang2020investigation}. Some prior work suggests that the gap between between A2W and subword-based models can be significantly narrowed when using larger training sets, and our future work includes investigations with larger data. At the same time, A2W models retain the benefit of being very simple, truly end-to-end models that avoid using a decoder and are less dependent on a language model. 

\begin{table}
\centering
\caption{WER (\%) results on SWB/CH evaluation sets.}
\setlength{\tabcolsep}{3pt}
\begin{tabular}{lccc}\\\toprule
  {\bf System} & \multicolumn{3}{c}{Vocab} \\
                          & 4k/5k & 10k & 20k \\
  \midrule
  \footnotesize{Seg, AWE+AGWE init} & 14.0/24.9 & 12.8/23.5 & 12.5/24.5 \\ 
  \footnotesize{\hspace{0.25cm}+SpecAugment} & \textbf{12.8/22.9} & \textbf{10.9/20.3} & 12.0/21.9 \\ \midrule
  \footnotesize{CTC, phone init (Chapter~\ref{ch:ctc_a2w})~\cite{settle2019_a2w}} &  16.4/25.7 & 14.8/24.9 & 14.7/24.3 \\
  \footnotesize{CTC, AWE+AGWE init (Chapter~\ref{ch:ctc_a2w})~\cite{settle2019_a2w}} & 15.6/25.3 & 14.2/24.2 & 13.8/24.0 \\
  \footnotesize{\hspace{0.25cm}+reg (Chapter~\ref{ch:ctc_a2w})~\cite{settle2019_a2w}} & 15.5/25.4 & 14.0/24.5 & 13.7/23.8 \\
  \footnotesize{CTC, AWE+AGWE rescore (Chapter~\ref{ch:ctc_a2w})~\cite{settle2019_a2w}} & 15.0/25.3 & 14.4/24.5 & 14.2/24.7 \\
  \footnotesize{S2S \cite{palaskar2018_s2s}} &  - & 22.4/36.1 & 22.4/36.2 \\
  \footnotesize{Curriculum \cite{yu2018multistage}}  & - & - & 13.4/24.2 \\
  \footnotesize{\hspace{0.25cm}+Joint CTC/CE\cite{yu2018multistage}}  & - & - &  13.0/23.4 \\
  \footnotesize{\hspace{0.25cm}+Speed Perturbation\cite{yu2018multistage}}  & - & - &  \textbf{11.4/20.8} \\
  \bottomrule
\end{tabular}
\label{tab:exp_eval2000}
\end{table}

\section{Conclusion}

We have introduced an end-to-end whole-word segmental model, which to our knowledge is the first to perform large-vocabulary speech recognition competitively and efficiently. Our model uses a simple segment score function based on a dot product between written word label embeddings and acoustic segment embeddings, which both improves efficiency and enables us to pretrain the model with jointly trained AWE and AGWE models. We find that our proposed A2W segmental model outperforms previous A2W approaches, and is much more efficient than previous segmental models. The key aspects that are most important to the performance improvements are (1) pretraining the acoustic segment embedding model for AWEs and (2) regularizing the word label embeddings toward pretrained AGWEs.

%% file: text/9_joint_a2w.tex
\chapter{Joint training of whole-word embedding and recognition models}
\label{ch:joint_a2w}

End-to-end acoustic-to-word (A2W) speech recognition systems are conceptually simpler than subword-based systems as they avoid the need for subword decoding. However, large-vocabulary whole-word recognizers have trouble matching the performance of subword-based systems unless very large training sets are available to ensure adequate representation learning for all words in the vocabulary. Recent work~\cite{yu2018multistage,audhkhasi2018a2w,settle2019_a2w,shi2021seg,collobert2019wordlevel,collobert2020word} reduces this gap with pretraining techniques and use of a dynamic word-level lexicon. In this chapter, we extend our prior work on A2W CTC and A2W segmental models from Chapters~\ref{ch:ctc_a2w} and~\ref{ch:seg_a2w}, respectively, by incorporating joint training of AWE and AGWE models alongside A2W recognition. Our work combines the recognition loss with a contrastive loss over the acoustic word embeddings and written word label embeddings implicit in the recognizer. We explore this approach on datasets in the low (Babel) to moderate (Switchboard-$300$) resource regimes, and we find that joint training consistently improves recognition performance, especially for lower-resource settings. Finally, our best A2W CTC-based models are competitive with prior subword CTC models of similar complexity. 

\subsubsubsection{Collaboration} This work relies on collaboration with Bowen Shi and Yushi Hu. Bowen performs all segmental modeling experiments, Yushi Hu performs the final set of Babel experiments, and my contributions focus on everything else (including preliminary tuning, CTC-based experiments, and data preparation).

\section{Introduction}
End-to-end acoustic-to-word (A2W) models for speech recognition predict word labels directly from acoustic input~\cite{audhkhasi2017a2w,audhkhasi2018a2w,yu2018multistage,li2018advancing,settle2019_a2w,gaur2019acoustic_phrase,collobert2019wordlevel,collobert2020word}. When enough data is available, it is feasible to replace the output vocabulary of an end-to-end subword-based system with word labels and maintain competitive performance~\cite{soltau2016a2w,audhkhasi2017a2w,collobert2019wordlevel,collobert2020word}. However, given more modest data resources, training whole-word models is more challenging, due in part to the much larger label inventory of whole-word models with many of the labels seen rarely (or never) in the training data.

Several techniques have been developed to combat these challenges. Yu {\it et al.}~\cite{yu2018multistage} use curriculum learning to incrementally grow the vocabulary, while another approach is to pretrain some or all of the model with phone or character recognition objectives~\cite{audhkhasi2017a2w,audhkhasi2018a2w,yu2018multistage}. In Chapters~\ref{ch:ctc_a2w} and~\ref{ch:seg_a2w}, we find that pretraining the acoustic encoder of an A2W model with an acoustic word embedding (AWE) model and initializing the rows of the prediction layer with acoustically grounded word embeddings (AGWEs) improves recognition performance~\cite{settle2019_a2w,shi2021seg}. Recent work has also begun to address the problem of recognizing words unseen in training. Settle {\it et al.}~\cite{settle2019_a2w} pretrain the prediction layer with a parametric embedding model, which can be used to generate weight vectors for new words at test time, and Collobert {\it et al.}~\cite{collobert2019wordlevel} introduce a \emph{dynamic} lexicon where the parametric embedding model is learned along with the recognition model. This approach has competitive results on LibriSpeech, even when training on only a $100$-hour training subset, compared to subword models at the time~\cite{collobert2019wordlevel,collobert2020word}.

In this chapter, we train whole-word recognizers with both a recognition loss and a contrastive word embedding loss jointly to directly target the discriminability of word representations. We explore multiple types of whole-word recognition modeling, specifically CTC~\cite{settle2019_a2w} and segmental~\cite{shi2021seg}, with both static and dynamic lexicons. We study these models in the context of different languages and data resources, ranging from low-resource Babel languages with as little as $20$ hours of training data~\cite{babel_data} to the medium-resource $300$ hour Switchboard setting. Finally, prior work on pretraining with acoustic word embeddings assumes the existence of ground-truth or high-quality word alignments produced with near-state-of-the-art systems, but in this work we study the effect of relaxing alignment quality and show that we can successfully use alignments produced by a weak monophone hidden Markov model-based system.

The primary contribution of this work is to show that A2W recognition models trained jointly with acoustic and written word embeddings improve over systems trained solely for recognition, and furthermore that these whole-word systems are competitive with CTC-based subword systems on the Switchboard-300h and Babel datasets. This is to our knowledge the first set of competitive results with whole-word models in low-resource multilingual settings.

\section{Approach}

An end-to-end neural acoustic-to-word (A2W) recognition model takes as input a sequence of acoustic frames $\mats{X}$ and outputs a sequence of words $\mats{L}$ supported by a vocabulary $\mathcal{V}$. We consider two recognition approaches from prior work: connectionist temporal classification (CTC)~\cite{audhkhasi2018a2w,yu2018multistage,settle2019_a2w} (Chapter~\ref{ch:ctc_a2w}) and whole-word segmental~\cite{shi2021seg} models (Chapter~\ref{ch:seg_a2w}). Both of these models include an acoustic encoder $f$ and a prediction layer parameterized by a weight matrix $\mats{W}$ whose rows can be seen as word embeddings from the vocabulary. 

CTC-based models~\cite{graves2006connectionist} make one label prediction per frame with output probabilities
\begin{flalign*}
\text{Softmax}(f(\mats{X})^T \mats{W}) \in \mathbb{R}^{T \times \vert\mathcal{V}\vert + 1}
\end{flalign*}
\noindent where $T$ is the number of frames in $\mats{X}$ and $\vert\mathcal{V}\vert + 1$ is the word-level vocabulary size (including the blank symbol $\epsilon$). The model is trained with a marginal log loss that sums over possible alignments between frame labels and word transcripts (Equation~\ref{eq:ctc}).

Segmental models~\cite{zweig2009segmental,shi2021seg} evaluate multi-frame segments to output a score tensor
\begin{flalign*}
f({\bf X})^T {\bf W} \in \mathbb{R}^{T \times S \times \vert\mathcal{V}\vert}
\end{flalign*}
\noindent where $S$ is the maximum allowed segment length and $\mathcal{V}$ is the output vocabulary. The model is trained with a marginal log loss by assigning probabilities to segment paths, conditioned on the input acoustic sequence, normalizing the path scores, and summing over the valid segmentation paths under the model (Equation~\ref{eq:segmental_log_loss}).

\subsection{Parameterization under static vs.~dynamic lexicons}

Following Chapter~\ref{ch:ctc_a2w} and Collobert {\it et al.}~\cite{collobert2019wordlevel,collobert2020word}, we explore A2W recognition using both \emph{static} and \emph{dynamic} lexicon approaches. A \emph{static} lexicon approach refers to na\"ive A2W models with standard matrix-based prediction vocabularies that cannot be extended, while a \emph{dynamic} lexicon refers to a prediction vocabulary which is model-based and can be easily extended. 
The baseline approach using a \emph{static} lexicon optimizes 
\begin{flalign*}
&\mathcal{L}_{asr}(\tens{X}, \tens{L}; \theta^f, {\bf W})
\end{flalign*}
\noindent where $(\tens{X}, \tens{L})$ is a batch of $N$ utterance and word-level transcript pairs, $\theta^f$ are parameters of the acoustic encoder, $\mats{W} \in \mathbb{R}^{\vert\mathcal{V}\vert \times d}$ is the prediction layer weight matrix, and $\mathcal{L}_{asr}$ is the marginal log loss for either a CTC or segmental model. In a \emph{dynamic} lexicon approach~\cite{collobert2019wordlevel}, the prediction layer is constructed by a parameterized word embedding model $g$, so we optimize 
\begin{flalign*}
&\mathcal{L}_{asr}(\tens{X}, \tens{L}; \theta^f, \theta^g)
\end{flalign*}
\noindent where the word-level prediction layer $\mats{W} = \left[ g(v_1); \ldots; g(v_{\vert\mathcal{V}\vert}) \right] \in \mathbb{R}^{\vert\mathcal{V}\vert \times d}$ is now explicitly constructed from $g$. By parameterizing the prediction layer this way, a dynamic lexicon can help to better model rare words, extend the vocabulary to unseen words at test time, and resize the vocabulary for more efficient training.

\subsection{Acoustic and acoustically grounded word embeddings}
\label{ssec:mv}

In Chapters~\ref{ch:ctc_a2w} and~\ref{ch:seg_a2w}, we use jointly trained acoustic word embedding (AWE) and acoustically grounded word embedding (AGWE) models as pretraining to initialize A2W recognition systems and improve their performance, especially for rare words. Now, we investigate this line of work further by applying AWE+AGWE techniques for both pre- {\it and} joint training.

We again pretrain AWE+AGWE models using the multi-view approach~\cite{he2017multiview,settle2019_a2w,shi2021seg}, in which we jointly train both an acoustic ``view" embedding model $f$ and a written ``view" embedding model $g$ using a contrastive loss. Several slight variations on contrastive multi-view training have been used in prior work, but we use the same objective described in Chapter~\ref{ch:seg_a2w}. 
\begin{flalign}
\mathcal{L}_{emb}(\tens{X}, \tens{L}) = \sum_{n=1}^{N}\sum_{i=1}^{\vert \mats{L}^{(n)} \vert} \sum_{obj=0}^{2} \mathcal{L}_{obj}(\mats{X}_i^{(n)}, v_i^{(n)})
\tag{See Equation~\ref{ch:seg_a2w:eq:multiview}}
\end{flalign}
\noindent Multiple other approaches have been proposed for learning acoustic word embeddings and/or acoustically meaningful written word embeddings~\cite{settle2016rnn_awe,kamper2016cnn_awe,settle2017query,chung2016audio,yuan2018learning_awe_for_qbe,audhkhasi2017asr_free_kws,bengio2014word,ghannay2020study}, but we follow the multi-view approach as it directly optimizes discriminability between words using the same dot product as in the A2W recognizers and because of its success in prior work~\cite{settle2019_a2w,shi2021seg}. 

The acoustic encoder $f$ used for A2W CTC has an identical set of trainable parameters $\theta^f$ as the acoustic view embedding model $f$, but differs slightly in its form. This is because, while CTC predictions are made at the frame-level, average pooling over multi-frame segments is necessary to get fixed dimensional embeddings for AWE training. However, when referring to the A2W segmental model, both the acoustic encoder $f$ used for recognition and the AWE model used in pretraining are identical as they both operate on variable-length segments. In all settings, rows of the prediction layer $\mats{W}$ that correspond to special symbols, such as the out-of-vocabulary (OOV) symbol $<$unk$>$ and the blank symbol $\epsilon$, are randomly initialized.

\subsubsection{Pretraining with a static lexicon}
When using a \emph{static} lexicon, we initialize the parameters of $f$ with the pretrained AWE model and the prediction layer weights $W$ with the output of the learned AGWE model, i.e.~$\mats{W}_v = g(v)$ where $\mats{W}_v$ is the row of $\mats{W}$ corresponding to the word $v$, and then train the recognizer with a loss combining the baseline recognition loss with a regularizer that encourages the prediction layer to remain close to the pretrained AGWEs:
\begin{flalign*}
&\mathcal{L}_{asr}(\tens{X}, \tens{L}; \theta^f, \mats{W}) + \lambda_{reg} \sum_{v \in \cup_{n=1}^N \mats{L}^{(n)}} \Vert \mats{W}_v - g(v)\Vert
\end{flalign*}
\noindent Note that the embedding model $g$ is kept fixed during recognizer training. A hyperparameter $\lambda_{reg}$ is used to adjust the contribution of the regularizing term to the overall objective.

\subsubsection{Joint training with a static lexicon}
In the proposed \emph{joint} training approach, we pretrain as above, but add the embedding loss $\mathcal{L}_{emb}$ to recognizer training:
\begin{flalign*}
\mathcal{L}_{asr}(\tens{X}, \tens{L}; \theta^f, \mats{W}) + \lambda_{emb} \mathcal{L}_{emb}(\tens{X}, \tens{L}; \theta^f, \theta^g) + \lambda_{reg} \sum_{v \in \cup_{n=1}^N \mats{L}^{(n)}}\Vert \mats{W}_v - g(v)\Vert
\end{flalign*}
Now $g$ is updated during recognizer training through $\mathcal{L}_{emb}$, so the regularization term encourages the prediction layer to match the {\it current} outputs from $g$ rather than the initialization. Hyperparameters $\lambda_{emb}$ and $\lambda_{reg}$ are added to adjust the contribution of the embedding loss and the regularizing term, respectively, to the overall joint objective.

\subsubsection{Pretraining with a dynamic lexicon}
When using a \emph{dynamic} lexicon, we build the prediction layer from outputs of $g$, which removes the need for a regularization term and we simply optimize with the recognizer loss
\begin{flalign*}
&\mathcal{L}_{asr}(\tens{X}, \tens{L}; \theta^f, \theta^g)
\end{flalign*}
where AWE+AGWE pretraining is used to initialize the parameters $\theta^f$ and $\theta^g$ from the acoustic and written view models $f$ and $g$, respectively. Since we are using the dynamic lexicon approach, both $\theta^f$ and $\theta^g$ will continue to be updated during recognizer training from $\mathcal{L}_{asr}$.

\subsubsection{Joint training with a dynamic lexicon} 
The \emph{joint} setting is again similar, with the addition of $\mathcal{L}_{emb}$:
\begin{flalign*}
&\mathcal{L}_{asr}(\tens{X}, \tens{L}; \theta^f, \theta^g) + \lambda_{emb} \mathcal{L}_{emb}(\tens{X}, \tens{L}; \theta^f, \theta^g)
\end{flalign*}
where parameters are initialized exactly as when pretraining above. Both $f$ and $g$ are then updated based on the joint objectives of AWE+AGWE and ASR rather than ASR alone. Hyperparameter $\lambda_{emb}$ is used to adjust the embedding loss contribution to the overall objective.

\section{Experimental setup}

\subsection{Data}
Experiments are conducted on Switchboard-300h~\cite{godfrey1992switchboard}, LibriSpeech~\cite{panayotov2015librispeech}, and 11 language datasets from Babel~\cite{babel_data}.~\footnote{We use 11 languages (Cantonese (zh), Assamese (as), Bengali (bn), Pashto (ps), Turkish (tr), Tagalog (tl), Tamil (ta), Zulu (zu), Lithuanian (li), Guarani (gu), and Igbo (ig)), and the training data available for each language ranges from 20-120 hours.} The input acoustic features used for Switchboard (108-dimensional), LibriSpeech (240-dimensional), and Babel (108-dimensional) are log-Mel spectra $+\Delta+\Delta\Delta$s extracted using Kaldi~\cite{povey2011kaldi}.  For Babel, 3-dimensional pitch features and their first and second derivatives are also used. We use frame stacking over pairs of consecutive frames, giving inputs that are 216-, 480-, and 234-dimensional, respectively. Word alignments are needed for all experiments that use AWE and AGWE models. For Switchboard, both ground-truth and monophone alignments are used (see details below), and for Babel we use triphone alignments. All monophone and triphone alignments used are produced by the respective Kaldi recipes. The vocabulary size $\vert\mathcal{V}\vert$ is $20$k for Switchboard, $74$k for LibriSpeech,  and ranges from $4$k to $12$k for Babel languages. These vocabulary sizes correspond to minimum word occurrence counts in their respective training sets of $2$ for Switchboard, $1$ for LibriSpeech, and $3$ for each of the Babel languages. We use no language models in any of our experiments. As in prior work~\cite{conneau2020unsupervised,inaguma2019transfer}, we use the released Babel development data as the test set since the eval data has not been released, and we construct a heldout development set from $10\%$ of the training data. We report the word error rate (WER) given by the NIST SCTK scoring toolkit. PyTorch~\cite{pytorch} is used for all experiments.

\subsection{Acoustic-to-word (A2W) speech recognizers}

\begin{figure}
  \centering
\includegraphics[width=0.725\linewidth]{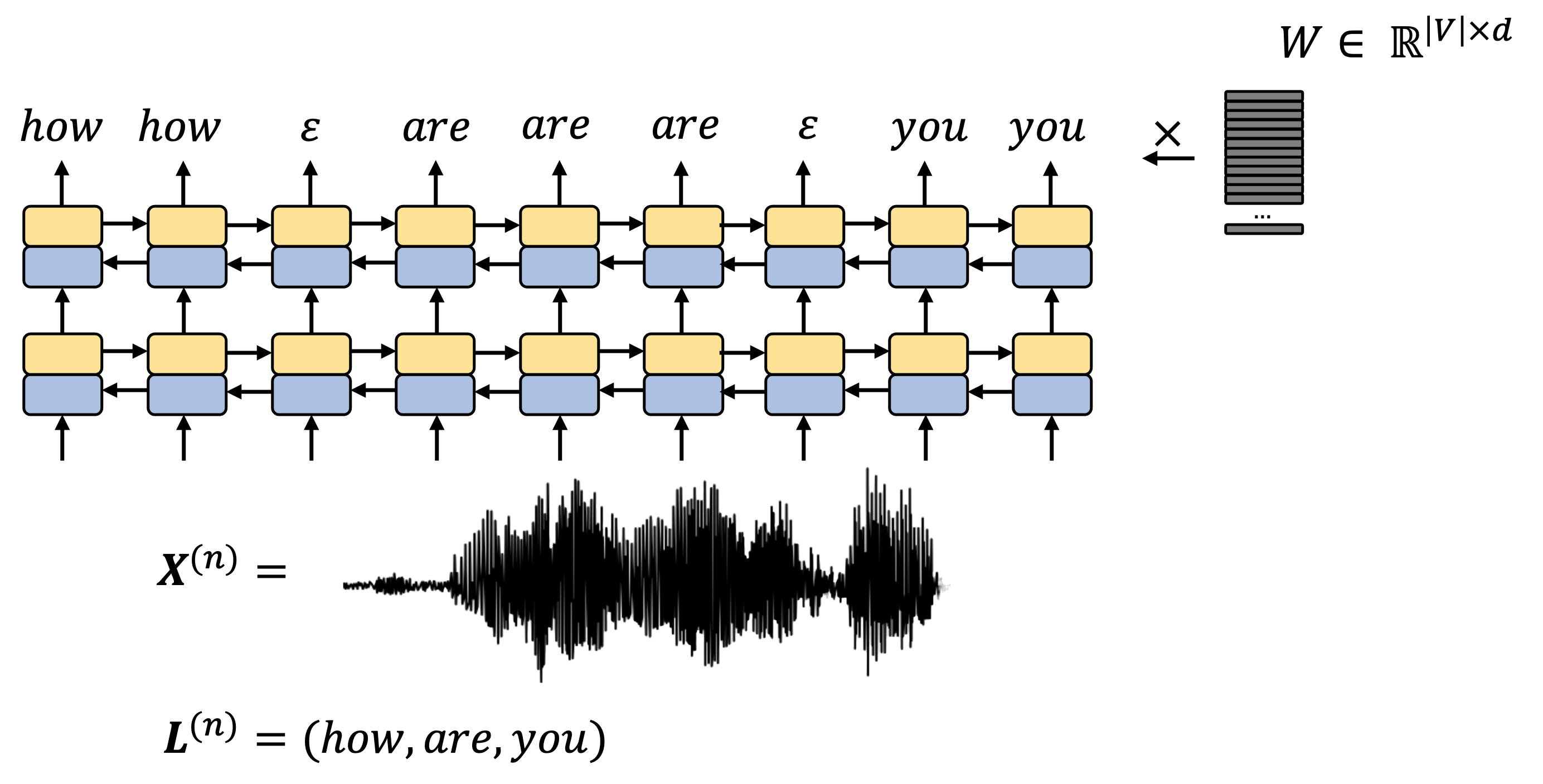}
  \caption{Acoustic-to-word (A2W) CTC modeling for speech recognition.}
\label{ch:joint_a2w:fig:ctc}
\end{figure}

\subsubsection{Connectionist temporal classification (CTC)-based}

The encoder (and acoustic ``view" model)  $f$ is a $6$-layer bidirectional long short-term memory~\cite{hochreiter1997lstm} (BiLSTM) network (subsampled by $2$ after layer $3$), with $512$-dimensional hidden states, dropout of $0.4$, and followed by a $128$-dimensional projection layer. The AGWE model (i.e.~written view) $g$ consists of a character (for English) or phone (for Babel languages) embedding layer to map input tokens into 32-dimensional embeddings, a 1D convolutional layer with kernel width $3$ and dimension $512$, max-pooling over time, and a $128$-dimensional linear layer.

Following~\cite{collobert2019wordlevel}, $10$k words are randomly sampled to construct the prediction layer during each iteration of CTC training. During AWE+AGWE training, the minimum acoustic segment duration is $8$ frames, the margin $m$ is $0.45$, and the number of offending examples per word starts at $k=128$ and reduces by $1$ per batch until $k=6$. An AWE for a given spoken word segment is contextual as it is the average over projection layer outputs from $f$ within the word boundaries according to the alignments. AWE+AGWE training follows directly from Chapter~\ref{ch:seg_a2w} with one update. The vocabulary $\mathcal{V}$ considered for each batch now consists not only of the unique word labels from that batch but also an additional $200$ randomly sampled word labels from outside the batch to better approximate the full vocabulary.

The optimization schedule is the same for CTC and AWE+AGWE training. Variable batching is used with a $20000$-frame maximum where each utterance is $\geq8$ frames. We use the Adam~\cite{kingma2014adam} optimizer with learning rate $5 \times 10^{-4}$ and weight decay $10^{-4}$ ($1.25 \times 10^{-4}$ for the prediction layer). The maximum gradient norm is $10$. The learning rate is reduced by a factor of $0.7$ when held-out performance does not improve in $5$ epochs. Training stops when the learning rate is below $5 \times 10^{-6}$ (or $5 \times 10^{-7}$ when using SpecAugment~\cite{park2019specaugment}).

Early stopping and hyperparameter tuning for AWE+AGWE training are done according to development set performance on a cross-view word discrimination task, measured with average precision (AP), as in previous work on AWE+AGWEs~\cite{he2017multiview,settle2019_a2w}, and according to development set WER for CTC training. Further details on this tuning criterion can be found in the background under preliminaries in Section~\ref{ch:back:prelims:crossview_ap}.

For experiments on Babel data, we use only the \emph{static} lexicon approach, and we consider two variants of pretrained AWE and AGWE models: monolingual and multilingual. Monolingual AWE+AGWE pretraining involves training 11 setups separately, one for each language. Multilingual AWE+AGWE pretraining involves training one setup model with all 11 languages, following from~\cite{hu2020multilingual}. For Babel experiments, the AGWE model $g$ uses phone sequences as input following Hu {\it et al.}~\cite{hu2020multilingual} to allow for straightforward multlilingual AWE and AGWE pretraining across all Babel languages~\cite{hu2020multilingual}. All fine-tuning for Babel ASR is monolingual, including for joint AWE+AGWE and ASR training.

\begin{figure}
\centering
\includegraphics[width=0.7\linewidth]{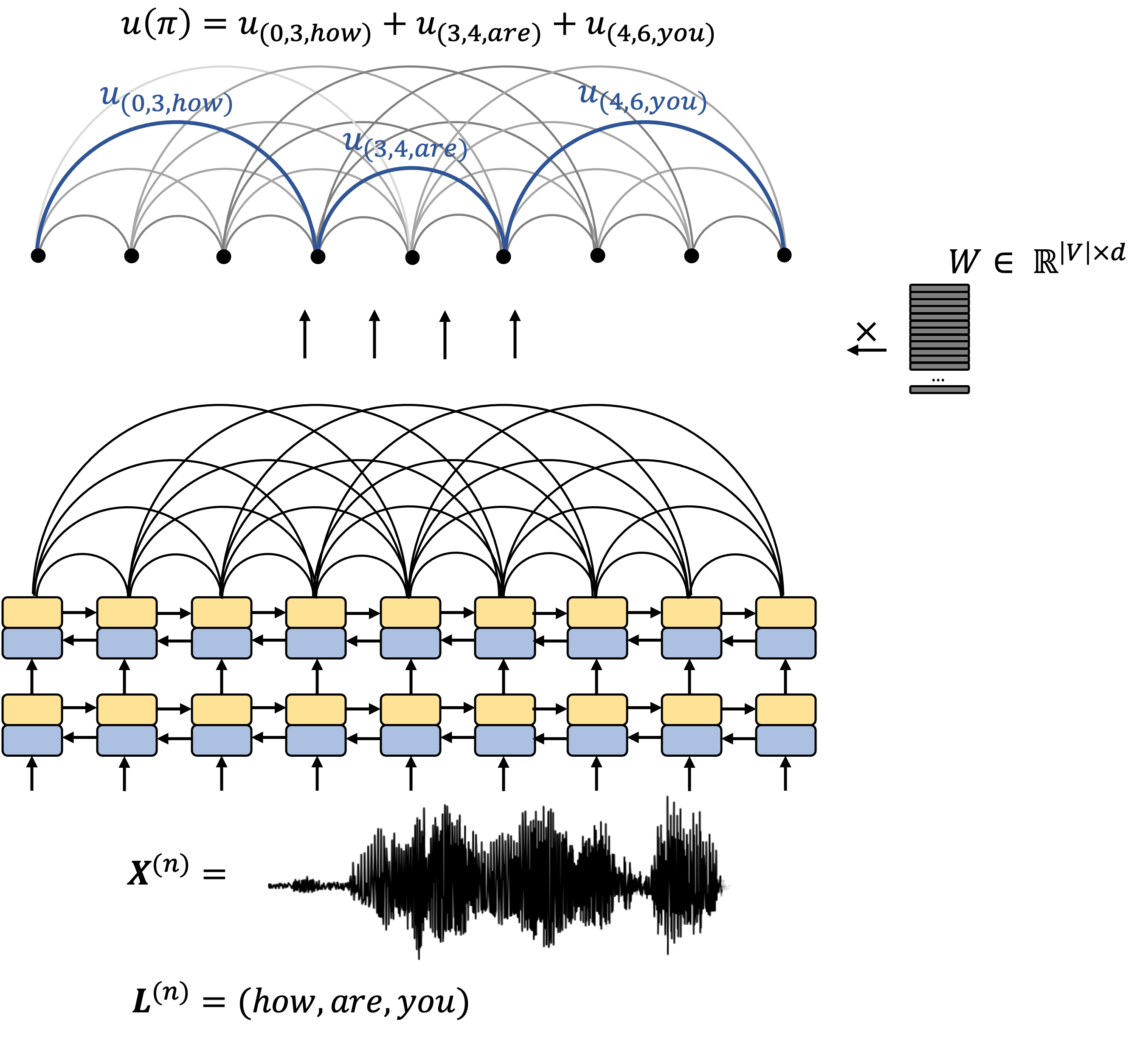}
\caption{Acoustic-to-word (A2W) segmental modeling for speech recognition.}
\label{ch:joint_a2w:fig:segmental}
\end{figure}

\subsubsection{Segmental modeling-based}

Similarly to the A2W CTC experiments, the backbone encoder of the acoustic view is a 6-layer BiLSTM with 512-dimensional hidden states with dropout of $0.45$ added between layers. Following Chapter~\ref{ch:seg_a2w}, a convolutional layer of kernel size 5 followed by average pooling with stride 4 is added on top of the BiLSTM. Acoustic segment embeddings result from concatenating the convolutional layer outputs at the two endpoints of the segment and passing this vector through an additional $512$-dimensional linear layer followed by a ReLU nonlinearity. The AGWE model $g$ is the same as in the A2W CTC model described above. The maximum segment length is $32$ frames, corresponding to an approximately $2.4s$ maximum word duration, and we reduce the maximum segment size per batch during training to $\min\{2*\max\{\frac{\text{input length}}{\text{\# words}}\}, 32\}$. We randomly sample only $3$k words from the vocabulary to construct the prediction layer in each batch. For AWE+AGWE pretraining, we set the minimum allowable acoustic segment duration to $8$ frames, the margin $m$ in $\mathcal{L}_{emb}$ to $0.45$ and the number of offending examples used for each word segment to $k=6$. The model is trained for $40$ epochs with the Adam optimizer~\cite{kingma2014adam} with $24$-utterance mini-batches and an initial learning rate of $5\times 10^{-4}$. The learning rate is halved when the development set WER stops descreasing for 2 epochs. The gradients are clipped to maximum norm of $25$. 

In AWE+AGWE pretraining for the A2W segmental model, we use a variable batch size with up to $18,000$ frames per batch. We use early stopped based on the cross-view AP on the development set (Section~\ref{ch:back:prelims:crossview_ap}). The optimization and scheduling scheme are the same as for ASR training.

\section{Results}

This section details our experimental results. Subsections are broken down by dataset. We start with LibriSpeech for A2W-only results to establish that our methods are competitive with prior and concurrent work~\cite{collobert2019wordlevel,collobert2020word}. Next, we move to comparisons on Switchboard where the bulk of prior work, including our own, is covered. Finally, we perform low-resource experiments a variety of Babel languages to showcase how far these methods have come with the additions of AWE+AGWE pre- and joint training.

\subsection{Librispeech}

\begin{table}[H]
\caption{A2W CTC \emph{baseline} word error rates (WERs) on LibriSpeech.}
\label{tab:libri}
\begin{center}
\begin{tabular}{lrrrr}
\toprule
{\bf System}  &\multicolumn{2}{c}{\bf dev} &\multicolumn{2}{c}{\bf test}\\
              & clean & other & clean & other\\
\midrule
dynamic~\cite{collobert2019wordlevel}  & 5.6 & 15.5 & 5.5 & 16.0\\
dynamic (transformer)~\cite{collobert2020word} & 3.0 & 7.7 & 3.2 & 7.7\\
\midrule
\emph{static} (\emph{baseline}) & 5.4 & 14.4 & 5.6 & 14.8\\
\emph{dynamic} (\emph{baseline}) & 5.2 & 15.1 & 5.5 & 15.3\\
\bottomrule
\end{tabular}
\end{center}
\end{table}

\subsubsection{Baseline comparisons of static vs. dynamic lexicon performance}

Table~\ref{tab:libri} compares our A2W CTC baselines with prior work~\cite{collobert2019wordlevel,collobert2020word} introducing the dynamic lexicon approach. Our baseline \emph{static} and \emph{dynamic} lexicon models perform competitively with the more comparable original work~\cite{collobert2019wordlevel}, but lag behind the follow-up work which utilizes a transformer-based model architecture. The \emph{static} lexicon approach outperforms the \emph{dynamic} lexicon on the noisier (other) dev and test partitions, while underperforming it on the clean subset. This hints at the tendency for the \emph{dynamic} lexicon approach where the prediction layer is model-based has a higher tendency to overfit. 

\subsection{Switchboard-300h}

\subsubsection{Effect of training set size}
\begin{figure}[H]
  \centering
  \includegraphics[width=0.65\linewidth]{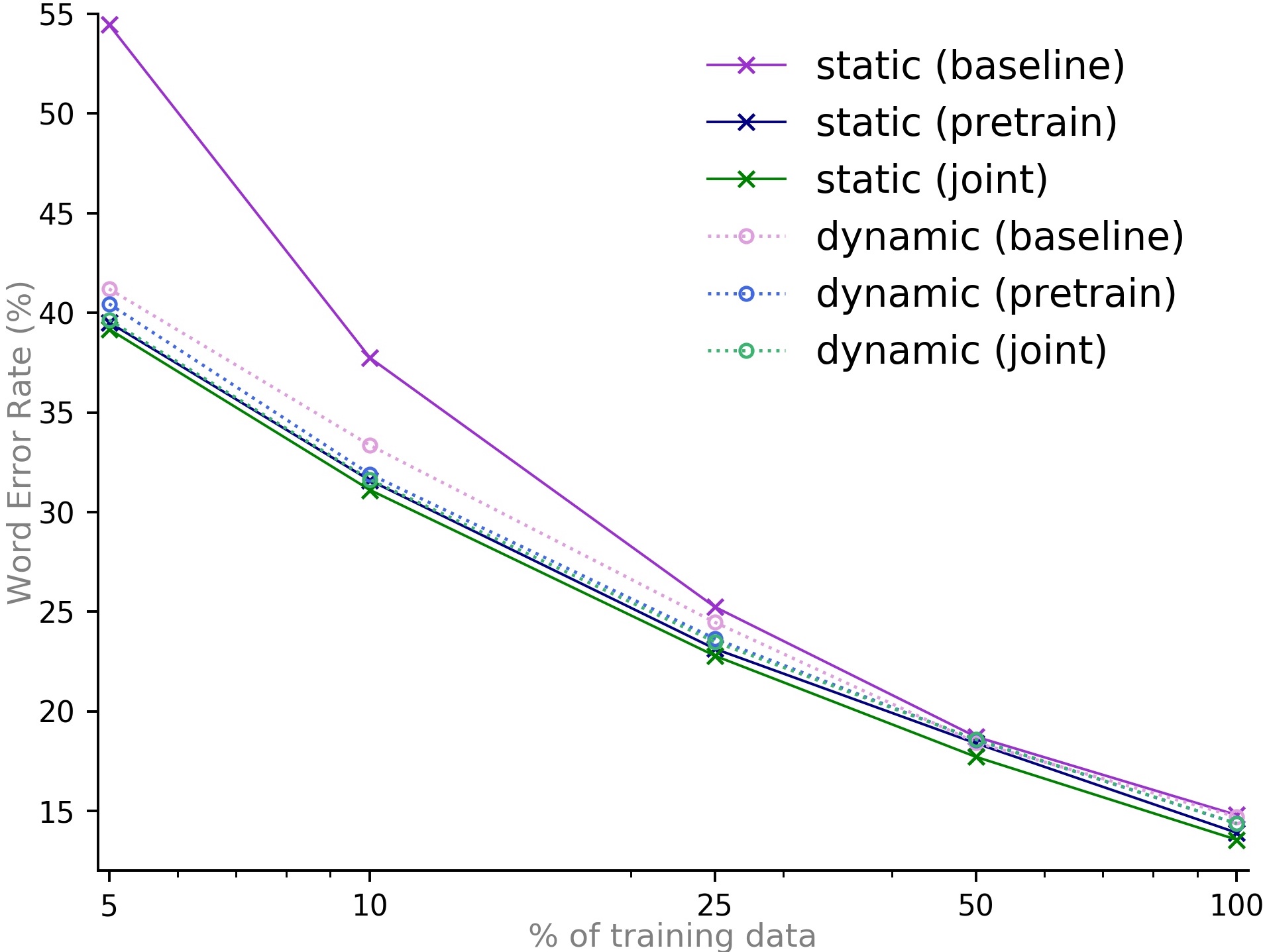}
  \caption{Switchboard development set WER when varying \%  of training data.}
\label{fig:vary_data}
\end{figure}

Figure~\ref{fig:vary_data} shows the effect on Switchboard development set WER of training the CTC-based A2W models with varying amounts of training data.  For the \emph{static} lexicon model, AWE+AGWE pretraining and joint training has a much more dramatic effect as the training set size is reduced.  Since the \emph{dynamic} lexicon baseline model is stronger overall than the \emph{static} baseline across training set sizes, the \emph{dynamic} approach correspondingly improves less with embedding-based pre- and joint training.

Despite the strength of the baseline \emph{dynamic} system, the single best-performing system at all operating points is the \emph{static} model with joint training, which motivates its use in the notably low-resource Babel experiments with results detailed in Section~\ref{ch:joint_a2w:results:babel}.

\subsubsection{Effect of word alignment quality}
One of the challenges of multi-view AWE+AGWE pretraining methods is the need for word-level alignments. Table~\ref{tab:align} studies the effect of using ground-truth (human-annotated) word alignments versus forced alignments obtained with recognition systems of differing quality. Despite disparities in alignment model quality, even monophone alignments computed from a small portion of the training set~\footnote{See Kaldi recipe for further details.} offer similar benefits on downstream recognition performance as ground-truth word alignments. While cross-view word discrimination performance (AP) is impacted in absolute terms, this does not seem to affect downstream ASR performance in terms of WER. When using SpecAugment~\cite{park2019specaugment}, performance improves further, both on the word discrimination task as well as ASR. This is likely because of the nature of SpecAugment that we are often masking out large portions of words, so minor accuracy improvements in word boundaries are less consequential.

\begin{table}[H]
\caption{Switchboard development set WER vs. cross-view average precision (AP) when pretraining with word-level alignments of differing quality.}
\label{tab:align}
\begin{center}
\begin{tabular}{llll}
\toprule
{\bf System} & {\bf Alignment} & {\bf AP} &{\bf WER}\\
\midrule
\emph{baseline} & n/a && 14.8\\
\emph{joint} & monophone & 0.81 & 13.8 \\
\hspace{0.5cm}+SpecAug & monophone & 0.85 & 12.4\\
\emph{joint} & ground-truth & 0.88  & 13.6 \\
\bottomrule
\end{tabular}
\end{center}
\end{table}

\subsubsection{Effect of AWE+AGWE training}
Table~\ref{tab:swbd} shows our main Switchboard-300h results demonstrating the effect of word embedding pretraining and joint training on multiple types of A2W recognizers. 

While our baseline \emph{static} and \emph{dynamic} A2W CTC systems are already competitive with prior work on A2W models, AWE+AGWE pre- and joint training improves them further. AWE+AGWE training has a larger effect on the \emph{static} lexicon model as it produces absolute WER improvements $\geq1\%$ over the baseline on both the SWB and CH test sets with pretraining and another $\geq0.5\%$ with joint training. The \emph{dynamic} approach sees less improvement, with absolute reductions of $0.4\%$ and $0.8\%$ for SWB and CH, respectively, from pretraining and even smaller benefits from joint training. However, the \emph{dynamic} approach to A2W CTC is the best performing baseline indicating the benefits of the AGWE-style representation of the word-level vocabulary.

\begin{table}[H]
\caption{Word error rates (WER) on Switchboard-$300$h development set and Eval2000 test sets. All our models here use SpecAugment~\cite{park2019specaugment} unless otherwise stated. A2W CTC \emph{pretrain} and \emph{joint} experiments use monophone alignments, while A2W Segmental \emph{pretrain} and \emph{joint} use ground-truth alignments.}
\label{tab:swbd}
\begin{center}
\begin{tabular}{lllllll}
\toprule
{\bf System}             & \multicolumn{3}{l}{\emph{static}} & \multicolumn{3}{l}{\emph{dynamic}}\\
                         & dev & SWB & CH & dev & SWB & CH\\
\midrule
Phone CTC (SF)~\cite{audhkhasi2019forget} & - & 10.6 & 19.5 &&\\
 +sMBR+LM~\cite{audhkhasi2019forget} & - & 9.6 & 17.7 &&\\
+speed perturb.~\cite{audhkhasi2019forget} & - & 9.1 & 17.4 &&\\
\midrule
A2W MultiStage-CTC~\cite{yu2018multistage} & - & 11.4 & 20.8 &&\\
A2W CTC+AWE~\cite{settle2019_a2w} & - & 13.7 & 23.8 &&\\
A2W Seg+AWE~\cite{shi2021seg} & - & 11.9 & 21.2 &&\\
\midrule
\underline{A2W CTC}&&&&&&\\
\emph{baseline w/o SpecAug}\tablefootnote{Note even without SpecAugment the \emph{static} baseline is strong relative to prior A2W work.} & 14.8 & 11.8 & 21.7 &&\\
\emph{baseline} & 14.2 & 11.5 & 20.1 & 13.8 & 11.1 & 20.0\\
\emph{pretrain} & 13.1 & 10.5 & 18.9 & 13.4 & {\bf 10.7} & {\bf 19.2}\\
\emph{joint} & 12.4 & {\bf 10.0} & {\bf 18.3} & 13.6 & 11.0 & 19.4\\
\underline{A2W Seg}&&&&&&\\
\emph{baseline} & 18.6 & 15.1 & 25.3 & 18.3 & 14.7 & 24.8\\
\emph{pretrain}  & 15.6 & 12.2 & 21.2 & 14.8 & 11.9 & 20.7\\
\emph{joint}  & 14.3 & 11.4 & 20.5 & 13.8 & 10.9 & 19.8\\
\bottomrule
\end{tabular}
\end{center}
\end{table}

Compared to A2W CTC models, the A2W segmental model benefits more from AWE+AGWE training both with \emph{static} and \emph{dynamic} lexicons, which implies a higher tendency to overfit when training segmental models from scratch. In contrast to CTC, the segmental models that use a \emph{dynamic} lexicon gain more from AWE+AGWE training than those that use a \emph{static} lexicon. This is potentially because in A2W segmental models $f$ (the AWE model) is already built directly into the segment score function such that when using a \emph{dynamic} lexicon, segmental model scoring with $f$ and $g$ now perfectly match the form of embedding comparisons within AWE+AGWE pre- and joint training. Overall our joint training method outperforms the previous best A2W segmental model by $\sim 1\%$ (absolute) without increasing model size; however, we find that our joint A2W CTC model with a \emph{static} lexicon is the highest performing model.

The top rows of Table~\ref{tab:swbd} also includes the best prior results we are aware of with an end-to-end subword-based recurrent CTC recognizer on this domain, showing that our best models are only slightly behind despite not using a language model or other add-ons such as sMBR training.  There are much better subword-based results, for example using transformers~\cite{wang2020investigation}, but we leave such improvements to future work.

\subsection{Babel languages}
\label{ch:joint_a2w:results:babel}

\subsubsection{Incorporating multilingual pretraining into low-resource recognition}

Following from the best A2W setup in Figure~\ref{fig:vary_data}, 
Table~\ref{tab:babel_ctc_a2w} shows the performance of our A2W CTC systems on a number of Babel datasets~\footnote{We use 11 languages (Cantonese (zh), Assamese (as), Bengali (bn), Pashto (ps), Turkish (tr), Tagalog (tl), Tamil (ta), Zulu (zu), Lithuanian (li), Guarani (gu), and Igbo (ig)), and the training data available for each language ranges from 20-120 hours.} when trained with a \emph{static} lexicon. To the best of our knowledge, these are the first results for A2W recognition applied to a variety of low-resource languages. Our first two systems, \emph{baseline} and \emph{joint}, are monolingual systems trained in the same way as our English \emph{static} lexicon \emph{baseline} and \emph{joint} training models. Joint training always improves performance, with larger improvements for lower-resource languages. In addition, we experiment with multilingual pretraining of embedding models, that is pretraining on the union of all of the languages' training sets, as suggested in prior work~\cite{hu2020multilingual}, which almost always further improves performance.

\begin{table}[H]
\small
\caption{Word error rates (WER) on Babel test sets. Our models use SpecAugment.}
\hspace{-1cm}\begin{tabular}{lccccccccccc}
    \toprule
    Lang & zh & as & bn & ps & tr  & tl & ta & zu & li & gu & ig\\
    \# hrs & 122 & 50 & 51 & 68 & 68 & 74 & 56 & 52 & 33 & 34 & 33\\
    \midrule
    \textbf{Prior} (subword) &&&&&&&&&&&\\
    \footnotesize{mono BLSTM-CTC~\cite{inaguma2019transfer}}&&64.5&&&&56.4&&69.5&&&\\
    \footnotesize{multi-15 + RNNLM~\cite{inaguma2019transfer}} &&53.4&&&&46.1&&58.8&&&\\
    \footnotesize{BLSTM-HMM + LM~\cite{inaguma2019transfer}}  &&49.1&&&&46.3&&61.1&&&\\
    \footnotesize{XLSR-10 (large) (no LM)~\cite{conneau2020unsupervised}}&&49.1&&&&40.6&&&\\
    \midrule
    \textbf{Ours} (A2W CTC) &&&&&&&&\\
    \emph{baseline} & 44.7 & 57.9 & 59.0 & 48.3 & 52.8 & 47.7 & 69.5 & 65.9 & 57.9 & 59.1 & 65.6\\
    \emph{joint} (monolingual pretrain) & 43.1 & \textbf{50.6} & 53.7 & 44.6 & 48.7 & 47.1 & 64.2 & 60.1 & 54.0 & 51.6 & 58.8\\
    \emph{joint} (multilingual pretrain) & \textbf{42.0}& 50.9 & \textbf{53.0} & \textbf{44.5} & \textbf{47.5} & \textbf{43.6} & \textbf{63.1} & \textbf{59.3} & \textbf{50.9} & \textbf{50.0} & \textbf{57.4}\\
    \bottomrule
  \end{tabular}
\label{tab:babel_ctc_a2w}
\end{table}

Compared to previous work on several of the languages~\cite{inaguma2019transfer, conneau2020unsupervised},\footnote{These are the only three of our Babel languages for which we are aware of pubished WERs (for other languages only character error rates are published).} our \emph{joint} models for Assamese outperform subword-based end-to-end models, while coming close to the best hybrid model~\cite{inaguma2019transfer} as well as a recent multilingual transformer approach~\cite{conneau2020unsupervised} when evaluated without a language model. Our results on Tagalog are also competitive, and are outperformed only by~\cite{conneau2020unsupervised}. On Zulu we outperform all but one model that also uses an external language model. Overall, we find that with the use of monolingual or multilingual pre-training coupled with monolingual fine-tuning through jointly training AWE+AGWE and CTC, we can build simple and competitive A2W recognition models for a number of low-resource Babel languages.

\section{Conclusion}
This work further closes the gap between whole-word and subword recognition systems, improves on the best previous whole-word recognition results on  Switchboard-300h, and establishes strong baselines for whole-word models applied to a variety of low-resource Babel languages.  We find that whole-word models, whether using a \emph{static} or \emph{dynamic} lexicon, benefit from pretraining and/or joint training with acoustic and acoustically grounded word embeddings. Future work includes extending this work to transformer-based models, scaling up the training data alongside our modeling improvements, and applying whole-word models beyond speech recognition to natural language understanding.

%% file: text/10_ssl_awe.tex
\chapter{Comparison to self-supervised speech representations}
\label{ch:ssl_awe}

Self-supervised representation learning has taken off within the speech research community~\cite{mohamed2022ssl_review} and offers significant benefits to downstream modeling across multiple tasks. These self-supervised approaches address the reliance on significant labeled resources when training high-performing models for speech recognition as well as other popular tasks such as query-by-example speech search, keyword search, and speech diarization. Given the aforementioned benefits of acoustic word embedding (AWE) techniques within resource constrained domains, we now compare our AWE methods with the most prominent self-supervised pretrained models of recent literature, including Wav2Vec2~\cite{baevski2020wav2vec2}, HuBERT~\cite{hsu2021hubert}, and WavLM~\cite{chen2022wavlm}. In this chapter, we use these models (1) to perform acoustic word discrimination using their output representations out-of-the-box and (2) as input features to our multi-view RNN training approach (from Chapter~\ref{ch:joint_a2w}).

\section{Introduction}

Acoustic word embeddings (AWEs) are an efficient alternative to dynamic time warping (DTW) for performing segment comparison with high precision and recall, most notably for query-by-example speech search. However, DTW is most frequently used in domains with limited labeled training data resources. While unsupervised AWE approaches~\cite{kamper2019truly_unsupervised_awe,holzenberger2018unsupervised_awe} continue to improve, there are still significant performance gaps when compared with supervised methods~\cite{settle2016rnn_awe,he2017multiview} since approaches using off-domain supervised pretraining (such as many multilingual methods~\cite{hu2020multilingual,kamper2020multilingual}) often outperform unsupervised training done directly on the target domain data. 

Self-supervised learning (SSL)~\cite{mohamed2022ssl_review} offers performance benefits to a variety of speech tasks as these approaches provide powerful pretrained representations from entirely unlabeled audio data that can then be fine-tuned on much less data for downstream applications. Three prominent and popular approaches in this area include Wav2Vec2~\cite{baevski2020wav2vec2}, HuBERT~\cite{hsu2021hubert}, and WavLM~\cite{chen2022wavlm}. These models learn high quality representations capturing both phone- and word-level information at different layers within the models~\cite{pasad2021layerwise}. 

With the success of these pretraining approaches, the community has already started to investigate their effectiveness in popular tasks~\cite{yang2021superb}, but their their utility as AWEs has not been explored. Van Staden and Kamper~\cite{van2021ssl_comparison_for_awe} compare different self-supervised methods for input feature learning, namely contrastive predictive coding (CPC)~\cite{oord2018cpc}, autoregressive predictive coding (APC)~\cite{chung2019apc}, and frame-level correspondence auto-encoders (CAE), and use these self-supervised input features to learn zero-resource AWEs using unsupervised CAE-RNN~\cite{kamper2019truly_unsupervised_awe} training. Van Staden and Kamper find improvement with self-supervised frame-level features over using traditional mel-frequency cepstral coefficients (MFCCs) as input features. Jacobs {\it et al.}~\cite{jacobs2021acoustic} explore alternative self-supervised learning methods for AWEs, particularly in the multilingual context. Prior multilingual work~\cite{hu2020multilingual,kamper2020multilingual} finds that when the target language does not have available labeled data, supervised AWE training on non-target languages outperforms unsupervised AWE training on the target language alone. Jacobs {\it et al.} find further performance improvement with an adaptation stage following supervised non-target language training where contrastive training on the target language is done using word segments pairs derived from an unsupervised term discovery (UTD) system.

Concurrent with and most similar to our own work, Sanabria {\it et al.}~\cite{sanabria2022analyzing} compare AWEs produced from self-supervised model outputs with prior unsupervised AWEs on English, Xitsonga, Mandarin, and French data. Sanabria {\it et al.} generate AWEs by taking the frame-level outputs of Wav2Vec2~\cite{baevski2020wav2vec2} and HubERT~\cite{hsu2021hubert} in combination with dimensionality reduction, subsampling, and pooling techniques to produce fixed-dimensional embeddings. Sanabria {\it et al.} find HuBERT~\cite{hsu2021hubert} and Wav2Vec2~\cite{baevski2020wav2vec2,conneau2020unsupervised} capture significant sequential and contextual information that allows for model outputs to be used effectively as embeddings within the acoustic word discrimination task, and offer competitive performance with other purpose-built unsupervised AWE techniques.

In this chapter, we investigate the effectiveness of combining these frame-level self-supervised model representations with simple pooling operations to learn AWEs with strong acoustic word discrimination performance. In addition, we show that even further gains can be made when fine-tuning for acoustic word discrimination using our multi-view contrastive objective with very little labeled data on top of the self-supervised frame-level representations.

\section{Preliminaries}

This section provides a brief background on the motivation and training for the self-supervised models we adopt in this chapter. While there are also subtle architectural differences between Wav2Vec2, HuBERT, and WavLM models, this section will focus on the primary differences in their training setups. Further transformer architecture differences as well as datasets used for self-supervised pretraining will be described in Section~\ref{ch:ssl_awe:setup}. Each model contains a convolutional frontend acoustic encoder that accepts raw waveforms as input and produces learned feature frames every $20$ms. Additional lower-level details such as differences in these frontend encoders are left to the relevant source material (i.e. Wav2Vec 2.0~\cite{baevski2020wav2vec2}, Wav2Vec 2.0 XLSR-53~\cite{conneau2020unsupervised}, HuBERT~\cite{hsu2021hubert}, and WavLM~\cite{chen2022wavlm}).

\subsection{Wav2Vec 2.0}
\label{ch:ssl_awe:wav2vec2}

The Wav2Vec2~\cite{baevski2020wav2vec2} model consists of a frontend encoder followed by a transformer network. Following the frontend encoder, frame-level outputs are passed both to the transformer network as well as to a vector quantization module. Similar to the methods popularized by SpecAugment~~\cite{park2019specaugment} and BERT~\cite{devlin2018bert}, randomized masking of spans of these frame-level features is incorporated before passing through the transformer network but not when passing to the quantizer. The quanitized outputs are then used within a contrastive loss that compares them with the corresponding outputs from the transformer, but the loss considers only output positions that were masked before passing through the transformer.  Additionally, a diversity loss is used to encourage the use of a variety of codes within the vector quantization codebook. One key point to keep in mind is that throughout training the loss is derived from comparison between the transformer outputs and the relatively low-level quantized representations.

\subsection{HuBERT}
\label{ch:ssl_awe:hubert}

Similar to Wav2Vec2~\cite{baevski2020wav2vec2}, HuBERT~\cite{hsu2021hubert} is composed of a convolutional frontend encoder followed by a transformer network. However, the self-supervised training loss and methodology differs. Instead of using a vector quantization module, offline clustering is done on the input frame-level acoustic features (i.e. MFCCs) to get the frame-level cluster IDs to use as training targets for self-supervision. As during Wav2Vec2 training, randomized masking is applied between the frontend encoder and the transformer model. The frame-level transformer outputs are used to predict the IDs from offline clustering, and the cross entropy loss is computed for the frames corresponding to the masked spans. Periodically during training, offline clustering is recomputed (and the cluster IDS updated) using frame-level feature representations extracted from intermediate layers within the transformer network. These periodic updates of the cluster IDs provide another important distinction between HuBERT and Wav2Vec2 as it means that the targets of the loss are derived from higher-level representations as training progresses, while loss targets are derived from lower-level features throughout Wav2Vec2 training.

\subsection{WavLM}
\label{ch:ssl_awe:wavlm}

Building on the work of Wav2Vec2~\cite{baevski2020wav2vec2} and HuBERT~\cite{hsu2021hubert}, WavLM~\cite{chen2022wavlm} makes further modifications to differentiate itself and improve performance on a variety of downstream applications (e.g., SUPERB~\cite{yang2021superb}). Outside of small changes to the transformer module architecture used by HubERT, the primary change to training is in the incorporation of input utterance mixing either with a random utterance or with noise. Mixing in combination with masked prediction improves model robustness to the quality and domain of the acoustic training data. This update to training also widens the scope of potential downstream applications through implicit incorporation of denoising and speech separation as subtasks of masked prediction.

\section{Approach}

\begin{figure}
\centering
\includegraphics[width=0.45\linewidth]{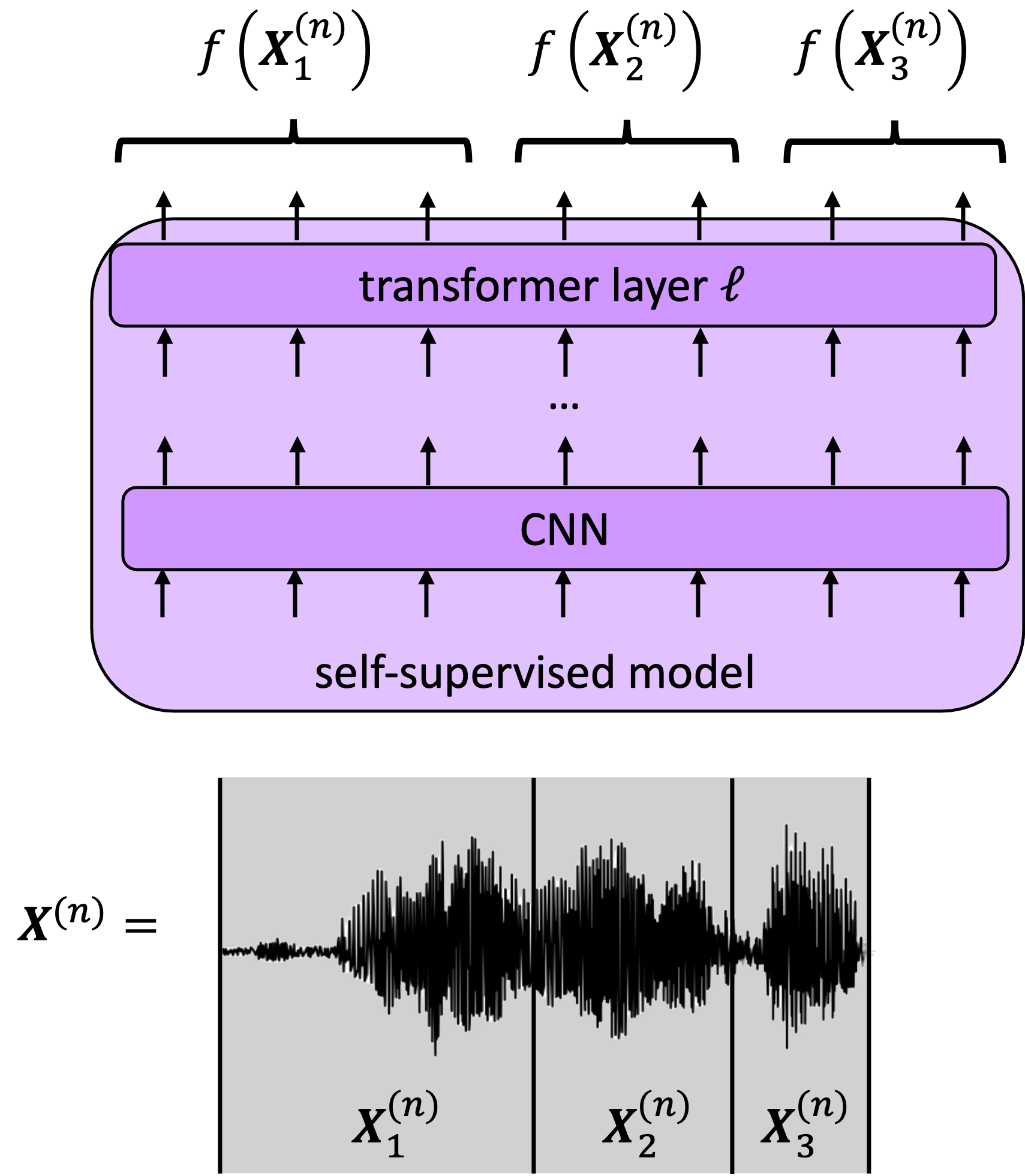}
\caption{Acoustic word embeddings (AWEs) extracted using self-supervised pretrained models out-of-the-box by mean-pooling over word segments.}
\label{ch:ssl_awe:out-of-the-box}
\end{figure}

\begin{figure}
\centering
\includegraphics[width=0.825\linewidth]{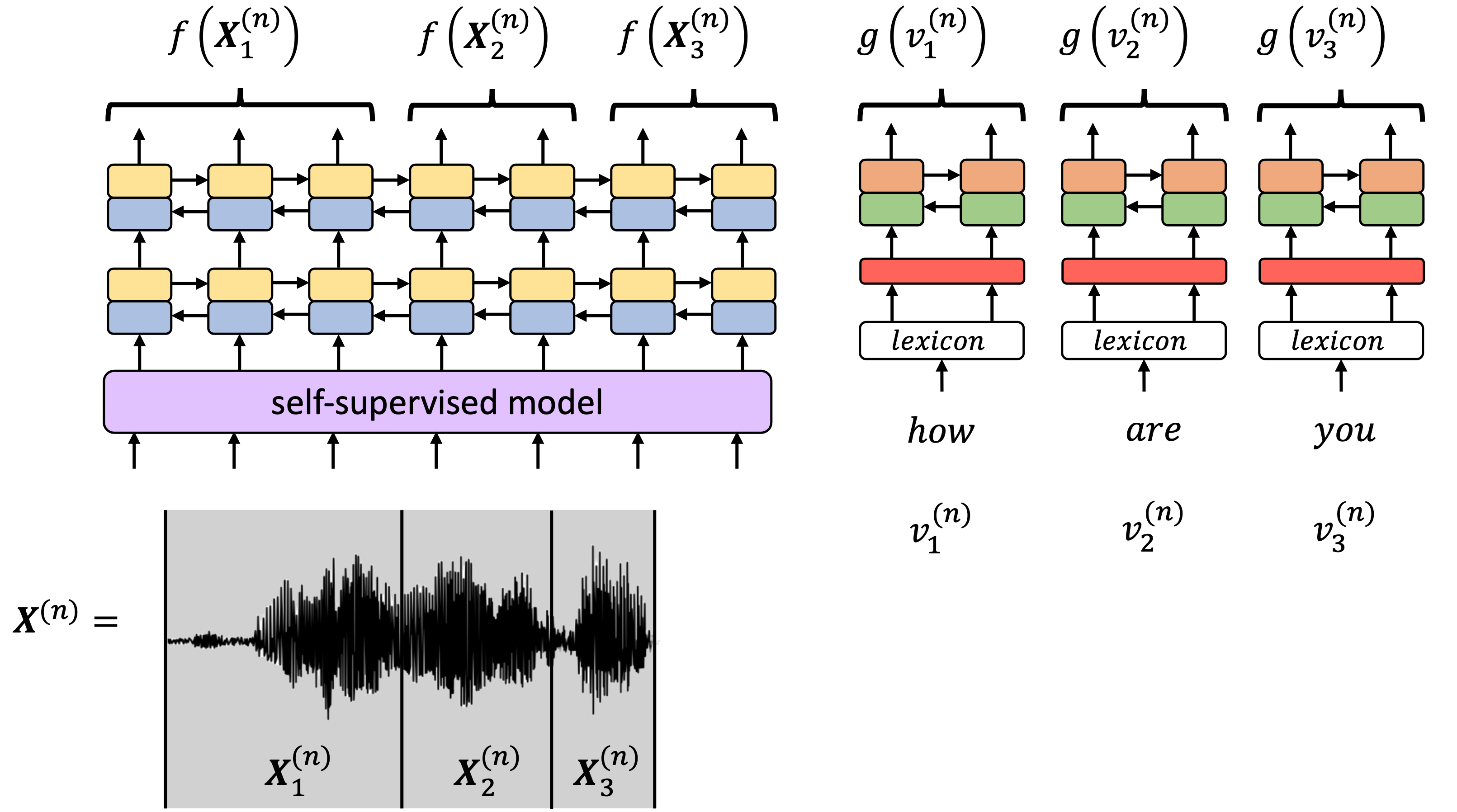}
\caption{Acoustic word embeddings (AWEs) extracted by multi-view training of RNNs on top of frame-level features output by self-supervised pretrained models.}
\label{ch:ssl_awe:fine-tuned}
\end{figure}

The goal of our work is to study the utility of pretrained self-supervised representations in acoustic word discrimination. We investigate this in two ways. The first approach (1) uses the pretrained models out-of-the-box, and their frame-level representations are mean-pooled according to pre-segmented word boundaries to obtain word-level representations (Figure~\ref{ch:ssl_awe:out-of-the-box}). When evaluating the pooled AWEs, we compare frame-level outputs across different transformer layers and models. The second method (2) investigates using the frame-level output representations as {\it input} features (Figure~\ref{ch:ssl_awe:fine-tuned}). The best performing layer representations of each model in (1) are used as inputs within our best RNN-based multi-view method from Chapter~\ref{ch:joint_a2w}. We then compare these learned AWEs to those learned with the same multi-view approach but using log-Mel spectral features as input.

\section{Experimental setup}
\label{ch:ssl_awe:setup}

We now detail the experimental setup, which includes specifying the self-supervised transformer models as well as defining the hyperparameters of our multi-view approach. 

\subsection{Model details}

\subsubsection{Wav2Vec 2.0}

We explore two pretrained model variants of Wav2Vec 2.0, namely {\it wav2vec2-base} and {\it wav2vec2-large-xlsr-53}. The {\it wav2vec2-base} model is composed of a convolutional frontend followed by a 12-layer transformer model architecture, and it is trained on the canonical $960$ hours LibriSpeech~\cite{panayotov2015librispeech} audio corpus of read English speech using the self-supervised loss described in Section~\ref{ch:ssl_awe:wav2vec2}. The {\it wav2vec2-large-xlsr-53} model is composed of a convolutional frontend\footnote{Note the convolutional frontend used for {\it wav2vec2-base}~\cite{baevski2020wav2vec2} and {\it wav2vec2-large-xlsr-53}~\cite{conneau2020unsupervised} differ, but not in a way that necessitates detailing here. For such details, see Baevski {\it et al.}~\cite{baevski2020wav2vec2} and Conneau {\it et al.}~\cite{conneau2020unsupervised}.} followed by a 24-layer transformer model architecture, and it is trained using the same general procedure as {\it wav2vec-base} (further details for this approach can be better found here~\cite{conneau2020unsupervised}). The data used to train {\it wav2vec2-large-xlsr-53}, however, is drawn from Multilingual LibriSpeech~\cite{pratap2020mls}, CommonVoice~\cite{ardila2019commonvoice}, and Babel~\cite{babel_data}, which includes $56,000$ hours of speech data spanning $53$ languages.

\subsubsection{HuBERT}

We explore only the base model variant of HuBERT (i.e. {\it hubert-base}). This model, similar to {\it wav2vec2-base}, is again composed of a convolutional frontend encoder followed by a 12-layer transformer model, and it is trained using the self-supervised loss described above in Section~\ref{ch:ssl_awe:hubert} on the same LibriSpeech-$960$h~\cite{panayotov2015librispeech} corpus used by Wav2vec2.

\subsubsection{WavLM}

We explore two model variants of the WavLM approach, including {\it WavLM-base} and {\it WavLM-large}. {\it WavLM-base} is another 12-layer transformer approach trained only on LibriSpeech-$960$h, which makes for a fair comparison with both {\it wav2vec2-base} and {\it hubert-base}. Of the three self-supervised approaches we investigate, WavLM is the most recent and highest performing on the SUPERB~\cite{yang2021superb} benchmarks, so we chose {\it WavLM-large} to investigate how increases to training data and model size may impact performance on acoustic word discrimination. The {\it WavLM-large} network includes a convolutional frontend followed by a 24-layer transformer, and it is trained on $94$k hours of English speech data drawn from LibriLight~\cite{kahn2020librilight} audiobook data ($60$k hours), English-only data from VoxPopuli~\cite{wang2021voxpopuli} ($24$k hours), and a subset of the GigaSpeech~\cite{chen2021gigaspeech} data corpus ($10$k hours). This $94$k hours of data is English-only, but it does include both read and spontaneous speech coming from a variety of settings.

\subsubsection{Multi-View RNN}

To best contextualize the performance of these self-supervised approaches with respect to our own contributions, we include our own multi-view RNN approach from Chapter~\ref{ch:joint_a2w}. We train our multi-view RNN approach using both traditional input acoustic features (log-Mel spectra) and the best-performing layer's frame-level outputs from each pretrained self-supervised model. Table~\ref{ch:ssl_awe:tab:hyperparams} details the non-default hyperparameters used in our experiments. Additional multi-view training details follow directly from Chapters~\ref{ch:seg_a2w} and~\ref{ch:joint_a2w} with the same incorporation of phone sequences when training on Babel data as in Chapter~\ref{ch:multi_qbe}.

\begin{table}
\centering
\small
\caption{Hyperparamters (non-defaults) used for Multi-View RNN experiments with log-Mel spectra input features as well as the pretrained self-supervised learning model output features.}
\begin{tabular}{lcc}
\toprule
{\bf Hyperparameters}   &   \multicolumn{2}{c}{input features}\\
                        &   log-Mel spectra & SSL model outputs\\
\midrule
\underline{AWE model (GRU)} &&\\
\# layers & 6 & 2\\
hidden size & 512 & 512 \\
input size & 240 & 768 \text{ (base) /} 1024 \text{ (large)}\\
dropout & $0.4$ & $0.4$ \\
\underline{AGWE model (GRU)} &&\\
\# layers & 2 & 1\\
hidden size &   512 & 512\\
input size &  64 & 64\\
dropout & $0.4$ & n/a\\
\underline{Loss} & &\\
$m$ (margin) & 0.45 & 0.45\\
$k$ (negative sample set size) & 8 & 8\\
\# extras (word labels outside batch added to $\mathcal{V}$) & 200 & 200\\
\underline{Optimization (Adam~\cite{kingma2014adam})} & &\\
$\eta$ & 0.0005 & 0.001\\
$\lambda$ ($L_2$ weight decay) & 0.0001 & 0.0001\\
max. grad norm & 10 & 10\\
\underline{Scheduler (ReduceLROnPlateau~\cite{pytorch})} &&\\
$\gamma$ ($\eta$ decay) & 0.1 & 0.5 \\
min. $\eta$ & $10^{-7}$ & $10^{-6}$\\
patience & 4 & 4\\
\bottomrule
\end{tabular}
\label{ch:ssl_awe:tab:hyperparams}
\end{table}

\subsection{Datasets}

\subsubsection{Training}

We only perform training for the Multi-View RNN models, which are indicated by the horizontal lines in Figures~\ref{ch:ssl_awe:fig:librispeech},~\ref{ch:ssl_awe:fig:swbd}, and~\ref{ch:ssl_awe:fig:babel}. This training is done on small subsets of the LibriSpeech~\cite{panayotov2015librispeech}, Switchboard~\cite{godfrey1992switchboard}, and Babel~\cite{babel_data} datasets. For each dataset, different word segment restrictions are used for training and evaluation, which are detailed in Table~\ref{ch:ssl_awe:tab:data}.

\subsubsection{Evaluation}

Our experiments involve model evaluation on the development sets of three datasets: 
LibriSpeech~\cite{panayotov2015librispeech} (both {\it clean} and {\it other}), Switchboard (the acoustic word discrimination partition from Chapters~\ref{ch:rnn_awe} and~\ref{ch:multi_awe}), and Babel~\cite{babel_data} (our development sets\footnote{It should be noted that the test set has not been released, so the officially released development set from Babel~\cite{babel_data} is often used as a test set (see Chapters~\ref{ch:multi_awe} and~\ref{ch:joint_a2w}), but here we are using our own partitioned development set taken from the training set.}). As in Chapters~\ref{ch:multi_awe},~\ref{ch:multi_qbe}, and~\ref{ch:joint_a2w}, the same 11 Babel languages are used, including Cantonese, Assamese, Bengali, Pashto, Turkish, Tagalog, Tamil, Zulu, Lithuanian, Guarani, and Igbo.

\begin{table}[H]
\centering
\small
\caption{Training sets are constructed according to a word segment duration range as well as a minimum word occurrence count in the training vocabulary.}
\begin{tabular}{lccccccc}
\toprule
{\bf Dataset}   & total (hours)   & \multicolumn{2}{c}{dur. (seconds)}     & \multicolumn{2}{c}{min train occ. count} & \multicolumn{2}{c}{approx. vocab size}\\ 
                & train               & train & eval                          & train & eval & train & eval\\
\midrule
LibriSpeech & 5.0 & [0.25, 3.0] & [0.5, 2.0] & 1 & 0 & $9$k & $5$k\\
Switchboard & 1.8 & [0.5, 2.0] & [0.5, 2.0] &  1 & 0 & $2$k & $4$k\\
Babel       & [8.3, 22.3] & [0.25, 5.0] & [0.5, 5.0] & 3 & 0 & $2$-$10$k & $0.25$-$2.5$k\\
\bottomrule
\end{tabular}
\label{ch:ssl_awe:tab:data}
\end{table}

\section{Results}
\label{ch:ssl_awe:results}

\subsection{LibriSpeech}

\begin{figure}
\centering
\includegraphics[width=0.725\linewidth]{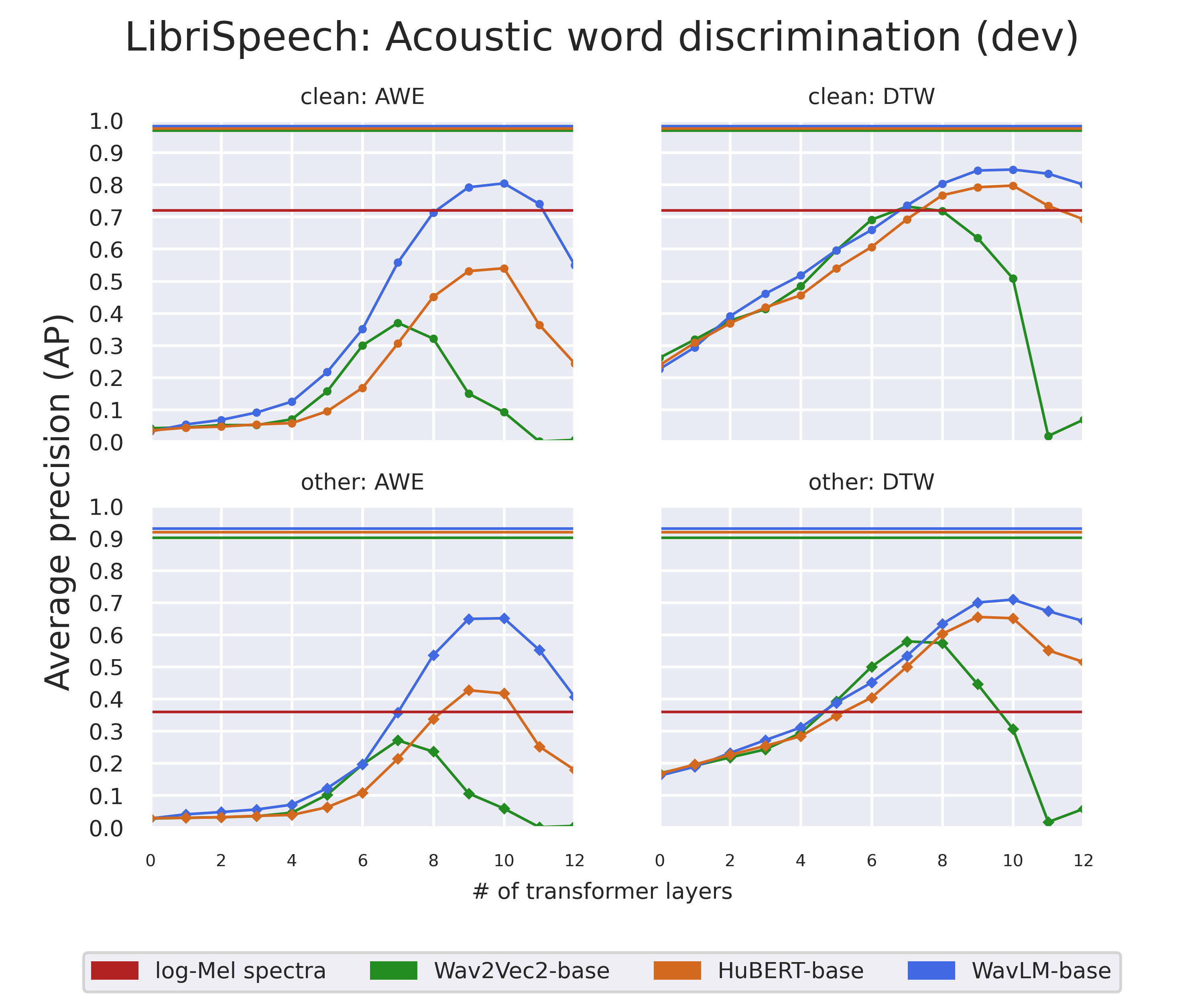}
\caption{Acoustic word discrimination results on the ``clean" and ``other" LibriSpeech development sets. Horizontal lines indicate Multi-View RNN training with color indicating the input features used during training.}
\label{ch:ssl_awe:fig:librispeech}
\end{figure}

Our first set of results on LibriSpeech measure in-domain AWE performance. We compare our multi-view approach with the base model variants of the three self-supervised approaches. In Figure~\ref{ch:ssl_awe:fig:librispeech}, AWE indicates acoustic word discrimination results using cosine distance computed between embeddings and DTW indicates results when using dynamic time warping distance between variable-length speech segments. On both the ``clean" and ``other" subsets of the LibriSpeech development set, we find HuBERT and WavLM style cross entropy training to outperform Wav2Vec2, with WavLM outperforming HuBERT and Wav2Vec2 at every layer for this task. The performance of both HuBERT and WavLM peaks around layers 9 and 10, while Wav2Vec2 acoustic word discrimination performance peaks earlier at layer 7. The peaks for all models are consistent across the ``clean" and ``other" subsets of the development set.

Since HuBERT and WavLM use later layers to produce cluster IDs as training progresses, their targets become higher level allowing them to better capture word-level information. In contrast, Wav2Vec2 uses quantized outputs produced from the convolutional frontend acoustic encoder, and so its targets are from a lower level layer throughout training. While higher word-level information appears to be extracted in the middle layers in all cases, using training targets from middle layers as training progresses appears to help learn information relevant for acoustic word discrimination. This insight suggests that the frame-level information learned by Wav2Vec2 is likely lower level with a higher acoustic granularity than that of HubERT and WavLM. In line with this thinking, the DTW results pictured in column two of Figure~\ref{ch:ssl_awe:fig:librispeech} indicate that the performance gap between the features from the three models can be reduced by using dynamic time warping over the speech segments rather than using simple pooling to produce AWEs. This suggests that a significant amount of word discriminating information is still present in the frame-level features produced by Wav2Vec2 and HuBERT that can be recovered by using DTW. WavLM, however, appears to be implicitly aggregating the same information across nearby frames since there is not as large of an improvement in average precision (AP) when moving from the AWE to the DTW approach here. 

When comparing with our Multi-View RNN approach trained with log-Mel spectra features, we see that in the case of ``clean", our model outperforms the embeddings produced by pooling outputs from Wav2Vec2 and HuBERT, but is surpassed in performance by WavLM. In the case of ``other", our approach outperforms Wav2Vec2 again, but falls just behind HuBERT and further from WavLM. When using DTW to compare pretrained segment representations, we find that now Wav2Vec2 is also competitive with our multi-view AWEs on ``clean", while HuBERT and WavLM representations surpass them. Meanwhile, all models using DTW on ``other" consistently outperform our embeddings at the peak performing layers. The strong performance of these unsupervised methods relative to our supervised approachc is likely due in part to their training on the LibriSpeech-$960$h dataset since this makes their training and this evaluation incredibly well-aligned by acoustic domain. By comparison, the multi-view RNN approach using log-Mel features is trained on only a small $10$ hour subset drawn primarily from the clean portion of the training set, so while it is supervised, it is also trained on a much smaller portion of the training set.

In addition to our baseline Multi-View RNN approach using log-Mel spectra inputs, we also investigate the use of model outputs from Wav2Vec2, HuBERT, and WavLM as indicated in Figure~\ref{ch:ssl_awe:fig:librispeech} by the horizontal lines matching the colors of each method in the legend. We find that the effectiveness of our multi-view approach combined with the quality of the frame-level features output by these models produces significant improvements in AP. Training the multi-view RNN with these pretrained representations outperforms all other approaches and records acoustic APs of approximately $0.97$ and $0.92$ on the ``clean" and ``other" sets, respectively.

\subsection{Switchboard}

\begin{figure}
\centering
\includegraphics[width=0.725\linewidth]{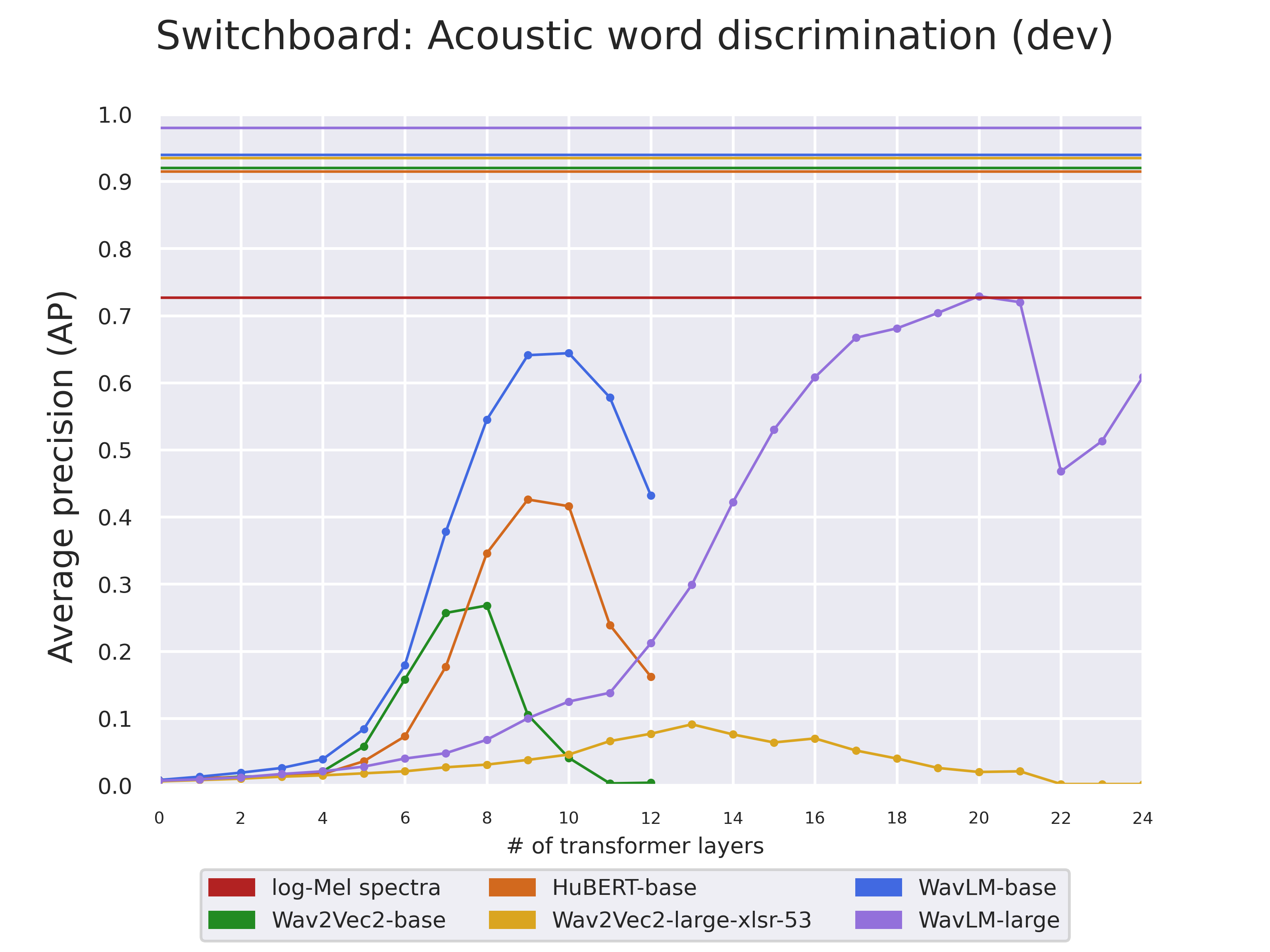}
\caption{Acoustic word discrimination results on the development set of the Switchboard word discrimination partition. Horizontal lines indicate Multi-View RNN training with color indicating the input features used during training.}
\label{ch:ssl_awe:fig:swbd}
\end{figure}

Figure~\ref{ch:ssl_awe:fig:swbd} shows model performance on conversational English data with Switchboard acoustic word discrimination. With the exception of our Multi-View RNN approach, none of the models are trained using the Switchboard data, though portions of {\it wavlm-large} include other conversational and spontaneous English speech. We include all of the models in this comparison since it has been the primary dataset of choice in this thesis. We see that {\it wav2vec2-base}, {\it hubert-base}, and {\it wavlm-base}, as well as their layers, roughly maintain their performance ordering from Figure~\ref{ch:ssl_awe:fig:librispeech}. Also, we see that the training approach of WavLM benefits from increasing the model size as well as expanding the training data as {\it wavlm-large} sees the best performance among all models tested, even managing to match our own Multi-View RNN (log-Mel spectra) performance. We consider the multilingual variant of Wav2Vec2 here as well ({\it wav2vec2-large-xlsr-53}), and see that despite an increase in model size the domain mismatch between multilingual pretraining and English-only evaluation leads to poor generalization to this task.

When using the frame-level pretrained model outputs within our Multi-View RNN, we again see significant improvements in Switchboard acoustic word discrimination performance as all such models surpass $0.9$ AP and {\it wavlm-large} now reaching over $0.97$ AP. These results greatly surpass the prior best results reported earlier in Table~\ref{ch:multi_awe:tab:baselines}, and indicate significant potential for the combination of these techniques to be applied to not only word discrimination but also query-by-example search and acoustic-to-word speech recognition as well.

\subsection{Babel}

\begin{figure}
\centering
\includegraphics[width=0.75\linewidth]{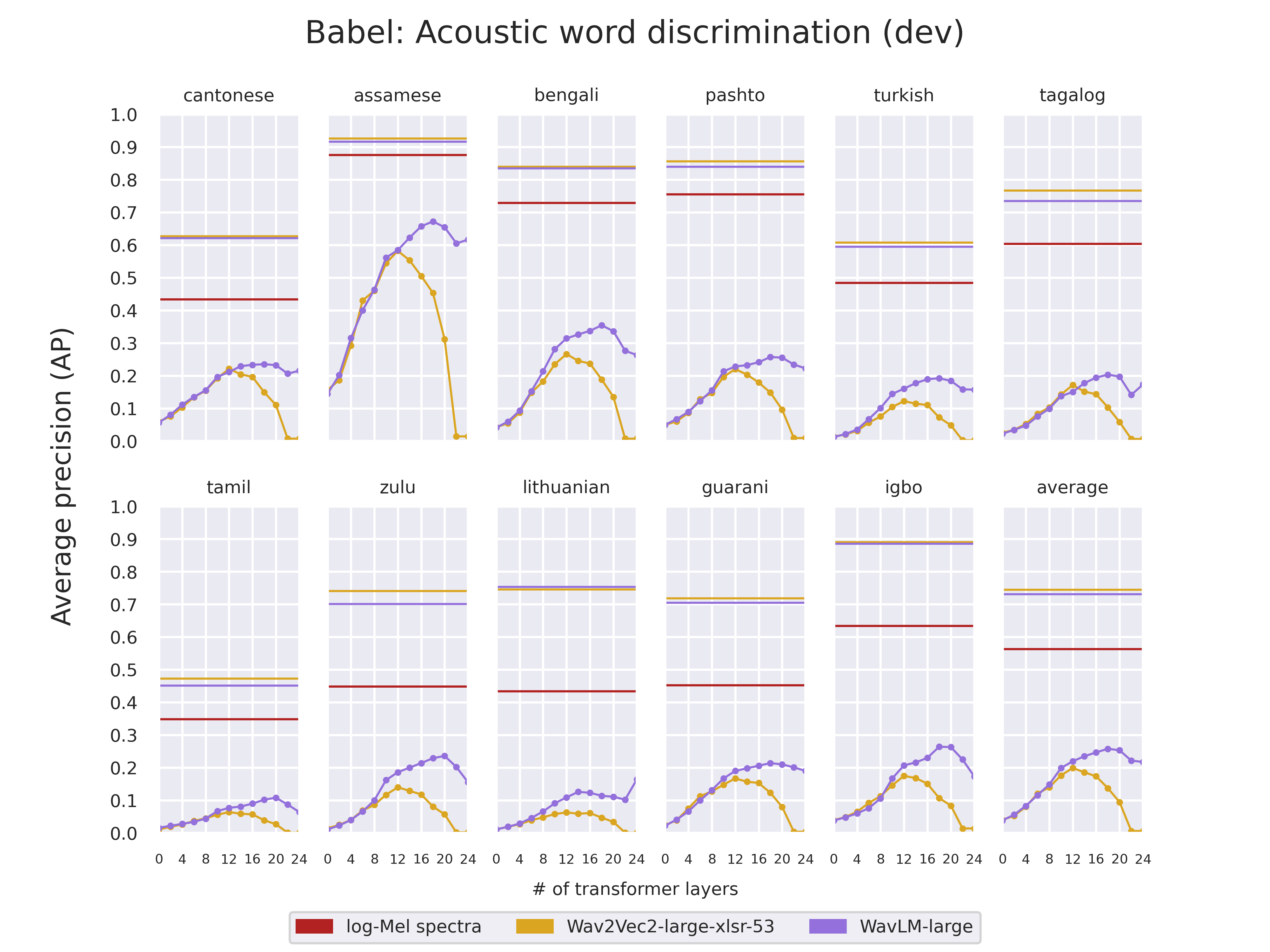}
\caption{Acoustic word discrimination results across 11 babel languges as well as average performance. Horizontal lines indicate Multi-View RNN training with color indicating different input features used during training.}
\label{ch:ssl_awe:fig:babel}
\end{figure}

Our final experiments consider the application of self-supervised models to acoustic word discrimination across conversational speech data from a number of languages. In particular, we compare the multilingually trained Wav2Vec2 model {\it wav2vec2-large-xlsr-53} with the English-only trained WavLM model {\it wavlm-large}\footnote{Note that {\it wav2vec2-large-xlsr-53} is trained on $56,000$ hours of multilingual data, while {\it wavlm-large} is trained on $94,000$ hours of English-only speech data.}. In Figure~\ref{ch:ssl_awe:fig:babel}, we see that despite the potential domain mismatch between the pretraining and testing conditions of {\it wavlm-large}, we find that {\it wavlm-large} consistently outperforms {\it wav2vec2-large-xlsr-53}. On average the gap between the best layer performances is about $0.1$ AP, with the peak {\it wavlm-large} and {\it wav2vec2-large-xlsr-53} acoustic word discrimination performances around $0.3$ and $0.2$ AP, respectively. Despite training only on English, WavLM continues to perform well relative to Wav2Vec2 even in this multilingual setting.

When using multi-view RNN training, we find that our AWEs, even with log-Mel spectra input features, significantly outperform the AWEs created from pooled frame-level model representations. Furthermore, the multi-view AWEs learned from the pretrained model outputs give an additional boost of almost $0.2$ AP on average over using log-Mel spectra features. This indicates that our model performance in Chapter~\ref{ch:multi_qbe} on the QUESST 2015 task could be well improved with the use of either {\it wav2vec2-large-xlsr-53} or {\it wavlm-large} features during training. Also, since the domain of our evaluation is a better match for {\it wav2vec2-large-xlsr-53}, the modeling improvements of {\it wavlm-large} make less of an impact on performance after multi-view fine-tuning. The AWEs learned with frame-level outputs from either {\it wav2vec2-large-xlsr-53} or {\it wavlm-large} perform similarly on average around $0.77$ AP, though {\it wav2vec2-large-xlsr-53} consistently can be seen edging out {\it wavlm-large} performance on many languages.

\section{Conclusion}

We apply several pretrained self-supervised models to the task of acoustic word discrimination, and compare them with our low-resource (but supervised) multi-view RNN training approach.

Overall, the best performing AWEs by a significant margin come from using the pretrained self-supervised model features as input to our multi-view training approach. In this setting, the performance is quite high on English regardless of which pretrained features are used (typically close to $1$ AP for LibriSpeech and $>0.9$ AP for Switchboard), and much higher than our log-Mel spectra trained multi-view models when applied on Babel (by approximately $0.2$ AP). In general, WavLM features (either from {\it wavlm-base} or {\it wavlm-large}) are best whether evaluating using only the pooled representations or training with the multi-view approach on top. Due to the domain mismatch of WavLM pretraining in the multilingual setting, it is possible to get similar or slightly better performance when multi-view training by using outputs from {\it wav2vec2-large-xlsr-53} but if one set of features needs to be chosen {\it wavlm-large} appears to have the most utility across domains.

%% file: text/11_conc.tex
\chapter{Conclusion}

We conclude by summarizing our primary contributions, including model design and training of acoustic and acoustically grounded word embeddings (AWEs/AGWEs) as well as their application to downstream query-by-example (QbE) speech search and acoustic-to-word (A2W) speech recognition. Additionally, we identify possible next steps for future work, in particular following our most recent exploration combining multi-view AWE modeling with self-supervised representation learning.

\section*{Acoustic and acoustically grounded word embeddings}

We pioneer RNN-based single-view AWEs and multi-view AWE+AGWEs that outperform prior approaches on the acoustic and cross-view word discrimination tasks (Sections~\ref{ch:back:prelims:acoustic_ap} and~\ref{ch:back:prelims:crossview_ap}). Our best single-view AWE results are when using deep LSTM RNNs with several recurrent and fully-connected layers optimized with a contrastive Siamese loss (Chapter~\ref{ch:rnn_awe}). Siamese RNN training is convenient because it can be done in a zero-resource setting by using unsupervised term discrovery (UTD) to provide example pairs. However, when labels are available, even for a very small training set, we find that significant improvements can be seen when training using both an acoustic view (accepting acoustic segment features as input) and a written view (accepting a character sequence as input). This approach allows for data efficient learning of high-quality AWE models with considerable improvement in acoustic word discrimination performance over prior work. Our work extends this powerful approach to many languages to produce a multilingual AWE model that can be effectively fine-tuned in low-resource settings and used out-of-the-box in zero-resource settings (Chapter~\ref{ch:multi_awe}). In addition, by learning an additional model for the written view of the data, we can extend our approach to be used for not only query-by-example speech search, but also text-based keyword search and speech recognition tasks as well.

\section*{Query-by-example speech search}

Our first embedding-based approach (Chapter~\ref{ch:rnn_qbe}) to QbE speech search is the first such method to use neural AWEs, following prior work with template-based AWEs from Levin {\it et al.}~\cite{levin2015srails}. Our RNN-based AWEs outperform template-based AWEs both on the acoustic word discrimination task as well as QbE speech search, offering the first comparison of neural and template-based AWEs on both tasks. This work highlights the effectiveness of neural AWEs not only for acoustic word discrimination but also for downstream application to QbE search. Our approach offers significant performance benefits to precision and recall over template-based AWEs with the same benefits in querying speed over dynamic time warping (DTW)-based methods.

Building on the success of our neural AWE approach to English QbE speech search, we further investigate the application of our AWEs to a more challenging setting. Using our novel multilingual joint training approach (Chapter~\ref{ch:multi_awe}) of AWEs and AGWEs, we are also the first to apply these models in the multilingual setting (Chapter~\ref{ch:multi_qbe}) with more challenging query definitions (e.g., approximate matches) as well as varying speech conditions (e.g., noisy audio). While these conditions create complications to DTW-based QbE search, often necessitating the ensembling of many different systems, our single multilingual multi-view AWE model can still be competitive with many DTW approaches. Then, by introducing a straightforward modification to our training approach, we allow the learning of both single- and multi-word acoustic span embeddings (ASE). This innovation allows our models to even outperform the DTW ensembles, particularly on the more challenging query settings. Finally, both our AWE and ASE approaches are trained on a number of languages with available language data, but are zero-resource with respect to the languages used in the multilingual QbE test set (QUESST 2015).

\section*{Acoustic-to-word speech recognition}

Our contributions to acoustic-to-word (A2W) speech recognition target two model training paradigms: frame-level prediction of whole-words with connectionist temporal classification (CTC) and segment-level prediction of whole-words with segmental modeling. Additionally, we consistently improve A2W recognition performance by using pretrained AWE and AGWE models. We use the AWE model weights as an initialization of the acoustic encoder, and we use the AGWEs to initialize the row vectors in the word-level prediction layer weight matrix. Our first work (Chapter~\ref{ch:ctc_a2w}) also showcases that the prediction layer can actually be frozen after initialization to improve performance in the low-vocabulary regime, while allowing for out-of-vocabulary prediction using the AGWE model to generalize beyond the training vocabulary. Our second work (Chapter~\ref{ch:seg_a2w}) shows that pretraining is even more crucial within the segmental modeling paradigm, allowing us to achieve state-of-the-art results on conversational speech recognition for A2W recognizers at the time. 

Beyond pretraining, our follow-up work (Chapter~\ref{ch:joint_a2w}) explores the use of both pretraining and joint training with the multi-view objective. We find consistent benefits when using joint training for CTC-based and segmental model training. Due to the similarity in the structure of the segmental model function and the AWE model used for multi-view training, we find the segmental model to benefit most significantly from pre- and joint training. However, despite this alignment between the segmental and multi-view modeling approaches, the jointly trained CTC-based approach is the highest performing model overall. With the benefits of pre- and joint training, our work significantly closes the performance gap between whole-word and subword recognition systems in low- to modest-resource regimes, offering the first competitive application of A2W models to Switchboard-300h and several low-resource Babel languages. 

Additionally our work provides the first comprehensive comparison between static and dynamic lexicon approaches to A2W speech recognition. We find that many of the disadvantages of the static lexicon approach can be overcome by using AWE+AGWE pre- and joint training, though use of the dynamic lexicon approach provides an elegant method for test time extension and a more direct connection between multi-view pretraining and recognition joint training, particlularly in the case of the whole-word segmental model.

\section*{Self-supervised pretraining, transformers, and future work}

We find that pretrained self-supervised speech models learn high-quality AWE representations that can be extracted through simple normalization and mean pooling operations after segmentation. However, our models, while relying on labeled data, can often achieve competitive performance training on only a very small amount of data. Due to the effectiveness of both our own training approach as well as the high quality representations learned through self-supervised pretraining, we find that if we combine these techniques by using the self-supervised representations as input to our multi-view model, we can offer significant further improvements over either method alone. This is an area for future exploration in the context of many of the tasks we have already discussed in this thesis, including but not limited to both query-by-example speech search as well as whole-word speech recognition. In fact, the performance improvements are so drastic in many cases that the average precision is very close to $1$ on tasks that only a few years ago were seeing results under $0.5$~\cite{carlin2011rapid_eval_spoken_term_detect}.

In addition, our Multi-View method, while requiring discrete character (or phone) sequences in our case, could be easily trained to use any sequence of discrete units as the ``written" view. This means that simple adaptations of our method could be applied to the outputs of pretrained self-supervised systems such as HuBERT~\cite{hsu2021hubert} and WavLM~\cite{chen2022wavlm} to train on discrete cluster ID sequences rather than groundtruth character or phone sequences. This kind of hybrid method could even be explored as a new self-supervised approach to multi-view training.

%% file: main.bbl
\begin{thebibliography}{100}
\providecommand{\url}[1]{#1}
\csname url@samestyle\endcsname
\providecommand{\newblock}{\relax}
\providecommand{\bibinfo}[2]{#2}
\providecommand{\BIBentrySTDinterwordspacing}{\spaceskip=0pt\relax}
\providecommand{\BIBentryALTinterwordstretchfactor}{4}
\providecommand{\BIBentryALTinterwordspacing}{\spaceskip=\fontdimen2\font plus
\BIBentryALTinterwordstretchfactor\fontdimen3\font minus
  \fontdimen4\font\relax}
\providecommand{\BIBforeignlanguage}[2]{{%
\expandafter\ifx\csname l@#1\endcsname\relax
\typeout{** WARNING: IEEEtran.bst: No hyphenation pattern has been}%
\typeout{** loaded for the language `#1'. Using the pattern for}%
\typeout{** the default language instead.}%
\else
\language=\csname l@#1\endcsname
\fi
#2}}
\providecommand{\BIBdecl}{\relax}
\BIBdecl

\bibitem{carlin2011rapid_eval_spoken_term_detect}
M.~A. Carlin, S.~Thomas, A.~Jansen, and H.~Hermansky, ``Rapid evaluation of
  speech representations for spoken term discovery,'' in
  \emph{Proc.~{I}nterspeech}, 2011.

\bibitem{he2017multiview}
W.~He, W.~Wang, and K.~Livescu, ``Multi-view recurrent neural acoustic word
  embeddings,'' in \emph{Proc.~Intl. Conf. on Learning Representations
  ({ICLR})}, 2017.

\bibitem{maaten2008visualizing}
L.~{van der Maaten} and G.~Hinton, ``Visualizing data using {t-SNE},''
  \emph{Journal of Machine Learing Research}, 2008.

\bibitem{jyothi-livescu-2014-revisiting}
P.~Jyothi and K.~Livescu, ``Revisiting word neighborhoods for speech
  recognition,'' in \emph{Proc.~Joint Meeting of {SIGMORPHON} and {SIGFSM}},
  Jun. 2014.

\bibitem{park2019specaugment}
D.~S. Park, W.~Chan, Y.~Zhang, C.-C. Chiu, B.~Zoph, E.~D. Cubuk, and Q.~V. Le,
  ``Spec{A}ugment: A simple data augmentation method for automatic speech
  recognition,'' in \emph{Proc.~{I}nterspeech}, 2019.

\bibitem{maas2012word}
A.~L. Maas, S.~D. Miller, T.~M. O’neil, A.~Y. Ng, and P.~Nguyen, ``Word-level
  acoustic modeling with convolutional vector regression,'' in
  \emph{Proc.~Intl. Conf. on Machine Learning ({ICML}), Workshop on
  Representation Learning}, 2012.

\bibitem{levin2013fixed}
K.~Levin, K.~Henry, A.~Jansen, and K.~Livescu, ``Fixed-dimensional acoustic
  embeddings of variable-length segments in low-resource settings,'' in
  \emph{Proc.~{IEEE} Workshop on Automatic Speech Recognition and Understanding
  ({ASRU})}, 2013.

\bibitem{bengio2014word}
S.~Bengio and G.~Heigold, ``Word embeddings for speech recognition,'' in
  \emph{Proc.~{IEEE} Intl. Conf. Acoustics, Speech and Signal Processing
  ({ICASSP})}, 2014.

\bibitem{kamper2016cnn_awe}
H.~Kamper, W.~Wang, and K.~Livescu, ``Deep convolutional acoustic word
  embeddings using word-pair side information,'' in \emph{Proc.~{IEEE} Intl.
  Conf. Acoustics, Speech and Signal Processing ({ICASSP})}, 2016.

\bibitem{rabiner1978dtw}
L.~Rabiner, A.~Rosenberg, and S.~Levinson, ``Considerations in dynamic time
  warping algorithms for discrete word recognition,'' \emph{{IEEE} Transactions
  on Acoustics, Speech, and Signal Processing}, 1978.

\bibitem{levin2015srails}
K.~Levin, A.~Jansen, and B.~Van~Durme, ``Segmental acoustic indexing for zero
  resource keyword search,'' in \emph{Proc.~{IEEE} Intl. Conf. Acoustics,
  Speech and Signal Processing ({ICASSP})}, 2015.

\bibitem{livescu2012subword}
K.~Livescu, E.~Fosler-Lussier, and F.~Metze, ``Subword modeling for automatic
  speech recognition: Past, present, and emerging approaches,'' \emph{{IEEE}
  Signal Processing Magazine}, 2012.

\bibitem{settle2016rnn_awe}
S.~Settle and K.~Livescu, ``Discriminative acoustic word embeddings: Recurrent
  neural network-based approaches,'' in \emph{Proc.~{IEEE} Workshop on Spoken
  Language Technology ({SLT})}, 2016.

\bibitem{hu2020multilingual}
Y.~Hu, S.~Settle, and K.~Livescu, ``Multilingual jointly trained acoustic and
  written word embeddings,'' in \emph{Proc.~{I}nterspeech}, 2020.

\bibitem{settle2017query}
S.~Settle, K.~Levin, H.~Kamper, and K.~Livescu, ``Query-by-example search with
  discriminative neural acoustic word embeddings,'' in
  \emph{Proc.~{I}nterspeech}, 2017.

\bibitem{hu2021ase}
Y.~Hu, S.~Settle, and K.~Livescu, ``Acoustic span embeddings for multilingual
  query-by-example search,'' in \emph{Proc.~{IEEE} Workshop on Spoken Language
  Technology ({SLT})}, 2021.

\bibitem{hou2015nni_qbe}
J.~Hou, C.-C.~L. Van Tung~Pham, C.-C. Leung, L.~Wang, H.~Xu, H.~Lv, L.~Xie,
  Z.~Fu, C.~Ni, X.~Xiao \emph{et~al.}, ``The {NNI} query-by-example system for
  {MediaEval} 2015,'' in \emph{{MediaEval}}, 2015.

\bibitem{proencca2016segmented}
J.~Proen{\c{c}}a and F.~Perdig{\~a}o, ``Segmented dynamic time warping for
  spoken query-by-example search.'' in \emph{Proc.~{I}nterspeech}, 2016.

\bibitem{leung2016toward}
C.-C. Leung, L.~Wang, H.~Xu, J.~Hou, V.~T. Pham, H.~Lv, L.~Xie, X.~Xiao, C.~Ni,
  B.~Ma \emph{et~al.}, ``Toward high-performance language-independent
  query-by-example spoken term detection for {MediaEval} 2015: Post-evaluation
  analysis.'' in \emph{Proc.~{I}nterspeech}, 2016.

\bibitem{settle2019_a2w}
S.~Settle, K.~Audhkhasi, K.~Livescu, and M.~Picheny, ``Acoustically grounded
  word embeddings for improved acoustics-to-word speech recognition,'' in
  \emph{Proc.~{IEEE} Intl. Conf. Acoustics, Speech and Signal Processing
  ({ICASSP})}, 2019.

\bibitem{shi2021seg}
B.~Shi, S.~Settle, and K.~Livescu, ``Whole-word segmental speech recognition
  with acoustic word embeddings,'' in \emph{Proc.~{IEEE} Workshop on Spoken
  Language Technology ({SLT})}, 2021.

\bibitem{audhkhasi2017a2w}
K.~Audhkhasi, B.~Ramabhadran, G.~Saon, M.~Picheny, and D.~Nahamoo, ``Direct
  acoustics-to-word models for {English} conversational speech recognition,''
  in \emph{Proc.~{I}nterspeech}, 2017.

\bibitem{audhkhasi2018a2w}
K.~Audhkhasi, B.~Kingsbury, B.~Ramabhadran, G.~Saon, and M.~Picheny, ``Building
  competitive direct acoustics-to-word models for {English} conversational
  speech recognition,'' in \emph{Proc.~{IEEE} Intl. Conf. Acoustics, Speech and
  Signal Processing ({ICASSP})}, 2018.

\bibitem{yu2018multistage}
C.~Yu, C.~Zhang, C.~Weng, J.~Cui, and D.~Yu, ``A multistge training framework
  for acoustic-to-word model,'' in \emph{Proc.~{I}nterspeech}, 2018.

\bibitem{audhkhasi2019forget}
K.~Audhkhasi, G.~Saon, Z.~T{\"u}ske, B.~Kingsbury, and M.~Picheny, ``Forget a
  bit to learn better: Soft forgetting for ctc-based automatic speech
  recognition.'' in \emph{Proc.~{I}nterspeech}, 2019.

\bibitem{inaguma2019transfer}
H.~Inaguma, J.~Cho, M.~K. Baskar, T.~Kawahara, and S.~Watanabe, ``Transfer
  learning of language-independent end-to-end {ASR} with language model
  fusion,'' in \emph{Proc.~{IEEE} Intl. Conf. Acoustics, Speech and Signal
  Processing ({ICASSP})}, 2019.

\bibitem{conneau2020unsupervised}
A.~Conneau, A.~Baevski, R.~Collobert, A.~Mohamed, and M.~Auli, ``Unsupervised
  cross-lingual representation learning for speech recognition,'' \emph{arXiv
  preprint arXiv:2006.13979}, 2020.

\bibitem{baevski2020wav2vec2}
A.~Baevski, Y.~Zhou, A.~Mohamed, and M.~Auli, ``wav2vec 2.0: A framework for
  self-supervised learning of speech representations,'' in \emph{Advances in
  Neural Information Processing Systems ({NeurIPS})}, 2020.

\bibitem{hsu2021hubert}
W.-N. Hsu, B.~Bolte, Y.-H.~H. Tsai, K.~Lakhotia, R.~Salakhutdinov, and
  A.~Mohamed, ``Hubert: Self-supervised speech representation learning by
  masked prediction of hidden units,'' \emph{{IEEE} Transactions on Acoustics,
  Speech, and Language Processing}, 2021.

\bibitem{chen2022wavlm}
S.~Chen, C.~Wang, Z.~Chen, Y.~Wu, S.~Liu, Z.~Chen, J.~Li, N.~Kanda,
  T.~Yoshioka, X.~Xiao \emph{et~al.}, ``Wavlm: Large-scale self-supervised
  pre-training for full stack speech processing,'' \emph{{IEEE} Journal of
  Selected Topics in Signal Processing}, 2022.

\bibitem{ostendorf1999beads}
M.~Ostendorf, ``Moving beyond the ‘beads-on-a-string’model of speech,'' in
  \emph{Proc.~{IEEE} Workshop on Automatic Speech Recognition and Understanding
  ({ASRU})}, 1999.

\bibitem{vintsyuk1968speech}
T.~K. Vintsyuk, ``Speech discrimination by dynamic programming,''
  \emph{Cybernetics}, 1968.

\bibitem{sakoe1978dynamic}
H.~Sakoe and S.~Chiba, ``Dynamic programming algorithm optimization for spoken
  word recognition,'' \emph{{IEEE} Transactions on Acoustics, Speech, and
  Signal Processing}, 1978.

\bibitem{dewachter2007template}
M.~De~Wachter, M.~Matton, K.~Demuynck, P.~Wambacq, R.~Cools, and
  D.~Van~Compernolle, ``Template-based continuous speech recognition,''
  \emph{{IEEE} Transactions on Acoustics, Speech, and Language Processing},
  2007.

\bibitem{heigold2012investigations}
G.~Heigold, P.~Nguyen, M.~Weintraub, and V.~Vanhoucke, ``Investigations on
  exemplar-based features for speech recognition towards thousands of hours of
  unsupervised, noisy data,'' in \emph{Proc.~{IEEE} Intl. Conf. Acoustics,
  Speech and Signal Processing ({ICASSP})}, 2012.

\bibitem{hazen2009query_posteriorgram_templates}
T.~J. Hazen, W.~Shen, and C.~White, ``Query-by-example spoken term detection
  using phonetic posteriorgram templates,'' in \emph{Proc.~{IEEE} Workshop on
  Automatic Speech Recognition and Understanding ({ASRU})}, 2009.

\bibitem{zhang2009unsupervised_spoken_keyword_spotting}
Y.~Zhang and J.~R. Glass, ``Unsupervised spoken keyword spotting via segmental
  dtw on gaussian posteriorgrams,'' in \emph{Proc.~{IEEE} Workshop on Automatic
  Speech Recognition and Understanding ({ASRU})}, 2009.

\bibitem{zhang2011piecewise_posteriorgram_dtw}
Y.~Zhang and J.~Glass, ``A piecewise aggregate approximation lower-bound
  estimate for posteriorgram-based dynamic time warping,'' in
  \emph{Proc.~{I}nterspeech}, 2011.

\bibitem{zhang2012fast}
Y.~Zhang, K.~Adl, and J.~Glass, ``Fast spoken query detection using lower-bound
  dynamic time warping on graphical processing units,'' in \emph{Proc.~{IEEE}
  Intl. Conf. Acoustics, Speech and Signal Processing ({ICASSP})}, 2012.

\bibitem{jansen2012rails}
A.~Jansen and B.~V. Durme, ``Indexing raw acoustic features for scalable zero
  resource search,'' in \emph{Proc.~{I}nterspeech}, 2012.

\bibitem{mantena2013speedup_dtw_hierarchical_kmeans}
G.~Mantena and X.~Anguera, ``Speed improvements to information retrieval-based
  dynamic time warping using hierarchical k-means clustering,'' in
  \emph{Proc.~{IEEE} Intl. Conf. Acoustics, Speech and Signal Processing
  ({ICASSP})}, 2013.

\bibitem{jansen2013weak_topdown}
A.~Jansen, S.~Thomas, and H.~Hermansky, ``Weak top-down constraints for
  unsupervised acoustic model training,'' in \emph{Proc.~{IEEE} Intl. Conf.
  Acoustics, Speech and Signal Processing ({ICASSP})}, 2013.

\bibitem{kamper2015unsupervised_weak_topdown}
H.~Kamper, M.~Elsner, A.~Jansen, and S.~Goldwater, ``Unsupervised neural
  network based feature extraction using weak top-down constraints,'' in
  \emph{Proc.~{IEEE} Intl. Conf. Acoustics, Speech and Signal Processing
  ({ICASSP})}, 2015.

\bibitem{xu2016approximate}
H.~Xu, J.~Hou, X.~Xiao, V.~T. Pham, C.-C. Leung, L.~Wang, V.~H. Do, H.~Lv,
  L.~Xie, B.~Ma, E.~S. Chng, and H.~Li, ``Approximate search of audio queries
  by using {DTW} with phone time boundary and data augmentation,'' in
  \emph{Proc.~{IEEE} Intl. Conf. Acoustics, Speech and Signal Processing
  ({ICASSP})}, 2016.

\bibitem{mikolov2013efficient}
T.~Mikolov, K.~Chen, G.~Corrado, and J.~Dean, ``Efficient estimation of word
  representations in vector space,'' \emph{arXiv preprint arXiv:1301.3781},
  2013.

\bibitem{pennington2014glove}
J.~Pennington, R.~Socher, and C.~Manning, ``{G}lo{V}e: Global vectors for word
  representation,'' in \emph{Proc.~Conf. on Empirical Methods in Natural
  Language Processing ({EMNLP})}, 2014.

\bibitem{bojanowski2017enriching}
P.~Bojanowski, E.~Grave, A.~Joulin, and T.~Mikolov, ``Enriching word vectors
  with subword information,'' \emph{Transactions of the Association for
  Computational Linguistics}, 2017.

\bibitem{shen2009comparison}
W.~Shen, C.~M. White, and T.~J. Hazen, ``A comparison of query-by-example
  methods for spoken term detection,'' in \emph{Proc.~{I}nterspeech}, 2009.

\bibitem{parada2009query}
C.~Parada, A.~Sethy, and B.~Ramabhadran, ``Query-by-example spoken term
  detection for oov terms,'' in \emph{Proc.~{IEEE} Workshop on Automatic Speech
  Recognition and Understanding ({ASRU})}, 2009.

\bibitem{szoke2015query}
I.~Sz{\"o}ke, L.~J. Rodriguez-Fuentes, A.~Buzo, X.~Anguera, F.~Metze,
  J.~Proenca, M.~Lojka, and X.~Xiong, ``Query by example search on speech at
  {MediaEval} 2015.'' in \emph{MediaEval}, 2015.

\bibitem{fiscus1970spoken_term_detection}
J.~Fiscus, J.~Ajot, J.~Garofolo, and G.~Doddington, ``Results of the 2006
  spoken term detection evaluation,'' in \emph{ACM SIGIR}, 1970.

\bibitem{waibel1989asr}
A.~Waibel, T.~Hanazawa, G.~Hinton, K.~Shikano, and K.~Lang, ``Phoneme
  recognition using time-delay neural networks,'' \emph{{IEEE} Transactions on
  Acoustics, Speech, and Signal Processing}, 1989.

\bibitem{gauvain2003conversational}
J.-L. Gauvain, L.~Lamel, H.~Schwenk, G.~Adda, L.~Chen, and F.~Lefevre,
  ``Conversational telephone speech recognition,'' in \emph{Proc.~{IEEE} Intl.
  Conf. Acoustics, Speech and Signal Processing ({ICASSP})}, 2003.

\bibitem{chen2006advances}
S.~F. Chen, B.~Kingsbury, L.~Mangu, D.~Povey, G.~Saon, H.~Soltau, and G.~Zweig,
  ``{Advances in speech transcription at IBM under the DARPA EARS program},''
  \emph{{IEEE} Transactions on Acoustics, Speech, and Language Processing},
  2006.

\bibitem{matsoukas2006advances}
S.~Matsoukas, J.-L. Gauvain, G.~Adda, T.~Colthurst, C.-L. Kao, O.~Kimball,
  L.~Lamel, F.~Lefevre, J.~Z. Ma, J.~Makhoul \emph{et~al.}, ``Advances in
  transcription of broadcast news and conversational telephone speech within
  the combined ears bbn/limsi system,'' \emph{{IEEE} Transactions on Acoustics,
  Speech, and Language Processing}, 2006.

\bibitem{jurafsky2009speech}
D.~Jurafsky and J.~H. Martin, \emph{Speech and language processing: an
  introduction to natural language processing, computational linguistics, and
  speech recognition}.\hskip 1em plus 0.5em minus 0.4em\relax Pearson Prentice
  Hall, 2009.

\bibitem{saon2015ibm_asr}
G.~Saon, H.-K.~J. Kuo, S.~Rennie, and M.~Picheny, ``{The IBM 2015 English
  conversational telephone speech recognition system},'' in
  \emph{Proc.~{I}nterspeech}, 2015.

\bibitem{xiong2016achieving}
W.~Xiong, J.~Droppo, X.~Huang, F.~Seide, M.~Seltzer, A.~Stolcke, D.~Yu, and
  G.~Zweig, ``Achieving human parity in conversational speech recognition,''
  \emph{arXiv preprint arXiv:1610.05256}, 2016.

\bibitem{rao2017exploring_rnn_transducer}
K.~Rao, H.~Sak, and R.~Prabhavalkar, ``Exploring architectures, data and units
  for streaming end-to-end speech recognition with rnn-transducer,'' in
  \emph{Proc.~{IEEE} Workshop on Automatic Speech Recognition and Understanding
  ({ASRU})}, 2017.

\bibitem{chiu2018sota_asr_with_seq2seq}
C.-C. Chiu, T.~N. Sainath, Y.~Wu, R.~Prabhavalkar, P.~Nguyen, Z.~Chen,
  A.~Kannan, R.~J. Weiss, K.~Rao, E.~Gonina \emph{et~al.}, ``State-of-the-art
  speech recognition with sequence-to-sequence models,'' in \emph{Proc.~{IEEE}
  Intl. Conf. Acoustics, Speech and Signal Processing ({ICASSP})}, 2018.

\bibitem{sanabria2018hierarchical}
R.~Sanabria and F.~Metze, ``{Hierarchical Multi Task Learning With CTC},'' in
  \emph{Proc.~{IEEE} Workshop on Spoken Language Technology ({SLT})}, 2018.

\bibitem{krishna2018hierarchical}
K.~Krishna, S.~Toshniwal, and K.~Livescu, ``{Hierarchical Multitask Learning
  for CTC-based Speech Recognition},'' \emph{arXiv preprint arXiv:1807.06234},
  2018.

\bibitem{yuan2016learning_from_bnf}
Y.~Yuan, C.-C. Leung, L.~Xie, B.~Ma, and H.~Li, ``Learning neural network
  representations using cross-lingual bottleneck features with word-pair
  information.'' in \emph{Proc.~{I}nterspeech}, 2016.

\bibitem{yuan2018learning_awe_for_qbe}
Y.~Yuan, C.-C. Leung, L.~Xie, H.~Chen, B.~Ma, and H.~Li, ``Learning acoustic
  word embeddings with temporal context for query-by-example speech search,''
  in \emph{Proc.~{I}nterspeech}, 2018.

\bibitem{zweig2009segmental}
G.~Zweig and P.~Nguyen, ``A segmental {CRF} approach to large vocabulary
  continuous speech recognition,'' in \emph{Proc.~{IEEE} Workshop on Automatic
  Speech Recognition and Understanding ({ASRU})}, 2009.

\bibitem{kamper2014unsupervised_lexical}
H.~Kamper, A.~Jansen, S.~King, and S.~Goldwater, ``Unsupervised lexical
  clustering of speech segments using fixed-dimensional acoustic embeddings,''
  in \emph{Proc.~{IEEE} Workshop on Spoken Language Technology ({SLT})}, 2014.

\bibitem{kamper2015unsupervised_small_vocab_asr}
H.~Kamper, A.~Jansen, and S.~Goldwater, ``Fully unsupervised small-vocabulary
  speech recognition using a segmental bayesian model,'' in
  \emph{Proc.~{I}nterspeech}, 2015.

\bibitem{lu2015study_rnn_lvsr}
L.~Lu, X.~Zhang, K.~Cho, and S.~Renals, ``A study of the recurrent neural
  network encoder-decoder for large vocabulary speech recognition,'' in
  \emph{Proc.~{I}nterspeech}, 2015.

\bibitem{chen2015qbe_kws_lstm}
G.~Chen, C.~Parada, and T.~N. Sainath, ``Query-by-example keyword spotting
  using long short-term memory networks,'' in \emph{Proc.~{IEEE} Intl. Conf.
  Acoustics, Speech and Signal Processing ({ICASSP})}, 2015.

\bibitem{chung2016audio}
Y.-A. Chung, C.-C. Wu, C.-H. Shen, H.-Y. Lee, and L.-S. Lee, ``Audio word2vec:
  Unsupervised learning of audio segment representations using
  sequence-to-sequence autoencoder,'' in \emph{Proc.~{I}nterspeech}, 2016.

\bibitem{holzenberger2018unsupervised_awe}
N.~Holzenberger, M.~Du, J.~Karadayi, R.~Riad, and E.~Dupoux, ``Learning word
  embeddings: Unsupervised methods for fixed-size representations of
  variable-length speech segments,'' in \emph{Proc.~{I}nterspeech}, 2018.

\bibitem{kamper2019truly_unsupervised_awe}
H.~Kamper, ``Truly unsupervised acoustic word embeddings using weak top-down
  constraints in encoder-decoder models,'' in \emph{Proc.~{IEEE} Intl. Conf.
  Acoustics, Speech and Signal Processing ({ICASSP})}, 2019.

\bibitem{chung2018speech2vec}
Y.-A. Chung and J.~Glass, ``Speech2vec: A sequence-to-sequence framework for
  learning word embeddings from speech,'' in \emph{Proc.~{I}nterspeech}, 2018.

\bibitem{palaskar2019learned}
S.~Palaskar, V.~Raunak, and F.~Metze, ``Learned in speech recognition:
  Contextual acoustic word embeddings,'' in \emph{Proc.~{IEEE} Intl. Conf.
  Acoustics, Speech and Signal Processing ({ICASSP})}, 2019.

\bibitem{yang2019linguistically_informed_awe}
Z.~Yang and J.~Hirschberg, ``Linguistically-informed training of acoustic word
  embeddings for low-resource languages,'' in \emph{Proc.~{I}nterspeech}, 2019.

\bibitem{kamper2020multilingual}
H.~Kamper, Y.~Matusevych, and S.~Goldwater, ``Multilingual acoustic word
  embedding models for processing zero-resource languages,'' in
  \emph{Proc.~{IEEE} Intl. Conf. Acoustics, Speech and Signal Processing
  ({ICASSP})}, 2020.

\bibitem{van2020improving}
L.~van Staden and H.~Kamper, ``Improving unsupervised acoustic word embeddings
  using speaker and gender information,'' in \emph{{Proc.~International
  SAUPEC/RobMech/PRASA Conference}}, 2020.

\bibitem{matusevych2020analyzing_AE_AWEs}
Y.~Matusevych, H.~Kamper, and S.~Goldwater, ``Analyzing autoencoder-based
  acoustic word embeddings,'' in \emph{{BAICS} Workshop at Intl. Conf. on
  Learning Representations ({ICLR})}, 2020.

\bibitem{kamper2021improved_multilingual}
H.~Kamper, Y.~Matusevych, and S.~Goldwater, ``Improved acoustic word embeddings
  for zero-resource languages using multilingual transfer,'' \emph{{IEEE}
  Transactions on Acoustics, Speech, and Language Processing}, 2021.

\bibitem{jacobs2021acoustic}
C.~Jacobs, Y.~Matusevych, and H.~Kamper, ``Acoustic word embeddings for
  zero-resource languages using self-supervised contrastive learning and
  multilingual adaptation,'' in \emph{Proc.~{IEEE} Workshop on Spoken Language
  Technology ({SLT})}, 2021.

\bibitem{audhkhasi2017asr_free_kws}
K.~Audhkhasi, A.~Rosenberg, A.~Sethy, B.~Ramabhadran, and B.~Kingsbury,
  ``End-to-end asr-free keyword search from speech,'' \emph{{IEEE} Journal of
  Selected Topics in Signal Processing}, 2017.

\bibitem{cui2015multilingual_representations}
J.~Cui, B.~Kingsbury, B.~Ramabhadran, A.~Sethy, K.~Audhkhasi, X.~Cui,
  E.~Kislal, L.~Mangu, M.~Nussbaum-Thom, M.~Picheny, Z.~Tüske, P.~Golik,
  R.~Schlüter, H.~Ney, M.~J.~F. Gales, K.~M. Knill, A.~Ragni, H.~Wang, and
  P.~Woodland, ``Multilingual representations for low resource speech
  recognition and keyword search,'' in \emph{Proc.~{IEEE} Workshop on Automatic
  Speech Recognition and Understanding ({ASRU})}, 2015.

\bibitem{allauzen2004general}
C.~Allauzen, M.~Mohri, and M.~Saraclar, ``General indexation of weighted
  automata-application to spoken utterance retrieval,'' in \emph{Proc.~Human
  Language Technology/Conf. of the North {A}merican Chapter of the Association
  for Computational Linguistics ({HLT/NAACL})}, 2004.

\bibitem{lafarga2015elirf}
M.~C. Lafarga, M.~Gim{\'e}nez, L.~F. Hurtado, E.~S. Arnal, and J.~A. G{\'o}mez,
  ``{ELiRF} at {MediaEval} 2015: Query by example search on speech task
  ({QUESST}),'' in \emph{{MediaEval}}, 2015.

\bibitem{lopez-otero2015gtm-uvigo_qbe}
P.~Lopez-Otero, L.~Docio-Fernandez, and C.~García-Mateo, ``{GTM-UVigo} systems
  for the query-by-example search on speech task at {MediaEval} 2015,'' in
  \emph{MediaEval}, 2015.

\bibitem{chen2016unsupervised_bnf_qbe}
H.~Chen, C.-C. Leung, L.~Xie, B.~Ma, and H.~Li, ``Unsupervised bottleneck
  features for low-resource query-by-example spoken term detection.'' in
  \emph{Proc.~{I}nterspeech}, 2016.

\bibitem{soltau2016a2w}
H.~Soltau, H.~Liao, and H.~Sak, ``Neural speech recognizer: Acoustic-to-word
  {LSTM} model for large vocabulary speech recognition,'' in
  \emph{Proc.~{I}nterspeech}, 2016.

\bibitem{levinson1983isolated_digit}
S.~Levinson, L.~Rabiner, and M.~Sondhi, ``Speaker independent isolated digit
  recognition using hidden markov models,'' in \emph{Proc.~{IEEE} Intl. Conf.
  Acoustics, Speech and Signal Processing ({ICASSP})}, 1983.

\bibitem{li2017acoustic}
J.~Li, G.~Ye, R.~Zhao, J.~Droppo, and Y.~Gong, ``{Acoustic-to-word model
  without OOV},'' in \emph{Proc.~{IEEE} Workshop on Automatic Speech
  Recognition and Understanding ({ASRU})}, 2017.

\bibitem{li2018advancing}
J.~Li, G.~Ye, A.~Das, R.~Zhao, and Y.~Gong, ``Advancing acoustic-to-word {CTC}
  model,'' in \emph{Proc.~{IEEE} Intl. Conf. Acoustics, Speech and Signal
  Processing ({ICASSP})}, 2018.

\bibitem{gaur2019acoustic_phrase}
Y.~Gaur, J.~Li, Z.~Meng, and Y.~Gong, ``Acoustic-to-phrase models for speech
  recognition,'' in \emph{Proc.~{I}nterspeech}, 2019.

\bibitem{collobert2019wordlevel}
R.~Collobert, A.~Hannun, and G.~Synnaeve, ``Word-level speech recognition with
  a dynamic lexicon,'' \emph{arXiv preprint arXiv:1906.04323v1}, 2019.

\bibitem{collobert2020word}
------, ``Word-level speech recognition with a letter to word encoder,'' in
  \emph{Proc.~Intl. Conf. on Machine Learning ({ICML})}, 2020.

\bibitem{palaskar2018_s2s}
S.~Palaskar and F.~Metze, ``Acoustic-to-word recognition with
  sequence-to-sequence models,'' in \emph{Proc.~{IEEE} Workshop on Spoken
  Language Technology ({SLT})}, 2018.

\bibitem{ostendorf1996hmm}
M.~Ostendorf, V.~Digalakis, and O.~Kimball, ``From {HMM}'s to segment models: a
  unified view of stochastic modeling for speech recognition,'' \emph{{IEEE}
  Transactions on Speech and Audio Processing}, 1996.

\bibitem{glass2003probabilistic}
J.~R. Glass, ``A probabilistic framework for segment-based speech
  recognition,'' \emph{Computer Speech \& Language}, 2003.

\bibitem{zweig2012_seg}
G.~Zweig, ``Classification and recognition with direct segment models,'' in
  \emph{Proc.~{IEEE} Intl. Conf. Acoustics, Speech and Signal Processing
  ({ICASSP})}, 2012.

\bibitem{he2012_sc}
Y.~He and E.~Fosler-Lussier, ``Efficient segmental conditional random fields
  for phone recognition,'' in \emph{Proc.~{I}nterspeech}, 2012.

\bibitem{hamid2013_dsnn}
O.~Abdel-Hamid, L.~Deng, D.~Yu, and H.~Jiang, ``Deep segmental neural networks
  for speech recognition,'' in \emph{Proc.~{I}nterspeech}, 2013.

\bibitem{tang2014_losses}
H.~Tang, K.~Gimpel, and K.~Livescu, ``A comparison of training approaches for
  discriminative segmental models,'' in \emph{Proc.~{I}nterspeech}, 2014.

\bibitem{lu2016_srnn}
L.~Lu, L.~Kong, C.~Dyer, N.~A. Smith, and S.~Renals, ``Segmental recurrent
  neural networks for end-to-end speech recognition,'' in
  \emph{Proc.~{I}nterspeech}, 2016.

\bibitem{schatz2013evaluating_abx}
T.~Schatz, V.~Peddinti, F.~Bach, A.~Jansen, H.~Hermansky, and E.~Dupoux,
  ``{Evaluating speech features with the minimal-pair ABX task: Analysis of the
  classical mfc/plp pipeline},'' in \emph{Proc.~{I}nterspeech}, 2013.

\bibitem{schatz2014evaluating_abx_2}
T.~Schatz, V.~Peddinti, X.-N. Cao, F.~Bach, H.~Hermansky, and E.~Dupoux,
  ``{Evaluating speech features with the minimal-pair ABX task (II): resistance
  to noise},'' in \emph{Proc.~{I}nterspeech}, 2014.

\bibitem{palaz2016keyword}
D.~Palaz, G.~Synnaeve, and R.~Collobert, ``Jointly learning to locate and
  classify words using convolutional networks,'' in \emph{Proc.~{I}nterspeech},
  2016.

\bibitem{metze2013spoken}
F.~Metze, X.~Anguera, E.~Barnard, M.~Davel, and G.~Gravier, ``The spoken web
  search task at {MediaEval} 2012,'' in \emph{Proc.~{IEEE} Intl. Conf.
  Acoustics, Speech and Signal Processing ({ICASSP})}, 2013.

\bibitem{anguera2012speaker}
X.~Anguera, ``Speaker independent discriminant feature extraction for acoustic
  pattern-matching,'' in \emph{Proc.~{IEEE} Intl. Conf. Acoustics, Speech and
  Signal Processing ({ICASSP})}, 2012.

\bibitem{szoke2015coping}
I.~Sz{\"o}ke, M.~Sk{\'a}cel, L.~Burget, and J.~{\v{C}}ernock{\`y}, ``Coping
  with channel mismatch in query-by-example-but quesst 2014,'' in
  \emph{Proc.~{IEEE} Intl. Conf. Acoustics, Speech and Signal Processing
  ({ICASSP})}, 2015.

\bibitem{voinea2014word}
S.~Voinea, C.~Zhang, G.~Evangelopoulos, L.~Rosasco, and T.~Poggio, ``Word-level
  invariant representations from acoustic waveforms,'' in
  \emph{Proc.~{I}nterspeech}, 2014.

\bibitem{hochreiter1997lstm}
S.~Hochreiter and J.~Schmidhuber, ``Long short-term memory,'' \emph{Neural
  Computation}, 1997.

\bibitem{chung2014gru}
J.~Chung, C.~Gulcehre, K.~Cho, and Y.~Bengio, ``Empirical evaluation of gated
  recurrent neural networks on sequence modeling,'' in \emph{Advances in Neural
  Information Processing Systems ({NIPS})}, 2014.

\bibitem{graves2013speech}
A.~Graves, A.~Mohamed, and G.~Hinton, ``Speech recognition with deep recurrent
  neural networks,'' in \emph{Proc.~{IEEE} Intl. Conf. Acoustics, Speech and
  Signal Processing ({ICASSP})}, 2013.

\bibitem{sak2014lstm_lvsr}
H.~Sak, A.~Senior, and F.~Beaufays, ``Long short-term memory based recurrent
  neural network architectures for large vocabulary speech recognition,''
  \emph{arXiv preprint arXiv:1402.1128}, 2014.

\bibitem{chorowski2015attention}
J.~Chorowski, D.~Bahdanau, D.~Serdyuk, K.~Cho, and Y.~Bengio, ``Attention-based
  models for speech recognition,'' in \emph{Advances in Neural Information
  Processing Systems ({NIPS})}, 2015.

\bibitem{nair2010rectified}
V.~Nair and G.~E. Hinton, ``Rectified linear units improve restricted boltzmann
  machines,'' in \emph{Proc.~Intl. Conf. on Machine Learning ({ICML})}, 2010.

\bibitem{synnaeve2014phonetic_embedding}
G.~Synnaeve, T.~Schatz, and E.~Emmanuel~Dupoux, ``Phonetics embedding learning
  with side information,'' in \emph{Proc.~{IEEE} Workshop on Spoken Language
  Technology ({SLT})}, 2014.

\bibitem{bromley1993signature_verification_siamese}
J.~Bromley, I.~Guyon, Y.~LeCun, E.~S{\"a}ckinger, and R.~Shah, ``Signature
  verification using a ``siamese" time delay neural network,'' in
  \emph{Advances in Neural Information Processing Systems ({NIPS})}, 1993.

\bibitem{godfrey1992switchboard}
J.~J. Godfrey, E.~C. Holliman, and J.~McDaniel, ``Switchboard: Telephone speech
  corpus for research and development,'' in \emph{Proc.~{IEEE} Intl. Conf.
  Acoustics, Speech and Signal Processing ({ICASSP})}, 1992.

\bibitem{povey2011kaldi}
D.~Povey, A.~Ghoshal, G.~Boulianne, L.~Burget, O.~Glembek, N.~Goel,
  M.~Hannemann, P.~Motlicek, Y.~Qian, P.~Schwarz \emph{et~al.}, ``The {Kaldi}
  speech recognition toolkit,'' in \emph{Proc.~{IEEE} Workshop on Automatic
  Speech Recognition and Understanding ({ASRU})}, 2011.

\bibitem{torch7}
R.~Collobert, K.~Kavukcuoglu, and C.~Farabet, ``Torch7: A matlab-like
  environment for machine learning,'' in \emph{Advances in Neural Information
  Processing Systems ({NIPS})}, 2011.

\bibitem{srivastava2014dropout}
N.~Srivastava, G.~E. Hinton, A.~Krizhevsky, I.~Sutskever, and R.~Salakhutdinov,
  ``Dropout: a simple way to prevent neural networks from overfitting.''
  \emph{Journal of Machine Learing Research}, 2014.

\bibitem{nesterov1983method}
Y.~Nesterov, ``A method for solving the convex programming problem with
  convergence rate {$O\left(\frac1{k^2}\right)$},'' in \emph{Dokl. Akad. Nauk
  SSSR}, 1983.

\bibitem{belkin2003laplacian_eigenmaps}
M.~Belkin and P.~Niyogi, ``Laplacian eigenmaps for dimensionality reduction and
  data representation,'' \emph{Neural Computation}, 2003.

\bibitem{chopra2005learning_similarity_face}
S.~Chopra, R.~Hadsell, and Y.~LeCun, ``Learning a similarity metric
  discriminatively, with application to face verification,'' in
  \emph{Proc.~{IEEE} Conf. Computer Vision and Pattern Recognition ({CVPR})},
  2005.

\bibitem{socher2014grounded}
R.~Socher, A.~Karpathy, Q.~V. Le, C.~D. Manning, and A.~Y. Ng, ``Grounded
  compositional semantics for finding and describing images with sentences,''
  \emph{Transactions of the Association for Computational Linguistics}, 2014.

\bibitem{indyk1998approximate_nearest_neighbors}
P.~Indyk and R.~Motwani, ``Approximate nearest neighbors: towards removing the
  curse of dimensionality,'' in \emph{Proc.~{ACM} symposium on {T}heory of
  {C}omputing}, 1998.

\bibitem{charikar2002similarity_from_rounding_algo}
M.~S. Charikar, ``Similarity estimation techniques from rounding algorithms,''
  in \emph{Proc. ACM symposium on {T}heory of {C}omputing}, 2002.

\bibitem{kingma2014adam}
D.~P. Kingma and J.~Ba, ``Adam: A method for stochastic optimization,'' in
  \emph{Proc.~Intl. Conf. on Learning Representations ({ICLR})}, 2014.

\bibitem{miller2007rapid_spoken_term_detect}
D.~R. Miller, M.~Kleber, C.-l. Kao, O.~Kimball, T.~Colthurst, S.~A. Lowe, R.~M.
  Schwartz, and H.~Gish, ``Rapid and accurate spoken term detection,'' in
  \emph{Proc.~{I}nterspeech}, 2007.

\bibitem{deerwester1990indexing}
S.~Deerwester, S.~Dumais, G.~Furnas, T.~Landauer, and R.~Harshman, ``Indexing
  by latent semantic analysis,'' \emph{Journal of the {A}merican society for
  information science}, 1990.

\bibitem{bengio2003neural}
Y.~Bengio, R.~Ducharme, P.~Vincent, and C.~Jauvin, ``A neural probabilistic
  language model,'' \emph{Journal of Machine Learing Research}, 2003.

\bibitem{mnih2007three}
A.~Mnih and G.~Hinton, ``Three new graphical models for statistical language
  modelling,'' in \emph{Proc.~Intl. Conf. on Machine Learning ({ICML})}, 2007.

\bibitem{chen2018phonetic}
Y.-C. Chen, S.-F. Huang, C.-H. Shen, H.-Y. Lee, and L.-S. Lee,
  ``Phonetic-and-semantic embedding of spoken words with applications in spoken
  content retrieval,'' \emph{Proc.~{IEEE} Workshop on Spoken Language
  Technology ({SLT})}, 2018.

\bibitem{ghannay2016evaluation_awe}
S.~Ghannay, Y.~Esteve, N.~Camelin, and P.~Deleglise, ``Evaluation of acoustic
  word embeddings,'' in \emph{Proc.~{ACL} Workshop on Evaluating Vector-Space
  Representations for NLP}, 2016.

\bibitem{pytorch}
A.~Paszke \emph{et~al.}, ``{PyTorch}: An imperative style, high-performance
  deep learning library,'' in \emph{Advances in Neural Information Processing
  Systems ({NeurIPS})}, 2019.

\bibitem{wells1995xsampa}
J.~Wells, ``Computer-coding the {IPA}: a proposed extension of {SAMPA},'' 1995.

\bibitem{phoible}
\BIBentryALTinterwordspacing
S.~Moran and D.~McCloy, Eds., \emph{PHOIBLE 2.0}.\hskip 1em plus 0.5em minus
  0.4em\relax Jena: Max Planck Institute for the Science of Human History,
  2019. [Online]. Available: \url{https://phoible.org/}
\BIBentrySTDinterwordspacing

\bibitem{babel_data}
{Philadelphia: Linguistic Data Consortium}, ``{IARPA Babel} language pack,''
  {IARPA}-babel101b-v0.4c, {IARPA}-babel102b-v0.5a, {IARPA}-babel103b-v0.4b,
  {IARPA}-babel104b-v0.4bY, {IARPA}-babel105b-v0.5, {IARPA}-babel106-v0.2g,
  {IARPA}-babel204b-v1.1b, {IARPA}-babel206b-v0.1e, {IARPA}-babel304b-v1.0b,
  {IARPA}-babel305b-v1.0c, {IARPA}-babel306b-v2.0c.

\bibitem{jansen2013jhu}
A.~{Jansen}, E.~{Dupoux}, S.~{Goldwater}, M.~{Johnson}, S.~{Khudanpur},
  K.~{Church}, N.~{Feldman}, H.~{Hermansky}, F.~{Metze}, R.~{Rose},
  M.~{Seltzer}, P.~{Clark}, I.~{ McGraw}, B.~{Varadarajan}, E.~{Bennett},
  B.~{Borschinger}, J.~{Chiu}, E.~{Dunbar}, A.~{Fourtassi}, D.~{Harwath},
  C.~{Lee}, K.~{Levin}, A.~{Norouzian}, V.~{Peddinti}, R.~{ Richardson},
  T.~{Schatz}, and S.~{Thomas}, ``A summary of the 2012 {JHU} {CLSP} workshop
  on zero resource speech technologies and models of early language
  acquisition,'' in \emph{Proc.~{IEEE} Intl. Conf. Acoustics, Speech and Signal
  Processing ({ICASSP})}, 2013.

\bibitem{anguera2014query}
X.~Anguera, L.~J. Rodriguez-Fuentes, I.~Sz{\"o}ke, A.~Buzo, and F.~Metze,
  ``Query by example search on speech at {MediaEval} 2014,'' in
  \emph{{MediaEval}}, 2014.

\bibitem{schroff2015facenet}
F.~Schroff, D.~Kalenichenko, and J.~Philbin, ``{FaceNet}: A unified embedding
  for face recognition and clustering,'' in \emph{Proc.~{IEEE} Conf. Computer
  Vision and Pattern Recognition ({CVPR})}, 2015.

\bibitem{graves2006connectionist}
A.~Graves, S.~Fern{\'a}ndez, F.~Gomez, and J.~Schmidhuber, ``Connectionist
  temporal classification: labelling unsegmented sequence data with recurrent
  neural networks,'' in \emph{Proc.~Intl. Conf. on Machine Learning ({ICML})},
  2006.

\bibitem{tang2015_cascade}
H.~Tang, W.~Wang, K.~Gimpel, and K.~Livescu, ``Discriminative segmental
  cascades for feature-rich phone recognition,'' in \emph{Proc.~{IEEE} Workshop
  on Automatic Speech Recognition and Understanding ({ASRU})}, 2015.

\bibitem{tang2017_e2e}
H.~Tang, L.~Lu, K.~Gimpel, C.~Dyer, and A.~S. Smith, ``End-to-end neural
  segmental models for speech recognition,'' \emph{{IEEE} Journal of Selected
  Topics in Signal Processing}, 2017.

\bibitem{kong2016_srnn}
L.~Kong, C.~Dyer, and N.~A. Smith, ``Segmental recurrent neural networks,'' in
  \emph{Proc.~Intl. Conf. on Learning Representations ({ICLR})}, 2016.

\bibitem{zweig2011speech}
G.~Zweig \emph{et~al.}, ``Speech recognition with segmental conditional random
  fields: A summary of the {JHU CLSP} 2010 summer workshop,'' in
  \emph{Proc.~{IEEE} Intl. Conf. Acoustics, Speech and Signal Processing
  ({ICASSP})}, 2011.

\bibitem{wang2020investigation}
W.~Wang, Y.~Zhou, C.~Xiong, and R.~Socher, ``An investigation of phone-based
  subword units for end-to-end speech recognition,'' in
  \emph{Proc.~{I}nterspeech}, 2020.

\bibitem{ghannay2020study}
S.~Ghannay, Y.~Est{\`e}ve, and N.~Camelin, ``A study of continuous space word
  and sentence representations applied to {ASR} error detection,'' \emph{Speech
  Communication}, 2020.

\bibitem{panayotov2015librispeech}
V.~Panayotov, G.~Chen, D.~Povey, and S.~Khudanpur, ``Librispeech: an asr corpus
  based on public domain audio books,'' in \emph{Proc.~{IEEE} Intl. Conf.
  Acoustics, Speech and Signal Processing ({ICASSP})}, 2015.

\bibitem{mohamed2022ssl_review}
A.~Mohamed, H.-Y. Lee, L.~Borgholt, J.~D. Havtorn, J.~Edin, C.~Igel,
  K.~Kirchhoff, S.-W. Li, K.~Livescu, L.~Maal{\o}e \emph{et~al.},
  ``Self-supervised speech representation learning: A review,'' \emph{{IEEE}
  Journal of Selected Topics in Signal Processing}, 2022.

\bibitem{pasad2021layerwise}
A.~Pasad, J.-C. Chou, and K.~Livescu, ``Layer-wise analysis of a
  self-supervised speech representation model,'' in \emph{Proc.~{IEEE} Workshop
  on Automatic Speech Recognition and Understanding ({ASRU})}, 2021.

\bibitem{yang2021superb}
S.-W. Yang, P.-H. Chi, Y.-S. Chuang, C.-I.~J. Lai, K.~Lakhotia, Y.~Y. Lin,
  A.~T. Liu, J.~Shi, X.~Chang, G.-T. Lin \emph{et~al.}, ``Superb: Speech
  processing universal performance benchmark,'' \emph{arXiv preprint
  arXiv:2105.01051}, 2021.

\bibitem{van2021ssl_comparison_for_awe}
L.~van Staden and H.~Kamper, ``A comparison of self-supervised speech
  representations as input features for unsupervised acoustic word
  embeddings,'' in \emph{Proc.~{IEEE} Workshop on Spoken Language Technology
  ({SLT})}, 2021.

\bibitem{oord2018cpc}
A.~van~den Oord, Y.~Li, and O.~Vinyals, ``Representation learning with
  contrastive predictive coding,'' \emph{arXiv preprint arXiv:1807.03748},
  2018.

\bibitem{chung2019apc}
Y.-A. Chung, W.-N. Hsu, H.~Tang, and J.~Glass, ``An unsupervised autoregressive
  model for speech representation learning,'' in \emph{Proc.~{I}nterspeech},
  2019.

\bibitem{sanabria2022analyzing}
R.~Sanabria, H.~Tang, and S.~Goldwater, ``Analyzing acoustic word embeddings
  from pre-trained self-supervised speech models,'' in \emph{Proc.~{IEEE} Intl.
  Conf. Acoustics, Speech and Signal Processing ({ICASSP})}, 2023.

\bibitem{devlin2018bert}
J.~Devlin, M.-W. Chang, K.~Lee, and K.~Toutanova, ``{BERT}: Pre-training of
  deep bidirectional transformers for language understanding,'' in
  \emph{Proc.~North {A}merican Chapter of the Association for Computational
  Linguistics ({NAACL})}, 2019.

\bibitem{pratap2020mls}
V.~Pratap, Q.~Xu, A.~Sriram, G.~Synnaeve, and R.~Collobert, ``{MLS}: A
  large-scale multilingual dataset for speech research,'' in
  \emph{Proc.~{I}nterspeech}, 2020.

\bibitem{ardila2019commonvoice}
R.~Ardila, M.~Branson, K.~Davis, M.~Henretty, M.~Kohler, J.~Meyer, R.~Morais,
  L.~Saunders, F.~M. Tyers, and G.~Weber, ``{Common Voice}: A
  massively-multilingual speech corpus,'' in \emph{Intl. Conf. on Language
  Resources and Evaluation ({LREC})}, 2020.

\bibitem{kahn2020librilight}
J.~Kahn, M.~Rivi{\`e}re, W.~Zheng, E.~Kharitonov, Q.~Xu, P.-E. Mazar{\'e},
  J.~Karadayi, V.~Liptchinsky, R.~Collobert, C.~Fuegen \emph{et~al.},
  ``Libri-light: A benchmark for asr with limited or no supervision,'' in
  \emph{Proc.~{IEEE} Intl. Conf. Acoustics, Speech and Signal Processing
  ({ICASSP})}, 2020.

\bibitem{wang2021voxpopuli}
C.~Wang, M.~Riviere, A.~Lee, A.~Wu, C.~Talnikar, D.~Haziza, M.~Williamson,
  J.~Pino, and E.~Dupoux, ``{VoxPopuli}: A large-scale multilingual speech
  corpus for representation learning, semi-supervised learning and
  interpretation,'' in \emph{Proc.~Association for Computational Linguistics},
  2021.

\bibitem{chen2021gigaspeech}
G.~Chen, S.~Chai, G.~Wang, J.~Du, W.-Q. Zhang, C.~Weng, D.~Su, D.~Povey,
  J.~Trmal, J.~Zhang \emph{et~al.}, ``{GigaSpeech}: An evolving, multi-domain
  asr corpus with 10,000 hours of transcribed audio,'' in
  \emph{Proc.~{I}nterspeech}, 2021.

\end{thebibliography}
